%% file: main.tex
\documentclass[MAL, biber]{nowfnt_arxiv} 

\usepackage[utf8]{inputenc}
\usepackage{amsmath,amsfonts,amssymb,amsthm}
\usepackage[dvipsnames]{xcolor} 
\usepackage{xurl} 
\usepackage{hyperref}
\usepackage[capitalise]{cleveref}
\usepackage{booktabs}
\usepackage{tikz}
\usepackage{listings}
\usepackage{multirow}
\usepackage{verbatim}
\usepackage{comment}
\usepackage{adjustbox}
\usepackage{enumitem}
\usepackage{blkarray}
\usepackage{mathtools}
\usepackage{chngcntr}
\usepackage{subcaption}
\usepackage{bm}
\usepackage{thmtools}
\usepackage{natbib}
\usepackage[nottoc]{tocbibind}

\usepackage{algorithm}
\counterwithin{algorithm}{chapter}
\usepackage[noend]{algpseudocode} 
\algrenewcommand\algorithmicrequire{\textbf{Input:}}
\algrenewcommand\algorithmicensure{\textbf{Output:}}
\algrenewcommand{\algorithmiccomment}[1]{\hfill\textcolor{blue}{$\blacktriangleright$ \textit{#1}}}  

\input{macros.tex}

\renewcommand{\cite}{\citep}

\definecolor{dkgreen}{rgb}{0,0.6,0}
\definecolor{gray}{rgb}{0.5,0.5,0.5}
\definecolor{mauve}{rgb}{0.58,0,0.82}
\newcommand{\myhighlight}[1]{\textcolor{C1!75}{#1}}

\usepackage{tikz,pgfplots}
\pgfplotsset{compat=1.14}
\usepackage{pgfplotstable}
\usetikzlibrary{shapes}
\usetikzlibrary{positioning}
\usetikzlibrary{plotmarks}
\usetikzlibrary{patterns}
\usetikzlibrary{intersections,shapes.arrows}
\usetikzlibrary{arrows.meta}
\usetikzlibrary{pgfplots.fillbetween}
\usepgfplotslibrary{groupplots}
\usetikzlibrary{fit}
\definecolor{darkpink}{rgb}{0.91, 0.33, 0.5}

\definecolor{C0}{HTML}{1F77B4}
\definecolor{C1}{HTML}{ff7f0e}
\definecolor{C2}{HTML}{2ca02c}
\definecolor{C3}{HTML}{d62728}
\definecolor{C4}{HTML}{9467bd}

\title{Correlated Noise Mechanisms for Differentially Private Learning}

\subtitle{A Tutorial on Mathematical Foundations and Practical Aspects}

\booltrue{authortwocolumn} 
\maintitleauthorlist{
Krishna Pillutla \\
IIT Madras
\and
Jalaj Upadhyay \\
Rutgers University
\and Christopher A. Choquette-Choo \\
Google DeepMind
\and 
Krishnamurthy Dvijotham \\ ServiceNow Research
\and 
Arun Ganesh \\ Google Research 
\and 
Monika Henzinger \\ IST, Austria
\and 
Jonathan Katz \\ Google 
\and
Ryan McKenna \\ Google Research 
\and
H. Brendan McMahan \\ Google Research 
\and
Keith Rush \\ Google DeepMind  
\and
Thomas Steinke \\ Google DeepMind  
\and
Abhradeep Thakurta \\ Google DeepMind  
}

\issuesetup
{%
 copyrightowner={TODO},
 volume        = xx,
 issue         = xx,
 pubyear       = 2025,
 isbn          = xxx-x-xxxxx-xxx-x,
 eisbn         = xxx-x-xxxxx-xxx-x,
 doi           = 10.1561/XXXXXXXXX,
 firstpage     = 1, 
 lastpage      = 18
 }

\usepackage{mwe}

\author[1]{Pillutla, Krishna}
\author[2]{Upadhyay, Jalaj}
\author[5]{Choquette-Choo, Christopher A.}
\author[3]{Dvijotham, Krishnamurthy}
\author[4]{Ganesh, Arun}
\author[6]{Henzinger, Monika}
\author[7]{Katz, Jonathan}
\author[4]{McKenna, Ryan}
\author[4]{McMahan, H. Brendan }
\author[5]{Rush, Keith}  
\author[5]{Steinke, Thomas} 
\author[5]{Thakurta, Abhradeep}

\affil[1]{Indian Institute of Technology, Madras}
\affil[2]{Rutgers University}
\affil[3]{ServiceNow Research}
\affil[4]{Google Research}
\affil[5]{Google DeepMind}
\affil[6]{Institute of Science and Technology Austria}
\affil[7]{Google}

\articledatabox{\nowfntstandardcitation}

\begin{document}

\makeabstracttitle

\begin{abstract}
\input{0_abstract}
\end{abstract}

\chapter*{Notation}

We summarize the main notation in \Cref{tab:notation,tab:notation:2}.

\begin{table}[hp]
\caption{Notation summary (part 1). Matrices and vectors are denoted in boldface.}\label{tab:notation}
\renewcommand{\arraystretch}{1.4}%
\begin{footnotesize}
\begin{tabular}{lp{3.2in}}
\toprule
\nd & Number of training steps. More generally, the number of steps on which private estimates are released.  \\
$\mdim$ & Model dimension. \\
$\workload \in \R^\ndnd$ &The \emph{workload matrix}. \\ 
$\prefix \in \R^\ndnd$ & The lower-triangular matrix of ones, defining an \emph{unweighted} prefix sum workload. \\
$\workload = \bfB \strategy$ & Matrix factorization of $\bfA$, where $\bfB$ is called the \emph{decoder matrix}, and $\strategy$ is the \emph{strategy} or \emph{encoder} matrix. \\
$\Cinv$ & The \emph{noise-correlating matrix}, the (pseudo)inverse of the strategy matrix $\strategy$. The matrix $\Cinv$ maps i.i.d. Gaussian noise to a non-i.i.d. distribution.\\
$\Gradients  \in \R^\ndmd$  &$= (\gradient_0, \ldots, \gradient_{n-1})$ stacked row-wise, the sequence of private inputs processed by the DP mechanism (typically $\bfg_t = \Gradients\idx{t}{:} \in \R^m$ is a gradient). \\
$\bfS \in \R^\ndmd$ & $=(\bfs_0, \ldots, \bfs_{n-1})$ stacked row-wise, the (weighted) prefix sums to be computed, $\bfS = \workload \Gradients$, equivalently $\bfs_t = \sum_{\tau=0}^t \workload\idx{t}{\tau} \bfg_\tau$. \\ 
$\dpGradients \in \R^\ndmd$  &$= (\dpgradient_0, \ldots, \dpgradient_{n-1})$ stacked row-wise, DP estimates of $\Gradients$. \\
$\Znoise \in \R^\ndmd$ & $=(\znoise_0, \dots, \znoise_{\nd-1})$ stacked row-wise, the source i.i.d. noise injected for DP; private estimates of $\bfG$ are given by $\dpGradients = \Gradients + \Cinv \Znoise$. \\
$\corrZnoise \in \R^\ndmd$ & $= \strategy^{-1}\Znoise$, the correlated noise injected into the learning algorithm. The noise injected in step $t$ is $\corrznoise = \corrZnoise[t, :]$. \\
$\hat\bfS \in \R^\ndmd$ & $=(\hat\bfs_0, \dots \hat\bfs_{\nd-1})$ stacked row-wise, DP estimates of $\bfS$, computed as $\widehat\bfS = \workload \dpGradients$. \\

\bottomrule
\end{tabular}
\end{footnotesize}
\end{table}

\begin{table}[hp]
\caption{Notation summary (part 2). Matrices and vectors are denoted in boldface.}\label{tab:notation:2}
\renewcommand{\arraystretch}{1.4}
\begin{footnotesize}
\begin{tabular}{lp{3.2in}}
\toprule
$[n]$ & $=\{0, \dots, n - 1\}$ for $n \ge 1$. Indexed quantities (matrices, vectors, sequences) are zero-indexed throughout.\\
$\N$ & $= \{0, 1, 2, \dots \}$, the natural numbers including zero. \\ 
$\bfM\tp$  & Transpose of a matrix $\bfM$. \\
$\bfM^\star$  & A matrix $\bfM$ that is ``optimal'' in a context-dependent sense. \\
$\bfM^\dagger$  & The Moore-Penrose pseudoinverse of matrix $\bfM$. \\
$\bfM\idx{i}{j}$ & The $(i, j)^{\text{th}}$ entry of matrix $\bfM$, zero-indexed.\\
$\bfM\idx{i}{:}$, $\bfM\idx{:}{j}$  & The $i^{\text{th}}$ row and $j^{\text{th}}$ column of $\bfM$, zero-indexed. The $i$th row of $\bfM$ is also often denoted $\bfm_i \coloneqq \bfM\idx{i}{:}$\\
$\lfrob{\bfM}$  & The Frobenius norm of a matrix $\bfM$. \\
$\rownorm{\bfM}$ & $ =\max_{t \in [n]} \norm{\bfM\idx{t}{:}}_2$, the maximum $L_2$ norm of the rows of $\bfM$. \\
$\colnorm{\bfM}$ & $ =\max_{t \in [n]} \norm{\bfM\idx{:}{t}}_2$, the maximum $L_2$ norm of the columns of $\bfM$. \\
$\ln$ & Natural logarithm, $\ln(2.718) \approx 1$. 
\\
$\ExP{Z \sim P}{f\br{Z}}$ & Expectation of the function $f$ under the distribution $P$. \\
$\Znoise \sim \normalnm{\mu}{\stddev^2}$ & $\Znoise$ is a random $n \times m$ matrix whose entries are i.i.d. $\normal(\mu, \stddev^2)$. \\
$\nbuf$ & Order of recurrence / degree / number of memory buffers. \\
\hline 
$\bfx \in \calX$ & ML training example.\\
$\ell: \Theta \times \calX \to \R$ & Loss function. \\ 
$\bftheta \in \Theta \subset \R^\mdim$ & Parameters of AI model to be learned. Typically,\newline $\bfg_t = \nabla \ell(\bftheta_t, \bfx)$. \\
$\bsz$ & The (mini)batch size for model training. \\
$\datasize$  & The total number of examples $\bfx$ in the training dataset $D$. \\

$\maxpart$ & (Maximum) number of times an example/user participates in training. \\
$\minsep$ & Minimum separation between the participations of an example/user. \\
\bottomrule
\end{tabular}
\end{footnotesize}
\end{table}

\chapter{Introduction: Private SGD and Prefix Sums with Correlated Noise}
\label{chap:intro}
\input{1_background.tex}

\chapter{Correlated Noise Mechanisms for Streaming Prefix Sums}
\label{chap:prefixsum}
\input{2a_prefix_sum_new}

\chapter{Correlated Noise Mechanisms for Learning Problems}\label{chap:ml}
\input{3_ml.tex}

\chapter{Implementation Details and Practical Recommendations}\label{chap:practical}

\input{4_practical}

\chapter{Challenges and Open Questions}\label{chap:open}

\input{5_open_new}

\bibliographystyle{plainnat}
\bibliography{ref}

\end{document}

%% file: macros.tex
\newcommand{\op}[1]{\operatorname{#1}}

\makeatletter
\def\renewtheorem#1{%
  \expandafter\let\csname#1\endcsname\relax
  \expandafter\let\csname c@#1\endcsname\relax
  \gdef\renewtheorem@envname{#1}
  \renewtheorem@secpar
}
\def\renewtheorem@secpar{\@ifnextchar[{\renewtheorem@numberedlike}{\renewtheorem@nonumberedlike}}
\def\renewtheorem@numberedlike[#1]#2{\newtheorem{\renewtheorem@envname}[#1]{#2}}
\def\renewtheorem@nonumberedlike#1{  
\def\renewtheorem@caption{#1}
\edef\renewtheorem@nowithin{\noexpand\newtheorem{\renewtheorem@envname}{\renewtheorem@caption}}
\renewtheorem@thirdpar
}
\def\renewtheorem@thirdpar{\@ifnextchar[{\renewtheorem@within}{\renewtheorem@nowithin}}
\def\renewtheorem@within[#1]{\renewtheorem@nowithin[#1]}
\makeatother

\definecolor{darkgreen}{rgb}{0,0.4,0.0}
\definecolor{C0}{HTML}{1F77B4}
\definecolor{C1}{HTML}{ff7f0e}
\definecolor{C2}{HTML}{2ca02c}
\definecolor{C3}{HTML}{d62728}
\definecolor{C4}{HTML}{9467bd}
\definecolor{C5}{HTML}{8c564b}
\definecolor{keynoteBlue}{HTML}{01A2FF}
\definecolor{ballblue}{rgb}{0.13, 0.67, 0.8}
\newcommand{\bnote}[1]{[{\color{C3}{\textbf{Brendan:} \emph{#1}}}]}

\newcommand{\jucomment}[1]{[{\color{ballblue}{\textbf{Jalaj:} \emph{#1}}}]}

\newcommand{\kpcomment}[1]{{\color{C1} \textbf{Krishna}: {#1}}}
\newcommand{\rycomment}[1]{{\color{purple} \textbf{Ryan}: {#1}}}

\newtheorem{question}{Question}[chapter]

\newcommand{\nd}{\ensuremath{n}} 
\newcommand{\mdim}{m}  
\newcommand{\ndnd}{{\nd\!\times\!\nd}}
\newcommand{\ndmd}{{\nd\!\times\!\mdim}}
\newcommand{\nbuf}{d}  

\newcommand{\datasize}{N}  
\newcommand{\bsz}{B}  

\newcommand{\txtloss}{loss\xspace}
\newcommand{\txtmaxloss}{max loss\xspace}
\newcommand{\txtrmsloss}{RMS-loss\xspace}

\newcommand{\txtmaxlossCap}{Max Loss\xspace}

\newcommand{\maxloss}{\mathcal{L}_\infty}
\newcommand{\maxlossn}{\bar{\mathcal{L}}_\infty}
\newcommand{\rmsloss}{\mathcal{L}_2}
\newcommand{\rmslossn}{\bar{\mathcal{L}_2}}

\newcommand{\workload}{\bfA}
\newcommand{\prefix}{\workload_{\mathsf{pre}}}
\newcommand{\Amom}{\workload_{\mathsf{mom}}}
\newcommand{\strategy}{\bfC}
\newcommand{\Cinv}{\inv{\bfC}}
\newcommand{\Znoise}{\bfZ}
\newcommand{\znoise}{\bfz}
\newcommand{\corrznoise}{\widetilde\bfz}
\newcommand{\corrZnoise}{\widetilde \bfZ}
\newcommand{\gradient}{\bfg}
\newcommand{\Gradients}{\bfG}
\newcommand{\dpgradient}{\widehat\bfg}
\newcommand{\dpGradients}{\widehat\bfG}
\newcommand{\Vector}{\mathsf{vec}}
\newcommand{\nmult}{\sigma}
\newcommand{\stddev}{\nu}
\newcommand{\clip}{\mathsf{clip}}
\newcommand{\clipnorm}{\zeta}

\newcommand{\normal}{\mathcal{N}}
\newcommand{\normaldim}[3]{\normal_{#1}\big(#2, #3\big)}
\newcommand{\normalm}[2]{\normaldim{m}{#1}{#2}}
\newcommand{\normalnm}[2]{\normaldim{\nd\!\times\!\mdim}{#1}{#2}}

\newcommand{\mech}{\mathcal{M}}

\newcommand{\minseppart}{Min-Sep participation\xspace}

\newcommand{\maxpart}{k}
\newcommand{\minsep}{b}
\newcommand{\bands}{b}

\newcommand{\BLT}{\mathsf{BLT}}
\newcommand{\bltlower}{\mathsf{blt}}

\newcommand{\idx}[2]{[{#1, #2}]}

\newcommand{\inv}[1]{{{#1}^{-1}}}

\newcommand{\restart}{\bfA}
\newcommand{\opt}{\mathsf{Opt}}
\newcommand{\optToep}{\mathsf{Opt}_\mathsf{Toep}}
\newcommand{\optNormalized}{\mathsf{Opt}_\mathsf{norm}}

\newcommand{\sparen}[1]{\left[ #1 \right]}
\newcommand{\distribution}{\tau}

\newcommand{\ones}{{\bm{1}}}
\newcommand{\zeros}{{\bm{0}}}

\newcommand{\model}{{\bftheta}}

\DeclareMathOperator*{\argmin}{arg\,min}
\newcommand{\poly}{\mathrm{poly}}
\newcommand{\sens}{\ensuremath{\mathsf{sens}}}
\newcommand{\cyclic}{\ensuremath{\mathsf{cyclic}}}
\newcommand{\MinSep}{\ensuremath{\mathsf{minSep}}}

\newcommand{\calA}{\ensuremath{\mathcal{A}}}
\newcommand{\calB}{\ensuremath{\mathcal{B}}}

\newcommand{\calE}{\ensuremath{\mathcal{E}}}

\newcommand{\calG}{\ensuremath{\mathcal{G}}}

\newcommand{\calN}{\ensuremath{\mathcal{N}}}

\newcommand{\calX}{\ensuremath{\mathcal{X}}}
\newcommand{\calY}{\ensuremath{\mathcal{Y}}}

\newcommand{\bfA}{{\bm{A}}}
\newcommand{\bfB}{{\bm{B}}}
\newcommand{\bfC}{{\bm{C}}}

\newcommand{\bfF}{{\bm{F}}}
\newcommand{\bfG}{{\bm{G}}}
\newcommand{\bfH}{{\bm{H}}}
\newcommand{\bfI}{{\bm{I}}}

\newcommand{\bfL}{{{\bm{L}}}}
\newcommand{\bfM}{{\bm{M}}}

\newcommand{\bfP}{{\bm{P}}}
\newcommand{\bfQ}{{\bm{Q}}}

\newcommand{\bfS}{{\bm{S}}}

\newcommand{\bfU}{{\bm{U}}}
\newcommand{\bfV}{{\bm{V}}}
\newcommand{\bfW}{{\bm{W}}}
\newcommand{\bfX}{{\bm{X}}}

\newcommand{\bfZ}{{\bm{Z}}}

\newcommand{\bfa}{{\bm{a}}}
\newcommand{\bfb}{{\bm{b}}}
\newcommand{\bfc}{{\bm{c}}}

\newcommand{\bff}{{\bm{f}}}
\newcommand{\bfg}{{\bm{g}}}

\newcommand{\bfm}{{\bm{m}}}

\newcommand{\bfp}{{\bm{p}}}
\newcommand{\bfq}{{\bm{q}}}
\newcommand{\bfr}{{\bm{r}}}
\newcommand{\bfs}{{\bm{s}}}

\newcommand{\bfu}{{\bm{u}}}
\newcommand{\bfv}{{\bm{v}}}

\newcommand{\bfx}{{\bm{x}}}
\newcommand{\bfy}{{\bm{y}}}
\newcommand{\bfz}{{\bm{z}}}

\newcommand{\bftheta}{{\ensuremath{\bm{\theta}}}}

\newcommand{\bfdelta}{\ensuremath{\bm{\delta}}}
\newcommand{\bfalpha}{\ensuremath{\bm{\alpha}}}

\newcommand{\bfLambda}{\ensuremath{\bm{\varLambda}}}
\newcommand{\bflambda}{\ensuremath{\bm{\lambda}}}
\newcommand{\bfpi}{\ensuremath{\bm{\pi}}}
\newcommand{\bfphi}{\ensuremath{\bm{\phi}}}
\newcommand{\bfxi}{\ensuremath{\bm{\xi}}}

\newcommand{\br}[1]{\left({#1}\right)}

\DeclareMathOperator*{\Exp}{\mathbb{E}}
\newcommand{\ExP}[2]{\Exp_{#1}\left[{#2}\right]}
\newcommand{\Pdata}{\mathbb{P}_{\mathsf{data}}}
\newcommand{\norm}[1]{\left\|{#1}\right\|}
\newcommand{\tp}{^\top} 

\newcommand{\ceiling}[1]{\left\lceil #1 \right\rceil}
\newcommand{\tr}{\ensuremath{\mathsf{Tr}}}
\newcommand{\diag}{\ensuremath{\mathsf{diag}}}

\newcommand{\ltwo}[1]{\left\|#1\right\|_2}
\newcommand{\lfrob}[1]{\left\|#1\right\|_{\op{F}}}

\newcommand{\rownormop}[1]{\left\| #1 \right\|_{\op{2 \to \infty}}}
\newcommand{\colnormop}[1]{\left\| #1 \right\|_{\op{1 \to 2}}}
\newcommand{\rownorm}[1]{\left\| #1 \right\|_{\mathsf{row}}}
\newcommand{\colnorm}[1]{\left\| #1 \right\|_{\mathsf{col}}}

\newcommand{\conv}{\mathsf{conv}}
\newcommand{\trace}[1]{\mathsf{Tr}(#1)}

\newcommand{\T}{^{\top}}  
\newcommand{\N}{\mathbb{N}}
\newcommand{\R}{\mathbb{R}}
\newcommand{\C}{\mathbb{C}}
\newcommand{\D}{\,\mathrm{d}}
\newcommand{\real}{\R}
\renewcommand{\epsilon}{\varepsilon}
\newcommand{\eps}{\varepsilon}
\DeclarePairedDelimiterX{\inp}[2]{\langle}{\rangle}{#1, #2}  
\DeclarePairedDelimiter{\ceil}{\lceil}{\rceil}

\newcommand{\thseq}{(\bftheta_t)_{t=1}^{\nd-1} }

\newcommand{\hFa}{\widehat \bfF_a}
\newcommand{\hbff}{\widehat \bff}

\DeclareMathOperator{\dist}{dist}
\DeclareMathOperator{\cost}{cost}
\newcommand{\lnor}{\left\|}
\newcommand{\rnor}{\right\|}
\DeclareMathOperator{\expectation}{\mathbb{E}}

\newcommand{\Blue}[1]{\textcolor{blue}{#1}}
\newcommand{\Red}[1]{\textcolor{red}{#1}}

\makeatletter
\let\c@proposition\c@theorem  
\makeatother
\makeatletter
\let\c@remark\c@theorem  
\makeatother
\makeatletter
\let\c@definition\c@theorem  
\makeatother
\makeatletter
\let\c@example\c@theorem  
\makeatother
\newtheorem{problem}[theorem]{Problem}
\newtheorem{fact}[theorem]{Fact}
\newtheorem{property}[theorem]{Property}
\newtheorem{conjecture}[theorem]{Conjecture}

\crefname{figure}{Fig.}{Figs.}
\Crefname{figure}{Fig.}{Figs.}
\crefname{lemma}{Lemma}{Lemmas}
\Crefname{lemma}{Lemma}{Lemmas}
\crefname{proposition}{Proposition}{Propositions}
\Crefname{proposition}{Proposition}{Propositions}
\crefname{definition}{Definition}{Definitions}
\Crefname{definition}{Definition}{Definitions}
\crefname{remark}{Remark}{Remarks}
\Crefname{remark}{Remark}{Remarks}
\crefname{example}{Example}{Examples}
\Crefname{example}{Example}{Examples}
\crefname{theorem}{Theorem}{Theorem}
\Crefname{theorem}{Theorem}{Theorem}
\crefname{equation}{Eq.}{Eqs.}
\creflabelformat{equation}{(#2#1#3)}
\Crefname{equation}{Eq.}{Eqs.}
\Crefname{chapter}{Section}{Sections}  

\usepackage{mdframed}
\definecolor{defcolor}{RGB}{234, 255, 225}  %
\mdfdefinestyle{defsty}{
    backgroundcolor=white,
    rightline=false,
    bottomline=false,
    topline=false,
    linewidth=1pt,
    }
\definecolor{thmcolor}{rgb}{0.94, 0.97, 1.0}  %
\mdfdefinestyle{thmsty}{
    backgroundcolor=white,
    rightline=false,
    bottomline=false,
    topline=false,
    linewidth=1pt,
    }
\newenvironment{btheorem}[1][\unskip]{%
  \ifx#1\unskip
    \begin{theorem} 
  \else
    \begin{theorem}[#1] 
  \fi
  \begin{mdframed}[style=thmsty]}
  {\end{mdframed}\end{theorem}} 
\crefname{btheorem}{Theorem}{Theorems}
\Crefname{btheorem}{Theorem}{Theorems}
\newenvironment{blemma}[1][\unskip]{%
  \ifx#1\unskip
    \begin{lemma} 
  \else
    \begin{lemma}[#1] 
  \fi
  \begin{mdframed}[style=thmsty]}
  {\end{mdframed}\end{lemma}} 
\crefname{blemma}{Lemma}{Lemmas}
\Crefname{blemma}{Lemma}{Lemmas}
\newenvironment{bcorollary}[1][\unskip]{%
  \ifx#1\unskip
    \begin{corollary} 
  \else
    \begin{corollary}[#1] 
  \fi
  \begin{mdframed}[style=thmsty]}
  {\end{mdframed}\end{corollary}} 
\crefname{bcorollary}{corollary}{corollaries}
\Crefname{bcorollary}{Corollary}{Corollaries}

\newenvironment{bproposition}[1][\unskip]{%
  \ifx#1\unskip
    \begin{proposition} 
  \else
    \begin{proposition}[#1] 
  \fi
  \begin{mdframed}[style=thmsty]}
  {\end{mdframed}\end{proposition}} 
\crefname{bproposition}{Proposition}{Propositions}
\Crefname{bproposition}{Proposition}{Propositions}

\newenvironment{bproblem}[1][\unskip]{%
  \ifx#1\unskip
    \begin{problem} 
  \else
    \begin{problem}[#1] 
  \fi
  \begin{mdframed}[style=thmsty]}
  {\end{mdframed}\end{problem}} 
\crefname{bproblem}{Proposition}{Propositions}
\Crefname{bproblem}{Proposition}{Propositions}

  {\end{mdframed}\end{fact}} 
\crefname{bfact}{Proposition}{Propositions}
\Crefname{bfact}{Proposition}{Propositions}

  {\end{mdframed}\end{property}} 
\crefname{bproperty}{Proposition}{Propositions}
\Crefname{bproperty}{Proposition}{Propositions}

\newenvironment{bconjecture}[1][\unskip]{%
  \ifx#1\unskip
    \begin{conjecture} 
  \else
    \begin{conjecture}[#1] 
  \fi
  \begin{mdframed}[style=thmsty]}
  {\end{mdframed}\end{conjecture}} 
\crefname{bconjecture}{Conjecture}{Conjectures}
\Crefname{bconjecture}{Conjecture}{Conjectures}

\newenvironment{bdefinition}[1][\unskip]{%
  \ifx#1\unskip
    \begin{definition} 
  \else
    \begin{definition}[#1] 
  \fi
  \begin{mdframed}[style=defsty]}
  {\end{mdframed}\end{definition}} 
\crefname{bdefinition}{Definition}{Definitions}
\Crefname{bdefinition}{Definition}{Definitions}

\definecolor{exmcolor}{RGB}{255, 238, 244} %
\mdfdefinestyle{exmsty}{
    backgroundcolor=white,
    rightline=false,
    bottomline=false,
    topline=false,
    linewidth=1pt,
    }
\newenvironment{bexample}[1][\unskip]{%
  \ifx#1\unskip
    \begin{example} 
  \else
    \begin{example}[#1] 
  \fi
  \begin{mdframed}[style=exmsty]}
  {\end{mdframed}\end{example}} 
\crefname{bexample}{Example}{Examples}
\Crefname{bexample}{Example}{Examples}

\definecolor{remcolor}{RGB}{255,254,224}  %
\mdfdefinestyle{remsty}{
    backgroundcolor=white,
    rightline=false,
    bottomline=false,
    topline=false,
    linewidth=1pt,
    }
\newenvironment{bremark}[1][\unskip]{%
  \ifx#1\unskip
    \begin{remark} 
  \else
    \begin{remark}[#1] 
  \fi
  \begin{mdframed}[style=remsty]}
  {\end{mdframed}\end{remark}} 
\crefname{bremark}{Remark}{Remarks}
\Crefname{bremark}{Remark}{Remarks}

\newenvironment{bquestion}[1][\unskip]{%
  \ifx#1\unskip
    \begin{question} 
  \else
    \begin{question}[#1] 
  \fi
  \begin{mdframed}[style=remsty]}
  {\end{mdframed}\end{question}} 
\crefname{bquestion}{Question}{Questions}
\Crefname{bquestion}{Question}{Question}

\mdfdefinestyle{recsty}{
    backgroundcolor=C0!10,
    linewidth=1pt,
    }

\newenvironment{recommendation}[1][]{%
    \begin{mdframed}[style=recsty,frametitle={Recommendation\ifstrempty{#1}{}{: #1}}]%
}{%
    \end{mdframed}%
}


%% file: 0_abstract.tex
This monograph explores the design and analysis of correlated noise mechanisms for differential privacy (DP), focusing on their application to private training of AI and machine learning models via the core primitive of estimation of weighted prefix sums. While typical DP mechanisms inject independent noise into each step of a stochastic gradient (SGD) learning algorithm in order to protect the privacy of the training data, a growing body of recent research demonstrates that introducing (anti-)correlations in the noise can significantly improve privacy-utility trade-offs by carefully canceling out some of the noise added on earlier steps in subsequent steps. Such correlated noise mechanisms, known variously as matrix mechanisms, factorization mechanisms, and DP-Follow-the-Regularized-Leader (DP-FTRL) when applied to learning algorithms,
have also been influential in practice, with industrial deployment at a global scale. 

Unfortunately, the rapid development of this field and complex mathematical foundations pose a high barrier to entry for researchers and practitioners. This monograph provides a pedagogical tutorial of correlated noise mechanisms, with an emphasis on the the theoretical principles governing their design and derivation, and practical considerations such as scalability and runtime.

We start with private prefix sum estimation in the streaming setting, where each example is only processed on a single training step. We focus on mechanisms with structural constraints designed to balance high utility with low time and space complexity.
Next, we allow each example to participate on multiple training steps, to accommodate practical private AI model training. While we focus on this example-level notion of privacy, we also discuss the straightforward generalization to uer-level privacy, where the DP guarantee extends to possibly multiple training examples all associated with the same user, as in practice user-level privacy is usually necessary.
We also discuss how to numerically find the precise correlations (that is, the noise cancellation schedule) and offer implementation details and guidance for practitioners.
Finally, we survey promising theoretical and applied open problems for researchers to contribute to this active and growing area.


%% file: 1_background.tex
Privacy-sensitive data is increasingly being collected and stored by both private entities and governments. Such data includes health records, business records, personal communications (including chats and emails), individuals' location history (enabled by smartphone GPS), internet content consumption records, and social media interactions. General trends and insights gleaned from such data hold the promise of great benefit: real-time estimates of traffic congestion, better user experiences across apps and websites, public health insights, and more.  With the rise of powerful AI models, including large language models (LLMs), machine learning is increasingly becoming the principal technique for unlocking the value latent in private datasets, but this approach also entails real privacy risks: it is imperative to ensure that models do not leak sensitive information contained in their training data.

Recent work has shown that this is not just a theoretical risk.
{There is a large body of work showing that exact copies of (sometimes sensitive) training data can be extracted from production language models, even with just black-box API access to these models. }
Analyzing the privacy of sophisticated data processing mechanisms is challenging, as privacy leakage can be subtle and hard to detect.

Differential privacy (DP) is the gold standard for analyzing privacy-sensitive data with rigorous and mathematically justified privacy guarantees. It has been adapted to AI and machine learning model training applications in recent years and has garnered significant interest in the research community. Models trained with rigorous DP guarantees also have industrial deployments at a global scale by companies such as Google and Apple.

The workhorse of private machine learning is a differentially private version of stochastic gradient descent, known as DP-SGD. This algorithm adds isotropic Gaussian noise to the updates of a gradient descent-based learning algorithm. While most work on DP-SGD injects independent Gaussian noise in each step of the learning algorithm, recent work has shown that it can be beneficial to use noise that has non-zero correlations across {training steps.}  In particular, correlated noise mechanisms power ``the first production neural network trained directly on user data announced with a formal DP guarantee.''\footnote{\url{https://research.google/blog/federated-learning-with-formal-differential-privacy-guarantees/}}

This monograph serves as a pedagogical introduction to such correlated noise mechanisms, with a focus on theoretical principles behind their design and development and practical aspects such as scalability and run time.
Due to the rapid development of this field, a fragmented prior literature, and complex mathematical foundations, researchers and practitioners face high barriers to entry into this topic. To bridge this gap, we present this tutorial aimed at a broad audience, ranging from early graduate students with basic machine learning knowledge to experts seeking a consolidated reference.

\paragraph{Outline}
We start this section with a brief introduction to differential privacy in \Cref{sec:introToDP} for readers not already familiar with this notion.
Next, we describe the problem of private machine learning and the DP-SGD algorithm in \Cref{sec:weightedPrefixSum}. We also describe its connection to the problem of private weighted prefix sum estimation; this forms the basis for the correlated noise mechanisms introduced in \Cref{sec:correlatedNoiseIntro} and later sections. Next, we develop some intuition as to why correlated noise mechanisms can yield better privacy-utility tradeoffs in \Cref{sec:whyCorrelatedNoise}. We then discuss what considerations have to be made in designing correlated noise mechanisms in \Cref{sec:designSpaceCorrelatedNoise}, ending the section with technical aspects of DP guarantees in \Cref{sec:ch1-priv-unit,sec:intro:different-adj}.

Sections and remarks marked with an asterisk ($*$) go into technical details or other advanced material---they can safely be skipped on a first reading.

\section{Introduction to Differential Privacy} \label{sec:introToDP}

\textit{Differential privacy} is a notion of privacy that is defined on a collection of individual data records. In the machine learning context, a \emph{data record} typically refers to an individual training example.\footnote{
    Other notions of what constitutes a data record are also sometimes used. For instance, {\em user-level} differential privacy defines a data record as all training examples containing data from one individual.
} 
For the purposes of this monograph, we will make this assumption and deal with datasets $D$ that are a collection of individual data records $x_1, \ldots, x_\nd$. We will assume that each $x_i$ comes from a universe $\mathcal{X}$ and so $D \in \mathcal{X}^\star$ is an ordered sequence 
of elements of $\mathcal{X}$. 

Differential privacy is defined for \emph{mechanisms} that map input datasets to possible outcomes. In machine learning, the outcomes are typically the weights of a trained AI model. Formally, a mechanism $\mech: \mathcal{X}^* \rightarrow \mathcal{Y}$ is a randomised function that takes as input a dataset $D \in \mathcal{X}^*$ and returns an output $\mech(D) \in \mathcal{Y}$.

Differential privacy (DP) is a formal condition that ensures that the output $\mech(D)$ of the mechanism does not leak ``too much'' information about any one ``data unit'' in the input dataset~$D$. A data unit can refer to an example, or a set of examples corresponding to a \emph{user}.

At a high level, DP is a stability condition on the mechanism $\mech$, i.e., the output $\mech(D')$ on a dataset $D'$ obtained by changing $D$ slightly should be \textit{nearly indistinguishable} from~$\mech(D)$. 

To make these ideas concrete, we use a notion of indistinguishability, and hence of DP, called \textit{approximate differential privacy} and \textit{Gaussian DP}.
As a default, we take the adjacency relation (i.e., what ``changing $D$ slightly'' means) to be the so-called ``replace-one'' relation at the example level.\footnote{Other notions of adjacency and units of privacy are also of interest. For the most part, the key ideas can be applied directly to those other notions. To keep this monologue concentrated on providing the fundamentals, we defer this discussion to \Cref{sec:intro:different-adj}.}

\begin{bdefinition}[Replace-One Example Adjacency]
\label{def:replaceone}
We say that two datasets $D, D' \in \calX^*$ are \textbf{adjacent} (denoted by $D \simeq D'$)  in the replace-one notion at the example level if $|D \setminus D'| = |D' \setminus D| = 1$, i.e., $D'$ is obtained from $D$ by replacing one element with another. 
\end{bdefinition}

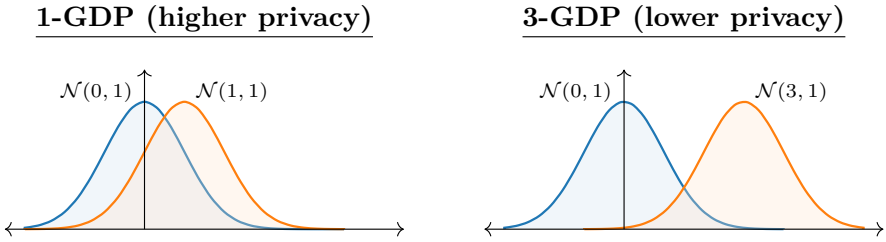
\begin{figure}[t]
    \centering
    \input{gdp}
    \caption{
    The notion of $\mu$-GDP means that distinguishing between the outputs $\mech(D)$ and $\mech(D')$ of the mechanism for two adjacent datasets $D$ and $D'$ is as hard as distinguishing between $\calN(0, 1)$ and $\calN(\mu, 1)$.
    Smaller $\mu$, as in the left subplot, means that an adversary will be less successful in distinguishing between the two distributions, corresponding to a higher level of privacy.
    }
    \label{fig:gdp}
\end{figure}

Equipped with this notion of adjacency between datasets, we can now formally define one of the most widely used formulations of differential privacy.

\begin{bdefinition}[$(\epsilon,\delta)$-Differential Privacy]\label{def.dp}
    Let $\mech:\mathcal{D}\rightarrow \mathcal{R}$ be a randomized algorithm, where $\mathcal{R}$ is the output domain. For fixed $\epsilon > 0$ and $\delta \in [0,1)$, we say that $\mech$ satisfies $(\epsilon,\delta)$-differential privacy if for all measurable sets $S \subset \mathcal{R}$ and for all pairs of adjacent datasets $D, D' \in \mathcal{D}$, it holds that
    \[
        \mathsf{Pr}[\mech(D) \in S] \leq e^\epsilon \cdot \mathsf{Pr}[\mech(D') \in S] + \delta.
    \]
\end{bdefinition}

A number of mechanisms are known to satisfy $(\epsilon,\delta)$-differential privacy, one of the most common being the \emph{Gaussian mechanism} (see Definition~\ref{def:gaussian-mechanism}). When analyzing this mechanism, it is often more intuitive to use an equivalent but analytically convenient alternative: \emph{Gaussian differential privacy} ($\mu$-GDP)\footnote{Throughout this monograph, we use the most convenient formalism depending on context. For instance, $\mu$-GDP is used in \Cref{chap:intro} and \Cref{chap:prefixsum}, while $(\epsilon,\delta)$-DP is used in \Cref{chap:ml}. Most of our results do not depend on the specific notion of differential privacy used.}.

\paragraph{Gaussian differential privacy}
Informally, a mechanism $\mech$ is {\em $\mu$-Gaussian differentially private} ($\mu$-GDP) if for any pair of adjacent datasets $D \simeq D'$, distinguishing between the distribution of $\mech(D)$ and $\mech(D')$ is \emph{no easier than} distinguishing between the Gaussian distributions $\calN(0, 1)$ and $\calN(\mu, 1)$ based on a single sample from each distribution. Thus, $\mu$-GDP is a stronger privacy guarantee for smaller $\mu$, as illustrated in \Cref{fig:gdp}. Note that $\mu = 0$ gives perfect indistinguishability (i.e., perfect privacy), $\mu = 1$ gives reasonably good indistinguishability, and $\mu = \infty$ gives no indistinguishability (i.e., no privacy). 
This intuition is captured formally by the following definition: 

\begin{bdefinition}[Gaussian Differential Privacy]\label{def:gdp}
A mechanism $\mech$ is {\em $\mu$-Gaussian differentially private} ($\mu$-GDP) if, for any adjacent datasets $D \simeq D'$, there is a (possibly randomized) function $g: \R \to \mathcal{Y}$ with:\footnotemark
\begin{align*}
g(Z)  \stackrel{\mathrm{d}}{=} \mech(D) \quad&\text{for} \quad Z \sim \calN(0, 1), \quad\text{and} \\
g(Z') \stackrel{\mathrm{d}}{=} \mech(D') \quad&\text{for} \quad Z' \sim \calN(\mu, 1),
\end{align*}
where $X \stackrel{\mathrm{d}}{=} Y$ means random variables $X$ and $Y$ are identically distributed.
\end{bdefinition}

\footnotetext{$\calN(\mu,\stddev^2)$ denotes the univariate Gaussian distribution with mean $\mu \in \R$ and variance $\stddev^2 > 0$. Its probability density function is $p(z)=\frac{1} {\sqrt{2\pi \stddev^2}} e^{-(z-\mu)^2/(2\stddev^2)}$.}

\paragraph{Gaussian Mechanism}
As mentioned above, the Gaussian mechanism is a canonical GDP mechanism: it creates a differentially private version of any deterministic function~$f$ by adding zero mean Gaussian noise (with the appropriate variance) to the output of~$f$.
This noise is scaled according to the \emph{sensitivity} of~$f$, which is a measure of how much the output of~$f$ is affected by altering a single record in the input dataset:

\begin{bdefinition} \label{def:l2-sens}
The \textbf{$\ell_2$-sensitivity} of
a vector-valued function $f:\calX^* \to \R^m$ is defined as $\sens(f) := \max_{D \simeq D'} \ltwo{f(D) - f(D')}$. 
\end{bdefinition}

The $\ell_2$-sensitivity determines how much noise we need to add to $f(D)$ to ensure  privacy. This is made formal by the Gaussian mechanism. 

\begin{bdefinition}\label{def:gaussian-mechanism}
Given a function $f: \calX^* \to \R^\mdim$ and noise multiplier~$\nmult$, set $\stddev = \nmult \cdot \sens(f)$. 
Then, the \textbf{Gaussian mechanism} on input $D$ outputs $f(D) + \znoise$, where $\znoise \sim \normalm{0}{\stddev^2}$ is a random vector in $\R^\mdim$ with i.i.d. entries drawn from $\normal\big(0, \stddev^2 \big)$.
\end{bdefinition}

This recipe remains unchanged if the function $f$ outputs a matrix or a higher order tensor: we conceptually flatten it into a vector, compute its sensitivity, add entry-wise i.i.d. Gaussian noise, and reshape this vector into its original shape.
To determine the privacy guarantees of the Gaussian mechanism, we only need to know the {\em noise multiplier} $\nmult$.\label{noisemultiplier}

\begin{blemma}\label[lemma]{lem:gdp-of-gaussian-mechanism}
For any function $f: \calX^* \to \R^m$, the Gaussian mechanism with noise multiplier $\nmult$ is ${1/\nmult}$-GDP.
\end{blemma}
\begin{proof}[Proof Sketch]
In the proof sketch, we only consider a scalar-valued function, $f$, to highlight the main idea behind the proof. The same proof idea extends to $\mdim > 1$ dimensions by utilizing the rotational invariance of the multivariate standard Gaussian distribution. We cover the full proof in the starred \Cref{sec:chap1-proof}.

Consider the special case of a scalar-valued function $f$ with $\mdim=1$.
Consider a pair of worst-case adjacent datasets $D \simeq D'$ such that $f(D') - f(D) = \sens(f)$. If $f(D)$ and $f(D')$ are closer, it only makes the mechanism outputs $\mech(D), \mech(D')$ more indistinguishable. We exhibit a function $g: \R \to \R$ such that $\mech(D) \stackrel{\mathrm{d}}{=} g(Z)$ and $\mech(D') \stackrel{\mathrm{d}}{=} g(Z + \mu)$ for $Z \sim \normal(0, 1)$, as required by \Cref{def:gdp}:
\[
    g(s) := f(D) \,+\, \nmult \cdot \sens(f) \cdot s  \,.
\]
Plugging in $s = Z \sim \calN(0, 1)$, we can verify that $g(Z) \stackrel{\mathrm{d}}{=} \mech(D)$.
Instead, with $s = Z + \mu$ for $\mu = 1/\nmult$, we get
\begin{align*}
    g(Z + \mu) &= f(D) + \nmult  \cdot \sens(f) \, \left(Z + \mu \right) \\
    &= f(D') + \nmult  \cdot \sens(f) \cdot Z \stackrel{\mathrm{d}}{=} \mech(D') \,,
\end{align*}
since $\mu = 1/\nmult$ and $\sens(f) = f(D') - f(D)$.
\end{proof}

\paragraph{Post-processing} 
One of the important properties of differential privacy is that it is preserved under arbitrary post-processing as long as the post-processing function does not use any part of the confidential data: 
\begin{blemma}
\label{lem:post-processing-lemma}
Let $\mech$ be a randomized
algorithm that is $\mu$-GDP. Let $g$ be an
arbitrary randomized mapping. Then $g \circ \mech(D) := g\big(\mech(D)\big)$ is also $\mu$-GDP.
\end{blemma}
Post-processing is a fundamental property of DP. It can be used to improve the utility or applicability of differentially private algorithms, such as reducing noise or enforcing domain constraints, without affecting the privacy guarantees.
 
\paragraph{Components of a DP Guarantee}
In real-world applications, we should specify three orthogonal components when providing a DP guarantee:\footnote{
In practical AI model training, a complete report of the DP guarantee should include additional details of the data access assumed and how the DP guarantee was computed.}
\begin{itemize}
    \item \textbf{The privacy unit} defines what entity's privacy is being protected by fixing the semantics of a one-unit change between $D$ and $D'$. For example, we could protect the privacy of individual examples in the dataset, or allow a one-unit change to modify all examples derived from a single user or entity.  \label{item:privacyUnit} 
    \item \textbf{The adjacency relation} formalizes what it means to obtain an adjacent dataset, in the context of a particular unit of privacy. We could replace one unit (as in \Cref{def:replaceone}), allow the addition or removal of one unit (\emph{add-or-remove} DP), or zero-out the contributions of a unit (Section~\ref{sec:ch1-zero-out}). 
    \item \textbf{The indistinguishability notion} is the mathematical formulation used to quantify indistinguishability. Common choices include Gaussian DP, $(\eps, \delta)$-DP, and zero-concentrated DP.
    Tight conversions exist between different indistinguishability notions (for a given privacy unit and adjacency; see the appendix for a brief review).
\end{itemize}

The fundamental principles discussed in this monograph are broadly applicable across various choices for each of these three components.  Unless explicitly stated otherwise, we default to examples as the privacy unit, replace-one as the adjacency unit, and Gaussian DP as the notion of indistinguishability.

\section{Private Learning via Weighted Prefix Sums}
\label{sec:weightedPrefixSum}

In this monograph, our key primitive will be the DP estimation of weighted prefix sums of a sequence of input vectors. In AI and machine learning, this arises most commonly in the privatization of stochastic gradient descent, where each vector is a single SGD model update.

\paragraph{Stochastic Gradient Descent (SGD)}
 Suppose we wish to find a model $\bftheta \in \Theta$ from a \textit{parameter space} $\Theta \subset \R^m$ that minimizes the objective, more commonly known as {\em stochastic optimization}, defined as 
\begin{align}
        \min_{\bftheta \in \Theta} \,\, \expectation_{\bfx \sim \Pdata}[\ell\br{\bftheta, \bfx}].
    \label{eq:poprisk}
\end{align}

Here, $\ell(\bftheta, \bfx)$ is the {\em loss} of making a prediction with model parameters~$\bftheta$ on a datapoint $\bfx$, which is in turn drawn i.i.d. from a data distribution $\Pdata$. 
In practice, we will have access only to a finite sample of datapoints $\bfx_0, \cdots, \bfx_{\nd-1}$ sampled i.i.d. from $\Pdata$, leading to an \emph{empirical risk minimization} problem:
\begin{equation}\label{eq:empiricalrisk}
\min_{\bftheta \in \Theta} \frac{1}{\nd}\sum_{i \in [\nd]} \ell(\bftheta,\bfx_i)\,,
\end{equation}
where we use shorthand $[\nd] \coloneqq \{0, \ldots, n-1\}$.

The standard workhorse algorithm used for solving the empirical risk minimization problem is  {\em stochastic gradient descent} (SGD). It is presented in \Cref{alg:sgd-1}.

\begin{figure}[t]
    \centering
    \begin{minipage}{.45\textwidth}
    \vspace{0pt}
    \centering
    \begin{algorithm}[H]
    \caption{\small \Blue{SGD}}
    \label{alg:sgd-1}
    \small
    \textbf{Inputs:} Dataset $D$, number of steps $n$,  learning rate $\eta$

    \begin{algorithmic}[1]
    \State Pick an initial model $\bftheta_0 \in \Theta$
    \For{$t = 0,\ldots, n-1$}
    \State Receive a fresh data
    \Statex \hspace{2.5em} point $\bfx_t \in D$
    \State \Blue{$\gradient_t \gets \nabla_\bftheta \ell(\bftheta_t,\bfx_t)$} 
    \State $\bftheta_{t+1} \gets \bftheta_t - \eta \, \Blue{\gradient_t}$
    \EndFor
    \State \textbf{Return $\bftheta_n$} \strut
    \vspace{0.1em}  
    \end{algorithmic}
    \end{algorithm}
    \end{minipage}%
    \hfill
    \begin{minipage}{0.5\textwidth}
    \vspace{0pt}
    \centering
    \begin{algorithm}[H]
    \caption{\small \Red{DP-SGD}}
    \label{alg:dpsgd-1}
    \small
    \textbf{Inputs:} 
    Inputs $D, n, \eta$ to SGD, 
    clip norm $\clipnorm$, noise variance $\stddev^2$

    \begin{algorithmic}[1]
    \State Pick an initial model $\bftheta_0 \in \Theta$
    \For{$t = 0,\ldots, n-1$}
    \State Receive fresh $\bfx_t \in D$
    \State \Red{$\gradient_t \gets \clip_\clipnorm\big( \nabla_\bftheta \ell(\bftheta_t, \bfx_t) \big)$}
    \label{line:dpsgd-clip-new}
    \State \Red{$\widehat \bfg_t \gets \gradient_t + \normalm{0}{\stddev^2}$}
    \label{line:dpsgd-noise-new}
    \State $\bftheta_{t+1} \gets \bftheta_t - \eta \, \Red{\dpgradient_t}$
    \label{line:dpsgd-gradient-addition}
    \EndFor
    \State \textbf{Return $\bftheta_n$} \strut
    \end{algorithmic}
    \end{algorithm}
    \end{minipage}
    \caption{\Blue{Stochastic Gradient Descent (SGD)} and \Red{differentially private stochastic gradient descent (DP-SGD)}, both with a batch size of $1$.
    Note that DP-SGD clips its gradients to norm at most $\clipnorm$ using the function $\clip_\clipnorm(\bfv) \coloneqq \bfv \cdot \min\{1, \clipnorm / \ltwo{\bfv}\}$
    }
    \label{alg:sgd-and-dp-sgd}
\end{figure}

\begin{bremark} \label{remark:large-batch}
In AI and machine learning applications, stochastic gradient optimization typically uses gradients computed on mini-batches of examples, and might use a different update rule than the fixed learning rate given in \Cref{alg:sgd-1} (for example, momentum or adaptively chosen learning rates). Furthermore, we typically make multiple passes through the data, in contrast to the streaming assumption that each datapoint is processed only once.
We will consider these extensions later, in \Cref{chap:ml}.
The discussions in this chapter directly generalize to larger batches and other first-order optimizers, while the lifting the streaming assumption requires significant extensions, and is the subject of \Cref{chap:ml}.
\end{bremark}

\paragraph{Differentially Private SGD (DP-SGD)}
Suppose we wish to solve the learning problem in \cref{eq:empiricalrisk} with a differential privacy constraint such as $\mu$-GDP.
This is achieved by a differentially private version of SGD, known as DP-SGD, and is contrasted with the non-private version of SGD in \Cref{alg:dpsgd-1}.

DP-SGD makes two key modifications to SGD:
\begin{itemize}
    \item \textbf{Gradient Clipping}:
    In order to restrict the $\ell_2$ sensitivity of a data point $\bfx_t$, each gradient $\nabla_\bftheta \ell(\bftheta_t, \bfx_t)$ is clipped so that its $\ell_2$ norm is at most some constant $\clipnorm > 0$. That is, if the gradient has a larger norm, we multiply it by the largest constant $\alpha > 0$ such that $\bfg_t = \alpha \, \nabla_\bftheta \ell(\bftheta_t, \bfx_t)$ satisfies $\norm{\bfg_t}_2 \le \clipnorm$. This is performed by Line~\ref{line:dpsgd-clip-new} of \Cref{alg:sgd-and-dp-sgd} (right) using the following clipping function: for a vector $\bfv$ and $\clipnorm>0$
    \begin{align}
    \clip_\clipnorm(\bfv) \coloneqq \bfv \cdot \min\left\{1, \frac{\clipnorm}{\ltwo{\bfv}}\right\}
    \label{eq:clippingfunction}      
    \end{align}
    \item \textbf{Noise Addition}: 
    DP-SGD adds independent zero-mean Gaussian noise $\znoise_t \sim \normalm{0}{\stddev^2}$ to the (clipped) gradients $\gradient_t$ in Line~\ref{line:dpsgd-noise-new} of \Cref{alg:sgd-and-dp-sgd} (right). 
    Here, the noise variance $\stddev^2$ is calibrated to the desired privacy level, e.g. $\mu$-GDP.
\end{itemize}

The main objective of this monograph is to study a family of correlated noise mechanisms which privatize the gradient $\dpgradient_t = \gradient_t + \corrznoise_t$ using noise $\corrznoise_t$ that is correlated across time. That is, $\corrznoise_t$ and $\corrznoise_{\tau}$ for $t \neq \tau$ are not required to be statistically independent.

\paragraph{From SGD to Prefix Sum Estimation}
We begin developing correlated noise mechanisms for learning by framing DP-SGD as a problem of privately estimating prefix sums.
For convenience, let us assume that a learning rate $\eta > 0$ is fixed throughout.
Unrolling the model parameters updates of non-private SGD yields
\begin{align} \label{eq:sgd-iterates-unrolled}
    \bftheta_t - \bftheta_0 = - \eta \sum_{\tau=0}^{t-1} \gradient_\tau \,.
\end{align}
Thus, the sequence of iterates of the SGD algorithm are obtained from the prefix sums $\bfs_t := \sum_{\tau=0}^{t} \gradient_\tau  \in \real^\mdim$ of the gradients $\gradient_0, \gradient_1, \ldots \in \real^\mdim$.

Now consider two matrices $\bfS \in \R^{n \times m}$ and $\bfG \in \R^{n \times m}$ formed by stacking the prefix sums and gradients row-wise, i.e.,
\begin{align}
\label{eq:SandG}
    \bfS = \begin{pmatrix}
    \bfs_0 \\ \bfs_2 \\ \vdots \\ \bfs_{n-1}
    \end{pmatrix} \in \real^{\nd \times \mdim} \quad \text{and} \quad 
    \bfG = \begin{pmatrix}
    \gradient_0 \\ \gradient_2 \\ \vdots \\ \gradient_{n-1}
    \end{pmatrix} \in \real^{\nd \times \mdim}.
\end{align}

\noindent Then it is easy to check that $\bfS$ can be obtained from $\bfG$ via a linear map $\bfS = \prefix {\bm G}$ corresponding to the lower-triangular matrix
\begin{equation} \label{eq:all-ones-workload}
\prefix\idx{t}{\tau} = \begin{cases}
1 & \text{ if } t \le \tau \\
0 & \text{ else};
\end{cases}
\qquad \text{e.g,} \qquad
\prefix^{4\times4} = 
\begin{psmallmatrix}
1 & 0 & 0 & 0 \\
1 & 1 & 0 & 0 \\
1 & 1 & 1 & 0 \\
1 & 1 & 1 & 1 
\end{psmallmatrix}.
\end{equation}
The matrix $\prefix$ is also known as the \emph{prefix sum workload matrix}.

Vanilla SGD with a non-constant learning rate and
other first-order optimizers for the empirical risk minimization problem such as SGD with \emph{momentum} lead to different lower triangular workload matrices $\workload$ that may not necessarily have only binary entries; we give a few more examples in \Cref{chap:ml}. Thus, we use $\prefix$ when referring specifically to prefix sums (vanilla SGD), but whenever possible, we state results in terms of a arbitrary lower-triangular workload matrix $\workload$, allowing the extension to other first-order optimizers and {learning rate schedules}.
We note that prefix sums arise naturally in other contexts such as counting items in a streaming setting or density estimation, as we discuss in \Cref{sec:chap1-notes}.

In this monograph, we are interested in differentially private learning algorithms. This translates to estimating these (weighted) prefix sums with diffential privacy guarantees. We formalize this next.

\paragraph{General Problem: Private Weighted Prefix Sum Estimation} 
Generalizing the above, we can consider the problem of computing differentially private estimates of \emph{weighted} prefix sums $\bfs_0, \ldots, \bfs_{n-1}$ of a sequence of vectors representing
(functions of) datapoints $\gradient_0, \gradient_1, \ldots, \gradient_{\nd-1} \in \R^m$ weighted by a \emph{workload matrix} $\workload \in \R^{\nd \times \nd}$:
\begin{align} \label{eq:mainProblem}
    \text{Estimate }\,\, \bfs_t 
    \coloneqq \sum_{\tau=0}^t \workload[t, \tau] \, \gradient_{\tau}    
    \,\,\,\quad  \forall t \in [n] \,\, \text{ with DP}.
\end{align}

Since only the lower triangular part of $\workload$ shows up in \cref{eq:mainProblem}, we assume without loss of generality that $\workload$ is a lower triangular matrix, so equivalently we may write $\bfs_t = (\workload \Gradients)[t, :]$ and our goal is to estimate the rows of $\bfS = \workload\Gradients$ with DP.
Recall that, in the case of vanilla SGD, $\workload=\prefix$ while it can have different form for other first-order optimizers.
Finally, we assume throughout that the inputs $(\gradient_t)_{t=0}^{n-1}$ are bounded in $\ell_2$ norm: 
\begin{align} \label{eq:ch1-grad-norm-bound}
    \norm{\gradient_t}_2 \le 1\,.
\end{align}

In the context of DP-SGD, this is equivalent to taking the clip norm $\clipnorm=1$ in \cref{eq:clippingfunction}.
This assumption is without loss of generality: 
since the output $\bfs_t$ in \cref{eq:mainProblem} is a linear function of $\gradient_t$, we can take $\norm{\gradient_t}_2 \le 1$ and scale $\bfs_t$ by $\clipnorm$ instead.

\begin{bremark}
We will give an alternative (though equivalent) perspective in \Cref{sec:noisecancel}: we can view the problem as providing DP estimates $\dpgradient_t = \gradient_t + \corrznoise_t$ of the $\gradient_t$, with (correlated) noise $\corrznoise_t$ chosen such that beneficial cancellation occurs when we compute $\workload \dpGradients$ as post processing.
\end{bremark}

\begin{bremark}
Though the matrix $\workload$ is flexible enough to describe more complicated first-order optimizers and learning rate schedules, we remark that this can be achieved either by directly describing the algorithm in $\workload$ as described above or by using the standard prefix sums $\prefix$ with the corresponding choice of inner optimizer. This can lead to different privacy-utility Pareto frontiers for machine learning applications as may be observed in \Cref{chap:ml}.
\end{bremark}

\begin{figure*}
    \centering
    \adjustbox{max width=\linewidth}{
    \input{adjacency_intro}}
    \caption{Illustration of adjacency of gradient sequences $\gradient_t = \nabla \ell(\bftheta_t, \bfx_t)$ and $\gradient_t' = \nabla \ell(\bftheta_t, \bfx_t')$, where equal datapoints are indicated by having the same color cell. Following \cref{thm:gdp-adaptivity}, we assume gradients are computed in the non-adaptive setting for a fixed sequence $\thseq$.
    \textbf{Left}: In the streaming setting, each datapoint participates only once. Thus,
    the sequences of gradients on adjacent datasets differ in only one item (in this example, we have $\gradient_1 \neq \gradient_1'$) and are equal everywhere else.
    \textbf{Right}: In the multiple-participation setting considered in \cref{chap:ml}, each datapoint can participate multiple times (e.g. by taking mulitple passes through the dataset). Thus, sequences of gradients on adjacent datasets can differ in multiple indices. In this example, the blue datapoint is replaced by the green one, changing both $\gradient_1$ and $\gradient_3$---note that the same (changed) datapoint is used for both $\gradient_1$ and $\gradient_3$. 
    In this section, we focus on the streaming setting.
    }
    \label{fig:adjacency-intro}
\end{figure*}

\section{Correlated Noise Mechanisms}
\label{sec:correlatedNoiseIntro}

We now turn our attention to the main focus of this monograph: correlated noise mechanisms. 
Such mechanisms can be described as producing differentially private estimates of the prefix sums $\workload \Gradients$.

In particular, given a factorization $\workload=\bfB \strategy$ of the workload matrix $\workload \in \R^{n \times n}$, we define a \textbf{correlated noise mechanism} as
\begin{align} \label{eq:correlatedMechDef}
    \mech(\Gradients) :=
    \bfB \br{\strategy \Gradients + \Znoise}
    = \workload \Gradients + \bfB \Znoise \,,
\end{align} 
where $\bfB \in \R^{\nd \times \nd}$ and $\strategy \in \R^{\nd \times \nd}$ are lower triangular matrices, and  $\Znoise \sim \calN_{\nd \times \mdim} (0,{\stddev^2})$ is an $n \times \mdim$ matrix of component-wise i.i.d. Gaussian noise with appropriate variance $\stddev^2$. 

\paragraph{Correlated Noise for DP-SGD}
To map these correlated noise mechanisms directly to DP-SGD (\Cref{alg:sgd-and-dp-sgd}, right),
we can equivalently write \Cref{eq:correlatedMechDef} as 

\begin{align} \label{eq:correlatedMechCinv}
    \mech(\Gradients) = \workload(\Gradients + \Cinv \Znoise) \,,
\end{align}
assuming the matrix $\strategy$ is invertible.
This equation involves two components: (1) computing the noisy gradients $\dpGradients = \Gradients + \Cinv \Znoise$, and (2) multiplying it by the workload matrix $\workload$ to return $\workload\dpGradients$.

First, the noisy gradients $\dpgradient_t = \dpGradients\idx{t}{:} = \gradient_t + \corrznoise_t$ are computed by injecting the correlated noise 
\begin{align} \label{eq:corr_znoise_t}
    \corrznoise_t \coloneqq \big(\Cinv\Znoise\big)\idx{t}{:} = \sum_{\tau = 0}^{t} (\Cinv)\idx{t}{\tau} \, \znoise_{\tau} \,,
\end{align}
where $\znoise_t \sim \normalm{0}{\stddev^2}$ is i.i.d. seed noise. This is a direct replacement to the i.i.d. noise update in DP-SGD (Line~\ref{line:dpsgd-noise-new} in \Cref{alg:dpsgd-1}).

\begin{algorithm}[t]
\caption{DP-SGD with \Red{Correlated Noise}}
\label{alg:dpsgd-corr}
\textbf{Inputs:} 
Dataset $D$, number of steps $n$, learning rate $\eta$, clip norm $\clipnorm$, noise variance $\stddev^2$,
noise correlating matrix $\Cinv$ (standard DP-SGD uses $\Cinv = \bfI_{n\times n}$)
\begin{algorithmic}[1]
\State Pick an initial model $\bftheta_0 \in \Theta$
\For{$t$ \textbf{in} $0, 1, \dots, n - 1$}
\State Receive a fresh data point $\bfx_t \in D$
\State \label{line:clip-dpsgd-intro}
$\gradient_t \gets \clip\big(\nabla_\bftheta\ell(\bftheta_{t}; \bfx_t)\big)$ 
\Comment{$\clip(\bfv) := \bfv \cdot \min\{1, \clipnorm / \ltwo{\bfv}\}$}
\State Sample i.i.d. seed noise $\znoise_t \sim \normalm{0}{\stddev^2}$   
\State Calculate the  \Red{correlated noise $$\corrznoise_t \gets \sum_{\tau = 0}^{t} (\Cinv)\idx{t}{\tau} \, \znoise_{\tau}$$ }
\State \label{line:dpgrad-dpsgd-intro}
$\dpgradient_t \gets \gradient_t + \Red{\corrznoise_t}$
\State $\bftheta_{t+1} \gets \bftheta_t - \eta \dpgradient_t$
\EndFor
\end{algorithmic}
\end{algorithm}

Second,
multiplication by the matrix $\workload=\prefix$ (for the case of vanilla SGD) is achieved by the gradient step $\bftheta_{t+1} = \bftheta_t - \eta \dpgradient_t$ (Line~\ref{line:dpsgd-gradient-addition} in \Cref{alg:dpsgd-1}); this can be established by unrolling the noisy SGD update, similar to \cref{eq:sgd-iterates-unrolled}. This gradient step is usually implemented programmatically using an optimizer step.
The resulting algorithm is given in \Cref{alg:dpsgd-corr}.
Here, the matrix $\strategy$ is also known as the \emph{strategy matrix} or \emph{encoder matrix}, while the matrix $\Cinv$ is known as the \emph{noise correlating matrix}. 

\subsection{Privacy Guarantees of Correlated Noise Mechanisms}
\label{sec:ch1-privacy-correlated-noise}
Recall that two datasets $D$ and $D'$ satisfy replace-one example adjacency if $D'$ can be obtained by replacing one element of $D$, as per \Cref{def:replaceone}.
Since the correlated noise mechanism in \Cref{eq:correlatedMechDef} takes in the gradients as input, we need to reason about how adjacency of the underlying datasets affects the corresponding (clipped) gradients 
$\gradient_t = \clip\big(\nabla_{\bftheta} \ell(\bftheta_t, \bfx_t)\big)$ and $\gradient_t' = \clip\big(\nabla_{\bftheta} \ell(\bftheta_t, \bfx_t')\big)$ computed using $\bfx_t \in D$ and $\bfx_t' \in D'$ respectively for $t = 0, \ldots, n-1$.

Suppose $D$ and $D'$ differ in the $t$\textsuperscript{th} example where $\bfx_t \neq \bfx_t'$. This affects not only the gradient $\gradient_t$ in step $t$, but also \emph{all subsequent gradients} $\gradient_{t+1}, \ldots, \gradient_{n-1}$ via the updated model parameters $\bftheta_{t+1}$.
Fortunately, for the purposes of privacy analysis, it turns out that we can ignore the effect of of $\bfx_t$ on all future gradients $\gradient_{t+1}, \ldots, \gradient_{n-1}$. In particular, as we discuss next, the privacy analysis of a correlated noise mechanism coincides with a similar non-adaptive procedure.

\begin{figure}[t]
    \centering
    \fbox{
    \begin{minipage}{.45\textwidth}
    \centering \small
    \underline{\Blue{Adaptive Setting}}
\begin{algorithmic}
    \State Choose $\bftheta_0 \in \Theta$
    \State Let $f_0, \dots, f_{\nd - 1}$ be functions 
    \Statex \hspace{2em} $f_t: \calX^* \times \Theta^* \to \R^\mdim$. 
    \For{$t = 0, \ldots, n-1$}
        \State $\bff_t = f_t\big(D, (\bftheta_\tau)_{\tau=0}^t \big)$
        \State $\hbff_t = \bff_t + \normalm{0}{\stddev^2}$
        \State  \Blue{$\bftheta_{t+1} \gets $ an arbitrary function} \State \hspace{2.5em}\Blue{of $(\hbff_\tau)_{\tau=0}^{t}$ and $(\bftheta_\tau)_{\tau=0}^t$}
        \EndFor
        \State \textbf{Return} $\Blue{\hFa} \coloneqq \big(\hbff_0, \dots, \hbff_{n-1}\big)$
    \end{algorithmic}
    \end{minipage}%
    
    \begin{minipage}{0.45\textwidth}
    \centering \small
    \underline{\Red{Non-adaptive Setting}}
    \begin{algorithmic}
    \State Choose $\bftheta_0 \in \Theta$.
    \State Let $f_0, \dots, f_{\nd - 1}$ be functions 
    \Statex \hspace{2em} $f_t: \calX^* \times \Theta^* \to \R^\mdim$. 
    \State \Red{Fix a sequence from $\Theta^{n-1}$}
    \State \hspace{2em} \Red{$\thseq \coloneqq (\bftheta_1, \dots, \bftheta_{\nd-1})$}
    \For{$t = 0, \ldots, n-1$}
        \State $\bff_t = f_t\big(D, (\bftheta_\tau)_{\tau=0}^t \big)$
        \State $\hbff_t = \bff_t + \normalm{0}{\stddev^2}$
    \EndFor
    \State \textbf{Return} $\Red{\widehat \bfF} \coloneqq \big(\hbff_0, \dots, \hbff_{n-1}\big)$
    \end{algorithmic}

    \end{minipage}
    }
    \caption{\Blue{ Adaptive} and \Red{non-adaptive}
    iterative procedures with state denoted by $\bftheta_t$. DP-SGD can be expressed as an instance of the adaptive setting; however,  \cref{thm:gdp-adaptivity} shows that we achieve the same GDP guarantee in the non-adaptive setting, which greatly simplifies privacy analysis.
    }
    \label{fig:dp-sgd-real-and-ideal}
\end{figure}

\paragraph{Adaptive vs. Non-Adaptive GDP}
\Cref{fig:dp-sgd-real-and-ideal} gives an abstract iterative procedure with state denoted by $\bftheta_t$. The full sequence of states $\bftheta_0, \ldots, \bftheta_{n-1}$ is fixed ahead of time in the non-adaptive setting. In contrast, the state $\bftheta_{t+1}$ in the adaptive setting is allowed to depend arbitrarily on all \emph{past} information, including the previous states $\bftheta_0, \ldots, \bftheta_t$ as well as the privatized outputs $\hbff_0, \ldots, \hbff_t$. In both settings, the state $\bftheta_t$ is used to compute a data-dependent output $\bff_t$, whose privatized version $\widehat\bff_t$ (via the Gaussian mechanism) is then released.

The following general theorem shows that the privacy analysis with Gaussian additive noise in the adaptive case can be reduced to the non-adaptive case, though take care to note that analogous result does not hold for all mechanisms. This result underpins the privacy analysis of all the algorithms in this monograph:

\begin{btheorem}[Adaptive vs. Non-Adaptive Gaussian mechanism]
\label{thm:gdp-adaptivity}
$\quad$
Consider the two mechanisms defined in \Cref{fig:dp-sgd-real-and-ideal} with functions $f_0, \dots, f_{\nd - 1}: \calX^* \times \Theta^* \to \R^\mdim$, whose (combined) $\ell_2$-sensitivity as a function of the first argument is bounded by $s$ for \emph{any fixed inputs} to the second argument as:
\begin{gather*}
    \sup_{\bftheta_0, \ldots, \bftheta_{n-1} \in \Theta} \,\, \sup_{D \simeq D'} \,\,
    \lfrob{
    F\big(D, (\bftheta_t)_{t=0}^{n-1}\big) - F\big(D', (\bftheta_t)_{t=0}^{n-1}\big)
    } \le s\,, \quad\text{with} \\
    F\big(D, (\bftheta_t)_{t=0}^{n-1}\big) \coloneqq \begin{bmatrix} 
    f_0(D, \bftheta_0) &  \cdots & f_{n-1}\big(D, (\bftheta_t)_{t=0}^{n-1} \big)\end{bmatrix}\T \in \R^{\nd \times \mdim}
\end{gather*}
denoting a matrix whose rows are the outputs of $f_0, \ldots, f_{n-1}$, and $\lfrob{\bfM} = \sqrt{\sum_{i, j} \bfM\idx{i}{j}}$ denotes the Frobenius norm of the matrix $\bfM$.\footnotemark \,
Then, for any fixed noise variance $\stddev^2 > 0$, the output \Blue{$\hFa$ of the adaptive setting} satisfies the same GDP guarantee as the output \Red{$\widehat \bfF$ of the non-adaptive setting}.
\end{btheorem}
\footnotetext{This is the worst-case $\ell_2$ sensitivity in the non-adaptive setting. The Frobenius norm of a matrix is exactly the $\norm{\cdot}_2$ norm of the vector obtained by flattening it, and is used to compute the $\ell_2$ sensitivity of matrix-valued maps.}

Let $\clip(\cdot)$ denote the clipping funtion $\clip_\clipnorm(\cdot)$ defined in  \cref{eq:clippingfunction} instantiated with $\clipnorm=1$. Then 
\Cref{thm:gdp-adaptivity} captures DP-SGD with independent noise (\Cref{alg:dpsgd-1}): it is simply an adaptive procedure where the function $f_t$ denotes a computation of the clipped gradient 
\[
f_t\big(D, (\bftheta_\tau)_{\tau=0}^t\big) = \clip\big(\nabla_\bftheta \ell(\bftheta_t, \bfx_t)\big) \,,
\]
while the state update is the gradient update $\bftheta_{t+1} = \bftheta_t - \eta \, \widehat \bff_t$. 
In particular, note that $f_t\big(D, (\bftheta_\tau)_{\tau=0}^t\big)$ depends only on current model $\bftheta_t$ and not on the previous $\theta_\tau$'s for $\tau < t$. We need the additional generality of~\Cref{thm:gdp-adaptivity} to establish the privacy guarantee of \Cref{alg:dpsgd-corr}, as we will momentarily see in  \Cref{thm:gdp-of-correlated-noise-adaptive}.

In the adaptive setting, changing one example $\bfx_t$ can influence all of the following gradients, $\gradient_t, \dots, \gradient_{n-1}$. Fortunately, \cref{thm:gdp-adaptivity} allows us to instead analyze the privacy properties of the corresponding non-adaptive setting; here, by design, changing the data point $\bfx_t$ can \emph{only} change the gradient $\gradient_t$, while all other gradients are unchanged. 

Hence, we can easily extend the definition of adjacency to sequences of gradients by assuming the non-adaptive setting. In the streaming setting where each example is processed only once, we say that two sequences of gradients $\Gradients = (\gradient_t)_{t=0}^{n-1}$ and $\Gradients' = (\gradient_t')_{t=0}^{n-1}$ are adjacent if we have that
$\gradient_\tau = \gradient_\tau'$ for all $\tau \in [n]$, except possibly at some step $t$ (where $\bfx_\tau \neq \bfx_\tau'$).

\paragraph{GDP Bound of Correlated Noise Mechanisms}
We are now ready to tackle the privacy analysis of correlated noise mechanisms. 

In order to apply \Cref{thm:gdp-adaptivity}, our first step will be to derive a privacy analysis for the correlated noise mechanism of  \cref{eq:correlatedMechDef} in the non-adaptive setting. 
The key ingredient is bounding the sensitivity induced by the strategy matrix:
\begin{equation}
\sens(\strategy)
        \coloneqq \sens(\Gradients \mapsto \strategy \Gradients) =  \max_{\Gradients \simeq \Gradients'} \lfrob{\strategy(\Gradients - \Gradients')}.
\end{equation}
(Since the map $\Gradients\mapsto \strategy\Gradients$ returns a matrix, its $\ell_2$-sensitivity is computed using the Frobenius norm.)
This quantity is tightly related to the maximum column norm of $\strategy$:
\[
    \colnorm{\strategy} \coloneqq \max_{t \in [n]} \norm{\strategy[:, t]}_2 \,.
\]

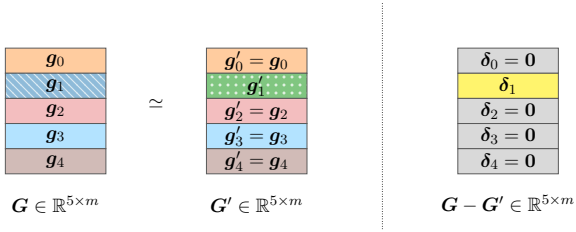
\begin{figure*}
    \centering
    \adjustbox{max width=0.75\linewidth}{
    \input{adjacency_intro_2}}
    \caption{For two adjacent sequences of gradients $\bfG \simeq \bfG'$ in the streaming setting, we have that $\bfG - \bfG'$ is non-zero only for one row. 
    }
    \label{fig:adjacency-2}
\end{figure*}

\begin{blemma}\label[lemma]{lem:sensbound}
Under replace-one-example adjacency (\cref{def:replaceone}) with gradients clipped to norm 1 (cf. \cref{eq:ch1-grad-norm-bound}), we have
\[\sens(\strategy) =  2 \cdot \colnorm{\strategy}\,.\]
\end{blemma}
\begin{proof}
    Suppose $\Gradients, \Gradients'$ differ in the $t$\textsuperscript{th} row (cf.\ \Cref{fig:adjacency-2}). Recall that $\Gradients$ ($\Gradients'$, respectively) is formed by stacking the vectors $\gradient_t \in \R^{\nd}$ ($\gradient_t' \in \R^{\nd}$, respectively) row-wise. Therefore $\strategy(\Gradients - \Gradients') = \strategy[:, t] (\gradient_t - \gradient'_t)\T \in \R^{p \times\nd}$. Since each row of $\Gradients, \Gradients'$, i.e., $\gradient_t, \gradient_t'$ for all $t \in [\nd]$ is bounded in $\ell_2$ norm, we have the $\ell_2$ norm of their difference, $\bfdelta_t := \gradient_t - \gradient_t'$, bounded as $\norm{\bfdelta_t}_2 \le 2$ by the triangle inequality. Thus, we have
    \begin{align}
        \sens(\strategy)
        &= \max_{\Gradients \simeq \Gradients'} \lfrob{\strategy(\Gradients - \Gradients')}
        \,\,=\,\, \max_{t \in [t], \,\, \norm{\bfdelta_t}_2 \le 2} \, \lfrob{\strategy[:, t] \, \bfdelta_t\T} \notag \\
        &\stackrel{(*)}{=} \max_{t \in [t]} \norm{\strategy[:, t]}_2 \,\cdot\, \max_{\norm{\bfdelta}_2 \le 2} \norm{\bfdelta}_2 
        \,\,=\,\, 2 \cdot \colnorm{\strategy} \,, \label{eq:senscolC}
    \end{align}
    where $(*)$ follows from using $\|\bfu \bfv\T\|_{\mathrm{F}} = \norm{\bfu}_2 \norm{\bfv}_2$ for any vectors $\bfu \in \R^n$ and $\bfv \in \R^m$:
    \begin{align*}
        \norm{\bfu\bfv^\top}_{\op F}^2 &= \sum_{i \in [\nd]} \sum_{j \in [m]} \big(\bfu[i]\, \bfv[j]\big)^2 
        = \sum_{i \in [\nd]} \bfu[i]^2 \sum_{j \in [m]} \bfv[j]^2 = 
        \norm{\bfu}^2_2 \norm{\bfv}^2_2 \,.
    \end{align*}
This completes the proof of \Cref{lem:sensbound}.
\end{proof}
We will see in the starred \Cref{sec:intro:different-adj} that the constant $2$ in \Cref{lem:sensbound} is specific to the replace-one-example adjacency. 

With this lemma in hand, the privacy result for the non-adaptive streaming case is straightforward:
\begin{blemma}
\label[lemma]{thm:gdp-of-correlated-noise} 
Fix a noise multiplier $\nmult > 0$.
Consider the replace-one-example adjacency (\cref{def:replaceone}) of gradients $\Gradients, \Gradients' \in \R^{n \times m}$ in the streaming non-adaptive setting, i.e.
\begin{enumerate}[label=(\alph*)]
    \item \label{item:gdp-corr-intro-a}
    any adjacent $\Gradients \simeq \Gradients'$ satisfy $\gradient_\tau = \gradient_\tau'$ for all rows $\tau \in [\nd]$ except possibly at some row $t \in [\nd]$, and
    \item \label{item:gdp-corr-intro-b}
    the rows $\gradient_t, \gradient_t'$ are clipped to norm 1 (cf. \cref{eq:ch1-grad-norm-bound}).
 \end{enumerate}
Then, the mechanism $\mech(\Gradients) = \bfB(\strategy \Gradients + \Znoise)$
for any matrices $\bfB, \strategy \in \R^{\nd \times\nd}$ (possibly non-lower-triangular and non-invertible) and i.i.d. Gaussian noise $\Znoise \sim \calN_{\nd \times \mdim}(0,{\stddev^2})$ satisfies $\frac{1}{\nmult}$-GDP if we choose the noise standard deviation as $\stddev =  2 \nmult \colnorm{\strategy}$.
\end{blemma}
\begin{proof}
The mechanism $\mech$ is a simple post-processing of the mechanism
\begin{equation}\label{eq:mechprime}
\mech'(\Gradients) \coloneqq \strategy\Gradients + \Znoise, 
\end{equation}
as $\mech(\Gradients) = \bfB \cdot \mech'(\Gradients)$. So it suffices to prove the GDP guarantee for $\mech'$. We have (after flattening the matrices to vectors), that $\mech'$ is instance of the Gaussian mechanism (\cref{def:gaussian-mechanism}) and so the result follows from  \Cref{lem:gdp-of-gaussian-mechanism} using the sensitivity bound from \cref{lem:sensbound}.
\end{proof}

Finally, we extend this non-adaptive privacy guarantee to DP-SGD with correlated noise:
\begin{btheorem} \label{thm:gdp-of-correlated-noise-adaptive}
    Fix a noise multiplier $\nmult$ and consider \Cref{alg:dpsgd-corr} with an invertible lower triangular strategy matrix $\strategy$, clip norm $\clipnorm = 1$, and noise standard deviation $\stddev = 2 \nmult \, \colnorm{\strategy}$. Then, the privatized gradients $(\dpgradient_t)_{t \in [\nd]}$ and
    iterates $(\bftheta_t)_{t \in [\nd]}$ produced by \cref{alg:dpsgd-corr} satisfy $\frac{1}{\nmult}$-GDP under replace-one-example adjacency in the streaming setting.
\end{btheorem}
\begin{proof}
We instantiate \Cref{alg:dpsgd-corr} as an adaptive iterative process in the framework of \Cref{thm:gdp-adaptivity}.
Concretely, we take $f_t$ to be functions which return the rows of $\strategy\Gradients$:
\begin{align} \label{eq:corr-noise-as-adaptive-process}
    f_t\big(D, (\bftheta_\tau)_{\tau=0}^t\big)
    = \big(\strategy\Gradients\big)\idx{t}{:}
    \quad\text{where}\quad 
    \gradient_\tau = \nabla_\bftheta\ell(\bftheta_\tau, \bfx_\tau) \,
\end{align}
for $\bfx_\tau \in D$.
Since $\strategy$ is a lower triangular matrix, we have that $\big(\strategy\Gradients\big)\idx{t}{:}$ depends only on the previous gradients $\gradient_0, \ldots, \gradient_t$, which in turn depend (only) on $(\bftheta_0,\cdots, \bftheta_t)$ and $D$.
It follows then that $\widehat \bff_t = \big(\strategy\Gradients + \Znoise\big)\idx{t}{:}$. We instantiate the state update of \cref{alg:dpsgd-corr} with 
\begin{align*}
    &\dpgradient_t \gets (\Cinv \hFa)\idx{t}{:}
    = \big(\Gradients + \Cinv\Znoise\big)\idx{t}{:}\,, \\
     &\bftheta_{t+1} \gets \bftheta_t - \eta \dpgradient_t.
\end{align*}

Since $\Cinv$ is lower-triangular, $\dpgradient_t$ is post-processing of $\hbff_0, \dots, \hbff_t$, the first $t$ rows of $\hFa$, and so this is a valid update. 
By the same post-processing argument, the privatized gradients $(\dpgradient_t)_{t \in [\nd]}$ and
    iterates $(\bftheta_t)_{t \in [\nd]}$ satisfy the same privacy guarantee as $\hFa$. Thus, with this choice of the functions $f_t$ (and appropriate update), we have that \cref{alg:dpsgd-corr} is in fact an instance of the adaptive setting of \cref{fig:dp-sgd-real-and-ideal}. 

Hence, applying \cref{thm:gdp-adaptivity}, it remains to analyze this choice of $f_t$ in the non-adaptive setting. By construction, the output of \cref{fig:dp-sgd-real-and-ideal} in the non-adaptive setting is
\[
  \widehat \bfF = \strategy \Gradients + \Znoise \,.
\] 
This is exactly the mechanism analyzed by \Cref{thm:gdp-of-correlated-noise} taking $\bfB = \bfI$, and the conditions \ref{item:gdp-corr-intro-a}-\ref{item:gdp-corr-intro-b} are satisfied by the assumption of replace-one-example adjacency, the choice of clipping parameter $\clipnorm = 1$, and non-adaptivity. Finally, the choice of $\stddev$ matches, and so the result follows.
\end{proof}

\subsection{Baseline Correlated Noise Mechanisms}

The family of correlated noise mechanisms defined by \Cref{eq:correlatedMechDef} includes two natural baselines to privately estimate the weighted prefix sums as in \cref{eq:mainProblem}: adding noise to the input or the output.

\paragraph{Baseline I: Input Perturbation} 
The first baseline corresponds to the strategy matrix $\strategy=\bfI_{\nd \times \nd}$, which is the choice taken by standard DP-SGD in \Cref{alg:dpsgd-1}. Here, we \emph{privatize each input} with the Gaussian mechanism $\dpgradient_t = \gradient_t + \znoise_t$ with $\znoise_t \sim \normalm{0}{\stddev^2}$ for some standard deviation $\stddev$. Then, it follows that the post-processed (weighted) prefix sums
\begin{align} \label{eq:baselineIndepZ}
    \widehat \bfs_t = \sum_{\tau=0}^t \workload[t, \tau] \,\dpgradient_\tau =  \sum_{\tau=0}^t \workload[t, \tau]\, \gradient_\tau +  \sum_{\tau=0}^t \workload[t, \tau]\, \znoise_\tau
\end{align}
satisfy the \emph{same} DP guarantees as the privatized inputs $\dpgradient_t$ for all choices of the workload matrix $\workload$.
In particular, taking the standard deviation $\stddev = 2 \nmult$ yields the desired $(1/\nmult)$-GDP guarantee.\footnote{
    Again, the factor of $2$ in the standard deviation $\stddev$ here and in the rest of this section is specific to the replace-one notion of adjacency. This vanishes with other notions of adjacency, as we discuss in \Cref{sec:intro:different-adj}.
}

\paragraph{Baseline II: Output Perturbation}
The second baseline corresponds to $\strategy=\workload$ (so that the matrix $\bfB = \bfI_{\nd \times \nd}$).
This is equivalent to \emph{privatizing each output} directly with the Gaussian mechanism to release
\begin{align} \label{eq:baselineIndepSum}
    \widehat \bfs_t = \left(\sum_{\tau=0}^t \workload[t, \tau]\, \gradient_\tau \right) + \znoise_\tau
\end{align}
for $\znoise_\tau \sim \normalm{0}{\stddev^2}$ with component-wise standard deviation $\stddev$. 
Unlike the input perturbation baseline, $\stddev$ depends on the choice of the workload matrix $\workload$ as it affects the sensitivity of the operation $\Gradients \mapsto \workload\Gradients$. 
For example, with the prefix sum workload $\workload=\prefix$, \Cref{lem:sensbound} yields that $\sens(\prefix) = 2 \sqrt{n}$.
Thus, to obtain a $(1/\nmult)$-GDP guarantee on the noisy prefix sums $\widehat \bfs_t$ in this case, we need a significantly larger standard deviation of $\stddev = 2\sqrt{n} / \mu$ compared to the input perturbation baseline.

Thus, the best correlated noise mechanism (for a given objective) is no worse than either of the two baselines.
However, an important question remains: can general correlated noise mechanisms (with \mbox{$\bfB, \bfC \neq \bfI_{\nd \times \nd}$}) \emph{strictly} improve over the baselines?
In other words, is there a strong reason to choose \emph{non-trivial} noise correlations across time steps?

\section{Why Correlated Noise Mechanisms?}
\label{sec:whyCorrelatedNoise}

We now develop some intuition regarding why correlated noise mechanisms might lead to improved privacy-utility tradeoffs for DP prefix sum estimation.   

In particular, if $\bfs_1, \ldots, \bfs_\nd$ are the true prefix sums and $\widehat \bfs_1, \ldots, \widehat \bfs_\nd$ are the outputs of a correlated noise mechanism, then we will argue that the maximum prefix sum loss
\[
\max_{t \in [n]} \, \mathbb{E} \norm{\widehat \bfs_t - \bfs_t}_2^2
\]
is smaller than the baseline described in the previous section at each privacy level.

\subsection{Why Estimate Prefix Sums Directly?}
\label{sec:trajectory-ex}

\begin{figure}[p]
\centering
    \includegraphics[width=0.95\linewidth]{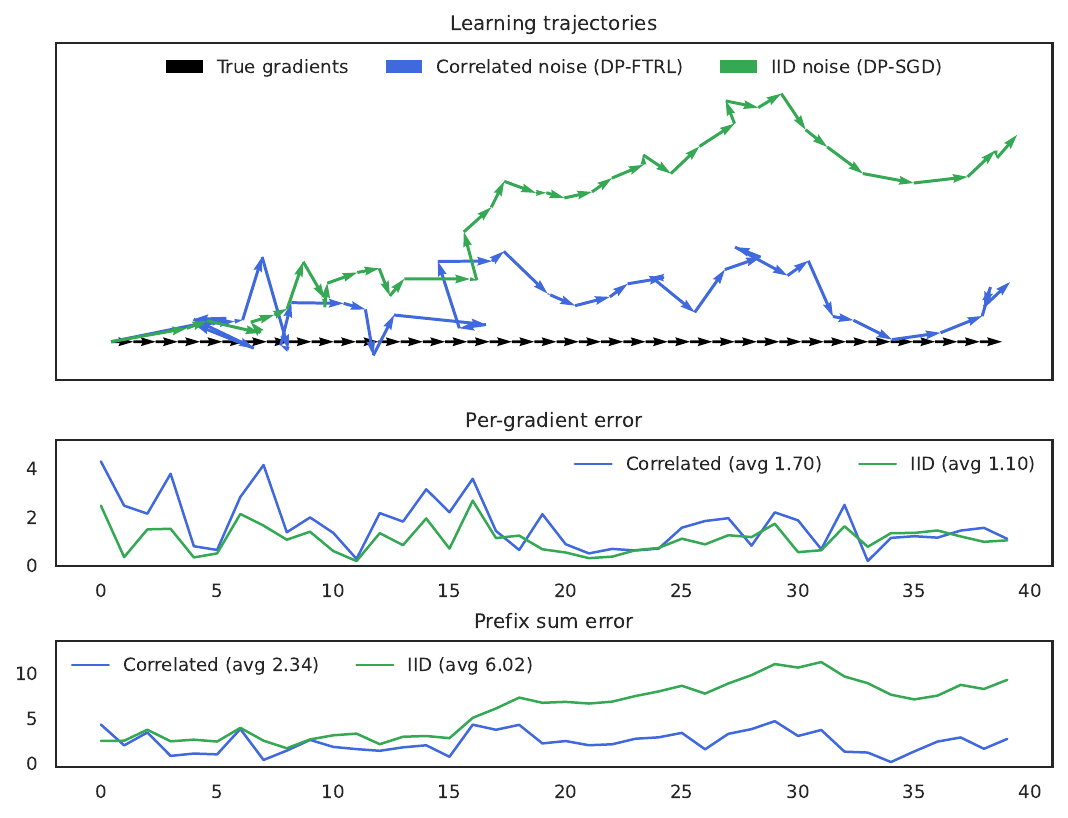}
\caption{
Example trajectories for learning with correlated noise vs. i.i.d. noise for $\nd=40$ steps in $\mdim=2$ dimensions. This figure considers a simple example where the black line is the non-private gradient descent baseline with learning rate $\eta = 1$, where the true gradients obtained at each step is $\gradient_t = (-1, 0)$. The {\color{blue}blue} line uses a correlated noise mechanism (optimized to achieves the best worst-case performance in the prefix sums; cf. \Cref{sec:optimized-dense-mech}) 
and the {\color{green}green} line corresponds to i.i.d. noise via DP-SGD (i.e., the input perturbation mechanism with $\bfC = \bfC^{-1} = \bfI_{n \times n}$).  We calibrate these mechanisms to represent equivalent privacy guarantees at any given noise level as per \Cref{thm:gdp-of-correlated-noise}; here we use Gaussian noise with $\stddev = 1$. The middle plot shows that i.i.d. noise is better at estimating individual gradients (1.10 vs 1.70 average root mean squared error in estimating the individual gradients $(-1, 0)$). However, correlated noise results in a trajectory that on average stays closer to the baseline trajectory, thanks to lower error in prefix sum estimates shown in the bottom plot (2.34 vs 6.02 average root mean squared error in estimating the prefix sums $(-t, 0)$ for $t \in [40]$).
} 
\label{fig:max-error}
\end{figure}

We consider a simple example in \Cref{fig:max-error} comparing a non-trivial correlated noise mechanism to the input perturbation mechanism, where $\strategy = \bfI_{n \times n}$ (i.e., DP-SGD, which uses i.i.d.\ noise). 
Consider minimization of the simple linear function $f(\bftheta_1, \bftheta_2) = -\bftheta_1$ in $\mdim = 2$ dimensions over the bounded set $\Theta = \{(\bftheta_1, \bftheta_2) \, :\, \bftheta_1^2 + \bftheta_2^2 \le M\}$, for a sufficiently large number $M$. At each step of the optimization, the gradient of the objective function is $\nabla f(\bftheta_1, \bftheta_2) = (-1, 0)$.

The top plot in \Cref{fig:max-error} illustrates (noisy) gradient descent with a learning rate of $\eta = 1$ under different noise conditions for $\nd=40$ steps starting from $\bftheta_0 = (0, 0)$. The black arrows depict the noiseless case, which produces iterates $\bftheta_t^{\mathsf{GD}} = (t, 0)$ as expected.
The {\color{green}green} arrows represent the sample path obtained under i.i.d. noise:
\[
 \bftheta_t^{\mathsf{IID}} = \left(t + \sum_{\tau=0}^{t-1} w_{\tau, 1},  \sum_{\tau=0}^{t-1} w_{\tau, 2}\right) \,, \quad \text{where} \quad \znoise_\tau \sim \normaldim{2}{0}{1} \,.
\]

The {\color{blue}blue} arrows depict that of a correlated noise mechanism corresponding to a certain decomposition $\prefix = \bfB_\star\strategy_\star$ (described in \Cref{sec:optimized-dense-mech}):
\[
    \bftheta_t^{\mathsf{corr}} = \left(t, 0\right) + \big(\bfB_\star \Znoise\big)[t, :] \,.
\]
where $\Znoise \sim \normalnm{0}{\colnorm{\strategy_\star}^2}$.
Note that the noise is scaled according to \Cref{thm:gdp-of-correlated-noise} so that both mechanisms represent equivalent privacy guarantees. We can then calculate:
\begin{align*}
      \bftheta_t^{\mathsf{IID}} - \bftheta_t^{\mathsf{GD}} &\sim \normaldim{2}{\zeros}{t} \,, \quad \text{and} \\
      \bftheta_t^{\mathsf{corr}} - \bftheta_t^{\mathsf{GD}} &\sim
      \normal_2\left(\zeros, \colnorm{\strategy_\star}^2 \norm{\bfB_\star[t, :]}_2^2 \right) \,.
\end{align*}

The correlated noise mechanism can yield a trajectory that is closer in the worst-case to the noiseless black trajectory (in expectation) if we can have
\begin{align} \label{eq:example-ch1}
    \max_{t \in [\nd]} \colnorm{\strategy_\star}^2 \norm{\bfB_\star[t, :]}^2 \le  \max_{t\in [\nd]} \,\, t \, = \nd - 1
\end{align}
for some factorization $\prefix = \bfB_\star \strategy_\star$. This is indeed possible, and we will see a quantitative example next.

The reason behind this improvement is the direct estimation of prefix sums rather than the gradients. Indeed, while i.i.d. noise is the optimal strategy to minimize the total squared error in the noisy \emph{gradients} $\dpGradients = \Gradients + \strategy^{-1} \Znoise$, correlated noise mechanisms are able to produce better estimates of the prefix sums $\workload\dpGradients = \workload\Gradients + \bfB \Znoise$, resulting in better worst-case noisy \emph{trajectories}.

\subsection{Quantitative Benefits of Using Correlated Noise}

We now build on the example of \Cref{sec:trajectory-ex} by constructing a simple correlated noise mechanism that outperforms the input-perturbation and output-perturbation baselines. For simplicity, we compare all three approaches under a fixed privacy requirement of $1$-GDP.
(As we will see, our calculations will actually hold for any level of $\mu$-GDP.)

Computing the gradient descent iterates in the previous setting reduces to estimating prefix sums with differential privacy, Problem~\ref{eq:mainProblem} for the prefix-sum matrix $\prefix$, as in \cref{eq:all-ones-workload}. For simplicity, suppose that the dimension of the stream is $\mdim=1$, that is $\gradient= (\gradient_0, \ldots, \gradient_{n-1}) \in \R^\nd$. 

Next, we need to quantify exactly how good our private prefix sum estimates are. As we saw in \cref{eq:example-ch1}, a reasonable objective is to minimize the maximum squared error between the private estimate $\widehat s_t = \mech(\gradient)[t]$ and the actual prefix sum $s_t$.\footnote{
    Other objectives are possible, as we discuss in \Cref{chap:prefixsum}
}  
Formally, we wish to minimize
\begin{align}
\label{eq:whycorrelate}
\max_{t \in [n]} \, \mathbb{E}{( \widehat s_t - s_t)^2} =
\max_{t \in [n]} \, \mathbb{E}{\Big(\br{\bfB\znoise}[t,:]\Big)^2}
= \stddev^2 \,\, \max_{t \in [n]} \norm{\bfB[t, :]}_2^2,
\end{align}
the maximum squared row norm of $\bfB$ scaled by the noise variance necessary to achieve $1$-GDP. Note that we have dropped the dependence on $\strategy$ via~\cref{thm:gdp-of-correlated-noise-adaptive}.

To  build some intuition why correlated noise helps, let us consider a simple family of correlated noise mechanisms parameterized by $0 \le \lambda < 1$ with
\begin{align} \label{eq:oneBuffertoeplitzC}
\strategy =\begin{pmatrix} 1 & 0 & 0 & \ldots & \ldots & 0 \\
\lambda & 1 & 0 & \ddots & \ddots & 0 \\
\lambda^2 & \lambda & 1 & \ddots & \ddots & 0\\
\lambda^3 & \lambda^2 & \lambda & \ddots & \ddots & 0\\
\vdots & \ddots & \ddots & \ddots & \ddots & \vdots
\end{pmatrix} \,.
\end{align}

These constructions provides a simple interpolation between the baselines introduced in \Cref{sec:correlatedNoiseIntro}: $\lambda=0$ recovers input perturbation, and $\lambda=1$ would recover output perturbation (though for technical reasons, our analysis requires $\lambda < 1$).

It can be easily checked 
that 
\begin{align} \label{eq:cinv:one-param-family}
    \bfC^{-1} 
    = \begin{pmatrix}
    1 & 0 & 0 & \ldots & 0 \\
    -\lambda & 1 & 0 & \ddots & 0 \\
    0 & -\lambda & 1  & \ddots & 0\\
    0 & 0 & -\lambda & \ddots & 0 \\
    \vdots & \ddots & \ddots & \ddots & \vdots 
    \end{pmatrix}.
\end{align}
Then, we get
\begin{align}
\label{eq:oneBuffertoeplitz}
    \bfB=\prefix \inv{\strategy}
    &= \begin{pmatrix}1 & 0 & 0 & \ldots & 0 \\
1-\lambda & 1 & 0 & \ddots & 0 \\
1-\lambda & 1-\lambda & 1  & \ddots & 0\\
1-\lambda & 1-\lambda & 1-\lambda & \ddots & 0 \\
\vdots & \ddots & \ddots & \ddots & \vdots \end{pmatrix}.
\end{align}

To achieve $1$-GDP, we need from \Cref{thm:gdp-of-correlated-noise} that
$\stddev \ge 2 \colnorm{\strategy}$. Since the first column of $\strategy$ has the highest norm, we have
\begin{equation} \label{eq:exampleConstraint}
    \colnorm{\strategy}^2 
    = \sum_{t = 0}^{\nd-1} \lambda^{2t} = \frac{1-\lambda^{2n}}{1-\lambda^2}
\end{equation}
and so for any $\nd$ we can achieve $1$-GDP with $\stddev^2 = 4(1 - \lambda^{2\nd})/(1 - \lambda^2)$.
Similarly, the last row of $\bfB$ has the highest norm, so the maximum error from \cref{eq:whycorrelate} is
\begin{align} 
    \max_{t \in [n]} \,\, \mathbb{E}(\widehat s_t - s_t)^2
      &= \stddev^2\br{1 + (n-1)(1-\lambda)^2} \notag \\ 
      &= 4 \br{\frac{1 - \lambda^{2n}}{1-\lambda^2}} \br{1 + (n-1)(1-\lambda)^2} \label{eq:exampleObjective} \\     
      &\le 4 \br{\frac{1}{1-\lambda^2}} \br{1 + (n-1)(1-\lambda)^2} \,.
      \label{eq:exampleObjBound}
\end{align}
Both baselines correspond to sub-optimal choices for $\lambda$ in this loss expression: input perturbation  \cref{eq:baselineIndepZ} corresponds to $\lambda = 0$ while the output perturbation baseline \cref{eq:baselineIndepSum} corresponds to $\lambda \to 1$ and $\stddev^2=\Theta(\nd)$.  

We can do better. One could minimize \cref{eq:exampleObjective} to choose the optimal $\lambda$ for a specific $n$; instead, to obtain an asymptotic result, we consider the upper bound of \cref{eq:exampleObjBound} with $\lambda = 1 - 1/\sqrt{\nd}$. With this choice, direct computation shows
\[
 \max_{t \in [\nd]} \,\, \mathbb{E}(\widehat s_t - s_t)^2
  \le \frac{4(2\nd - 1)}{2\sqrt{\nd} - 1}
  = \Theta(\sqrt{\nd}).
\]
This substantially improves on the $\Theta(\nd)$ error of the baselines, and is in fact the optimal maximum error (up to constants) for the one-parameter family defined in \cref{eq:oneBuffertoeplitzC,eq:oneBuffertoeplitz}.

In fact, as we shall see in \Cref{chap:prefixsum}, it is possible to attain a significantly lower objective value of $\Theta(\poly\ln(n))$ with appropriately-defined correlated noise mechanisms.

This is our first example of a general pattern we will follow regularly: we first design a class of correlated noise mechanisms, generally in terms of some parameterization or structural assumption on $\strategy$ or $\Cinv$, and then show how to find mechanisms in that class that minimize a certain notion of error, subject to a constraint on privacy.

\begin{bremark}[Error Analysis in $\mdim > 1$ Dimensions]
The error analysis for the mechanism defined above directly extends to $\mdim > 1$ dimensions. First, we note from \Cref{thm:gdp-of-correlated-noise} that the noise calibration does not depend on the dimension. Second, the noise is independent across dimensions, so that the error \cref{eq:whycorrelate} simply adds up across dimensions, leading to an error which is $\mdim$ times worse. In \Cref{chap:prefixsum} and later, we will consider the case of $\mdim > 1$ in full generality.
\end{bremark}

\begin{bremark}[Non-Square Factorizations]
\label{remark:nonsquare-factorizations}
We could define a correlated noise mechanism based on a non-square factorization $\workload = \bfB \strategy$ with $\bfB \in \R^{n \times p}$ and $\strategy \in \R^{p \times n}$. The privacy guarantee of \Cref{thm:gdp-of-correlated-noise} holds without any modification. All algorithms and key results, such as the correlated noise DP-SGD (\Cref{alg:dpsgd-corr}) and its privacy guarantee (\Cref{thm:gdp-of-correlated-noise-adaptive}) can be adapted to work with the pseudoinverse $\strategy^\dagger$ instead of $\strategy^{-1}$.
We stick to square and invertible $\bfB, \strategy$ for ease of presentation.
\end{bremark}

\begin{bremark}[Lower Triangular Factorizations]
\label{remark:lower-triangular-factorizations}
The factors $\bfB$ and $\strategy$ are both lower-triangular in the above example. At times, it may be computationally convenient to work with non-lower-triangular matrices, particularly when considering implementing sampling from the stream $\strategy^{-1}\Znoise$.
Conveniently, we can always get a lower triangular factorization from any factorization without altering either the error (as defined above) or the privacy properties of the mechanism. This can be achieved, for example,
by taking the LQ decomposition of $\bfB = \bfL \bfQ$ with a lower-triangular matrix $\bfL$ and an orthonormal matrix $\bfQ$. We then set $\bfB' =\bfL$ and $\strategy' = \bfL^{-1} \workload$. It follows from the rotational invariance of Gaussian distribution that we get the same error bound and privacy properties whether we use the factors $\bfB', \strategy'$ or $\bfB, \strategy$. Thus, we will restrict ourselves to lower triangular factorizations for the rest of the monograph. 
\end{bremark}

\begin{figure}[tp]
    \centering
    \adjustbox{max width=0.8\linewidth}{\input{noise_addition_independent}}
    \vspace{3em}
    
    \adjustbox{max width=0.8\linewidth}{\input{noise_addition_correlated}}
    \caption{An illustration of the noise accumulation in standard DP-SGD (with independent noise), which corresponds to $\Cinv = \bfI_{n \times n}$. On the other hand, DP-SGD with correlated noise (with $\Cinv\neq \bfI_{n \times n}$) can cancel a part of the injected noise due to \emph{anti}-correlations. In the lower plot, the inverse of the strategy matrix is given by 
    $\strategy^{-1} = \begin{psmallmatrix} 1 & 0 & 0 \\ -c_{10}' & 1 & 0 \\ -c_{20}' & -c_{21}' & 1 \end{psmallmatrix}$.}
    \label{fig:noise_cancellation}
\end{figure}
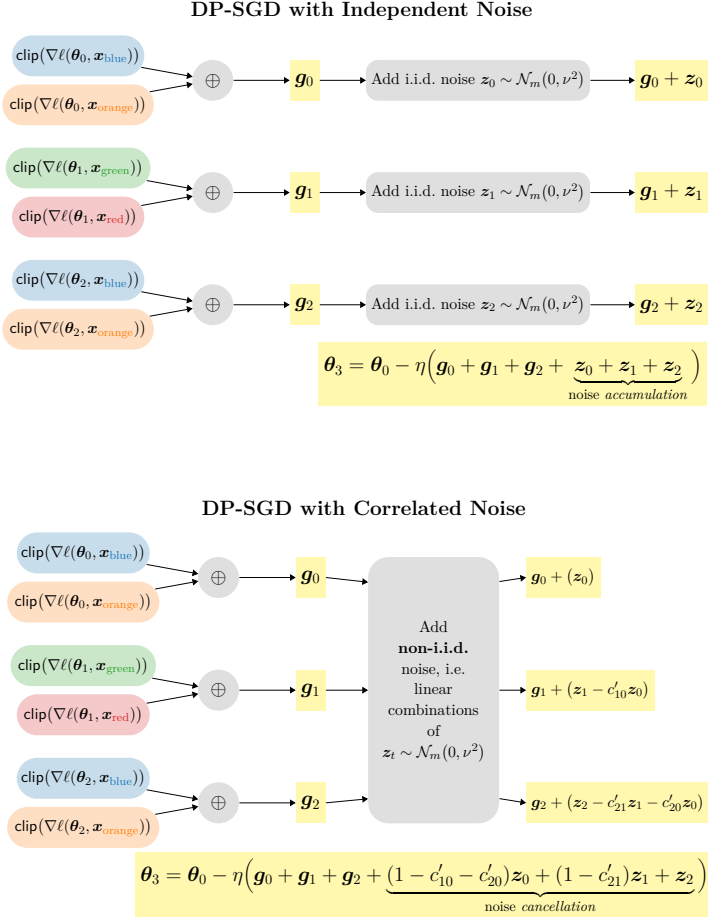

\subsection{A Noise-Cancellation View of Correlated Noise Mechanisms}
\label{sec:noisecancel}

To gain intuition for how correlated noise can improve the estimation error, we can give a noise cancellation view for a mechanism define by  matrix $\bfC$; this interpretation applies to any $\bfC$, but for concreteness we focus on the one-parameter family defined in \cref{eq:oneBuffertoeplitzC}. We consider the resulting private estimates $\widehat \bfG = \bfG + \bfC^{-1} \Znoise$ of the data $\bfG$, where $\Znoise \sim \normalnm{0}{ \stddev^2}$. In this case, $\bfC^{-1}$ is as in \cref{eq:cinv:one-param-family}. 

This directly gives us that 
\begin{equation}\label{eq:noisygrads}
\begin{pmatrix}
\dpgradient_0 \\ \dpgradient_1 \\  \vdots \\ \dpgradient_{\nd -1}
\end{pmatrix} = \begin{pmatrix}
 \gradient_0 \\  \gradient_1 \\  \vdots \\  \gradient_{\nd-1}
\end{pmatrix} + 
     \begin{pmatrix}
    1 & 0 &  \ldots & 0 \\
    -\lambda & 1  & \ldots & 0 \\
    \vdots & \ddots  & \ddots & \vdots \\
    0 & \cdots & \cdots & 1
    \end{pmatrix} \cdot 
    \begin{pmatrix}
    \znoise_0 \\ \znoise_1 \\  \vdots \\ \znoise_{\nd-1}
    \end{pmatrix}, 
\end{equation}
which can be rewritten as 
\[
\dpgradient_t = \begin{cases}
\gradient_1 + \znoise_1 & t= 0 \\
\gradient_t + \znoise_t - \lambda \znoise_{t-1} & t \ge 1 \,.
\end{cases}
\]

In the context of private stochastic gradient descent (\Cref{sec:weightedPrefixSum}), $\gradient_t$ is the gradient obtained in time step $t$. Then, the privatized gradient $\dpgradient_t$ can be interpreted as \emph{canceling out} a $\lambda$-fraction of the noise $\znoise_{t-1}$ injected in the previous time step; see \Cref{fig:noise_cancellation} for an illustration for general $\strategy$ matrices.

In general, a large $\lambda$ leads to better estimation at a fixed variance $\stddev^2$ (of each component of the noise $\znoise$) as more noise is canceled out. However, this leads to a larger $\colnorm{\bfC}$,
implying an increased privacy cost (\Cref{thm:gdp-of-correlated-noise}). Thus, to guarantee a given level of privacy such as 1-GDP, this in turn requires a larger variance $\stddev^2$.
In general, the correction applied via the correlated noise needs to be balanced with the privacy constraint to obtain improved privacy-utility tradeoffs over the baselines.

As we will see in the rest of this monograph, this is generally possible in a range of prefix sum estimation and machine learning tasks. 
In particular, we study details of how to optimally choose the noise correlations for various objectives and under various constraints on the correlation structure such that we obtain efficient implementations.

\section{The Design Space of Correlated Noise Mechanisms}
\label{sec:designSpaceCorrelatedNoise}

The design space of correlated noise mechanisms for gradient-based learning algorithms involves a complex interplay of several factors, requiring careful consideration of the workload, desired privacy guarantees, acceptable error levels, and computational constraints. We give a brief description of each of these design factors and describe how \Cref{chap:prefixsum,chap:ml,chap:practical} disentangle these factors, culminating in a set of practical recommendations by the end of the monograph.

A practitioner interested in employing correlated noise mechanisms in their model training tasks must make five key design decisions shown in \Cref{fig:outline}.

\begin{enumerate}
    \item {\bf Workload.} The starting point is the workload matrix $\workload$ to factorize. When implementing DP-SGD with correlated noise (\Cref{alg:dpsgd-corr}), we only need to specify the noise correlating matrix $\Cinv$; the privacy calibration depends on the matrix $\strategy$ alone (\Cref{thm:gdp-of-correlated-noise-adaptive}). On the other hand, the $\workload$ play a key role (via $\bfB = \workload\Cinv$) in the surrogate objective (e.g. \Cref{eq:whycorrelate}) we optimize to find the $\strategy$ matrix.

The viewpoint of \cref{sec:weightedPrefixSum} suggests a heuristic to select the workload matrix implied by the base (non-private) optimizer. For example, vanilla SGD corresponds to the prefix sum workload $\workload = \prefix$ (\Cref{eq:all-ones-workload}). We will use this a canonical example throughout this monograph. All key considerations we discuss generalize directly to other first-order optimization algorithms whose iterates can be obtained as linear combinations of gradients. This class includes momentum variants such as heavy ball and Nesterov momentum---we briefly discuss these other workloads in \Cref{chap:ml}.

    \item {\bf Error Metric.}
The practitioner also determines the error metric as a proxy to measure the utility of the correlated noise mechanisms. 
The error metric and the sensitivity (which depends on the participation pattern) together determine the objective we optimize to find the factorization $\workload=\bfB\strategy$. For instance, the objective \cref{eq:exampleObjective}  in the example of \Cref{sec:whyCorrelatedNoise} is a product of the sensitivity term with the error term; its general form is as shown in \Cref{fig:outline}.

We use the worst-case expected error in prefix sums (as in \cref{eq:whycorrelate} of \Cref{sec:whyCorrelatedNoise}) as a default error metric for concreteness. We briefly discuss the use of the average expected error across prefix sums towards the end of \Cref{chap:prefixsum} (and return to it in \Cref{chap:practical}). \Cref{chap:ml} also considers using the learning objectives as the utility measures for simple problems such as mean estimation and linear regression.

\begin{figure}[p]
    \centering
    \adjustbox{max width=0.99\linewidth}{\input{outline}}
    \caption{An outline of the topics defined and covered in each of the sections of this monograph.  }
    \label{fig:outline}
\end{figure}
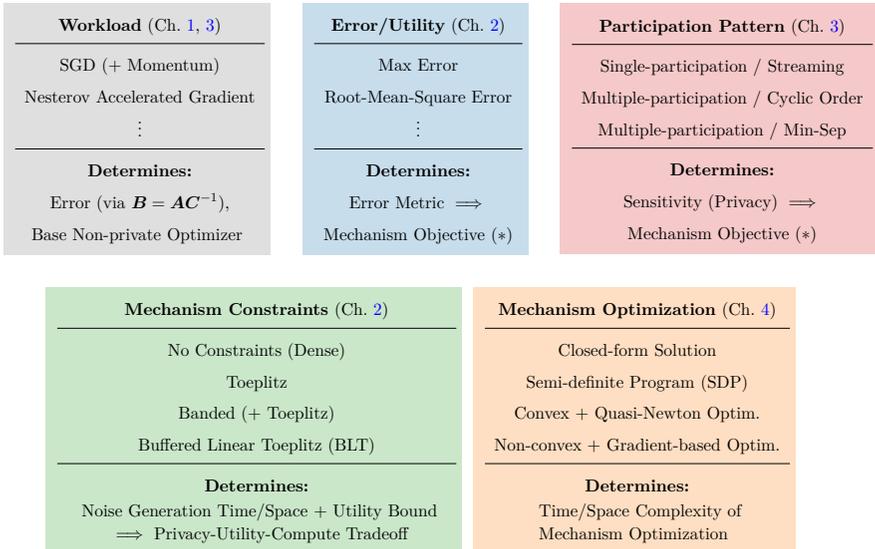

\item {\bf Participation Patterns.}
The third factor is to determine the \emph{participation pattern} of datapoints: 
\begin{quote}
    Do we process each datapoint only once (i.e., the streaming setting of \Cref{sec:weightedPrefixSum}) or do we make multiple passes over the data? In the latter case, are there any restrictions on the order in which the datapoints are processed (e.g. cyclic order)? 
\end{quote} 

As illustrated in \Cref{fig:adjacency-intro}, this determines the effect of changing one datapoint (in an adjacent dataset) on the output of the mechanism. In other words, the participation pattern is key to quantifying the sensitivity, and hence, the privacy guarantee, of the correlated noise mechanism.

The streaming setting is straightforward to handle, as we saw in \Cref{thm:gdp-of-correlated-noise}. Expanding upon the illustration of \Cref{fig:adjacency-intro}, \Cref{chap:ml} explains the challenges and nuances of tightly bounding the sensitivity in different multiple-participation scenarios, as well as its efficient computation in \Cref{chap:ml}.

\item {\bf Mechanism Constraints.}
The practitioner also decides the class of factorizations $\workload=\bfB\strategy$ to optimize over. This design decision can lead to different privacy-utility-compute tradeoffs of the correlated noise mechanism. For example, generating the correlated noise $(\strategy^{-1}\Znoise)[t, :]$ (or equivalently, $(\bfB\Znoise)[t, :]$) in step $t$ can take $O(\nd \mdim)$ time per-step, and simply storing the correlating noise matrix $\strategy \in \R^{\nd \times \nd}$  would require $O(\nd^2)$ space. This is impractical for long training runs (i.e., with $\nd$ large, especially for larger parameter model where $\mdim$ can range in billions). 
We explore families of structured matrices that allow reduced space and time complexity of this noise generation in \Cref{chap:prefixsum}. However, if the class of matrices is too restrictive, it can have sub-optimal utility.

We will see in \Cref{chap:prefixsum} that it is possible to attain favourable privacy-utility-compute tradeoffs with carefully designed families of structured matrices. We then adapt these mechanisms to the multiple-participation setting in \Cref{chap:ml}.

\item {\bf Mechanism Optimization.}
 Finally, the practitioner has to determine the optimization algorithm to find the factorization $\workload=\bfB\strategy$ over the determined set of structured matrices to minimize the objective (which is the product of the error metric and sensitivity); see \Cref{fig:outline}. This is a one-time cost, provided the obtained factors $\bfB, \strategy$ can be cached. \Cref{chap:practical} discusses the approaches to perform this optimization and approximations required to make to practical and scalable.
\end{enumerate}

\section{The Privacy Unit: Example-level vs. User-level DP} 
\label{sec:ch1-priv-unit}

As we discussed at the end of \Cref{sec:introToDP}, the definition of differential privacy can be instantiated at different granularity by specifying the unit of change between adjacent $D$ and $D'$. This includes 
\begin{enumerate}
    \item [(i)] \textbf{example-level DP}, where $D$ and $D'$ differ by one example (as specified by the adjacency notion); and 
    \item [(ii)] \textbf{user-level DP} refers to the setting where $D$ and $D'$ differ by all examples derived from a single user or entity. 
\end{enumerate}

In practice, the privacy unit must be selected based on the problem at hand. For example, if multiple examples derived from a single user or entity might have common features or attributes, user-level DP can align more closely with the intuitive notion of individual privacy protection. This is generally the case for AI models that are trained or fine-tuned directly on user-written emails, documents, or text messages. In this case, example-level DP may fail to adequately mitigate the risk of information leakage due to the inherent inter-dependencies among a user's examples.

To a large extent, the correlated noise mechanisms we develop are  agnostic to the privacy unit. For the sake of clarity and simplicity in exposition, we default to example-level DP in this monograph. Notably, the algorithmic modifications made to adapt independent noise DP-SGD from example-level to user-level DP are largely applicable to correlated noise mechanisms.

We can adapt \Cref{alg:dpsgd-1,alg:dpsgd-corr} to user-level DP as follows. It is convenient to describe the dataset $D$ as a collection of user datasets: 
\[
    D = \{D_u \, :\, u \in [N]\}\,,
\] 
where $D_u = \{\bfx_{u, 0}, \bfx_{u, 1}, \ldots\}$ is the dataset of user $u \in [N]$.
We define the loss on one user's data as the average loss over
\[
    \ell(\bftheta, D_u) := \frac{1}{|D_u|} \sum_{\bfx \in D_u}  \ell(\bftheta, \bfx) \,.
\]
Finally, in each iteration $t$, we sample/select a user $u_t$ and calculate the clipped user gradient:
\[
    \gradient_t = \clip_\clipnorm\big(
        \nabla \ell(\bftheta_t, D_{u_t})
    \big) \,.
\]

For the purposes of privacy analysis, this $\gradient_t$ can actually be \emph{any clipped function} of the user data such as a stochastic user gradient $\nabla \ell(\bftheta_t, \bfx_{u_t})$ for some arbitrary $\bfx_{u_t} \in D_{u_t}$. Another popular choice in the context of federated learning is a clipped \emph{pseudo-gradient}. 
This is the delta of $k$ stochastic gradient steps on $D_{u_t}$: 
\begin{align} \label{eq:dp-fedavg-update}
\begin{aligned}
    \gradient_t &= \clip_\clipnorm\left(
    \frac{1}{\eta}(\bftheta_t - \bftheta_{t, k}^{(u)}) \right)\,, \quad \text{where} \\
    \bftheta_{t, l+1}^{(u)} &= \bftheta_{t, l}^{(u)} - \eta \nabla \ell(\bftheta_{t, l}^{(u)}, \bfx_{u, l}) \text{ for } l=0,\ldots, k-1 \,,
\end{aligned}
\end{align}
and $\bfx_{u, l}$ sampled/selected from $D_u$ and the stochastic gradient steps are initialized at $\bftheta_{t, 0}^{(u)} = \bftheta_t$.

Importantly, the streaming assumption in user-level DP translates to each user appearing only once during training. If any user's data is used more than once (even if each example might be processed only once), the streaming assumption is violated for the purpose of user-level DP. In such a case, we have to use the multiple-participation techniques of \Cref{chap:ml}.

\section{Other Notions of Adjacency*}
\label{sec:intro:different-adj}

While we use the replace-one adjacency (Definition~\ref{def:replaceone}) in this section, in practical deployment and literature, two other preferences are the \textit{zero-out} adjacency and \textit{add-or-remove} adjacency. 

\subsubsection{Zero-out Adjacency}
\label{sec:ch1-zero-out}
The zero-out adjacency is usually used when working with algorithms operating on example gradients. In this setting, the zero-out adjacency says $D \simeq D'$ if $D$ and $D'$ are the same, except one example in $D$ is replaced with a special example in $D'$ whose gradients are zero everywhere or vice-versa:

\begin{bdefinition}\label{def:zeroout}
We say two datasets $D = \{\bfx_1, \ldots, \bfx_{\datasize}\}$ and $D' = \{\bfx_1', \ldots, \bfx_{\datasize}'\}$ are \textbf{zero-out adjacent} (denoted $D \simeq D'$) if $\bfx_i = \bfx_i$ for all $i \in [n] \setminus \{j\}$ for some index $j \in [n]$ and $\bfx_j = \perp$ or $\bfx_j' = \perp$, where $\perp$ is a special \emph{null} element such that $\nabla \ell(\bftheta, \perp) = \zeros$ for all $\bftheta$.
\end{bdefinition}

\paragraph{Replace-one vs. Zero-out Adjacency}
To compare replace-one and zero-out, consider $f(D) = \sum_{\bfx \in D} \nabla \ell(\bftheta, \bfx)$, the model gradient summed over a dataset $D$, where the parameter $\bftheta$ is 1-dimensional and the loss $\bftheta \mapsto \ell(\bftheta, \bfx)$ is $1$-Lipschitz, i.e., the per-example gradients are bounded as $|\nabla \ell(\bftheta, \bfx)| \le 1$.

Then, we have that $\sens(f) = 2$ under the replace-one adjacency, because when replacing an example, a gradient can switch from e.g. $1$ to $-1$. Under the zero-out adjacency, however, we have $\sens(f) = 1$, since the replaced gradient can switch from $1$ to $0$, but not to $-1$. A consequence of the doubled sensitivity is that an algorithm that is $\mu$-GDP with respect to the replace-one adjacency is $\mu/2$-GDP with respect to the zero-out adjacency and vice-versa, giving us an easy way to directly compare the two definitions. 

In particular, we have the following analogue to \Cref{lem:sensbound}:
\begin{blemma}\label[lemma]{lem:sensbound-2}
Under zero-out adjacency (\cref{def:zeroout}) with gradients clipped to norm 1 (cf. \cref{eq:ch1-grad-norm-bound}), we have
$\sens(\strategy) = \colnorm{\strategy}$.
\end{blemma}
\begin{proof}
Same as the proof of \Cref{lem:sensbound}, except that $\norm{\bfdelta}_2 \le 1$ due to the different notion of the adjacency.
\end{proof}

\noindent This leads to analogous privacy results in the zero-out case:

\begin{btheorem} \label{thm:gdp-corr-adaptive-zero-out}
    Consider the setting of \Cref{thm:gdp-of-correlated-noise-adaptive}. The gradients and iterates of DP-SGD with correlated noise (\Cref{alg:dpsgd-corr}) satisfy $\frac{1}{\sigma}$-GDP in the zero-out-example adjacency in the streaming setting if we take $\stddev = \nmult \, \colnorm{\strategy}$.
\end{btheorem}

\subsubsection{Advantages of Zero-out Adjacency and Add-or-Remove Adjacency}
Since zero-out and replace-one is only a factor of 2 off from each other, one might wonder why zero-out, which is restricted to gradient-like queries, is used instead of replace-one in practice. One reason is that zero-out is more comparable to the popular add-or-remove-one adjacency described below:
\begin{bdefinition}\label{def:add-remove}
Two datasets $D, D'$ are adjacent under the \textbf{add-or-remove adjacency} if $D = D' \cup \{\bfx\}$ for $D' = D \cup \{\bfx\}$ for some example $\bfx$.
\end{bdefinition}

The add-or-remove-one adjacency is popular in part because it is very natural. For instance, it gives privacy guarantees to a user who might want opt their data out of model training. The add-or-remove model is also more natural for analyzing  privacy amplification by Poisson sampling, discussed in \cref{sec:learning:amplification}.  
However, for model training, add-or-remove adjacency is somewhat inconvenient because adjacent datasets have different sizes. In turn, the dataset size itself becomes a private quantity, which greatly complicates privacy analyses. Zero-out adjacency maintains semantic similarity to add-or-remove-one, but maintains that $D$ and $D'$ are the same size and e.g. batches are formed the same way for both datasets, i.e., effectively allows us to publish the dataset size without violating DP guarantees.

For these reasons, we will use the zero-out notion of adjacency later in \Cref{chap:ml}.

\begin{bremark}[Event-Level Privacy]
In the setting of continual counting and in settings such as \Cref{ex:range-queries}, the notion of adjacency usually considered is the {\em event-level privacy}: there is exactly one time step where the neighboring streams differ. 
Thus, in the streaming machine learning setting, event-level privacy of gradients coincides with example-level (or user-level) privacy of their underlying datasets.
\end{bremark}

\section{Other Common Notions of Differential Privacy*}

In this chapter, we discuss two differential privacy (DP) definitions, Gaussian DP and approximate DP, which are more commonly reported in practice. We state a method for converting a Gaussian DP guarantee (or its equivalent description via a noise multiplier for the Gaussian mechanism) to another notion known as zero-concentrated DP. In particular, this conversion can be done by either computing a simple formula, or calling an existing open-source library.

\subsection{Zero-Concentrated DP (zCDP)}

Zero-concentrated differential privacy (zCDP) is defined in terms of the R\'enyi divergence:

\begin{bdefinition}
The \textbf{$\alpha$-R\'enyi divergence} between two distributions $P$ and $Q$ for $\alpha > 1$ is defined as:

\[R_\alpha(P, Q) = \frac{1}{\alpha-1} \log \mathbb{E}_{X \sim Q}\left(\frac{P(X)}{Q(X)}\right)^\alpha.\]
The definition extends to $\alpha \in \{1, \infty\}$ by continuity.
\end{bdefinition}

\begin{bdefinition}
A mechanism $\mech$ is $\rho$-zCDP if for all $D \simeq D'$, it holds that
\[R_\alpha(\mech(D), \mech(D')) \leq \rho \alpha\]
for all $\alpha > 1$.
\end{bdefinition}

We can convert from GDP (or its equivalent noise multiplier description) to zCDP using the following result:

\begin{blemma}
\label{lem:muGDPtozCDP}
Any $\mu$-GDP mechanism is $(\mu^2/2)$-zCDP. In particular,
the Gaussian mechanism with noise multiplier $\sigma$ is $(1/2\sigma^2)$-zCDP.
\end{blemma}

\section{Additional Proofs*}
\label{sec:chap1-proof}
Earlier on page~\pageref{lem:gdp-of-gaussian-mechanism}, we gave a proof sketch of \Cref{lem:gdp-of-gaussian-mechanism} in the sense that it was only for the $1$-dimensional case. However, in the case of correlated noise mechanisms, we would be applying the Gaussian mechanism on high-dimensional vectors.  
We now give a general proof of the GDP bound of the Gaussian mechanism (Lemma~\ref{lem:gdp-of-gaussian-mechanism}) in high-dimensions, which at a high level, uses the isometry of multi-variate Gaussian distribution to reduce the general high-dimensional case to $1$-dimensional case. For readers familiar with the proof of  $(\eps,\delta)$-DP for the Gaussian mechanism, the proof is analogous. 

\begin{proof}[Proof of Lemma~\ref{lem:gdp-of-gaussian-mechanism}]
The proof proceeds similarly to the 1-dimensional case.
Consider a pair of worst-case adjacent datasets $D \simeq D'$ such that $\norm{f(D) - f(D')}_2 = \sens(f)$. If $f(D)$ and $f(D')$ are closer, it only makes the mechanism outputs $\mech(D), \mech(D')$ more indistinguishable. We will exhibit a randomized map $g$ such that $\mech(D) \stackrel{\mathrm{d}}{=} g(Z)$ and $\mech(D) \stackrel{\mathrm{d}}{=} g(Z + \mu)$ for $Z \sim \normal(0, 1)$, as required by \Cref{def:gdp}.

Denote $\mu = 1/\nmult$ and $\bfu$ as the unit vector along $f(D') - f(D)$:
\[
    \bfu := \frac{f(D') - f(D)}{\norm{f(D') - f(D)}_2} = \frac{1}{\sens(f)}\big(f(D') - f(D)\big) \,.
\]

The idea is to construct $g$ such that $\mathbb{E}[g(Z)] = f(D)$ and that $g(Z+\mu)$ will introduce a bias along the unit vector $\bfu$, yielding $f(D')$ in expectation. All other directions are not relevant, as both distributions are essentially Gaussian noise with identical variance.

Concretely, consider the randomized function $g:\R \to \R^m$ given by
\[
    g(s) := f(D) \,+\, \nmult \cdot \sens(f) \cdot \br{ s \bfu + \br{\bfI_{m \times m} - \bfu\bfu\T} \bfxi }
    \,\, \text{for} \,\,
    \bfxi \sim \normalm{0}{1} \,.
\]
If we take $s = Z \sim \calN(0, 1)$, then the second term is just component-wise i.i.d. Gaussian noise with variance $\stddev^2 = \nmult^2 \, \sens(f)^2$, since
\[
Z \bfu + \br{\bfI_{m \times m} - \bfu \bfu\T} \bfxi \stackrel{\mathrm{d}}{=} \normalm{0}{1} \,,
\]
by the rotational invariance of the Gaussian distribution.
Thus, for $\znoise \sim \calN(0, \nmult^2 \, \sens(f)^2)$, we get
$g(Z) \stackrel{\mathrm{d}}{=} f(D) + \znoise  \stackrel{\mathrm{d}}{=} \mech(D)$, as required.
Instead, if we plug in $s = Z+ \mu$, only the mean changes, while the variance remains unchanged:
\begin{align*}
    g(Z + \mu) &= f(D) + \nmult  \cdot \sens(f) \, \left(\mu \bfu\right) + \znoise  \\
    &= f(D) + \sens(f) \cdot \bfu + \znoise 
    = f(D') + \znoise \,,
\end{align*}
since $\mu = 1/\nmult$ and $\sens(f) \cdot \bfu = f(D') - f(D)$.
\end{proof}

\section{Chapter Notes*}
\label{sec:chap1-notes}

We give some additional context and details on some of the topics covered here. 

\subsection{The Workload and Strategy Matrices}

In the context of correlated noise mechanisms, the matrix $\workload\in R^{\nd \times \nd}$ is called the \emph{workload matrix} or \emph{query matrix}. Similarly, in the factorization $\workload=\bfB\strategy$, the matrix $\strategy$ is referred to as the \emph{strategy matrix}. This terminology originates from the database literature, where early research on correlated noise mechanisms (detailed in \Cref{sec:ch1-biblio}) focused on counting records (or bins) satisfying a given predicate in a database (or histogram). A concrete example is provided in the upcoming \Cref{ex:range-queries}.

We map this problem to our notation.
Let $\Gradients \in \R^{n \times 1}$ represent the input database with $n$ records (here, the dimension is $m=1$). Each counting query to the database can be denoted by a vector $\bfa \in \{0, 1\}^{n}$, where $\bfa[t] = 1$ indicates the records to be counted. The result of this query is simply $\bfa\T \bfG$, the dot product of $\bfa$ and the database $\bfG$.
Early work on correlated noise mechanisms considered answering a batch of queries denoted by vectors $\bfa_0, \bfa_1, \ldots, \bfa_{N-1}$, which can be arranged row-wise into a matrix $\workload \in \R^{N \times n}$. The goal is to estimate $\workload \Gradients$ under differential privacy (we take $N=\nd$ in our setting). A key motivation of the early work was to show how answering all the queries collectively---with appropriate noise correlation---yields better private estimates than answering them individually (which corresponds to the output perturbation mechanism).

In databases, the term ``workload'' refers to a set of database requests or queries that share common characteristics (e.g., on the same table or histogram), allowing for the application of workload management controls to optimize system performance. Since the $\workload$ matrix denotes a collection of the queries $\bfa_0, \ldots, \bfa_{N-1}$ on the same input $\Gradients$, it is naturally called the \emph{workload matrix}.

Similarly, database management systems optimize query execution by grouping and refactoring queries to avoid redundant work. In correlated noise mechanisms, the matrix $\strategy$ captures a \emph{strategy} for performing some shared work $\strategy \Gradients$ on the input $\Gradients$ before adding noise $\Znoise \sim \normalnm{0}{\nu^2}$ for DP. Thus, it is analogously referred to as the \emph{strategy matrix}.

\subsection{DP-FTRL: DP-SGD with Correlated Noise}
\label{sec:dpftrl-name-background}

DP-SGD with correlated noise, as given in \Cref{alg:dpsgd-corr} is referred to in the literature as \emph{DP-FTRL}. This stands for a differentially private version of the \textit{follow-the-regularized-leader} (FTRL) algorithm. Given a sequence of past gradients $\gradient_0, \ldots, \gradient_{t-1}$, the FTRL algorithm (also known as \textit{dual averaging} or \textit{lazy mirror descent}) sets the next iterate $\bftheta_{t}$ as
\[
    \bftheta_{t} := \argmin_{\bftheta \in \Theta} \left\{ \bfs_t\T \bftheta + \frac{1}{2\eta} \norm{\bftheta -\bftheta_0}_2^2 \right\}\,, \quad \text{where} \quad \bfs_t = \sum_{\tau=0}^{t-1} \gradient_\tau
\]
is the prefix sum of gradients,  $\bftheta_0$ is the first iterate, and $\Theta$ is the constraint set. For an unconstrained optimization problem with $\Theta=\R^\mdim$, these iterates coincide exactly with the SGD iterates in \cref{eq:sgd-iterates-unrolled}.
FTRL differs from SGD when $\Theta$ is a strict subset of $\R^\mdim$ or when other penalties are used in the place of the squared Euclidean norm $\norm{\bftheta -\bftheta_0}_2^2$.

Historically, the name DP-FTRL has been used to emphasize that correlated noise mechanisms are used to privately estimate the prefix sums $(\bfs_t)_{t=0}^{n-1}$ (as opposed to independent noise DP-SGD with $\strategy=\bfI_{n\times n}$).
Throughout this monograph, we stick with the nomenclature \emph{DP-SGD with correlated noise}.

\subsection{More Examples of Weighted Prefix Sums}
\label{sec:examplesWeightedPrefixSums}

Weighted prefix sums are ubiquitous in data science, in addition to the setting of gradient descent considered in \Cref{sec:weightedPrefixSum}.
We give a few more examples here.

 \begin{bexample}[Online $k$-means Clustering]
  
  In this example, given input data that consists of a stream of  points $P$ in $\R^d$,  the goal is to output after the arrival of a point  a set $S$ of $k$ points (called the \emph{centers}) in $\R^d$ in order to minimize the following cost function:
$$\cost(P, S) = \sum_{p\in P} \min_{s \in S} \dist(p, s)^2,$$
where $\dist(p,s)$ is the Euclidean distance between $p$ and point $s$ while remaining differentially private with respect to the stream of datapoints. There is a reduction of this problem to a finite collection prefix sum problem, where each problem is $d$-dimensional.
\end{bexample}

\begin{bexample}[Range Queries and CDF Estimation]
 \label{ex:range-queries}
In this problem, we are given $n$ data points, $D=(x_1, \ldots, x_n)\in [0,C]^n$ for some fixed $C \in \mathbb R$. The goal is to construct a data structure so that, given a query range $[a,b] \in [0,C]$, the output is an estimate of 
\[
R_{[a,b]}=\left| \{ x_i \in D \text{ such that } a \leq x_i \leq b\} \right|.
\]
A notable application of range queries in the estimation of the cumulative distribution function (CDF) $F(b):= \mathbb{P}(X \le b)$ of a random variable $X$. In particular, if the data points $x_1, \ldots, x_n$ are i.i.d. samples of the random variable $X$, then the empirical CDF estimator $\widehat F(b) = \frac{R_{[0, b]}}{n}$ can be computed via a range query.
In general, range queries are relevant in applications ranging from online analytics such as financial markets or sensor networks to contact tracing, mobility analysis and urban planning.

One can reduce range queries to prefix sum computation, allowing us to leverage correlated noise mechanisms for better differentially private estimators. In particular, we divide the domain $[0,C]$ into disjoint buckets, each with size equal to the allowed accuracy. Then, each bucket stores the number of points that are in the range covered by the bucket. 
To estimate the value of $R_{[a,b]}$ for a given $0\leq a \leq b \leq C$, we use the prefix sum until the bucket that contains $a$ and the prefix sum until the bucket that contains $b$. The difference of the two gives the estimate of $R_{[a,b]}$.
 \end{bexample}

\section{Bibliographic Notes}
\label{sec:ch1-biblio}

\paragraph{Differential Privacy}
Differential privacy was introduced in the celebrated work of \citet*{dwork2016calibrating}. The Gaussian mechanism was introduced subsequently by \citet*{dwork06our} in which they also introduced the relaxation of differential privacy, now commonly known as {\em approximate differential privacy}.  Gaussian differential privacy was introduced by \citet*{dong22gaussian} and zero-concentrated differential privacy by \citet*{bun16concentrated} and \citet*{dwork2016concentrated}. Lemma~\ref{lem:muGDPtozCDP} was shown as Proposition 1.6 in  \citet*{bun16concentrated}.
The proof of \Cref{thm:gdp-adaptivity} is due to \citet*{denisov2022improved}.

\paragraph{Gradient Descent with DP}
Differentially private gradient descent was first introduced by \citet*{song2013stochastic}. \citet*{bassily2014private} showed that it achieves optimal empirical risk guarantee and \citet*{abadi2016deep} developed a better privacy accounting method (via the ``moments accountant'') using the Gaussian mechanism, making differentially private SGD (DP-SGD) practically usable on deep nets.

\paragraph{Correlated Noise Mechanisms for Learning}
The use of correlated noise for differentially private learning was first studied in the context of online learning by \citet*{jain2012differentially}, and was later by \citet*{guha2013nearly} who considered {\em follow-the-approximate leader}. The variant considered in this monograph was first defined in  \citet*{kairouz2021practical} and later improved in a series of works by \citet*{denisov2022improved,choquette2022multi,choquette2023correlated}. In particular, \citet*{denisov2022improved} proved \Cref{thm:gdp-adaptivity} and these works framed correlated noise mechanism as deriving a differentially private version of the popular follow-the-regularized leader (FTRL) algorithm, also known as dual averaging. This family of algorithms was, in turn, developed through a series of works by \citet{nesterov2009primal}, \citet*{xiao2009dual}, \citet*{mcmahan2011follow} and  \citet*{duchi2011dual}.

\paragraph{Correlated Noise Mechanisms for Prefix Sum Estimation}
Computing prefix sum under the constraints of differential privacy was first studied by \citet*{dwork2010continual} and \citet*{chan2011private}. These works used a particular hand-crafted correlations, now known as the \textit{binary tree mechanism}; we discuss this further in \Cref{sec:treeAgg}. The general framework of correlated noise mechanisms we defined in \Cref{sec:correlatedNoiseIntro}
(i.e., using a factorization $\workload = \bfB \strategy$) was first introduced concurrently by \citet*{li2015matrix} under the name {\em matrix mechanism} and by \citet*{nikolov2016geometry} as the {\em factorization mechanism}.  

The use of the matrix mechanism for prefix sum estimation was first observed independently and concurrently by \citet*{denisov2022improved} and \citet*{fichtenberger2023constant}. The weighted version of the prefix sum was first studied by \citet*{bolot2013private}, and recently improved by \citet*{henzinger2023almost,henzinger2025improved}. 

\paragraph{Large-Scale Practical Deployments of DP}
The United States Census Bureau used DP for the 2020 Census to provide demographic insights~\cite{uscensus}. Google used DP in the Chrome browser to analyze user behavior~\cite{erlingsson2014rappor}, while Apple used DP in iOS and MacOS~\cite{appledp}.
In particular, correlated noise mechanisms have been industrially deployed to train next-word-prediction models for mobile keyboards by Google~\citep*{xu2023federated}. For best practices involving DP in machine learning, see 
\citet*[Sec 5.3.3]{ponomareva2023dp}.

\paragraph{Adjacency Notions and Privacy Units}
For a detailed discussion of adjacency notions and privacy unit, we refer to \citet*{ponomareva2023dp}. In particular,  we refer to their Section 2.1 and 5.1 for how to select a privacy unit.
A discussion on the pros and cons of add-or-remove-one adjacency vis-à-vis replace-one adjacency in the context of privacy amplification by sampling can be found in \citep[Sec. 6]{steinke2022compositiondifferentialprivacy}. 
The resulting user-level DP learning algorithm obtained by modifying independent noise DP-SGD (\Cref{alg:dpsgd-1}) with \cref{eq:dp-fedavg-update} is also known as ``DP-FedAvg''. This algorithm was proposed by \citet*{mcmahan2017learning} to make the standard federated averaging algorithm satisfy user-level DP guarantees.
In fact, much of the prior literature on correlated noise mechanisms in the learning setting, starting from \citet*{kairouz2021practical}, was motivated by user-level DP in the federated learning context; see \Cref{ex:fl} in \Cref{chap:ml} for more background context.

We note that it is also possible to promote example-level DP guarantees to user-level DP guarantees by using generic group privacy reductions~\cite[Lemma 2.2]{vadhan2017complexity}. While such reductions are usually not as tight as the approach presented in \Cref{sec:ch1-priv-unit}, it is possible to develop tighter user-level privacy accounting techniques for the example-level DP algorithm of DP-SGD with independent noise~\citep*{charles2024fine}.

\paragraph{Other Remarks}
The reduction of online $k$-means clustering to a finite collection of prefix sum problem was shown in \citet*{Dupre2024}.

%% file: gdp.tex
\begin{adjustbox}{width=0.45\linewidth}
\begin{tikzpicture}[xscale=0.5,yscale=4]

\def\normalP{\x, {
    0.39894 * exp( -((\x-0.)^2)/ 2
    }}
\def\normalQ{\x, {
    0.39894 * exp( -((\x-1.)^2)/ 2
    }}
\def\normalR{\x, {
    0.39894 * exp( -((\x-3)^2)/ 2
    }}
    
 \draw[color=C0, preaction={fill=C0!20, fill opacity=0.3}, fill opacity=0.2, pattern color=C0, domain=-3:5,smooth, thick] (-3, 0) -- plot (\normalP) -- (5, 0);
 
\draw[color=C1, preaction={fill=C1!20, fill opacity=0.3}, fill opacity=0.2, pattern color=C1, domain=-3:5,smooth, thick] (-3, 0) -- plot (\normalQ) -- (5, 0);

\draw[<->] (-3.5,0) -- (6.5,0) {};
\draw[->] (0,0) -- (0,0.5) {};

\node at (-1.2, 0.43) {\scriptsize $\mathcal{N}(0, 1)$};

\node at (2.2, 0.43) {\scriptsize $\mathcal{N}(1, 1)$};

\node at (1.5, 0.65) {\small 
\underline{\textbf{1-GDP (higher privacy)}}
};

\end{tikzpicture}
\end{adjustbox}
%
\hfill
\begin{adjustbox}{width=0.45\linewidth}
\begin{tikzpicture}[xscale=0.5,yscale=4]

\def\normalP{\x, {
    0.39894 * exp( -((\x-0.)^2)/ 2
    }}
\def\normalQ{\x, {
    0.39894 * exp( -((\x-1.)^2)/ 2
    }}
\def\normalR{\x, {
    0.39894 * exp( -((\x-3)^2)/ 2
    }}
    
 \draw[color=C0, preaction={fill=C0!20, fill opacity=0.3}, fill opacity=0.2, pattern color=C0, domain=-3:4,smooth, thick] (-3, 0) -- plot (\normalP) -- (4, 0);
 
\draw[color=C1, preaction={fill=C1!20, fill opacity=0.3}, fill opacity=0.2, pattern color=C1, domain=-1:6,smooth, thick] (-1, 0) -- plot (\normalR) -- (6, 0);

\draw[<->] (-3.5,0) -- (6.5,0) {};
\draw[->] (0,0) -- (0,0.5) {};

\node at (-1.2, 0.43) {\scriptsize $\mathcal{N}(0, 1)$};

\node at (4.2, 0.43) {\scriptsize $\mathcal{N}(3, 1)$};

\node at (1.5, 0.65) {\small 
\underline{\textbf{3-GDP (lower privacy)}}
};

\end{tikzpicture}
\end{adjustbox}

%% file: adjacency_intro.tex
\begin{tikzpicture}

\draw[white] (-9.5, 2.75) rectangle (6.5, 5.8) ; 

\tikzstyle{boxA}=[draw=black!65,fill=C0!50, pattern=north west lines, pattern color=C0!15, preaction={fill=C0!50}]
\tikzstyle{boxB}=[draw=black!65,fill=C1!40]
\tikzstyle{boxC}=[draw=black!65,fill=C2!60, pattern=dots, pattern color=C2!15, preaction={fill=C2!60}]
\tikzstyle{boxD}=[draw=black!65,fill=C3!30]
\tikzstyle{boxE}=[draw=black!65,fill=C5!40]
\tikzstyle{boxF}=[draw=black!65,fill=keynoteBlue!30]
\tikzstyle{boxG}=[draw=black!65,fill=gray!30]

\draw[densely dotted] (-1.5, 1.9) -- (-1.5, 6.4) ;

\node at (-8, 2.4) {\large $\Gradients \in \mathbb{R}^{5 \times m}$};

\node at (-4, 2.4) {\large $\Gradients' \in \mathbb{R}^{5 \times m}$};

\filldraw[boxB] (-9, 5) rectangle ++(2, 0.5) node[pos=0.5] {\large $\gradient_0$} ;
\filldraw[boxA] (-9, 4.5) rectangle ++(2, 0.5) node[pos=0.5] {\large $\gradient_1$} ;
\filldraw[boxD] (-9, 4) rectangle ++(2, 0.5) node[pos=0.5] {\large $\gradient_2$} ;
\filldraw[boxF] (-9, 3.5) rectangle ++(2, 0.5) node[pos=0.5] {\large $\gradient_3$} ;
\filldraw[boxE] (-9, 3) rectangle ++(2, 0.5) node[pos=0.5] {\large $\gradient_4$} ;

\node at (-6, 4.4) {${\simeq}$} ;

\node at (-6, 6) {\large \textbf{\underline{Streaming Setting}}} ;

\filldraw[boxB] (-5, 5) rectangle ++(2, 0.5) node[pos=0.5] {\large $\gradient_0' = \gradient_0$} ;
\filldraw[boxC] (-5, 4.5) rectangle ++(2, 0.5) node[pos=0.5] {\large $\gradient_1'$} ;
\filldraw[boxD] (-5, 4) rectangle ++(2, 0.5) node[pos=0.5] {\large $\gradient_2' = \gradient_2$} ;
\filldraw[boxF] (-5, 3.5) rectangle ++(2, 0.5) node[pos=0.5] {\large $\gradient_3' = \gradient_3$} ;
\filldraw[boxE] (-5, 3) rectangle ++(2, 0.5) node[pos=0.5] {\large $\gradient_4' = \gradient_4$} ;

\node at (1, 2.4) {\large $\Gradients \in \mathbb{R}^{5 \times m}$};

\node at (5, 2.4) {\large $\Gradients' \in \mathbb{R}^{5 \times m}$};

\filldraw[boxB] (0, 5) rectangle ++(2, 0.5) node[pos=0.5] {\large $\gradient_0$} ;
\filldraw[boxA] (0, 4.5) rectangle ++(2, 0.5) node[pos=0.5] {\large $\gradient_1$} ;
\filldraw[boxD] (0, 4) rectangle ++(2, 0.5) node[pos=0.5] {\large $\gradient_2$} ;
\filldraw[boxA] (0, 3.5) rectangle ++(2, 0.5) node[pos=0.5] {\large $\gradient_3$} ;
\filldraw[boxB] (0, 3) rectangle ++(2, 0.5) node[pos=0.5] {\large $\gradient_4$} ;

\node at (3, 4.4) {${\simeq}$} ;

\node at (3, 6) {\large \textbf{\underline{Multiple Participation Setting}}} ;

\filldraw[boxB] (4, 5) rectangle ++(2, 0.5) node[pos=0.5] {\large $\gradient_0' = \gradient_0$} ;
\filldraw[boxC] (4, 4.5) rectangle ++(2, 0.5) node[pos=0.5] {\large $\gradient_1'$} ;
\filldraw[boxD] (4, 4) rectangle ++(2, 0.5) node[pos=0.5] {\large $\gradient_2' = \gradient_2$} ;
\filldraw[boxC] (4, 3.5) rectangle ++(2, 0.5) node[pos=0.5] {\large $\gradient_3'$} ;
\filldraw[boxB] (4, 3) rectangle ++(2, 0.5) node[pos=0.5] {\large $\gradient_4' = \gradient_4$} ;

\end{tikzpicture}

%% file: adjacency_intro_2.tex
\begin{tikzpicture}

\draw[white] (-9.5, 2.75) rectangle (3.5, 5.8) ; 

\tikzstyle{boxA}=[draw=black!65,fill=C0!50, pattern=north west lines, pattern color=C0!15, preaction={fill=C0!50}]
\tikzstyle{boxB}=[draw=black!65,fill=C1!40]
\tikzstyle{boxC}=[draw=black!65,fill=C2!60, pattern=dots, pattern color=C2!15, preaction={fill=C2!60}]
\tikzstyle{boxD}=[draw=black!65,fill=C3!30]
\tikzstyle{boxE}=[draw=black!65,fill=C5!40]
\tikzstyle{boxF}=[draw=black!65,fill=keynoteBlue!30]
\tikzstyle{boxG}=[draw=black!65,fill=gray!30]
\tikzstyle{boxH}=[draw=black!65,fill=yellow!70]

\draw[densely dotted] (-1.5, 1.9) -- (-1.5, 6.4) ;

\node at (-8, 2.4) {\large $\Gradients \in \mathbb{R}^{5 \times m}$};

\node at (-4, 2.4) {\large $\Gradients' \in \mathbb{R}^{5 \times m}$};

\filldraw[boxB] (-9, 5) rectangle ++(2, 0.5) node[pos=0.5] {\large $\gradient_0$} ;
\filldraw[boxA] (-9, 4.5) rectangle ++(2, 0.5) node[pos=0.5] {\large $\gradient_1$} ;
\filldraw[boxD] (-9, 4) rectangle ++(2, 0.5) node[pos=0.5] {\large $\gradient_2$} ;
\filldraw[boxF] (-9, 3.5) rectangle ++(2, 0.5) node[pos=0.5] {\large $\gradient_3$} ;
\filldraw[boxE] (-9, 3) rectangle ++(2, 0.5) node[pos=0.5] {\large $\gradient_4$} ;

\node at (-6, 4.4) {${\simeq}$} ;


\filldraw[boxB] (-5, 5) rectangle ++(2, 0.5) node[pos=0.5] {\large $\gradient_0' = \gradient_0$} ;
\filldraw[boxC] (-5, 4.5) rectangle ++(2, 0.5) node[pos=0.5] {\large $\gradient_1'$} ;
\filldraw[boxD] (-5, 4) rectangle ++(2, 0.5) node[pos=0.5] {\large $\gradient_2' = \gradient_2$} ;
\filldraw[boxF] (-5, 3.5) rectangle ++(2, 0.5) node[pos=0.5] {\large $\gradient_3' = \gradient_3$} ;
\filldraw[boxE] (-5, 3) rectangle ++(2, 0.5) node[pos=0.5] {\large $\gradient_4' = \gradient_4$} ;

\node at (1, 2.4) {\large $\Gradients - \Gradients' \in \mathbb{R}^{5 \times m}$};

\filldraw[boxG] (0, 5) rectangle ++(2, 0.5) node[pos=0.5] {\large $\bfdelta_0 = \zeros$} ;
\filldraw[boxH] (0, 4.5) rectangle ++(2, 0.5) node[pos=0.5] {\large $\bfdelta_1$} ;
\filldraw[boxG] (0, 4) rectangle ++(2, 0.5) node[pos=0.5] {\large $\bfdelta_2 = \zeros$} ;
\filldraw[boxG] (0, 3.5) rectangle ++(2, 0.5) node[pos=0.5] {\large $\bfdelta_3 = \zeros$} ;
\filldraw[boxG] (0, 3) rectangle ++(2, 0.5) node[pos=0.5] {\large $\bfdelta_4 = \zeros$} ;

\end{tikzpicture}

%% file: noise_addition_independent.tex
\begin{tikzpicture}[
node distance = 5mm and 7mm,
box1/.style={rounded rectangle, semithick, minimum size=1cm, minimum width=3.5cm},
box2/.style={rectangle, semithick, dotted, minimum height=1cm},
op1/.style={circle, semithick, minimum size=1cm},
op2/.style={rectangle, semithick, minimum size=1cm, rounded corners=0.3cm},
boxOp/.style = {op1, fill=gray!25},
boxOp2/.style = {op2, fill=gray!25},
boxGrad1/.style = {box1, fill=C0!25},
boxGrad2/.style = {box1, fill=C1!25},
boxGrad3/.style = {box1, fill=C2!25},
boxGrad4/.style = {box1, fill=C3!25},
boxMain/.style = {box2, fill=yellow!40},
arr/.style = {-Triangle},
]

\node[boxMain] (g1) at (5, 5) {\Large $\bfg_0$};
\node[boxOp, left of=g1, xshift=-1.75cm] (plus1) {\Large $\oplus$};
\draw[arr] (plus1.east) -- (g1.west) ;

\node[boxGrad1, left of=plus1, xshift=-2.8cm, yshift=0.6cm] (g11) { $\mathsf{clip}\big(\nabla \ell(\bftheta_0, \bfx_{\text{\textcolor{C0}{blue}}})\big)$};
\node[boxGrad2, left of=plus1, xshift=-2.8cm, yshift=-0.6cm] (g12) { $\mathsf{clip}\big(\nabla \ell(\bftheta_0, \bfx_{\text{\textcolor{C1}{orange}}})\big)$};
\draw[arr] (g11) -- (plus1) ;
\draw[arr] (g12) -- (plus1) ;
\node[boxOp2, right of=g1, xshift=3.75cm] (noise1) {Add i.i.d. noise $\znoise_0 \sim \normalm{0}{\stddev^2}$ };
\draw[arr] (g1) -- (noise1) ;
\node[boxMain, right of=noise1, xshift=4.25cm] (gh1) {\Large $\bfg_0 + \znoise_0$};
\draw[arr] (noise1) -- (gh1);

\node[boxMain, below of=g1, yshift=-2.25cm] (g2) {\Large $\bfg_1$};
\node[boxOp, left of=g2, xshift=-1.75cm] (plus2) {\Large $\oplus$};
\draw[arr] (plus2) -- (g2) ;

\node[boxGrad3, left of=plus2, xshift=-2.8cm, yshift=0.6cm] (g21) { $\mathsf{clip}\big(\nabla \ell(\bftheta_1, \bfx_{\text{\textcolor{C2}{green}}})\big)$};
\node[boxGrad4, left of=plus2, xshift=-2.8cm, yshift=-0.6cm] (g22) { $\mathsf{clip}\big(\nabla \ell(\bftheta_1, \bfx_{\text{\textcolor{C3}{red}}})\big)$};
\draw[arr] (g21) -- (plus2) ;
\draw[arr] (g22) -- (plus2) ;
\node[boxOp2, right of=g2, xshift=3.75cm] (noise2) {Add i.i.d. noise $\znoise_1\sim \normalm{0}{\stddev^2}$};
\draw[arr] (g2) -- (noise2) ;
\node[boxMain, right of=noise2, xshift=4.25cm] (gh2) {\Large $\bfg_1 + \znoise_1$};
\draw[arr] (noise2) -- (gh2);

\node[boxMain, below of=g2, yshift=-2.25cm] (g3) {\Large $\bfg_2$};
\node[boxOp, left of=g3, xshift=-1.75cm] (plus3) {\Large $\oplus$};
\draw[arr] (plus3) -- (g3) ;

\node[boxGrad1, left of=plus3, xshift=-2.8cm, yshift=0.6cm] (g31) { $\mathsf{clip}\big(\nabla \ell(\bftheta_2, \bfx_{\text{\textcolor{C0}{blue}}})\big)$};
\node[boxGrad2, left of=plus3, xshift=-2.8cm, yshift=-0.6cm] (g32) { $\mathsf{clip}\big(\nabla \ell(\bftheta_2, \bfx_{\text{\textcolor{C1}{orange}}})\big)$};
\draw[arr] (g31) -- (plus3) ;
\draw[arr] (g32) -- (plus3) ;
\node[boxOp2, right of=g3, xshift=3.75cm] (noise3) {Add i.i.d. noise $\znoise_2\sim \normalm{0}{\stddev^2}$};
\draw[arr] (g3) -- (noise3) ;
\node[boxMain, right of=noise3, xshift=4.25cm] (gh3) {\Large $\bfg_2 + \znoise_2$};
\draw[arr] (noise3) -- (gh3);

\node[boxMain, below of=gh3, yshift=-1.2cm, xshift=-3.9cm] (result1) {\Large 
$\displaystyle
\bftheta_3 = \bftheta_0 - \eta 
\Big(
  \bfg_0 + \bfg_1 + \bfg_2 + \underbrace{\znoise_0 + \znoise_1 + \znoise_2}_{\text{noise \emph{accumulation}}}
\Big)
$
};

\node[above of=g1, yshift=1.2cm, xshift=1.4cm] (title) {\Large 
\textbf{DP-SGD with Independent Noise}
};

\end{tikzpicture}

%% file: noise_addition_correlated.tex
\begin{tikzpicture}[
node distance = 5mm and 7mm,
box1/.style={rounded rectangle, semithick, minimum size=1cm, minimum width=3.5cm},
box2/.style={rectangle, semithick, dotted, minimum height=1cm},
op1/.style={circle, semithick, minimum size=1cm},
op2/.style={rectangle, semithick, minimum width=2cm, rounded corners=0.5cm},
boxOp/.style = {op1, fill=gray!25},
boxOp2/.style = {op2, fill=gray!25},
boxGrad1/.style = {box1, fill=C0!25},
boxGrad2/.style = {box1, fill=C1!25},
boxGrad3/.style = {box1, fill=C2!25},
boxGrad4/.style = {box1, fill=C3!25},
boxMain/.style = {box2, fill=yellow!40},
arr/.style = {-Triangle},
]

\node[boxMain] (g1) at (5, 5) {\Large $\bfg_0$};
\node[boxOp, left of=g1, xshift=-1.75cm] (plus1) {\Large $\oplus$};
\draw[arr] (plus1.east) -- (g1.west) ;

\node[boxGrad1, left of=plus1, xshift=-2.8cm, yshift=0.6cm] (g11) { $\mathsf{clip}\big(\nabla \ell(\bftheta_0, \bfx_{\text{\textcolor{C0}{blue}}})\big)$};
\node[boxGrad2, left of=plus1, xshift=-2.8cm, yshift=-0.6cm] (g12) { $\mathsf{clip}\big(\nabla \ell(\bftheta_0, \bfx_{\text{\textcolor{C1}{orange}}})\big)$};
\draw[arr] (g11) -- (plus1) ;
\draw[arr] (g12) -- (plus1) ;

\node[boxMain, below of=g1, yshift=-2.25cm] (g2) {\Large $\bfg_1$};
\node[boxOp, left of=g2, xshift=-1.75cm] (plus2) {\Large $\oplus$};
\draw[arr] (plus2) -- (g2) ;

\node[boxGrad3, left of=plus2, xshift=-2.8cm, yshift=0.6cm] (g21) { $\mathsf{clip}\big(\nabla \ell(\bftheta_1, \bfx_{\text{\textcolor{C2}{green}}})\big)$};
\node[boxGrad4, left of=plus2, xshift=-2.8cm, yshift=-0.6cm] (g22) { $\mathsf{clip}\big(\nabla \ell(\bftheta_1, \bfx_{\text{\textcolor{C3}{red}}})\big)$};
\draw[arr] (g21) -- (plus2) ;
\draw[arr] (g22) -- (plus2) ;

\node[boxMain, below of=g2, yshift=-2.25cm] (g3) {\Large $\bfg_2$};
\node[boxOp, left of=g3, xshift=-1.75cm] (plus3) {\Large $\oplus$};
\draw[arr] (plus3) -- (g3) ;

\node[boxGrad1, left of=plus3, xshift=-2.8cm, yshift=0.6cm] (g31) { $\mathsf{clip}\big(\nabla \ell(\bftheta_2, \bfx_{\text{\textcolor{C0}{blue}}})\big)$};
\node[boxGrad2, left of=plus3, xshift=-2.8cm, yshift=-0.6cm] (g32) { $\mathsf{clip}\big(\nabla \ell(\bftheta_2, \bfx_{\text{\textcolor{C1}{orange}}})\big)$};
\draw[arr] (g31) -- (plus3) ;
\draw[arr] (g32) -- (plus3) ;

\node[boxOp2, right of=g2, xshift=2.5cm, minimum height=6.5cm] (noise0) {
\begin{tabular}{c}
Add \\ \textbf{non-i.i.d.} \\ noise, i.e. 
\\ 
linear \\ 
combinations \\
of \\
$\znoise_t \sim \normalm{0}{\stddev^2}$
\end{tabular}} ; 
\draw[arr] (g1.east) -- ([yshift=-0.4mm]noise0.121) ;
\node[boxMain, right of=g1, xshift=4.75cm, anchor=west] (gh1) {\large $\bfg_0 + \big(\znoise_0\big)$};
\draw[arr] ([yshift=-0.25mm]noise0.59) -- (gh1.west);

\draw[arr] (g2.east) -- (noise0) ;
\node[boxMain, right of=g2, xshift=4.75cm, anchor=west] (gh2) {\large $\bfg_1 + \big(\znoise_1 - c_{10}' \znoise_0\big)$};
\draw[arr] (noise0) -- (gh2.west);

\draw[arr] (g3.east) -- ([yshift=0.25mm]noise0.239) ;
\node[boxMain, right of=g3, xshift=4.75cm, anchor=west] (gh3) {\large $\bfg_2 + \big(\znoise_2 - c_{21}' \znoise_1 - c_{20}' \znoise_0\big)$};
\draw[arr] ([yshift=0.3mm]noise0.301) -- (gh3.west);

\node[boxMain, below of=gh3, yshift=-1.5cm, xshift=-4.72cm] (result1) {\Large 
$\displaystyle
\bftheta_3 = \bftheta_0 - \eta 
\Big(
  \bfg_0 + \bfg_1 + \bfg_2 + \underbrace{(1-c_{10}' - c_{20}') \znoise_0 + (1 - c_{21}')\znoise_1 + \znoise_2}_{\text{noise \emph{cancellation}}}
\Big)
$
};

\node[above of=g1, yshift=1.2cm, xshift=1.3cm] (title) {\Large 
\textbf{DP-SGD with Correlated Noise}
};

\end{tikzpicture}

%% file: outline.tex
\begin{tikzpicture}[
node distance = 5mm and 7mm,
box1/.style={rectangle, semithick, minimum size=1.25cm},
box2/.style={rectangle, semithick, minimum height=5cm},
boxWL/.style = {box1, fill=gray!25},
boxErr/.style = {box1, fill=C0!25},
boxSens/.style = {box1, fill=C3!25},
boxMech/.style = {box1, fill=C2!25},
boxOpt/.style = {box1, fill=C1!25},
boxWL2/.style = {box2, fill=gray!25},
boxErr2/.style = {box2, fill=C0!25},
boxSens2/.style = {box2, fill=C3!25},
boxMech2/.style = {box2, fill=C2!25},
boxOpt2/.style = {box2, fill=C1!25},
]

\node at (29, 2) {\Large\underline{\textbf{Finding a Correlated Noise Mechanism}}:};

\node[boxOpt] (min) at (25, 0) {\Large $\underset{\bm{B}, \bm{C}} {\mathrm{minimize}}$};
\node[boxErr, right=of min, xshift=3mm] (err) {\Large \,$\mathsf{Error}(\bm{B})$};
\node[right=of err, xshift=-7mm] (times) {\Large $\times$};
\node[boxSens,right= of times, xshift=-7mm] (sens) {\Large $\mathsf{Sensitivity}(\bm{C})$};
\node[below=of min] (st) {\Large $\mathrm{subject\,\, to}$};
\node[boxWL] (wl) at (err|- st) {\Large $\bm{B} \bm{C} = \bm{A}$};
\node[right=of wl, xshift=-5mm] (Andbox) {\Large and};
\node[right=of sens, xshift=2cm] {\Large $(*)$};
\node[boxMech, below=of wl, xshift=19.5mm] (constr) {\Large $\bm{C}$ satisfies some constraints};

\node[below=of constr, xshift=-1.5cm,yshift=-0.5cm] {\Large\underline{\textbf{Degrees of Freedom}}:};

\node[boxWL2, below left=of constr, yshift=-1.6cm, xshift=-1cm] (wlbox)
{
\renewcommand{\arraystretch}{1.4}
\begin{tabular}{c}
\textbf{Workload} (Ch. \ref{chap:intro}, \ref{chap:ml}) \\
\midrule 
SGD (+ Momentum) \\
Nesterov Accelerated Gradient \\
$\vdots$ \\
\midrule 
\textbf{Determines:} \\
Error (via $\bfB = \workload \strategy^{-1}$), \\
Base Non-private Optimizer
\textcolor{gray!25}{}
\end{tabular}
};

\node[boxErr2, right=of wlbox] (errbox)
{
\renewcommand{\arraystretch}{1.4}
\begin{tabular}{c}
\textbf{Error/Utility} (Ch. \ref{chap:prefixsum}) \\
\midrule 
Max Error \\
Root-Mean-Square Error \\
$\vdots$ \\
\midrule 
\textbf{Determines:} \\
Error Metric $\implies $\\
Mechanism Objective $(*)$
\end{tabular}
};

\node[boxSens2, right=of errbox] (sensbox)
{
\renewcommand{\arraystretch}{1.4}
\begin{tabular}{c}
\textbf{Participation Pattern} (Ch. \ref{chap:ml}) \\
\midrule 
Single-participation / Streaming  \\
Multiple-participation / 
Cyclic Order \\
Multiple-participation / Min-Sep \\ \midrule 
\textbf{Determines:}  \\
Sensitivity (Privacy) $\implies$ \\
Mechanism Objective $(*)$
\end{tabular}
};

\node[boxMech2, yshift=-6.5cm, xshift=-0.5cm] (mechbox) at ($(wlbox)!0.5!(errbox)$)
{
\renewcommand{\arraystretch}{1.4}
\begin{tabular}{c}
 
\textbf{Mechanism Constraints} (Ch. \ref{chap:prefixsum}) \\
\midrule 
No Constraints (Dense)  \\
Toeplitz \\
Banded (+ Toeplitz) \\
Buffered Linear Toeplitz (BLT)
\\ \midrule 
\textbf{Determines:}  \\
{\renewcommand{\arraystretch}{1}
\begin{tabular}{c}
Noise Generation Time/Space $+$ Utility Bound \\
$\implies$ Privacy-Utility-Compute Tradeoff
\end{tabular}} \\
\end{tabular}
};

\node[boxOpt2, yshift=-6.5cm, xshift=1.5cm] (optbox) at ($(errbox)!0.5!(sensbox)$)
{
\renewcommand{\arraystretch}{1.4}
\begin{tabular}{c}

\textbf{Mechanism Optimization} (Ch. \ref{chap:practical}) \\
\midrule 
Closed-form Solution  \\
Semi-definite Program (SDP) \\
Convex + Quasi-Newton Optim. \\
Non-convex + Gradient-based Optim.
\\ \midrule 
\textbf{Determines:}  \\
{\renewcommand{\arraystretch}{1}
\begin{tabular}{l}
Time/Space Complexity of \\
Mechanism Optimization
\end{tabular}} \\
\end{tabular}
};

\end{tikzpicture}

%% file: 2a_prefix_sum_new.tex
We develop some correlated noise mechanisms in the simplified \emph{streaming setting} to convey the key ideas. As we introduced in \Cref{chap:intro}, the streaming setting refers to the setting where each data point participates in training only once.

Recall that our goal is to factorize a lower-triangular workload matrix $\workload \in \R^{\ndnd}$ into $\workload = \bfB \strategy$.
We describe a general approach but instantiate it specifically with unweighted prefix sum matrix, $\prefix \in \{0, 1\}^{\ndnd}$, which is a lower-triangular matrix of all ones (see also \cref{eq:all-ones-workload}):
\[
    \prefix\idx{i}{j} = \begin{cases}
    1 & i \geq j \\
    0 & \text{otherwise}.
    \end{cases}
\]

Recall from \Cref{sec:weightedPrefixSum} that privatizing the iterates $\bftheta_1, \ldots, \bftheta_{\nd-1}$ obtained from the iterations of stochastic gradient descent (SGD; see \cref{alg:sgd-1}) with correlated noise mechanisms corresponds to factorizing the matrix $\prefix$. 

For all theorems and lemmas stated without proof, detailed references that include proofs are given in the bibliographic notes of \Cref{sec:ch2-biblio} at the end of the section.

\section{Design Considerations}

The primary design considerations for a correlated noise mechanism are its computational cost and its effectiveness in improving the utility of the algorithm.
We recall the setup and describe each in turn.

\paragraph{Setup}
As described in \Cref{chap:intro}, our goal is to privately compute the prefix sums $\Gradients \mapsto \workload \Gradients$ for an input sequence $\Gradients = (\gradient_0, \ldots, \gradient_{n-1}) \in \R^{\nd \times m}$ (representing gradients). Given a factorization $\workload = \bfB \strategy$ of a weighted prefix sum matrix with lower triangular factors $\bfB, \strategy \in \R^{n \times n}$, consider a correlated noise mechanism $\mech$ that returns 
\begin{align} \label{eq:ch2-mechanism}
    \mech(\Gradients) = \bfB(\strategy \Gradients + \Znoise) = \workload(\Gradients + \strategy^{-1} \Znoise) \,,
\end{align}
where $\Znoise$ is component-wise i.i.d. Gaussian noise. In \cref{sec:noisecancel}, we saw this was equivalent to a mechanism $\mech'$ which computes DP estimates of $\Gradients$, $\mech'(\Gradients) = \dpGradients = \Gradients + \Cinv \Znoise$, and then applying the workload as post-processing, $\workload \dpGradients = \workload \mech'(\Gradients) = \mech(\Gradients)$.

We know from \Cref{thm:gdp-of-correlated-noise} that $\Znoise \sim \normalnm{0}{\stddev^2}$ for 
\begin{equation}\label{eq:stddevnmult}
\stddev = \sens(\strategy) \cdot \nmult = 2 \colnorm{\strategy} \nmult 
\end{equation}
yields a $\left({1/ \nmult}\right)$-GDP mechanism in the streaming setting, where $\colnorm{\strategy}$
denotes the maximum column norm of the matrix $\strategy$. Thus, throughout this section, we choose $\stddev$ following \cref{eq:stddevnmult}; then $\nmult$ is interpreted as a noise multiplier parameter that is independent of the factorization used. 

\subsection{Time and Space Complexity of Noise Generation}  
As we saw in \Cref{alg:dpsgd-corr} in \Cref{chap:intro},
private optimization using the correlated noise mechanism requires us to compute the noisy  gradient 
\begin{align} \label{eq:noise-gen}
\dpgradient_t 
= \gradient_t + \left(\strategy^{-1} \Znoise\right)[t, :] 
= \gradient_t + \sum_{\tau=0}^{t} (\bfC^{-1})[t, \tau] \,\, \Znoise[\tau, :] 
\end{align}
in each step $t$ of the algorithm, where $\Znoise$ is component-wise i.i.d. Gaussian noise.\footnote{
    The summation over $\tau$ in the second expression in \cref{eq:noise-gen} runs only until $\tau=t$ because $\strategy^{-1}$ is lower triangular. 
    This follows from $\strategy$ being lower triangular.
} 
Consequently, a critical design consideration in correlated noise mechanisms is the time and space complexity of the \textbf{noise generation} process of \cref{eq:noise-gen}.
In the worst case, it can take $O(mn)$ time to compute a sum of up to $n$ vectors $\Znoise[\tau, :] \in \R^m$. Similarly, it can take up to $O(n^2)$ space to simply represent the matrix $\strategy^{-1}$ (or $O(\nd^3)$ time to compute, given $\strategy$).

In typical AI and machine learning settings, the model dimension $\mdim$ can be significantly larger than the number of steps $\nd$. Indeed, $\mdim$ can vary between a few millions for small on-device models to several billions or more for transformer language models. On the other hand, differentially private training entails large batches, so the number of steps $n$ is usually a few tens of thousands or smaller. Thus, the $O(m n)$ time complexity of noise generation is usually more of a bottleneck than the $O(n^2)$ memory. In some cases, even a noise generation cost $O(m n)$ may be tractable, as discussed in \cref{chap:practical} and \cref{rem:relativecomplexity}.
Further, as we will see in this section, it is often possible to do better with carefully designed structured matrices.

The cost of finding a factorization $\workload = \bfB\strategy$ is also relevant. This is a one-time cost as the matrices $\bfB, \strategy$ can be cached for future use. In this section, we only consider the cost of noise generation, as it is incurred in every step of the correlated noise mechanism. We will revisit the factorization cost in \Cref{chap:practical}.

\subsection{Loss Metric}

While the ultimate test of a correlated noise mechanism is its learning performance (e.g., accuracy for classification tasks), it is nevertheless useful to track the error in the prefix sum estimate as a surrogate measure of utility. \Cref{fig:max-error} illustrates the error in the prefix sums in a learning setting. 

A natural metric is the maximum squared error in the estimation of any prefix sum; we considered this in \cref{eq:whycorrelate} in \Cref{chap:intro}. In the literature, it is also  common to consider its square root:

\begin{bdefinition} \label{def:unnormalized-max-loss}
The \textbf{unnormalized max \txtloss} (also known as the \textbf{unnormalized $\ell_\infty$ \txtloss}) of a mechanism $\mech(\Gradients)$ estimating the (weighted) prefix sum $\workload \Gradients$ of an input $\Gradients \in \R^{n \times m}$ is defined as
\begin{align*}
    \maxloss(\mech) 
    & \coloneqq 
    \,\, \max_{t \in [n]} \sqrt{\mathbb{E} \left\| ( \workload\Gradients - \mech(\Gradients) )[t, :] \right\|_2^2} \,.
\end{align*}
\end{bdefinition}

It turns out that the \txtmaxloss for the correlated noise mechanisms has a simple characterization in terms of the maximum row norm $\rownorm{\bfB}$ of $\bfB$ and the maximum column norm $\colnorm{\strategy}$ of $\strategy$:\footnote{
In general, we use the notation $\norm{\bfM}_{p \to q} \coloneqq \max_{\bfu \neq \zeros} \frac{\norm{\bfM \bfu}_q}{\norm{\bfu}_p}$ to denote the induced matrix norm.
}

\begin{btheorem}
\label{thm:accuracyMatrixMechanism}
Consider a correlated noise mechanism
$\mech(\Gradients) = \bfB(\strategy \Gradients + \Znoise)$ for $\Znoise \sim \normalnm{0}{\stddev^2}$ 
with $\stddev = \sens(\strategy) \cdot \nmult = 2 \colnorm{\strategy} \nmult$ for noise multiplier $\nmult$, following \cref{lem:sensbound}. Then, we have that its \txtmaxloss, denoted as $\maxloss(\bfB, \strategy)$, equals
\begin{align*}
\maxloss(\bfB, \strategy) = 
    2\sqrt{m}\nmult \,\, \rownorm{\bfB} \colnorm{\strategy}  \,.
\end{align*}
\end{btheorem}
\begin{proof}[Proof Sketch]
Note that $\workload \Gradients - \mech(\Gradients) = \bfB \Znoise \sim \normal(\zeros, \stddev^2  \bfB \bfB^\top)$ 
is oblivious to the input $\Gradients$. Recalling $\bfs_t \coloneqq \workload[t,:]\Gradients$ and $\widehat\bfs_t = {\mech(\Gradients)[t,:]}$, we can measure the $\ell_2$ norm of the error of the $t$ prefix sum estimate $\widehat\bfs_t$ as
$\norm{\bfs_t - \widehat\bfs_t}_2^2 = \norm{(\bfB[t,:] \, \Znoise)}_2^2$, and so 
\begin{align*}
    \mathbb E_{\mech}  \norm{\bfs_t - \widehat\bfs_t}_2^2 
    &= {\mathbb E_{\Znoise} \norm{(\bfB[t,:] \, \Znoise)}_2^2} 
    = m\stddev^2 \left(\sum_{\tau \in [n]}\bfB[t,\tau]^2\right)
\end{align*}
because $\Znoise \sim \normalnm{0}{\stddev^2}$ is the only source of randomness in $\mech$. 
Therefore, we get that
\begin{align*}
    \maxloss(\bfB, \strategy)
    & = \sqrt{m} \stddev \,\, \max_{t \in [n]}  \sqrt{\sum_{\tau \in [n]}\bfB[t,\tau]^2} \\
    & = 2 \sqrt{m} \nmult \colnorm{\strategy}\rownorm{\bfB}
\end{align*}
as required.
\end{proof}

Since the dimension $m$ is a fixed constant for a given problem, and the noise multiplier $\nmult$ is independent of the factorization $\workload = \bfB \strategy$ defining the correlated noise mechanism, we will state our results in terms of the \textbf{(normalized) \txtmaxloss}; because this loss is the primary function considered when designing mechansimms, we will refer to the normalized \txtmaxloss as simply ``\txtmaxloss'' going forward.

\begin{bdefinition}[\txtmaxlossCap] \label{def:max-loss}
For a correlated noise mechanism
$\mech(\Gradients) = \bfB(\strategy \Gradients + \Znoise)$ for $\Znoise \sim \normalnm{0}{\stddev^2}$, we define the (normalized) \txtmaxloss by
\begin{equation} \label{eq:norm-max-err}
    \maxlossn(\bfB, \strategy) \coloneqq 
    \rownorm{\bfB} \, \colnorm{\strategy} 
    \,.
\end{equation}
\end{bdefinition}

Other \txtloss metrics can be used; we discuss root-mean-squared-loss (\txtrmsloss) in \cref{sec:prefix-L2-error}. Throughout, for a factorization $\workload = \bfB \strategy$, we will use the term \emph{loss} when we take into account the privacy of the mechanism, e.g. scaling the noise by the sensitivity of $\strategy$, and \emph{error} when we measure to the magnitude of the noise $\bfB \bfZ$ introduced in our estimates of $\workload \Gradients$. Further, the loss generally scales linearly in the noise multiplier $\nmult$ as in \cref{thm:accuracyMatrixMechanism}, and hence when evaluating the quality of a mechanism, we can ignore this term, leading to the notion of \emph{normalized loss}. Hence, in general we have
\begin{equation}\label{eq:loss-error-sens}
\mathsf{loss}(\bfB, \strategy) = \mathsf{error}(\bfB) \cdot \sens(\strategy).
\end{equation}
The error term can be varied, e.g. depending on whether we select the maximum per-iteration error (as in this section) or the mean error. In the streaming setting, $\sens(\strategy) = 2 \colnorm{\strategy}$, as in \cref{eq:senscolC}, but in \cref{chap:ml} we will generalize this.

\subsection{Baseline Mechanisms}
\label{sec:chap2-baselines}

Recall the two baseline mechanisms of \Cref{chap:intro}: the input perturbation ($\strategy=\bfI_{\nd \times \nd}$) and output perturbation ($\strategy=\workload$).  We review the complexity of noise generation and the utility bounds for both these cases.

\paragraph{Input Perturbation}
With $\strategy=\bfI_{\nd \times \nd}$ as the identity matrix, the noise generation process
in \cref{eq:noise-gen} is $\left(\strategy^{-1} \Znoise\right)[t, :] = \Znoise[t, :] \sim \normalm{0}{\nmult^2}$. In other words, it simplifies to generating independent Gaussian noise. This corresponds to the DP-SGD algorithm in the learning setting.

The per-step noise generation time is $O(m)$, which is required to sample $m$-dimensional white noise $\Znoise[t, :]$. This noise also requires $O(m)$ memory.
As established in \Cref{chap:intro}, its normalized \txtmaxloss for the unweighted prefix sum workload $\prefix$ is $\maxlossn(\prefix, \bfI) = \Theta(\sqrt{\nd})$.

\paragraph{Output Perturbation}
The $\strategy=\workload$ baseline corresponds to perturbing the output $\workload \Gradients$ with independent noise $\Znoise$. For the unweighted prefix sum workload $\workload=\prefix$, the noise generation process
in \cref{eq:noise-gen} takes the form 
\[
    \left(\prefix^{-1} \Znoise\right)[t, :] =
    \Znoise[t, :] - \Znoise[t-1, :] \,.
\]
This follows, for example, from the formula for $\strategy^{-1}$ in \cref{eq:cinv:one-param-family}.

The time complexity of the noise generation process remains $O(m)$ per-step even for output perturbation.
Thus, in each step $t$, we need to store the noise $\Znoise[t-1, :]$ sampled from the previous time step in addition to $\Znoise[t, :]$; we still have an $O(m)$ memory overhead. Its utility bound $\maxlossn(\bfI, \prefix) = \Theta(\sqrt{\nd})$
is the same as input perturbation, as we saw in \Cref{chap:intro}.

The time and space complexities of these baselines are the best possible: $O(m)$ memory is required to store the input vector $\gradient_t \in \R^m$ and $O(m)$ time is required to process it.
However, the $\Theta(\sqrt{n})$ utility is suboptimal as we saw in \Cref{sec:whyCorrelatedNoise}, where we showed that a (normalized) \txtmaxloss of $O(n^{1/4})$ is achievable. We show that this can, in fact, be significantly improved to $\Theta(\ln(n))$. We summarize the mechanisms we study in this section in \Cref{tab:mechanisms}.

\begin{table}[t]
\centering
\caption{Summary of mechanisms considered in \Cref{chap:prefixsum}. Here, $n$ is the number of steps, $m$ is the model dimension, and $c$ is an absolute constant that can change in each row. We describe the maximum time and space complexity of producing the correlated noise $(\strategy^{-1}\Znoise)[t, :]$ for any iteration $t \in [n]$. For most real-world applications, $m > n$ (possibly by orders of magnitude), and the time complexity becomes the dominating concern. The $O(m +n)$ space complexity for Toeplitz mechanisms holds for an arbitrary Toeplitz mechanism; when the Toeplitz coefficients can be easily (re)computed on the fly, e.g. as with the max-loss-optimal factorization of \cref{thm:optimal-toeplitz}, the space requirement reduces to $O(m)$. The Banded Toeplitz and BLT mechanisms are in fact Toeplitz and so could be implemented with the $O(nm)$ time and $O(m+n)$ space, but the given time/space complexities achieved by specializing noise generation are almost always preferable.
}
\label{tab:mechanisms}
\adjustbox{max width=0.99\linewidth}{
\renewcommand{\arraystretch}{1.5}
\begin{tabular}{llll}
\toprule
\textbf{Mechanism} & \textbf{Time}  & \textbf{Space} & \textbf{\txtmaxloss} $\maxlossn$ \\
\midrule
Input perturbation   & $O(m)$   & $O(m)$         & $\Theta(\sqrt{n})$ \\
Output perturbation  & $O(m)$   & $O(m)$         & $\Theta(\sqrt{n})$ \\
Dense                & $O(n m)$ & $O(m +n^2)$    & $\tfrac{1}{\pi}\ln(n) + c$ \\
Toeplitz             & $O(n m)$ & $O(m + n)$         & $\tfrac{1}{\pi}\ln(n) + c$ \\
$b$-Banded Toeplitz  & $O(b m)$ & $O(b m)$       & See \Cref{thm:max-err-banded} \\
$d$-buffer BLT, $d = \poly\ln(n)$
                     & $O(d m)$ & $O(d m)$       & $\tfrac{1}{\pi}\ln(n) + c$ \\
\bottomrule
\end{tabular}}
\end{table}

\section{Dense Mechanism}
\label{sec:optimized-dense-mech}

The poor utility of the baselines stems from their disregard for the error metric. This can be remedied  by optimizing for the normalized \txtmaxloss $\maxlossn(\bfB, \strategy)$ directly when finding $\bfB, \strategy$:
\begin{align} \label{eq:max-err-optim}
    \min
    \left\{
        \rownorm{\bfB} \colnorm{\strategy} \, :\, 
        \begin{matrix}
        \bfB \strategy = \workload, \quad \text{and} \\
        \bfB, \strategy \text{ are lower-triangular}
        \end{matrix}
    \right\} \,.
\end{align}
While this optimization problem is non-convex, it can be cast as a semi-definite program, enabling high-quality solutions for small problems (with a few thousand steps or less); we discuss considerations around scalability of this approach in \Cref{chap:practical}.

We refer to the mechanism defined by the resulting matrices, $\bfB$ and $\strategy$ as the \emph{dense mechanism}. (These matrices are dense; in contrast, some of the mechanisms we consider later are either sparse or can be parameterized using a small number of parameters.)

\paragraph{Complexity of Noise Generation}
To compute the linear combination of noises $\sum_{\tau=0}^t \br{\strategy^{-1}}[t, \tau]\,\, \Znoise[\tau, :]$ in the noise generation process of \cref{eq:noise-gen} in step $t$ requires $O(mt)$ time to generate and then sum over $t$ white noise vectors in $\R^m$.
Thus, the worst-case per-step time complexity of  noise generation is $O(mn)$. 

Storing the matrix $\strategy^{-1}$ requires $O(n^2)$ space. We assume the source i.i.d. noise $\Znoise\idx{\tau}{:}$ can be generated on the fly (e.g., via a random seed hashed with $\tau$), and hence materializing the noise $\Znoise\idx{\tau}{:}$ only takes space $O(m)$.

\paragraph{Utility and Error Bounds}
For the unweighted prefix sum workload $\prefix$, the optimal \txtmaxloss from \cref{eq:max-err-optim} can be precisely characterized with upper and lower bounds matching up to a small additive constant:

\begin{btheorem} \label{thm:dense-opt-max-err}
    Let $\bfB^\star, \strategy^\star$ be the minimizers of the optimization problem from \cref{eq:max-err-optim} when $\workload=\prefix$ is the unweighted prefix sum matrix. Let 
    \begin{align}
        \opt = \maxlossn(\bfB^\star, \strategy^\star) = \rownorm{\bfB^\star} \colnorm{\strategy^\star}
        \label{eq:optimizationbarE}
    \end{align} 
    denote the minimal normalized \txtmaxloss. Then, we have the nearly matching upper and lower bounds: 
    \[
     \frac{\ln(2n + 1)}{\pi}  \le \opt \le 1 + \frac{\ln(n)}{\pi}  \,,
    \]
    where $\pi \approx 3.14$ is the ratio of a circle's circumference to its diameter and $\ln(\cdot)$ is the natural logarithm function.
\end{btheorem}

While the full proof of this statement is out of the scope of this monograph, we give a detailed attribution of this theorem and the other upcoming results of this section in the bibliographic notes of \Cref{sec:ch2-biblio}.
While \Cref{thm:dense-opt-max-err} gives the asymptotic behavior of the \txtmaxloss, we can numerically find high-precision solutions for small $n$ (smaller than a few thousands; we discuss algorithms and their scalability issues in \Cref{chap:practical}). We such solutions, we can numerically compute the optimal \txtmaxloss for specific $n$ (together with a certificate of optimality, such as a duality gap).

\paragraph{Summary}
The optimized dense mechanism's normalized \txtmaxloss of ${\ln(n) / \pi} + c$ (for a small numerical constant $c$) is significantly better than that the $\Theta(\sqrt{n})$ error obtained from the baselines when factorizing $\prefix$. However, the time complexity of $O(mn)$ (and to a lesser extent, the space complexity of $O(n^2)$) can be prohibitive when the number of steps $n$ is large. 
The mechanisms we discuss in the rest of the section will address this issue.

\section{Toeplitz Mechanism} 
\label{sec:optimial-toeplitz-mech}
We start by addressing the high $O(n^2)$ space complexity of the optimized dense mechanism. While this is generally less of a concern than the $O(nm)$ time complexity in typical machine learning settings, it forms a good starting point to develop mechanisms with improved time complexity.

A common trick to overcome the $O(n^2)$ memory required to store an $n \times n$ matrix is to assume that it is \emph{Toeplitz}, where each diagonal parallel to the main diagonal is a constant.\footnote{
    Another common trick is to assume a low-rank factorization; we will return to that in \Cref{sec:ch2-blt}.
} 
An $n\times n$ lower-triangular Toeplitz matrix $\strategy$ can be described in terms of its first column of $n$ numbers:
\begin{align} \label{eq:Toeplitz-C}
    \strategy = \begin{pmatrix}
        c_0 & 0 & 0 & \cdots & 0 \\
        c_1 & c_0  & 0 & \cdots & 0 \\
        c_2 & c_1 & c_0 & \cdots & 0\\
        \vdots & \ddots & \ddots & \ddots & \vdots \\
        c_{n-1} & c_{n-2} & c_{n-3} & \cdots & c_0
    \end{pmatrix} \,.
\end{align}

Toeplitz matrices offer not only reduced memory usage but also two remarkable properties that make them ideal for this application:
(a) the optimal Toeplitz factorization can be determined analytically, and (b) they \emph{almost} achieve the optimal \txtmaxloss of \Cref{thm:dense-opt-max-err}, namely up to a small additive constant.

It is most interesting to restrict the strategy matrix $\strategy$ to be Toeplitz (and lower triangular) when the workload matrix $\workload$ is also Toeplitz; in this case the matrix $\bfB = \workload \strategy^{-1}$ is also Toeplitz (and lower triangular).
The unweighted prefix sum workload $\prefix$ satisfies this requirement. As we shall see in the upcoming \Cref{chap:ml}, other common first-order optimizers also lead to workload matrices $\workload$ that are Toeplitz and element-wise non-negative.

\paragraph{Mechanism Definition}
The max-loss-optimal Toeplitz mechanism in terms of the \txtmaxloss is the solution to the optimization problem:
\begin{align} \label{eq:max-err-optim-toeplitz}
    \min
    \left\{
        \rownorm{\bfB} \colnorm{\strategy} \, :\, 
        \begin{matrix}
        \bfB \strategy = \workload, \quad \text{and} \\
        \bfB, \strategy \text{  lower-triangular \& Toeplitz}
        \end{matrix}
    \right\} \,.
\end{align}
As in the earlier case, this is a non-convex optimization problem, which can be solved numerically.
We restrict our discussion to the unweighted prefix sum workload $\workload=\prefix$ in the following.
For this matrix, Problem \eqref{eq:max-err-optim-toeplitz} admits an analytical solution for this case:

\begin{btheorem}[Max-Loss-Optimal Toeplitz Factorization]
\label{thm:optimal-toeplitz}
    For the unweighted prefix sum workload $\workload = \prefix$,
    the optimization problem from \cref{eq:max-err-optim-toeplitz} is minimized by Toeplitz matrices $\bfB_{\mathsf{Toep}}=\strategy_{\mathsf{Toep}}=\prefix^{1/2}$, the square root matrix of $\prefix$.\footnotemark\,
    In particular, their first column $c_0^\star, c_1^\star, \ldots, c_{n-1}^\star$ is given by 
    \begin{align}
    \label{eq:opt-toeplitz-c}
    c_t^\star 
    = (-1)^t \binom{-1/2}{t} 
    = \begin{cases}
        1 \,, & \text{ if } t = 0, \\
        \Theta(t^{-1/2})\,, & \text{ else}\,,
        \end{cases}
    \end{align}
    where we denote the generalized binomial  coefficient $\binom{p}{t} \coloneqq \prod_{\tau=0}^{t-1} \frac{p - \tau}{t - \tau}$ for $t \in \N$ and non-integer $p \in \R$.
    Moreover, $\strategy_{\mathsf{Toep}}^{-1}$ is also a Toeplitz matrix whose first column $c_0', c_1', \ldots, c_{n-1}'$ is given by
    \begin{align}
    \label{eq:opt-toeplitz-cinv}
        c_t' = (-1)^t \binom{1/2}{t} = \begin{cases}
        1 \,, & \text{ if } t = 0, \\
        - \Theta(t^{-3/2})\,, & \text{ else}\,.
        \end{cases}
    \end{align}
\end{btheorem}
\footnotetext{
    The square root
    $\workload^{1/2}$ of the lower triangular matrix $\workload$ with positive diagonal entries
    is the unique lower triangular matrix with positive diagonal entries such that $\workload = \workload^{1/2} \workload^{1/2}$.
    } 

Interestingly, the entries of the optimal factorization, as given in \cref{eq:opt-toeplitz-c}, are independent of the size $n$ of the problem.
See \Cref{sec:approx-theory} for the key idea behind the proof.

\begin{bremark} \label{remark:opt-toep-library-fns}
The generalized binomial coefficients $\binom{-1/2}{t}$ and $\binom{1/2}{t}$ from \Cref{thm:optimal-toeplitz} can directly be computed with standard library functions such as \texttt{scipy.special.binom}. We also have the alternative expressions
\[
    c_t^\star = 2^{-2t} \binom{2t}{t} = 
    \left(\frac{2t-1}{2t}\right) c_{t-1}^\star\,,
\]
where the latter recursion starts from the base case of $c_0^\star = 1$. Similarly, using the relation $\strategy_{\mathsf{Toep}}^{-1} = \prefix^{-1} \strategy_{\mathsf{Toep}}$, we can express the coefficients $c_t'$ of $\strategy_{\mathsf{Toep}}^{-1}$ as 
\[
    c_t' = \begin{cases}
    1, & \text{ if } t = 0, \\
    c_{t+1}^\star - c_t^\star, & \text{ if } t > 0.
    \end{cases}
\]
\end{bremark}

\paragraph{Complexity of Noise Generation}
As with the dense mechanism, we assume we can re-generate the previous noises $\znoise_{t-1}, \znoise_{t-2}, \ldots$ in each iteration (using e.g. their random seeds). This approach suffers from a $O(mn)$ time complexity of the per-step noise generation, which is the same as the dense mechanism, as well as the same  $O(m)$ memory requirement.
For a general Toeplitz matrix, only $O(n)$ rather than $O(n^2)$ memory is required to store the strategy matrix $\strategy$. For the specific factorization of \cref{thm:toeplitz-opt-max-err}, the entries of $\strategy_{\mathsf{Toep}}^{-1}$ can be generated as needed from \cref{eq:opt-toeplitz-cinv} (e.g., using the standard library functions as in \Cref{remark:opt-toep-library-fns}), so in this case we do not need even the  $O(n)$ space to store them.

It is possible to obtain an improved $O(mn \ln(n))$ time complexity by implementing the (left) multiplication by a Toeplitz matrix using the fast Fourier transform. Unfortunately, this requires $O(mn)$ space complexity to materialize all previous rows of the noise matrix $\Znoise$, which is generally prohibitive for typical AI and machine learning  models.

\paragraph{Utility and Error Bounds}
This mechanism is near-optimal among the class of \emph{all} (possibly non-Toeplitz) factorizations of the unweighted prefix sum workload $\prefix$ up to a small additive constant:

\begin{btheorem}
\label{thm:toeplitz-opt-max-err}
    Let $\optToep = \maxlossn(\prefix^{1/2}, \prefix^{1/2})$ denote the normalized \txtmaxloss of the optimal lower triangular Toeplitz factors $\bfB_{\mathsf{Toep}} = \strategy_{\mathsf{Toep}}=\prefix^{1/2}$ from \Cref{thm:optimal-toeplitz}. We have, 
    \[
         \optToep \le \frac{\gamma + \ln(n)}{\pi} + 1 \,,
    \]
    where $\gamma\approx0.577$ is the Euler-Mascheroni constant. 
    In particular, we have $\optToep - \opt \le 1$, where $\opt$ is the best achievable normalized \txtmaxloss by any mechanism as defined in \Cref{thm:dense-opt-max-err}.
\end{btheorem}
\begin{proof}[Proof Sketch]
    The maximum row norm of $\prefix^{1/2}$ is that of its last row, while its maximum column norm is attained by its first column; this can be observed by the structure of the Toeplitz matrix in \cref{eq:Toeplitz-C}.
    Thus, we have,
    \[
    \rownorm{\prefix^{1/2}}^2
    = \colnorm{\prefix^{1/2}}^2 = 
    \sum_{t=0}^{n-1} \binom{-1/2}{t}^2 \le 1 + \sum_{k=1}^{n-1} \Theta\left( \frac{1}{k} \right) \,.
    \]
    Standard result of the harmonic sum gives $\sum_{k=1}^{n-1} {1\over k} \le \ln(n-1) + \gamma + 1/(2(n-1)) \le \ln(n) + \gamma$, while a careful analysis of the coefficient $\binom{-1/2}{t} = {1\over 2^{2t}} \binom{2t}{t}$ reveals that the constant hidden in the big-$\Theta$ is ${1\over\pi}$. 
\end{proof}

In \Cref{sec:ch2-empirical}, we also verify empirically that the max-loss-optimal Toeplitz mechanism nearly matches the dense mechanism in terms of empirical performance.

\begin{bremark}[Lower Triangular and Toeplitz Factorizations]
\sloppy    Recall from \Cref{remark:lower-triangular-factorizations} that we restricted ourselves to lower triangular factorizations because given any factorization $\bfB \strategy = \workload$, there exists a lower triangular factorization $\bfB' \strategy' = \workload$ such that $\maxlossn(\bfB, \strategy) = \maxlossn(\bfB', \strategy')$. However, this construction does not preserve the Toeplitz structure of the matrices. That is, optimal lower triangular and Toeplitz factorization (\cref{eq:max-err-optim-toeplitz}) may attain worse normalized \txtmaxloss than
    the optimal Toeplitz factorization \emph{without} the lower triangular constraint. Fortunately, we know these are close from \Cref{thm:toeplitz-opt-max-err} as the latter always lies in the interval $[\opt, \optToep]$, and $\optToep - \opt \le 1$ (independent of $n$).
\end{bremark}

\paragraph{Column Normalization to Improve the Mechanism}
We discuss a heuristic known as column normalization that can further improve the \txtmaxloss of the Toeplitz mechanism.
It is motivated by the observation 
that optimal dense $\strategy^\star$ from \Cref{thm:dense-opt-max-err} generally have equal column norms:

\begin{blemma} \label[lemma]{lem:column-norm}
    There exists optimal solutions $\bfB^\star, \strategy^\star$ to the dense mechanism satisfying \cref{eq:optimizationbarE} (from \Cref{thm:dense-opt-max-err}) such that $\strategy^\star$ is column-normalized, i.e. $\norm{\strategy^\star[:, t]}_2 = \colnorm{\strategy^\star}$ for all $t \in [\nd]$.
\end{blemma}

Unfortunately, Toeplitz matrices $\strategy$ (including those defined by \cref{eq:Toeplitz-C}) cannot in general satisfy this property. This can be remedied by column normalization:

\begin{bdefinition}[Column Normalization]
\label{def:col-norm}
    Given a correlated noise mechanism based on a factorization $\workload = \bfB\strategy$ with  $\strategy$ invertible, its column normalized version is given by the factorization $\workload = \bfB_{\mathsf{norm}} \strategy_{\mathsf{norm}}$ with
    \[
        \strategy_{\mathsf{norm}}[:, t] = \frac{\strategy[:, t]}{\norm{\strategy[:, t]}}_2 \,,
        \quad\text{and}\quad
        \bfB_{\mathsf{norm}} = \workload \strategy_{\mathsf{norm}}^{-1} \,.
    \]
\end{bdefinition}

As we discussed in \Cref{thm:toeplitz-opt-max-err}, the suboptimality of the original max-loss-optimal Toeplitz mechanism is quite small. Yet, it can be shown that this mechanism can be improved slightly by column normalization at the same time and memory cost. (Note that the resulting $\strategy_{\mathsf{norm}}$ matrix is no longer Toeplitz.)
\begin{btheorem}
\label{thm:normalized-opt-max-err}
    Let $\optNormalized = \maxlossn(\bfB_{\mathsf{norm}}, \strategy_{\mathsf{norm}})$ denote the normalized \txtmaxloss of the column normalized version  $ \strategy_{\mathsf{norm}}$ of the max-loss-optimal Toeplitz mechanism's strategy matrix $\prefix^{1/2}$ and its corresponding factor $\bfB_{\mathsf{norm}}=\prefix \strategy_{\mathsf{norm}}^{-1}$. We have, 
    \[
         \optNormalized \le \frac{\ln(n)}{\pi} + 1 \,,
    \]
\end{btheorem}
In particular, this removes an extra $\gamma/\pi$ term in  \Cref{thm:toeplitz-opt-max-err}, matching the upper bound of the dense mechanism in \Cref{thm:dense-opt-max-err}.

\paragraph{Summary}
The max-loss-optimal Toeplitz mechanism gives a tight \emph{additive approximation} to $\opt$: it has the right asymptotic dependence of $\ln n$ including the leading multiplicative constant of $1/\pi$. Its space complexity is also the best possible. The main drawback of this mechanism is the time complexity of noise generation. The next two mechanisms attain a better time complexity of noise generation, at the cost of increased space complexity.

\section{Banded Toeplitz Mechanism}
\label{sec:ch2-banded-Toeplitz}

The banded Toeplitz mechanism reduces the per-step cost of noise generation by requiring that the strategy matrix $\strategy$ be \emph{sparse} (in addition to lower triangular and Toeplitz) to allow for efficient noise generation (as we see in the upcoming \Cref{alg:banded-mult-by-cinv}).
Since the optimal Toeplitz coefficients from \Cref{thm:optimal-toeplitz} are monotonically decreasing, it is natural to require that only the first $b$ Toeplitz coefficients are non-zero. Such matrices are a special instance of the class of \emph{banded matrices}:
\begin{bdefinition}[$b$-Banded Matrix]
\label{def:banded}
    A lower triangular matrix $\bfM$ is said to be \emph{$b$-banded} if $\bfM[t, \tau] = 0$ for all $t-\tau \ge b$.\footnotemark
\end{bdefinition}
\footnotetext{
    Since $\bfM$ is lower triangular we have that $\bfM[t, \tau] = 0$ for all $t < \tau$ as well.
}

This banded structure lends itself to more efficient noise generation algorithms, as we will momentarily see.

\paragraph{Mechanism Definition}
The banded Toeplitz mechanism aims to find the factorization with the smallest \txtmaxloss subject to the band sparsity of the strategy matrix and Toeplitz constraints:
\begin{align} \label{eq:max-err-banded-toeplitz}
    \min
    \left\{
        \rownorm{\bfB} \colnorm{\strategy} \, :\, 
        \begin{matrix}
        \bfB \strategy = \workload, \quad
        \bfC[t, \tau] = 0 \,\, \forall \, t-\tau \ge b\,,
        \\
        \bfB, \strategy \text{  lower-triangular \& Toeplitz}
        \end{matrix}
    \right\} \,.
\end{align}

For example, for $\nd = 5$ and $\minsep=3$, such a mechanism can be defined by 3 parameters $(c_0, c_1, c_2)$ as
    \[
    \strategy = \begin{bmatrix}
    c_0 &   0 &   0 &   0 &   0 \\
    c_1 & c_0 &   0 &   0 &   0 \\
    c_2 & c_1 & c_0 &   0 &   0 \\
    0   & c_2 & c_1 & c_0 &   0 \\
    0   &   0 & c_2 & c_1 & c_0 \\
    \end{bmatrix} \,.
    \]

Similar to the optimal factorization problem of \cref{eq:max-err-optim}, this is a non-convex optimization problem; we return to the optimization aspect in \Cref{chap:practical}. We assume for now that a suitable factorization is available.

\begin{algorithm}[t]
\caption{Noise Generation with the Banded Toeplitz Mechanism}
\label{alg:banded-mult-by-cinv}
\begin{algorithmic}[1]
\Require{A $b$-banded lower-triangular Toeplitz matrix $\strategy$ whose first column has non-zero entries $c_0, c_1, \ldots, c_{b-1}$, i.i.d. noise $\Znoise \in \R^{n \times m}$.}
\Ensure{$\corrznoise_t = (\strategy^{-1} \Znoise)[t, :]$ for each $t$.}

\For{$t = 0, \ldots, n-1$}
\State Define $\znoise_t = \Znoise[t, :] \in \R^m$ and \label{line:banded-update}
$$\corrznoise_t = \znoise_t - \frac{1}{c_0} \left(
\sum_{\tau=1}^{\min\{t, b-1\}} c_\tau \corrznoise_{t-\tau}
\right) $$  

\State \textbf{Yield} correlated noise $\corrznoise_t$ 
\EndFor
\end{algorithmic}
\end{algorithm}

\paragraph{Complexity of Noise Generation}
A $b$-banded and lower-triangular linear system can be solved efficiently: each output element can be sequentially computed in $O(b)$ time, independent of the size $n$ of the problem. The same approach can be adapted for the noise generation $(\strategy^{-1}\Znoise)[t, :]$, as show in \Cref{alg:banded-mult-by-cinv}.
In each step $t$ of the algorithm, we need to maintain the $b-1$ previous outputs $\corrznoise_1, \ldots, \corrznoise_{t-b+1}$, leading to a space complexity of $O(mb)$. Each update in Line~\ref{line:banded-update} simply takes a linear combination of these $b$ previous outputs, so its time complexity is $O(mb)$ as well.

\begin{figure}
    \centering
    \includegraphics[width=4in]{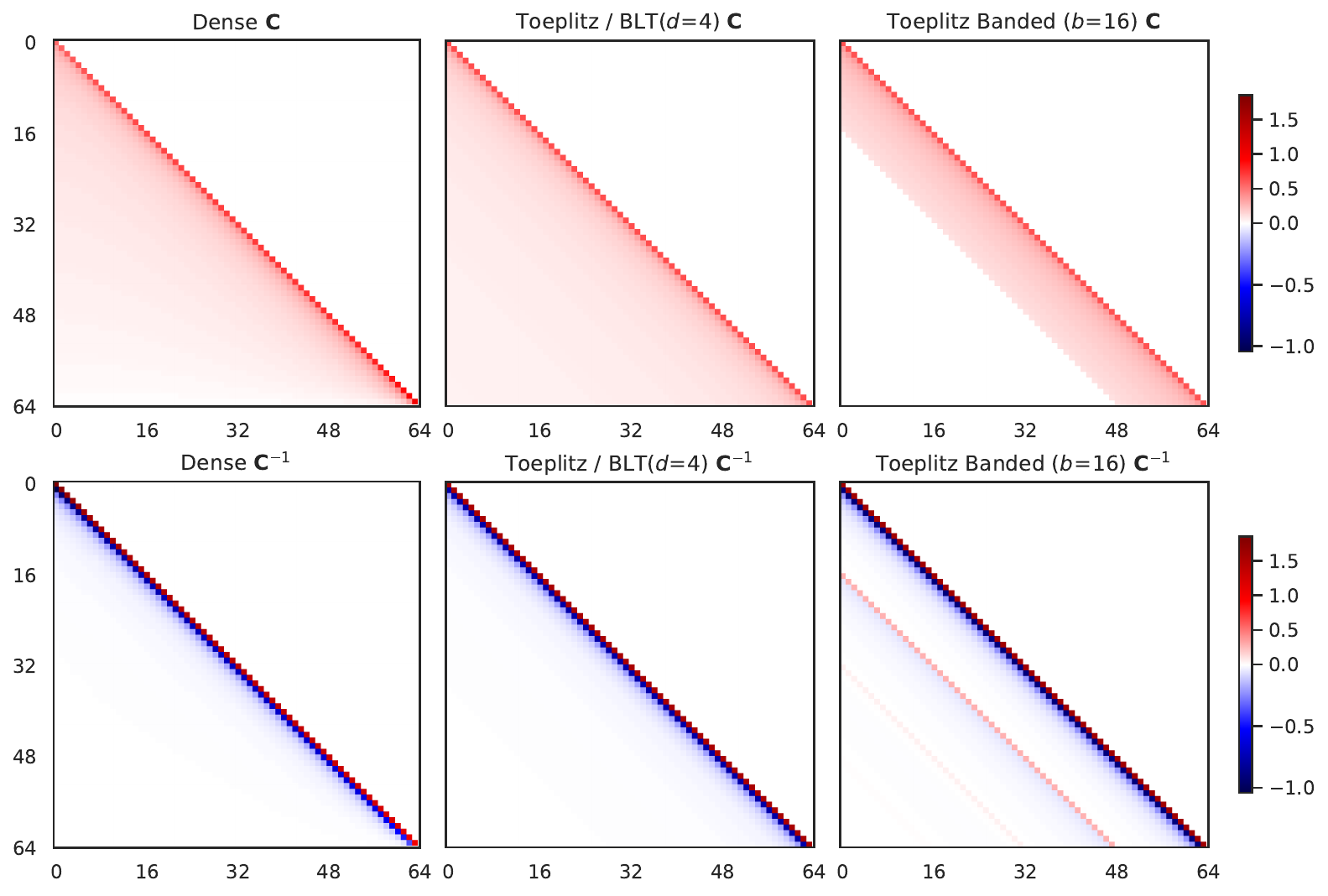}
    \caption{Examples of mechanisms as defined by $\big(\bfC, \bfC^{-1}\big)$ pairs for $n=64$: \textbf{(Left)} A dense (arbitrary lower-triangular) matrix (\cref{sec:optimized-dense-mech}), optimized for RMSE (see \cref{sec:prefix-L2-error}); \textbf{(Middle)} The max-loss-optimal Toeplitz mechanism of \cref{thm:optimal-toeplitz}, which is extremely well approximated (to the point of visual indistinguishability in this plot) by a BLT of order $d=4$; \textbf{(Right)} A banded Toeplitz mechanism optimized for \txtmaxloss (\cref{sec:ch2-banded-Toeplitz}). Optimizing mechanisms for scenarios with multiple participations can produce substantially different mechanisms, see \cref{fig:multi-participation-heatmaps}.}
    \label{fig:heatmaps}
\end{figure}

\paragraph{Utility and Error Bounds}
As previously, we focus on the unweighted prefix sum workload $\prefix$ for utility bounds.
While precise \txtmaxloss bounds for the solution of the optimization problem from \cref{eq:max-err-banded-toeplitz} have not been established (to the best of our knowledge), we can quantify the error of a feasible solution which is obtained by ``sparsifying'' the optimal Toeplitz factorization:

\begin{btheorem}
\label{thm:max-err-banded}
Fix the number of steps $n$ and the number of bands $1 \le b < n$.
Define the matrix $\widehat \strategy_{\mathsf{band}} \in \R^{n\times n}$ to be the lower-triangular and Toeplitz matrix whose first column is made up of $c_0^\star, c_1^\star, \ldots, c_{b-1}^\star, 0, \ldots, 0$, 
where $c_t^\star$ is the optimal Toeplitz coefficient defined in \cref{eq:opt-toeplitz-c} of \Cref{thm:optimal-toeplitz}.
Letting $\widehat \bfB_{\mathsf{band}} = \prefix \widehat \strategy_{\mathsf{band}}^{-1}$, we have
\[
    \maxlossn(\widehat \bfB_{\mathsf{band}}, \widehat \strategy_{\mathsf{band}}) \le O\left(
        \sqrt{\left(\frac{n }{b} - 1 \right) \ln b + \ln^2 b}
    \right) \,.
\]
\end{btheorem}

The above bound shows that the error decreases monotonically with an increasing number of bands $\minsep$.
In particular, $\minsep=\nd$ bands are optimal in the streaming setting, recovering the max-loss-optimal Toeplitz mechanism of \Cref{sec:optimial-toeplitz-mech}.
It is more practical to consider a small number of bands such as $b = O(1)$ or $b = O(\poly\ln(n))$. 
Unfortunately, the \txtmaxloss bound is suboptimal in this regime; the bound is comparable to the $\Theta(\sqrt{n})$ bound obtained by the baseline mechanisms and is exponentially worse than the optimal $\Theta(\ln n)$ scaling of \Cref{thm:dense-opt-max-err}.

\begin{bremark}[Bandedness and Amplification by Sampling*]
\label{remark:banded-advantages}
    While the banded Toeplitz mechanism appears to be suboptimal in terms of the \txtmaxloss, it has two key advantages in the learning setting. 
    First, the banded Toeplitz mechanism allows for significantly improved privacy guarantees via amplification by sampling, as we discuss in \cref{sec:learning:amplification}. 
    Second, the banded Toeplitz mechanism allows for a natural and interpretable structure in the multi-participation setting. In both these cases, it is advantageous to take $\bands < n$---we discuss these factors further in \Cref{chap:ml,chap:practical} (in particular, \Cref{fig:scaling_batch_size_bands} shows the empirical optimal number of bands). These reasons make the banded Toeplitz mechanism a compelling option in practice. We refer to \Cref{sec:recommendations} for rules of thumb regarding the selection of different mechanisms.
\end{bremark}

\paragraph{Summary}
While the banded Toeplitz mechanism improves the time complexity of noise generation from $O(mn)$ of the max-loss-optimal Toeplitz mechanism to $O(mb)$, it comes at the cost of a (potentially) exponentially worse \txtmaxloss. This would suggest that this mechanism requires a large number of bands $b$ to attain competitive empirical performance in the streaming setting, which in turn could make the $O(mb)$ space complexity prohibitively large.
The next mechanism overcomes these limitations.

\section{Buffered Linear Toeplitz (BLT) Mechanism} \label{sec:ch2-blt}

The Buffered Linear Toeplitz (BLT) mechanism takes a different approach to improving the time complexity of the max-loss-optimal Toeplitz mechanism in \Cref{sec:optimial-toeplitz-mech} while maintaining its near-optimal \txtmaxloss.

We develop some intuition.
Suppose the matrix $\strategy^{-1}$ takes the form
\[
    \strategy_\lambda^{-1} = \begin{pmatrix}
        1 & 0 & \cdots & 0  \\
        -\lambda & 1 & \cdots & 0\\
        \vdots & \ddots & \ddots & \vdots \\
        -\lambda^{n-1} & -\lambda^{n-2} & \cdots & 1
    \end{pmatrix} 
\]
for a parameter $\lambda \in [0, 1)$.\footnote{
    This is related to the one-parameter Toeplitz matrix considered in \cref{eq:oneBuffertoeplitzC}. The difference is that we parameterized the strategy matrix $\strategy$ with an exponentially decaying sequence in \Cref{chap:intro}, while we now parameterize $\strategy^{-1}$. We will see in Lemma~\ref{lem:inverse-blt} that this relationship can be made precise.
}

The exponential decay in the diagonals of $\strategy^{-1}_\lambda$ leads to an efficient recursive implementation of the noise generation. In particular, we can compute $\corrznoise_t = (\strategy_\lambda^{-1} \Znoise)[t, :]$ from $\znoise_t = \Znoise[t, :]$ recursively as 
\begin{align} \label{eq:recursive-noise-gen:simple-example}
    \corrznoise_0 = \znoise_0
    \quad \text{and} \quad
    \corrznoise_t = \znoise_t - \lambda \corrznoise_{t-1}  \,.
\end{align} 
Unfortunately, the optimal Toeplitz coefficients of $\bfC^{-1}$, which scale as $c'_0 = 1$ and $c'_t \approx - t^{-3/2}$ for $t > 1$ for the case of $\workload=\prefix$ (from \Cref{thm:optimal-toeplitz}), do not admit a similar efficiently-implementable recursion.\footnote{
    In particular, there exist no finite collection of numbers $q_1, \ldots, q_d\in \R$ for any $d < \infty$ such that $\sum_{i=0}^{d} q_i c'_{t-i} = 0$ holds (with $q_0 = 1$) for all $t \ge d$. This characterizes the general class of linear recurrences, for which an efficiently implementable recursion like \cref{eq:recursive-noise-gen:simple-example} exists.
    We return to this in \Cref{sec:approx-theory}.
} 

The BLT mechanism of order $d$ instead \emph{approximates} the optimal Toeplitz coefficients $c_t^\star$ (e.g. from \Cref{thm:optimal-toeplitz} for the unweighted prefix sum workload $\prefix$) with a linear combination of exponentials:
\begin{align} \label{eq:blt-approx}
c_t^\star \approx \sum_{i=1}^d \alpha_i \lambda_i^{t-1} 
\end{align}
for $t > 0$ using some scale parameters $\alpha_1, \ldots, \alpha_d$ and decay parameters $\lambda_1, \ldots, \lambda_d \in [0, 1)$.

There are two reasons to utilize this approximation.
First, it turns out that the BLT parameterization from \cref{eq:blt-approx} can effectively approximate decreasing non-negative sequences, including the optimal Toeplitz coefficients $c_t^\star$. We briefly discuss this in \Cref{remark:blt-fourier} below and present a more detailed intuition in \Cref{sec:approx-theory}.
Secondly, the BLT parameterization allows an efficient recursion for noise generation $(\strategy^{-1} \Znoise)[t, :]$ that can be implemented (as shown in the upcoming Lemma \ref{lem:blt-noise-gen}) with a per-step time complexity and total space complexity of $O(\mdim d)$. This can be viewed as $d$ buffers of the same size $\mdim$ as the model parameters.
In practice, a small constant order $d$ (e.g. $d<5$) can provide near-optimal performance empirically (see also \Cref{sec:ch2-empirical}), so this is effectively only a small constant-factor overhead of storing and computing on the model of size $\mdim$.

\begin{bremark}[BLT and Fourier Approximations*]
\label{remark:blt-fourier}
    For readers familiar with the Fourier transform, it is instructional to view its parallels with the BLT parameterization from \cref{eq:blt-approx}.
    Any sequence $(c_t)_{t=0}^\infty$ can be expressed in the Fourier basis $\omega \mapsto \big(\exp(\iota \, \omega t)\big)_{t=0}^\infty$ (where $\iota = \sqrt{-1}$ is the imaginary unit) as
    \[
        c_t = \frac{1}{2\pi} \int\limits_{-\pi}^\pi F_c(\omega) \exp( \iota \, \omega t) \,\D t\,,
    \]
    where $F_c: [-2\pi, 2\pi] \to \C$ is the discrete-time Fourier transform of the sequence $(c_t)_{t=0}^\infty$.
    It is common practice to approximate this integral as a sum over $d$ points $\omega_1, \ldots, \omega_d \in \C$ with weights $\beta_1, \ldots, \beta_d \in \C$:
    \begin{align} \label{eq:prony}
        c_t \approx \sum_{i=1}^d \beta_i \exp(\iota \, \omega_i t) \,.
    \end{align}
    The BLT approximation in \cref{eq:blt-approx} is analogous to this, with the key distinction that is a linear combination of real and decreasing exponential functions $\exp(-\mu_i t)$ (where $\mu_i = \ln(1/\lambda_i) > 0$), making it suitable to approximate decreasing functions.
\end{bremark}

\paragraph{Mechanism Definition}
A BLT mechanism of order-$d$ is parameterized by a scale parameter $\bfalpha = (\alpha_1, \ldots, \alpha_d) \in \R^d$
and a decay parameter $\bflambda = (\lambda_1, \ldots, \lambda_d) \in [0, 1)^d$. It represents the Toeplitz matrix $\strategy \in \real^{\nd \times \nd}$ as
\begin{align} \label{eq:blt-param}
    \BLT(\bfalpha, \bflambda)
    \coloneqq \begin{pmatrix}
    1 & 0 & \cdots & \cdots & 0 \\
    \sum_{i=1}^d \alpha_i & 1 & \cdots & \cdots & 0  \\
    \sum_{i=1}^d \alpha_i \lambda_i  & \sum_{i=1}^d \alpha_i & 1 & \cdots & \vdots \\ 
    \vdots  & \vdots & \vdots & \ddots & \vdots \\
    \sum_{i=1}^d \alpha_i \lambda_i^{n-2}  & \sum_{i=1}^d \alpha_i \lambda_i^{n-3}  & \sum_{i=1}^d \alpha_i \lambda_i^{n-4} & \cdots & 1 \\ 
    \end{pmatrix}  \,.
\end{align}

With slight abuse of nomenclature, we refer to matrices $\BLT(\bfa, \bflambda)$ parameterized in this fashion as ``BLT matrices''.  The best parameters $\bfa, \bflambda$ can be found numerically as solutions to the optimization problem:
\begin{align} \label{eq:max-err-BLT}
    \min
    \left\{
        \rownorm{\bfB} \colnorm{\strategy} \, :\, 
        \begin{matrix}
        \strategy = \BLT(\bfalpha, \bflambda), \,\, \bfB = \workload \strategy^{-1}, \\
        \bfalpha \in \R^{d}_+, \bflambda \in [0, 1)^d
        \end{matrix}
    \right\} \,,
\end{align}
where $\R_+$ denotes the set of positive real numbers. 
Note that we constrain the scale parameters $\bfalpha$ to be non-negative in Problem~\eqref{eq:max-err-BLT}. This is because the optimal Toeplitz coefficients $c_t^\star$ we wish to approximate are non-negative for typical workload matrices $\workload$ encountered in machine learning. For instance, we see from \Cref{thm:optimal-toeplitz} that this is true for the unweighted prefix sum workload $\prefix$.

While the optimization problem \eqref{eq:max-err-BLT} is non-convex, it turns out that we can obtain empirically high-quality solutions within a few seconds for $n$ up to several billions.  
We return to the optimization aspect in \Cref{chap:practical} and assume for now that suitable parameters $\bfalpha, \bflambda$ are available.

\paragraph{Noise Generation}
BLT noise generation is efficiently implementable via a suitable recursion. While  we wish to compute\footnote{
    The matrix $\BLT(\bfalpha, \bflambda)$ is invertible. Indeed, as it is a lower triangular matrix with all ones along the diagonal, we have that all of its eigenvalues equal 1.
} 
$(\strategy^{-1} \Znoise)[t, :]$ for $\strategy = \BLT(\bfalpha, \bflambda)$, it is instructive to first consider how to compute $\strategy[t, :] \Znoise$ for $t \in [\nd]$. Recalling $\znoise_t = \Znoise[t, :]$ is the $t$\textsuperscript{th} row of $\Znoise$, we have
\begin{align} \label{eq:blt-mult-by-C}
    \strategy[t, :] \Znoise
    = \znoise_t + \sum_{\tau=1}^{t} \left(\sum_{i=1}^d \alpha_i \lambda_i^{\tau-1}\right) \znoise_{t-\tau}
    = \znoise_t + \sum_{i=1}^d \alpha_i \bfm_{t, i}
\end{align}
where we have $d$ memory buffers $\bfm_{t,i}$ for $i \in 1, \dots d$ given by $\bfm_{0, i} \coloneqq \zeros$ and for $t\ge 1$, 
and \[
\bfm_{t,i} \coloneqq \sum_{\tau=1}^t \lambda_i^{\tau-1} \znoise_{t-\tau} 
= \znoise_{t-1} + \lambda_i \znoise_{t-2} + \dots + \lambda_i^{t-1} \znoise_0 \in \R^{m}.\]
The key to an efficient implementation is the following recursion akin to \cref{eq:recursive-noise-gen:simple-example}:
\[
     \bfm_{0, i} = \zeros 
     \quad\text{and}\quad
     \bfm_{t+1, i} = \znoise_t + \lambda_i \bfm_{t, i} \,.
\]

This recipe can directly be used to generate the noise $(\strategy^{-1} \Znoise)[t, :]$ since $\strategy^{-1}$ exists and is also BLT: 
\begin{blemma}[Inverse BLT parameterization] 
\label[lemma]{lem:inverse-blt}
    \sloppy Any matrix $\strategy=\BLT(\bfalpha, \bflambda) \in \R^{n \times n}$ that BLT-parameterized matrix is invertible. 
    Further, suppose the scale parameters are positive (i.e. $\alpha_i > 0$ for all $i$) with $\sum_{i=1}^d \alpha_i < 1$, and the decay parameters $\bflambda \in (0, 1)^d$ are unique (i.e. $\lambda_i \neq \lambda_j$ for $i \neq j$). 
    Then, there exist parameters $\widehat \bfalpha \in \R^d$ and $\widehat \bflambda \in [-1, 1]^d$ such that $\strategy^{-1} = \BLT(\widehat\bfalpha, \widehat\bflambda)$. Furthermore,
    the map $(\bfalpha, \bflambda) \mapsto (\widehat\bfalpha, \widehat\bflambda)$ is continuously differentiable and can be implemented in $O(d^3)$ time.
\end{blemma}

The scale parameter $\bfalpha$ of the BLT is positive element-wise while the corresponding parameter $\widehat \bfalpha$ of the inverse BLT is element-wise negative. This mimics the optimal Toeplitz coefficients of \Cref{thm:optimal-toeplitz}: we have that the coefficients $c_t^\star = \Theta(t^{-1/2})$ of the Toeplitz matrix $\strategy_\mathsf{Toep}$ are positive, while the corresponding coefficients $c_t' = - \Theta(t^{-3/2})$ of $\strategy_\mathsf{Toep}^{-1}$ are negative for $t \ge 1$.

The parameters $\widehat\bfalpha, \widehat\bflambda$ of the inverse BLT can be computed using an eigenvalue decomposition of a non-symmetric $d \times d$ matrix, a common sub-routine in numerical software. Therefore, Lemma~\ref{lem:inverse-blt} allows us to use \cref{eq:blt-mult-by-C} for efficient noise generation. 

Alternatively, instead of deriving the BLT parameters for $\strategy^{-1}$, we can directly compute $(\strategy^{-1}\Znoise)[t, :]$ from the BLT parameterization of $\strategy$. This approach, given as \Cref{alg:blt-mult-by-cinv}, may be preferable in practice as it avoids potential numerical issues in the computation of the inverse parameters.

\begin{algorithm}[t]
\caption{Noise Generation with BLTs}
\label{alg:blt-mult-by-cinv}
\begin{algorithmic}[1]
\Require{Degree-$d$ BLT $\bfC = \BLT(\bfalpha, \bflambda)$, i.i.d. noise $\Znoise \in \R^{n \times m}$.}
\Ensure{$\corrznoise_t = (\strategy^{-1} \Znoise)[t, :]$ for each $t$.}
\State Initialize buffer $\bfM_{0} = \zeros_{m \times d}$

\For{$t = 0, \ldots, n-1$}
\State  \label{line:blt-noise-gen}
$\corrznoise_t = \znoise_t - \bfM_{t} \bfalpha$ with $\znoise_t = \Znoise[t, :] \in \R^m$
\Comment{Noise generation}
\State \label{line:blt-buffer-update}
$\bfM_{t+1} = \bfM_{t} \bfLambda + \corrznoise_t \ones_d^\top$ with $\bfLambda = \diag(\bflambda)$
    \Comment{Buffer update}
\State \textbf{Yield} correlated noise $\corrznoise_t$ 
\EndFor
\end{algorithmic}
\end{algorithm}
\begin{blemma}[Correctness of BLT noise generation]
\label[lemma]{lem:blt-noise-gen}
    The output $\corrZnoise = (\corrznoise_0, \ldots, \corrznoise_{n-1}) \in \R^{n \times m}$ of \Cref{alg:blt-mult-by-cinv} with inputs $\strategy = \BLT(\bfalpha, \bflambda)$ and a matrix $\Znoise \in \R^{n \times m}$
    satisfies $\corrZnoise = \strategy^{-1} \Znoise$.
\end{blemma}
\begin{proof}[Proof Sketch]
We start by noting that the memory buffer $\bfM_t$ satisfies
\[
    \bfM_t = \begin{pmatrix}
    \sum_{\tau=1}^t \lambda_1^{\tau-1} \corrznoise_{t-\tau}
    &
    \cdots & 
    \sum_{\tau=1}^t \lambda_d^{\tau-1} \corrznoise_{t-\tau}
    \end{pmatrix} \in \R^{m \times d} \,.
\]
This can be proved, for instance, by induction.
Next, we use this formula together with Line~\ref{line:blt-noise-gen} of \Cref{alg:blt-mult-by-cinv} to expand out
\begin{align*}
    \znoise_t &= 
    \corrznoise_t + \bfM_{t} \bfalpha
    = \corrznoise_t + \sum_{i=1}^d \alpha_i
    \sum_{\tau=1}^{t}  \lambda_i^{\tau-1} \corrznoise_{t-\tau} \,.
\end{align*}
Comparing this with \cref{eq:blt-mult-by-C} gives $\Znoise = \strategy \corrZnoise$. Thus, we have $\corrZnoise = \strategy^{-1} \Znoise$ and \Cref{alg:blt-mult-by-cinv} gives the correctly correlated noise.
\end{proof}

\paragraph{Complexity of Noise Generation}
Noise generation with a $d$-buffer BLT requires $O(md)$ space to store the memory buffer $\bfM_t$ of \Cref{alg:blt-mult-by-cinv}. Its per-step time-complexity is $O(md)$ to generate the correlated noise and update the buffer. 
It typically suffices to take $d \ll n$ to get competitive performance (see the error bound below).
This is a huge improvement in the running time of $O(mn)$ of the max-loss-optimal Toeplitz mechanism and the banded Toeplitz mechanism.

\paragraph{Utility and Error Bounds}
As previously, we give utility bounds for the unweighted prefix sum workload $\workload=\prefix$.
It turns out to be sufficient to take $d=\poly(\ln(n))$ to get a competitive approximation guarantee:

\begin{btheorem}
\label{thm:blt-max-err}
    Fix the number of steps $n$, and error term $\delta>0$. There exists a $d$-buffer BLT matrix $\strategy_\bltlower$ with $d = O\big(\ln^2(n / \delta)\big)$ and its corresponding factor $\bfB_\bltlower = \prefix \strategy_\bltlower^{-1}$ such that
    \[
        \maxlossn(\bfB_\bltlower, \strategy_\bltlower) \le \optToep + \delta \,,
    \]
    where $\optToep = \maxlossn(\prefix^{1/2}, \prefix^{1/2})$ is the optimal normalized \txtmaxloss achievable by any Toeplitz factorization as defined in \Cref{thm:toeplitz-opt-max-err}.
\end{btheorem}

\Cref{thm:blt-max-err} implies that $\maxlossn(\bfB_\bltlower, \strategy_\bltlower) \le \ln(n) / \pi \, +\, \mathsf{constant}$ achieves the optimal asymptotic rate of $\ln(n)$ and optimal leading constant $1/\pi$. To get this \emph{additive} approximation on the optimal \txtmaxloss $\opt$, the explicit construction used in \Cref{thm:blt-max-err} requires $d=O(\ln^2(n))$ buffers, leading to a space and time complexity of $O(m \ln^2(n))$. However, directly optimizing $\strategy = \BLT(\bfalpha, \bflambda)$ for a specific $n$ can in practice produce BLTs with equivalent performance and substantially smaller number $d$ of dimensions. We discuss the direct optimization of BLTs in \Cref{sec:blt-opt}.

\begin{bremark}[Column Normalization of BLTs]
Column normalization, defined in \Cref{def:col-norm}, can also yield improvements to the \txtmaxloss BLT mechanism. In particular, given a fixed number of steps $n$, the efficient noise generation approach of \Cref{alg:blt-mult-by-cinv} can be extended to column-normalized BLTs at the same time and space complexity. We leave the details as an exercise to the reader.
\end{bremark}

\paragraph{Summary}
The BLT mechanism attains a favorable tradeoff between noise generation complexity and utility. It is the only mechanism (as of this writing) that simultaneously admits an \emph{additive} \txtmaxloss guarantee and a $\poly(\ln(n))$ time and space complexity of noise generation.

\section{Empirical Comparison of the Mechanisms}
\label{sec:ch2-empirical}

\begin{figure*}[p]
    \centering
        \includegraphics[width=0.8\linewidth]{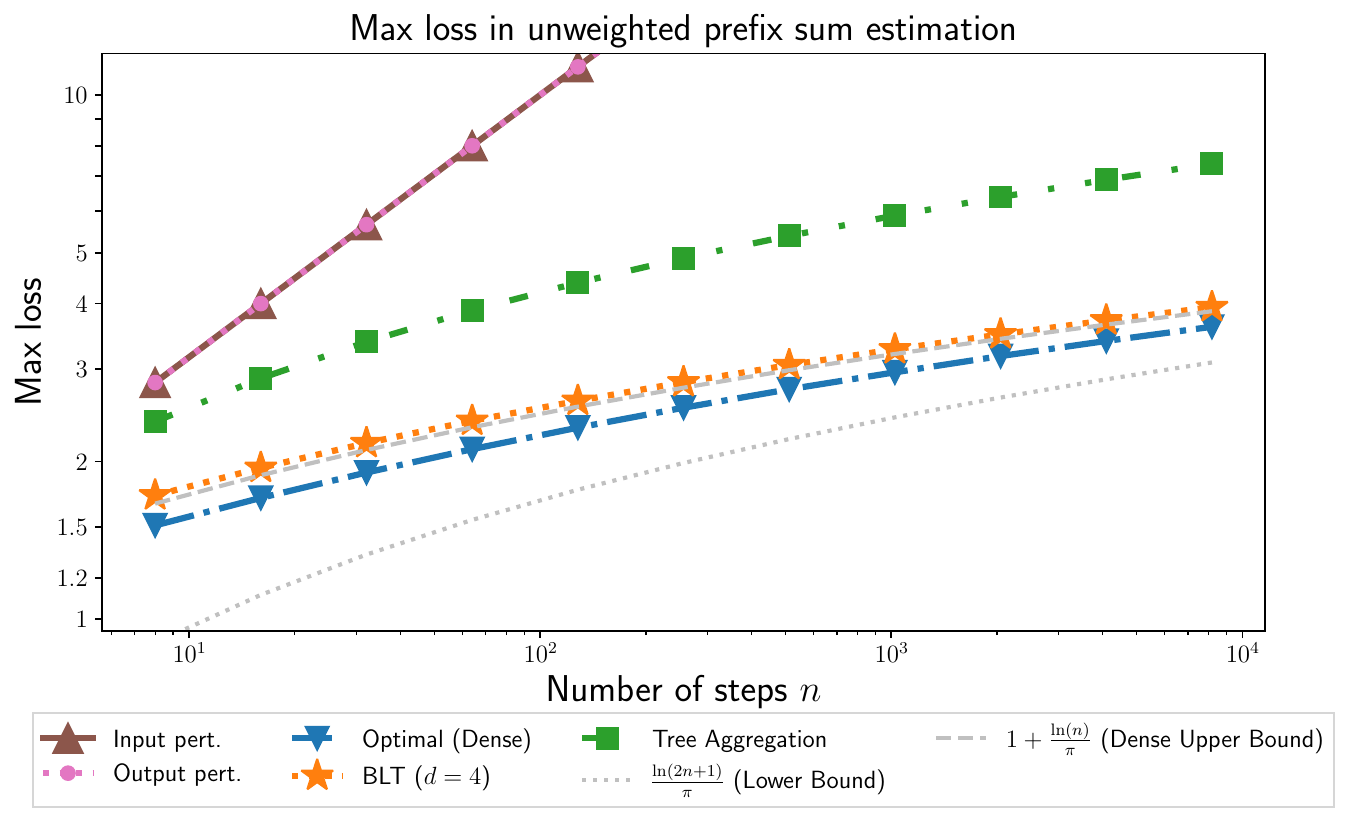}
        \includegraphics[width=0.8\linewidth]{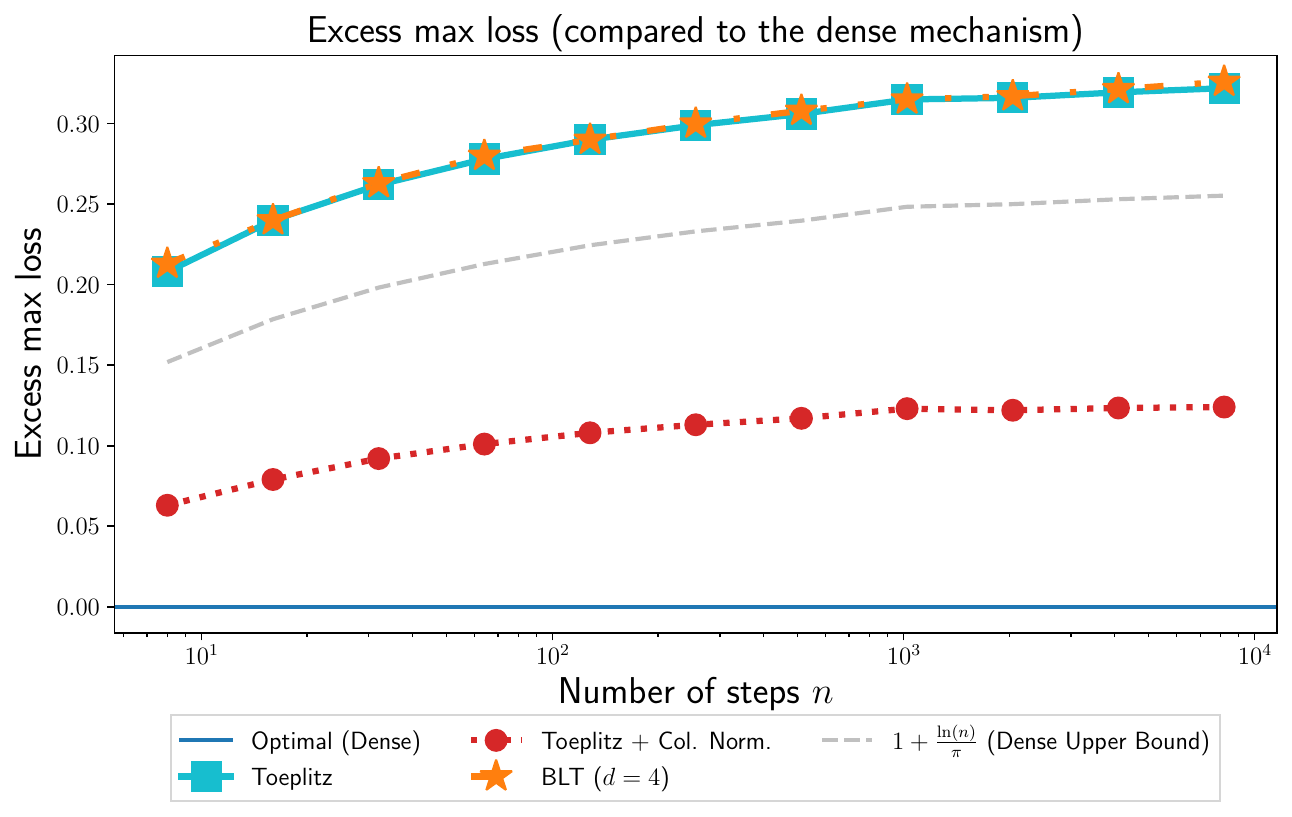}
    \caption{An empirical comparison of the \txtmaxloss for various mechanisms in the context of  streaming unweighted prefix sum estimation. 
    \textbf{Top}: A log-log plot of the \txtmaxloss of the baselines (input and output perturbation), tree aggregation (\Cref{sec:treeAgg}; optimal rate but suboptimal leading coefficient), BLT (\Cref{sec:ch2-blt}; 
    optimal rate and leading coefficient) with $d=4$ buffers, the optimal dense mechanism (\Cref{sec:optimized-dense-mech}). For reference, we also plot the lower/upper bounds on the dense \txtmaxloss from \Cref{thm:dense-opt-max-err}.  The BLT parameters are optimized for the \txtmaxloss (as described in \Cref{chap:practical}).
    \textbf{Bottom}:
    A comparison of the excess \txtmaxloss $\maxlossn(\bfB, \strategy) - \maxlossn(\bfB^\star, \strategy^\star)$ of mechanisms defined by the factorization $\prefix=\bfB\strategy$ compared to the dense mechanism $\prefix=\bfB^\star \strategy^\star$. Only mechanisms that attain the optimal leading constant (i.e., additive approximation guarantee), including column normalization (\Cref{def:col-norm}) are shown.
    See \Cref{sec:a:detailed-numerical} for the exact numerical values plotted here. Note the excess \txtmaxloss for all of these mechanisms is $o(n)$.
    }
    \label{fig:max-err-empirical}
\end{figure*}

We numerically compute the optimal strategy within each class of factorizations defined above, and report the \txtmaxloss for different values of the number of steps $n$ in the streaming setting in \Cref{fig:max-err-empirical}.\footnote{
    In the setting that we consider here, the recommended number of bands $b$ for the banded Toeplitz mechanism is equal to the number of steps $n$ (see also \Cref{remark:banded-advantages}), and hence banded Toeplitz mechanism of \Cref{sec:ch2-banded-Toeplitz} is equivalent to max-loss-optimal Toeplitz mechanism of \Cref{sec:optimial-toeplitz-mech}.
    Thus, we omit the banded Toeplitz mechanism from this comparison.}

In addition to the correlated noise mechanism described in this section, we also compare the \txtmaxloss\ of the so called {\em binary tree mechanism} (annotated as Tree Aggregation in \Cref{fig:max-err-empirical}), historically, the first mechanism for estimating prefix sums with $\poly\log(\nd)$ error and $\log(\nd)$ space and time requirement. This mechanism also has a correlated noise mechanism perspective. Since it is not the central aspect of this monograph, we cover it in more detail in \Cref{sec:treeAgg}.

The empirical findings closely mirror the theoretical \txtmaxloss bounds presented so far.
\begin{enumerate}
    \item First, we note the sub-optimality of the baselines from the top plot of \Cref{fig:max-err-empirical}. Indeed, the input/output perturbation lines are identical and have a slope of $0.5$ (up to an error of $10^{-5}$) in the log-log plot. This indicates a $\sqrt{n}$ \txtmaxloss scaling, as established.\footnote{
    The log-log plot of $f(n) = c\, x^a$ versus $n$ appears as a straight line with slope $a$.}
    
    \item Second, tree aggregation (\Cref{sec:treeAgg}) which achieves the optimal $O(\ln n)$ rate, but its leading constant is sub-optimal. This leads to significantly worse empirical performance than methods like the BLT mechanism that can attain an additive performance guarantee. Third, the BLT mechanism gives a tight approximation to the dense mechanism and its upper bound in the top plot. 
\end{enumerate}

We turn to the bottom plot of \Cref{fig:max-err-empirical} for more detail. Here, we see that the BLT mechanism's performance is nearly indistinguishable from the max-loss-optimal Toeplitz mechanism up to the resolution of this plot, while being computationally more efficient. We also note that column normalization improves the \txtmaxloss of the Topelitz mechanism, while maintaining the same running time.

In conclusion, the theoretical bounds presented in this section capture the true empirical behavior. For this streaming prefix sum setting, we recommend using the dense mechanism when its run-time overhead is not prohibitive, and the BLT mechanism otherwise.
We will revisit these recommendations in the context of training AI and machine learning models in \Cref{sec:recommendations}.

\section{Tree Aggregation*}
\label{sec:treeAgg}
\begin{figure}
\centering
\begin{adjustbox}{max width=0.9\linewidth}
\begin{tikzpicture}[every node/.style={circle, draw, fill=C0!20}, level/.style={sibling distance=60mm/#1}]
\node {$\bfs_{0,7}$}
  child {node {$\bfs_{0,3}$}
    child {node {$\bfs_{0,1}$}
      child {node {$\gradient_0$}}
      child {node {$\gradient_1$}}
    }
    child {node {$\bfs_{2,3}$}
      child {node {$\gradient_2$}}
      child {node {$\gradient_3$}}
    }
  }
  child {node {$\bfs_{4,7}$}
    child {node {$\bfs_{4,5}$}
      child {node {$\gradient_4$}}
      child {node {$\gradient_5$}}
    }
    child {node {$\bfs_{6,7}$}
      child {node {$\gradient_6$}}
      child {node {$\gradient_7$}}
    }
  };
\end{tikzpicture}    
\end{adjustbox}
\caption{The \textbf{tree aggregation} mechanism generates privatized prefix sums using a binary tree data structure.
The input vectors $\gradient_0, \ldots, \gradient_7$ are arranged as leaves of the binary tree and each non-leaf node contains the sum of the leaves within its sub-tree. For instance, $\bfs_{0, 1} = \gradient_0 + \gradient_1$, while $\bfs_{4, 7} = \gradient_4 + \cdots + \gradient_7$. The root node $\bfs_{0, 7} = \gradient_0 + \cdots + \gradient_7$ represents the sum of all input vectors.
The core idea behind this mechanism is that any prefix sum $\gradient_0 + \cdots + \gradient_{t-1}$ can be expressed as a sum of at most $\log_2(n)$ nodes. This is achieved using a dyadic partition of the interval $[0, t-1]$. For example, $\gradient_0 + \cdots + \gradient_6 = \bfs_{0, 3} + \bfs_{4, 5} + \gradient_6$ corresponds to the partition $[0, 6] = [0, 3] \cup [4, 5] \cup [6]$. 
Once we privatize each intermediate $\bfs_{i, j}$ with white Guassian noise $ \znoise_{i, j}$ as $\widehat \bfs_{i, j} = \bfs_{i, j} + \znoise_{i, j}$, we can compute any prefix sum by adding at most $\log_2(n)$ noise vectors.
}
    \label{fig:binarytree}
\end{figure}
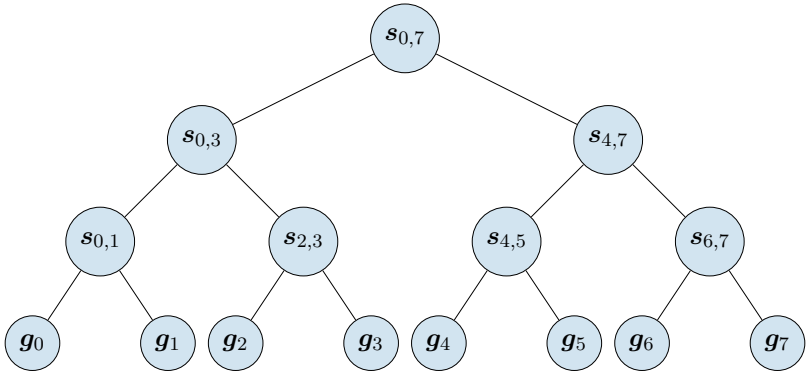

The tree aggregation mechanism alleviates the high space and time complexity of the optimized dense mechanism by leveraging \emph{sparsity}. Also known as the \emph{binary tree} mechanism, it constructs a factorization $\bfB_{\mathsf{tree}}  \strategy_{\mathsf{tree}}=\prefix$ with sparse rectangular matrices $\bfB_{\mathsf{tree}}  \in \{0,1\}^{n \times (2n-1)}$ and $\strategy_{\mathsf{tree}} \in \{0,1\}^{(2n-1) \times n}$ whose non-zero entries are given by a binary tree data structure, as illustrated in \Cref{fig:binarytree}.\footnote{
    While we construct a factorization with rectangular $\bfB_{\mathsf{tree}}, \strategy_{\mathsf{tree}}$ matrices, we can obtain square matrices $\bfB_{\mathsf{tree}}',  \strategy_{\mathsf{tree}}'$ by dropping some unused rows of $\strategy_{\mathsf{tree}}$ and columns of $\bfB_{\mathsf{tree}}$, as will see soon.
}

\paragraph{Mechanism Definition}
Consider an input sequence $\gradient_0, \ldots, \gradient_{n-1}$ where $\nd=2^k$ as the leaves of a (complete) binary tree.\footnote{This is just for the ease of presentation. For $n$ which is not a power of two, we can simply pad our input sequence with zero vectors to length $n' = 2^{\ceiling{\log_2(n)}}$
}.
Each intermediate node of the tree is assigned the sum of the leaves of its sub-tree. In the example of \Cref{fig:binarytree}, we have
\begin{align*}
 & \bfs_{0, 1} = \bfs_0 + \bfs_1, \quad \bfs_{2,3} = \bfs_2+\bfs_3, \\
& \bfs_{4,5}=\bfs_4+\bfs_5, \quad \bfs_{6,7}= \bfs_6 + \bfs_7, \\
& \bfs_{0, 3} = \sum_{t=0}^3 \gradient_t, \quad \bfs_{4,7} = \sum_{t=4}^7 \gradient_t, \quad \text{and} \quad \bfs_{0,7} = \sum_{t=0}^7 \gradient_t.
\end{align*}

Every intermediate node is of the form
$\bfs_{j2^\ell , (j+1)2^\ell-1}$ for some $j\ge 0$ and $\ell$ such that its sub-tree contains $2^l$ leaves. In fact, it represents a \emph{dyadic interval}: the $(\ell+1)$\textsuperscript{st} bit (from the right) in the binary representation of each of the leaves in the sub-tree is the same. 
For instance, the node $\bfs_{4, 5}$ corresponds to $\ell=1, j=2$, and its leaves are $4=(1\underline{0}0)_2$ and $5=(1\underline{0}1)_2$ with a common bit $\underline{0}$ (underlined). Similarly, the node $\bfs_{0, 3}$ corresponds to $\ell=2, j=0$ and $\bfs_{4, 7}$ corresponds to $\ell=2,j=1$.

The main idea underlying the binary tree mechanism is that any prefix sum $\gradient_0 + \cdots + \gradient_{t-1}$ can be expressed as a sum of at most $\log_2(n)$ nodes. This is achieved using a maximal \textit{dyadic partition} of the interval $[0, t-1]$. For example, $\gradient_0 + \cdots + \gradient_6 = \bfs_{0, 3} + \bfs_{4, 5} + \gradient_6$ corresponds to the partition $[0, 6] = [0, 3] \cup [4, 5] \cup [6]$. 

This procedure corresponds to a certain factorization $\bfB_{\mathsf{tree}} \strategy_{\mathsf{tree}} = \prefix$.
Indeed, the matrix $\strategy_{\mathsf{tree}} \in \{0, 1\}^{(2n-1) \times n}$ gives us all intermediate nodes in the tree, while the matrix $\bfB_{\mathsf{tree}} \in \{0, 1\}^{n \times (2n-1)}$ sums up the nodes forming a maximal dyadic partition of any prefix sum. 
To concretely write out these matrices, let us index the $2n-1$ nodes of the tree using a {\em postorder traversal}. For the example of \Cref{fig:binarytree}, this is
\[
    \gradient_0, \gradient_1, \bfs_{0, 1},  \gradient_2, \gradient_3, \bfs_{2,3}, \bfs_{0,3}, \gradient_4, \gradient_5, \bfs_{4,5} , \gradient_6 , \gradient_7 , \bfs_{6,7}, \bfs_{4,7} , \bfs_{0,7} \,.
\]
Denoting $\Gradients = (\gradient_0, \ldots, \gradient_{n-1}) \in \R^{n \times m}$ as the input matrix, we define $\strategy_{\mathsf{tree}}$ such that $(\strategy_{\mathsf{tree}} \gradient)[p, :]$ returns the value of the $p$\textsuperscript{th} node in the postorder traversal.
Then, we can express $\strategy_{\mathsf{tree}} = \bfP_k$ based on the recursive definition $\bfP_0 = (1) \in \R^{1 \times 1}$ and
\[
\bfP_{i+1} =
\begin{pmatrix}
\bfP_{i} & \zeros  \\
 \zeros & \bfP_{i} \\
\ones & \ones
\end{pmatrix} \,.
\]

Indeed, the two $\bfP_{i-1}$'s give the postorder traversal of the left and right sub-trees respectively, while the final row of ones produces the sum of the entire sub-tree, completing the postorder traversal.

Finally, we take $\bfB_{\mathsf{tree}}[t, p] = 1$ if the corresponding node $p$ in the postorder traversal is used in the computation of the prefix sum $\gradient_0 + \cdots + \gradient_{t-1}$.
For the example of \Cref{fig:binarytree}, we have (with the index nodes of the rows and columns shown in orange):
\begin{equation*}
\adjustbox{max width=0.99\linewidth}{$\displaystyle
\bfB =
\begin{blockarray}{c ccccccccccccccc}
& \myhighlight{\gradient_0} & \myhighlight{\gradient_1} & \myhighlight{\bfs_{0,1}} &  \myhighlight{\gradient_2} & \myhighlight{\gradient_3} & \myhighlight{\bfs_{2,3}} & \myhighlight{\bfs_{0,3}} & \myhighlight{\gradient_4} & \myhighlight{\gradient_5} & \myhighlight{\bfs_{4,5}}  & \myhighlight{\gradient_6}  & \myhighlight{\gradient_7}  & \myhighlight{\bfs_{6,7}} & \myhighlight{\bfs_{4,7}} & \myhighlight{\bfs_{0,7}} \\
\begin{block}{c(ccccccccccccccc)}
 \myhighlight{\gradient_0} & 1  &  0  &  0  &  0  &  0  &  0  &  0  &  0  &  0  &  0  &  0  &  0  &  0  &  0  &  0 \\
 \myhighlight{\gradient_1} & 0  &  0  &  1  &  0  &  0  &  0  &  0  &  0  &  0  &  0  &  0  &  0  &  0  &  0  &  0 \\
 \myhighlight{\gradient_2} & 0  &  0  &  1  &  1  &  0  &  0  &  0  &  0  &  0  &  0  &  0  &  0  &  0  &  0  &  0 \\
 \myhighlight{\gradient_3} &  0  &  0  &  0  &  0  &  0  &  0  &  1  &  0  &  0  &  0  &  0  &  0  &  0  &  0  &  0 \\
  \myhighlight{\gradient_4} & 0  &  0  &  0  &  0  &  0  &  0  &  1  &  1  &  0  &  0  &  0  &  0  &  0  &  0  &  0 \\
  \myhighlight{\gradient_5} & 0  &  0  &  0  &  0  &  0  &  0  &  1  &  0  &  0  &  1  &  0  &  0  &  0  &  0  &  0 \\
  \myhighlight{\gradient_6} & 0  &  0  &  0  &  0  &  0  &  0  &  1  &  0  &  0  &  1  &  1  &  0  &  0  &  0  &  0 \\
  \myhighlight{\gradient_7} & 0  &  0  &  0  &  0  &  0  &  0  &  0  &  0  &  0  &  0  &  0  &  0  &  0  &  0  &  1 \\
\end{block} 
\end{blockarray} \,\,\,\,.
$}  
\end{equation*}
The columns corresponding to some nodes such as $\Gradients_1$ and $\bfs_{2, 3}$ are all zeros because they never appear in any maximal dyadic partition; such nodes can be dropped to obtain a square factorization.

\paragraph{Complexity of Noise Generation}
Note that each row of $\bfB_{\mathsf{tree}}$ contains at most $ \ceiling{\log_2(\nd)}$ non-zeros, as this is the largest size of any maximal dyadic partition of $[0, t-1]$ for $1\leq t \leq \nd$. Thus, we can compute $(\bfB_{\mathsf{tree}} \Znoise)[t, :]$ in 
$O(m \, \ceiling{\log_2(n)})$ time
by summing up these non-zero noise vectors. This immediately gives us the correlated noise $\strategy_{\mathsf{tree}}^{-1}\Znoise$, since
\[
    (\strategy_{\mathsf{tree}}^{-1}\Znoise)[t, :] = (\prefix^{-1} \bfB_{\mathsf{tree}} \Znoise)[t, :] = (\bfB_{\mathsf{tree}} \Znoise)[t, :] - (\bfB_{\mathsf{tree}} \Znoise)[t-1, :] \,.
\]
This follows from the formula for $\prefix^{-1}$, which can be inferred from \cref{eq:cinv:one-param-family}.
Similarly, we need to materialize $\ceiling{\log_2(n)}$ noise vectors in $\R^m$, leading to a space complexity of $O(m \, \ceiling{\log_2(n)})$.

\paragraph{Utility and Error Bounds} 
The tree aggregation mechanism can attain the optimal error up to a multiplicative factor:

\begin{btheorem} \label{thm:treeAgg}
    For matrices $\bfB_{\mathsf{tree}}$ and $\strategy_{\mathsf{tree}}$ obtained by the tree aggregation algorithm, we have that 
    \[
        \maxlossn(\bfB_{\mathsf{tree}}, \strategy_{\mathsf{tree}})
        \le \sqrt{\ceiling{\log_2(n)} (1+ \ceiling{\log_2(n)})} \,.
    \]
\end{btheorem}
\begin{proof}[Proof Sketch]
We have already argued that the maximum number of non-zero entries in any row of $\bfB_{\mathsf{tree}}$ is  $\ceiling{\log_2(n)}$, so that $\rownorm{\bfB_{\mathsf{tree}}}^2 \le \ceiling{\log_2(n)}$. We can also argue (e.g. by induction) that the maximum number of non-zero entries in any column of $\strategy_{\mathsf{tree}}$ is $\ceiling{\log_2(n)}+1$, leading to $\colnorm{\strategy_{\mathsf{tree}}}^2 \le {1+\ceiling{\log_2(n)}}$.
\end{proof}

Thus, tree aggregation has the same asymptotic $\Theta(\ln n)$ \txtmaxloss as the optimized dense mechanism (\Cref{thm:dense-opt-max-err}). However, it is suboptimal by a multiplicative factor of approximately $\pi / \ln 2 \approx 4.5$.
Several improvements have been proposed to improve this leading constant (see Bibliographic Notes in \Cref{sec:ch2-biblio}), yet all of them remain suboptimal. In this problem, the multiplicative constants in the error have a larger practical impact on the mechanism's utility than an additive error: this is also illustrated by the empirical comparisons of \Cref{sec:ch2-empirical}. 

\paragraph{Summary}
In summary, the binary tree-based mechanisms have a near-optimal time and space complexity of $O(m \ln n)$, improving greatly over the dense mechanism. This even better than the BLT mechanism's $O(m \ln^2 n)$ by a $\ln n$ factor.
However, while it has the correct $\Theta(\ln n)$ asymptotic \txtmaxloss guarantee, 
the BLT mechanism attains better empirical performance due to the additive approximation guarantee.

\section{Other Loss Metrics*}
\label{sec:prefix-L2-error}
In this section, we mostly focussed on max-loss; however, in many machine learning applications of prefix sums, another common \txtloss metric which is considered is the root mean squared loss: the root-mean-squared-error (RMSE) of $\bfB$, scaled by the sensitivity of $\strategy$.

\begin{bdefinition} \label{def:unnormalizedrmse}
The \textbf{unnormalized root mean squared loss} of a mechanism $\mech(\Gradients)$ estimating the (weighted) prefix sum $\workload \Gradients$ of an input $\Gradients \in \R^{\nd \times m}$ is defined as
\begin{align*}
    \rmsloss(\mech) 
    & \coloneqq 
    \,\,  \sqrt{\mathbb{E}_{\mech}\sparen{ \frac{1}{\nd}\sum_{t=0}^{\nd-1}\left\| ( \workload\Gradients - \mech(\Gradients) )[t,:] \right\|_{2}^2}} \,.
\end{align*}
\end{bdefinition}
This is essentially a scaled version of the $\ell_2$ error of the prefix sums, while the \txtmaxloss measures the $\ell_\infty$ norm of the error. The scaling factor above is chosen to ensure that the \txtrmsloss and \txtmaxloss are of the same scale, as we see in the upcoming Lemma~\ref{lem:rmse-maxerr}.
Before that, we start with a closed form expression for the  \txtrmsloss.

\begin{btheorem}
\label{thm:rmse:closedform}
Consider a correlated noise mechanism
$\mech(\Gradients) = \bfB(\strategy \Gradients + \Znoise)$ for $\Znoise \sim \normalnm{0}{\stddev^2}$  with $\stddev = \sens(\strategy) \cdot \nmult = 2 \colnorm{\strategy} \nmult$.
Then, the unnormalized \txtrmsloss satisfies
\begin{align*}
\rmsloss(\bfB, \strategy) = 
     2\sqrt{\frac{\mdim}{\nd}}\, \nmult \,\, \norm{\bfB}_{\op F} \colnorm{\strategy}  \,.
\end{align*}
\end{btheorem}
\begin{proof}[Proof Sketch]
As in the proof of \Cref{thm:accuracyMatrixMechanism}, we have $\workload \Gradients - \mech(\Gradients) = \bfB \Znoise \sim \normal(\zeros, \stddev^2  \bfB \bfB^\top)$. 
Therefore, we get, 
\begin{align*}
    {\mathbb E_{\mech} { \norm{(\mech(\Gradients)[t, :] - \workload[t,:]\Gradients)}_2^2} } &= {\mathbb E_{\Znoise} {\norm{(\bfB[t,:] \, \Znoise)}_2^2}} 
    = \mdim \stddev^2 \left(\sum_{\tau =0}^{t}\bfB[t,\tau]^2\right)
\end{align*}
because $\Znoise \sim \mathcal N(0,\stddev^2 )^{n \times m}$. 
Therefore, from the linearity of expectation,
\begin{align*}
   \nd \cdot \rmsloss(\bfB, \strategy)^2
    & = \mathbb{E}_{\mech}{ \sum_{t=0}^{\nd-1}\left\| ( \workload\Gradients - \mech(\Gradients) )[t,:] \right\|_{2}^2} \\
    & =  \sum_{t=0}^{\nd-1}\mathbb{E}_{\mech}{\left\| ( \workload\Gradients - \mech(\Gradients) )[t,:] \right\|_{2}^2} \\
    & = {\mdim} \stddev^2 \,\,  {\sum_{t=0}^{\nd-1} \sum_{\tau =0}^{t} \bfB[t,\tau]^2} \\
    & = {\mdim} \stddev^2\norm{\bfB}_{\op F}^2\,,
\end{align*}
where the last equality follows from $\bfB$ being lower triangular. The result follows be re-arranging the above equation and plugging in $\stddev = 2 \colnorm{\strategy} \nmult$.
\end{proof}

Analogous to the normalized \txtmaxloss in \cref{eq:norm-max-err}, 
we also define (normalized) root mean-squared loss: 
\begin{bdefinition} \label{def:rms-loss}
Consider a correlated noise mechanism
$\mech(\Gradients) = \bfB(\strategy \Gradients + \Znoise)$ for $\Znoise \sim \normalnm{0}{\stddev^2}$. Then, we define \txtrmsloss by
\begin{equation}
    \rmslossn(\bfB, \strategy) 
    \coloneqq \frac{1}{\sqrt{\nd}} \lfrob{\bfB} \, \colnorm{\strategy} \,.
\end{equation}
\end{bdefinition}

\begin{figure*}[p]
    \centering
        \includegraphics[width=0.8\linewidth]{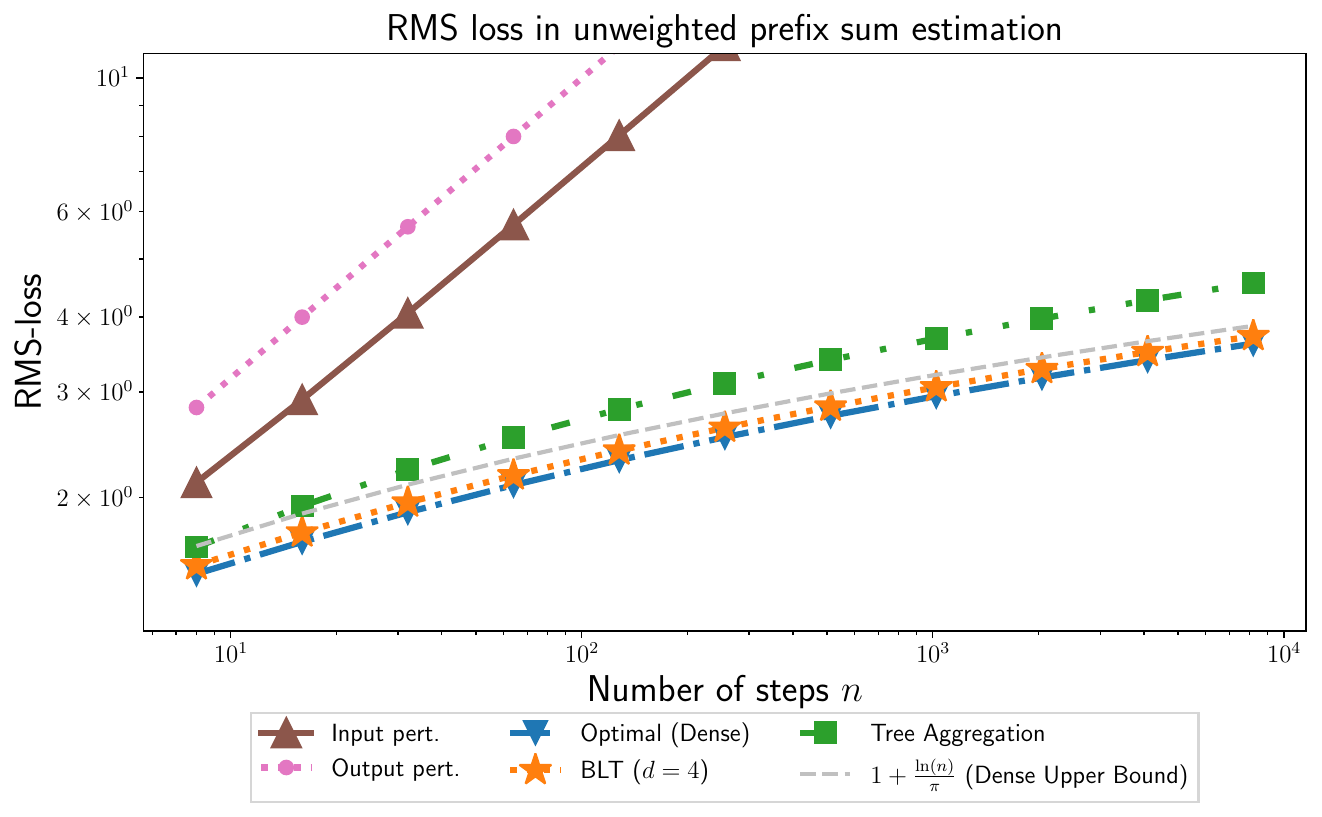}

        \includegraphics[width=0.8\linewidth]{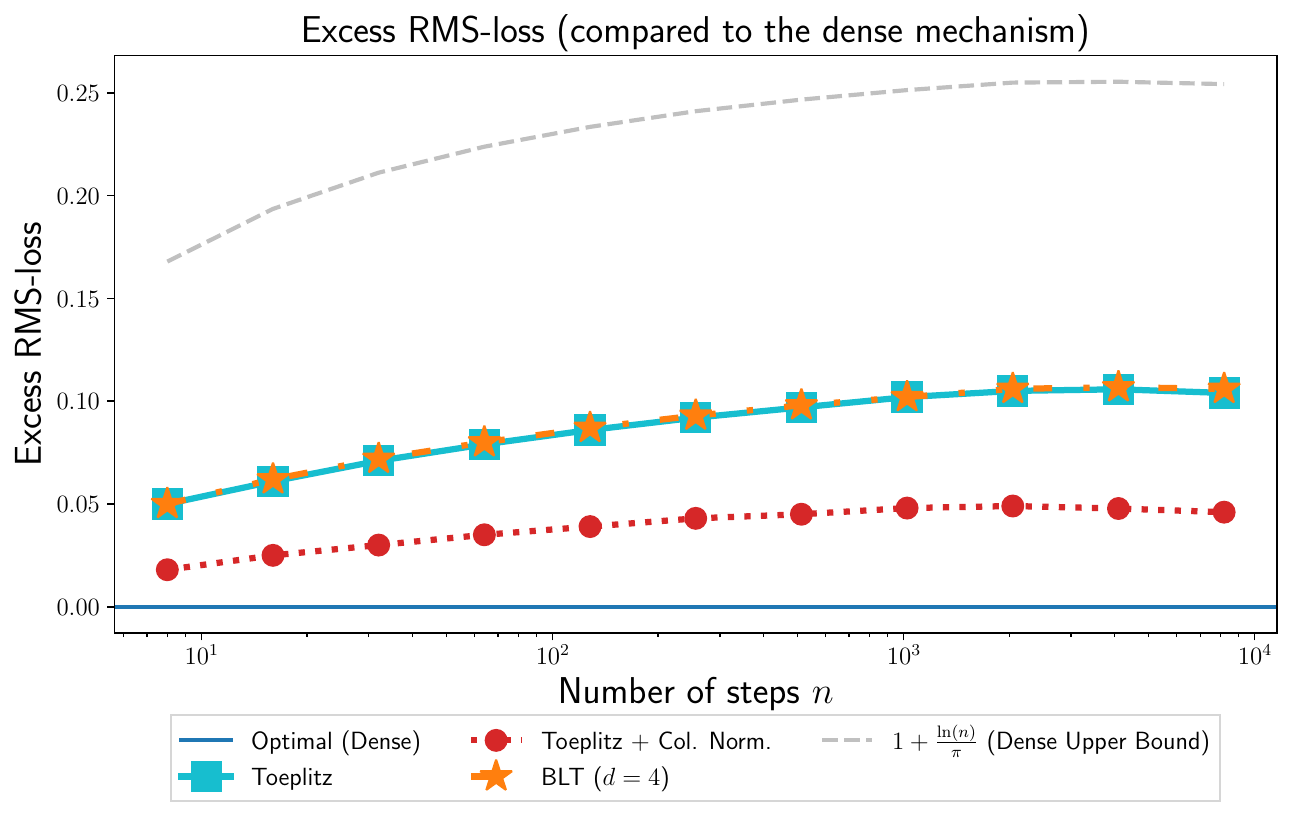}
    \caption{The \txtrmsloss counterpart to \Cref{fig:max-err-empirical}. We empirically compare various mechanisms in the streaming unweighted prefix sum estimation in terms of their \txtrmsloss. 
    The upper bound of $1 + \ln(n) / \pi$ on the \txtmaxloss is also a valid upper bound on the RMSE by 
    Lemma~\ref{lem:rmse-maxerr}.
    See \Cref{sec:a:detailed-numerical} for the exact numerical values plotted here.     
    }
    \label{fig:rmse-empirical}
\end{figure*}

\noindent The normalized \txtrmsloss is always upper bounded by the normalized \txtmaxloss:
\begin{blemma} \label[lemma]{lem:rmse-maxerr}
    For any matrices $\bfB, \strategy\in\R^{n \times n}$, we have $\rmslossn(\bfB, \strategy) \le \maxlossn(\bfB, \strategy)$.
\end{blemma}
\begin{proof}
    Since each row norm is at most the largest row norm, we have, 
    \[
    \lfrob{\bfB}^2
    = 
    \sum_{t \in [\nd]} \norm{\bfB[t,:]}_2^2 \leq \nd \, \max_{t \in [\nd]} \norm{\bfB[t,:]}_2^2 =   \nd \, \rownorm{\bfB}^2 \,.
    \]
Thus, we have that
\[
\rmslossn(\bfB,\strategy) = {\frac{1}{\sqrt{n}}} \lfrob{\bfB} \colnorm{\strategy} 
\leq
\rownorm{\bfB} \colnorm{\strategy} 
= \maxlossn(\bfB,\strategy)
\]
as required.
\end{proof}

\section{An Approximation Theory Viewpoint*}
\label{sec:approx-theory}

The properties of an (infinite) Toeplitz matrix with first column given by $\bfc = (c_0, c_1, \ldots)$ can be understood through its \textbf{ordinary generating function}, which is the formal power series:\footnote{
    The term ``formal'' here refers to the fact that we regard $x$ as a placeholder symbol rather than a number, meaning that we disregard issues of convergence.
}
\[
    f_\bfc(x) \coloneqq \sum_{t=0}^\infty c_t x^t \,.
\]
For instance, the generating function of the all-ones worklad matrix $\prefix$ is
\[
    f_{\mathsf{pre}}(x) = \sum_{t=0}^\infty x^t = \frac{1}{1-x} \,.
\]
The Toeplitz coefficients $\bfc=(c_0, c_1, \ldots)$ can be reconstructed from the Taylor expansion of the generating function around $x=0$:
\[
    f_\bfc(x) = \sum_{t=0}^\infty \frac{f_\bfc^{(t)}(0)}{t!}   \, x^t  = \sum_{t=0}^\infty c_t x^t 
    \quad \iff \quad 
    c_t = \frac{f_\bfc^{(t)}(0)}{t!} \,,
\]
where $f_\bfc^{(t)}$ denotes the $t$\textsuperscript{th} derivative of the function $f_\bfc$.

The key relationship governing this connection is that the product of two Toeplitz matrices is equivalent to the product of the generating functions of their respective coefficients:
\begin{blemma} \label[lemma]{lem:generating-fn}
    Consider real-valued sequences $\bfa=(a_t)_{t=0}^\infty, \bfb=(b_t)_{t=0}^\infty, \bfc=(c_t)_{t=0}^\infty$. The following properties are equivalent:
    \begin{enumerate}[label=(\roman*),nosep]
        \item The sequence $\bfa$ is the convolution of $\bfb$ and $\bfc$, i.e. $a_t = \sum_{\tau=0}^t b_t c_{t-\tau}$;
        \item For any size $n > 0$, the $n \times n$ lower triangular Toeplitz matrices $\bfM_\bfa, \bfM_\bfb, \bfM_\bfc$ with respective first columns given by sequences $(a_t)_{t=0}^{n-1}, (b_t)_{t=0}^{n-1}, (c_t)_{t=0}^{n-1}$ satisfy $\bfM_\bfa = \bfM_\bfb \bfM_\bfc$;
        \item Their respective generating functions $f_\bfa, f_\bfb, f_\bfc$ satisfy $f_\bfa(x) = f_\bfb(x) f_\bfc(x)$.
    \end{enumerate}
\end{blemma}
Thus, the factorization $\prefix=\bfB\strategy$ underlying a correlated noise mechanism can be understood by looking at the implied factorization of the generating function $f_{\mathsf{pre}} = 1/(1-x)$ of the workload matrix $\prefix$. 

The optimal Toeplitz factorization of \Cref{sec:optimial-toeplitz-mech} with $\bfB_{\mathsf{Toep}} = \strategy_{\mathsf{Toep}} = \prefix^{1/2}$ corresponds to the factorization
\begin{align} \label{eq:opt-toep-gen-fn}
    f_{\mathsf{pre}}(x) = \frac{1}{1-x} = \frac{1}{\sqrt{1-x}} \,\cdot\, \frac{1}{\sqrt{1-x}} = f_\bfb(x) \, f_\bfc(x)\,.
\end{align}
Indeed, the coefficients $c_t^\star$ of \Cref{thm:optimal-toeplitz} are the Taylor coefficients of
\[
    f_{\bfc^\star}(x) = \frac{1}{\sqrt{1-x}} = \sum_{t=0}^\infty \binom{-1/2}{t} (-x)^t = \sum_{t=0}^\infty c_t^\star x^t \,.
\]

Similarly, the first column of $\strategy_{\mathsf{Toep}}^{-1}$ can be found as the Taylor coefficients of
\[
\frac{1}{f_{\bfc^\star}(x)} = \sqrt{1-x} = \sum_{t=0}^\infty (-1)^t \binom{1/2}{t} x^t\,,
\]
yielding \cref{eq:opt-toeplitz-cinv}.

\paragraph{Generating Function Approximations and Utility Bounds}
Given any generating function $r(x)$, we can obtain a factorization 
\[
    f_\bfb(x) = \frac{r(x)}{1-x}, \quad\text{and}\quad f_\bfc(x) = \frac{1}{r(x)} \,,
\]
such that the $n \times n$ Toeplitz matrix $\strategy^{-1}$ is made up of the first $n$ Taylor coefficients of $r(x)$ and $\bfB=\prefix \bfC^{-1}$ for any size $n > 0$. 
The optimal factorization  \cref{eq:opt-toep-gen-fn} corresponds to $r(x) = \sqrt{1-x}$. Thus, we can expect that $r(x) \approx \sqrt{1-x}$ will lead to a good factorization with tighter approximations leading to better factorizations in terms of the \txtmaxloss: 
\begin{btheorem} \label{thm:approx-theory-error}
    Fix a size $n \in \N$ and consider a complex-valued function $r: \{x \in \C \,:\, |x| < 1\} \to \C$ defined on the open unit disc in the complex plane. Define the $n \times n$ Toeplitz matrices $\bfB_r$ and $\strategy_r$ whose first columns are given by the first $n$ Taylor coefficients of the generating functions $f_\bfb(x) = r(x) / (1-x)$ and $f_\bfc(x) = 1/r(x)$ respectively. Then, we have:
    \[
        \maxlossn(\bfB_r, \strategy_r) \le \maxlossn(\prefix^{1/2}, \prefix^{1/2}) +  O\big(  n \, \cdot\, \mathsf{err}(r) \big) \,.
    \]
    where $\mathsf{err}(r)$ is the approximation error of $r(x) \approx \sqrt{1-x}$:
    \[
         \mathsf{err}(r) \coloneqq 
         \max_{x \in \C \,:\, |x|=1-n^{-1}} \big|r(x) - \sqrt{1-x}\big|\,,
    \]
    and is assumed to satisfy $\mathsf{err}(r) < 1$. Here, the notation $|x|$ denotes the \textit{absolute value} or \textit{modulus} of the complex number $x\in\C$.
 \end{btheorem}

\paragraph{Examples}
The baselines of input perturbation ($r(x) = 1$ for all $x$) and output perturbation ($r(x) = 1-x$) are clearly poor approximations to $\sqrt{1-x}$. Indeed, we showed $\mathsf{err}(r) = \Theta(\sqrt{n})$ in this case in \Cref{chap:intro} and \Cref{sec:chap2-baselines}. (Note that \Cref{thm:approx-theory-error} only gives an upper bound.)

\begin{figure}[t]
    \centering
    \includegraphics[width=0.95\linewidth]{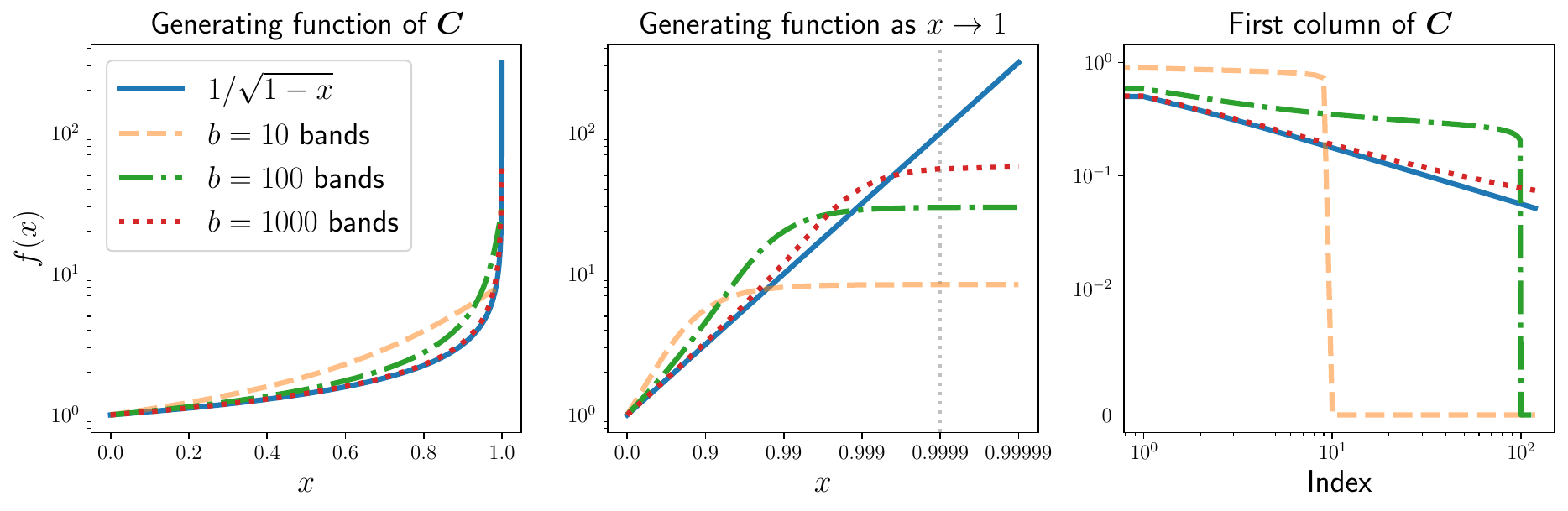} 
    
    \includegraphics[width=0.95\linewidth]{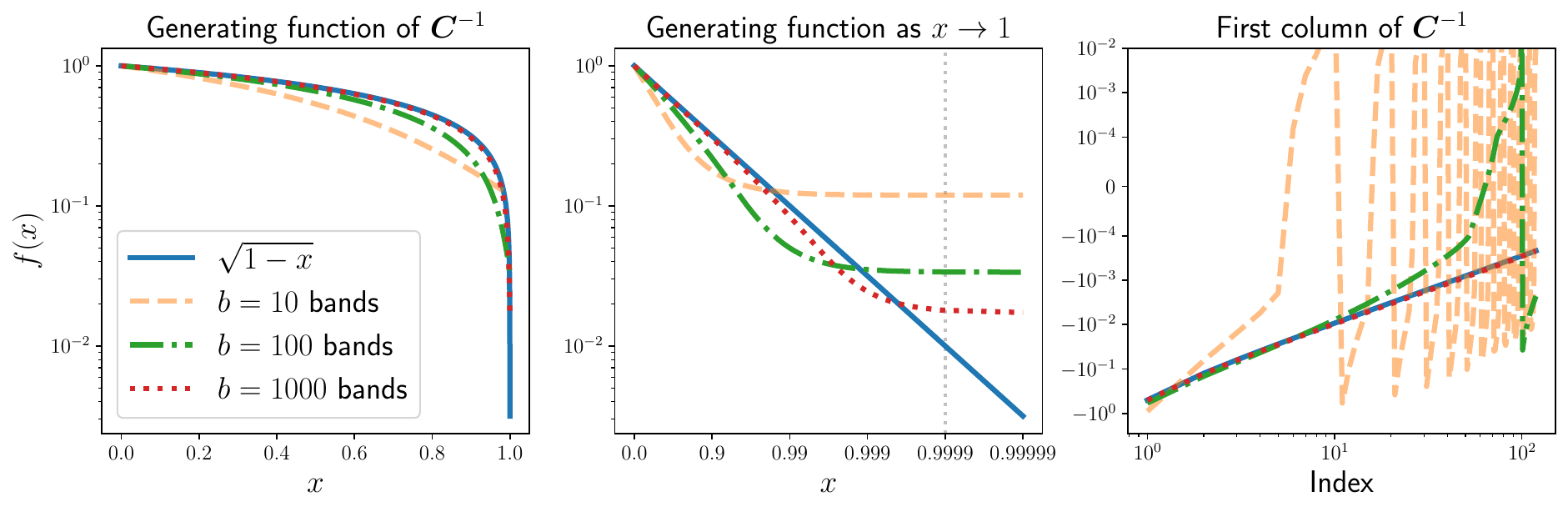}
    \caption{An illustration of the polynomial generating function approximation (corresponding to $\strategy$) provided by the banded Toeplitz mechanism compared to the optimal generating function of $1/\sqrt{1-x}$ of the max-loss-optimal Toeplitz mechanism (\textbf{top}) and the correspond approximation provided by $\strategy^{-1}$ to $\sqrt{1-x}$ (\textbf{bottom}).
    The generating function corresponding to the $b$-banded $\strategy$ matrix is a degree-$(b-1)$ polynomial: $f(x) = \sum_{t=0}^{b-1} c_t x^t$. Here, the Toeplitz coefficients $c_t$ are optimized to minimize \cref{eq:max-err-banded-toeplitz} (using the techniques of \Cref{chap:practical}) for $n=10^4$.
    Notice the approximation quality of the generating function as $x \to 1$ (middle plots) and the Toeplitz coefficients (right plots), are not as tight as the BLT mechanism, shown in \Cref{fig:gen_fn_rational}.
    The dotted line in the middle plot shows $x=1-n^{-1}$; 
    \Cref{thm:approx-theory-error} shows that the approximation error in the generating function at $|x| = 1-n^{-1}$ (in the complex plane) determines the \txtmaxloss.
    }
    \label{fig:gen_fn_bands}
\end{figure}

The $b$-banded Toeplitz mechanism with coefficients $c_0, \ldots, c_{b-1}$ corresponds to the (inverse) polynomial of degree $b-1$:
\[
    f_\bfc(x) = \frac{1}{r(x)} = \sum_{t=0}^{b-1} c_t x^t \,.
\]
Then, $\mathsf{err}(r)$ is related to how well a degree-$b$ polynomial can approximate the function $1/\sqrt{1-x}$ at $|x| = 1-n^{-1}$. Unfortunately, polynomial approximations can be quite bad, especially around the poles\footnote{For a rational function in reduced form, the \textit{poles} are the values of the function where the denominator is equal to zero.} of the function being approximated; $1/\sqrt{1-x}$ has a pole at $1$. Indeed, it turns out that approximating $1/\sqrt{1-x}$ with a degree-$b$ polynomial up to error $\delta$ requires $b$ to be larger than $1/\poly(\delta)$. 
See \Cref{fig:gen_fn_bands} for an illustration.

\begin{figure}[t]
    \centering
    \includegraphics[width=0.95\linewidth]{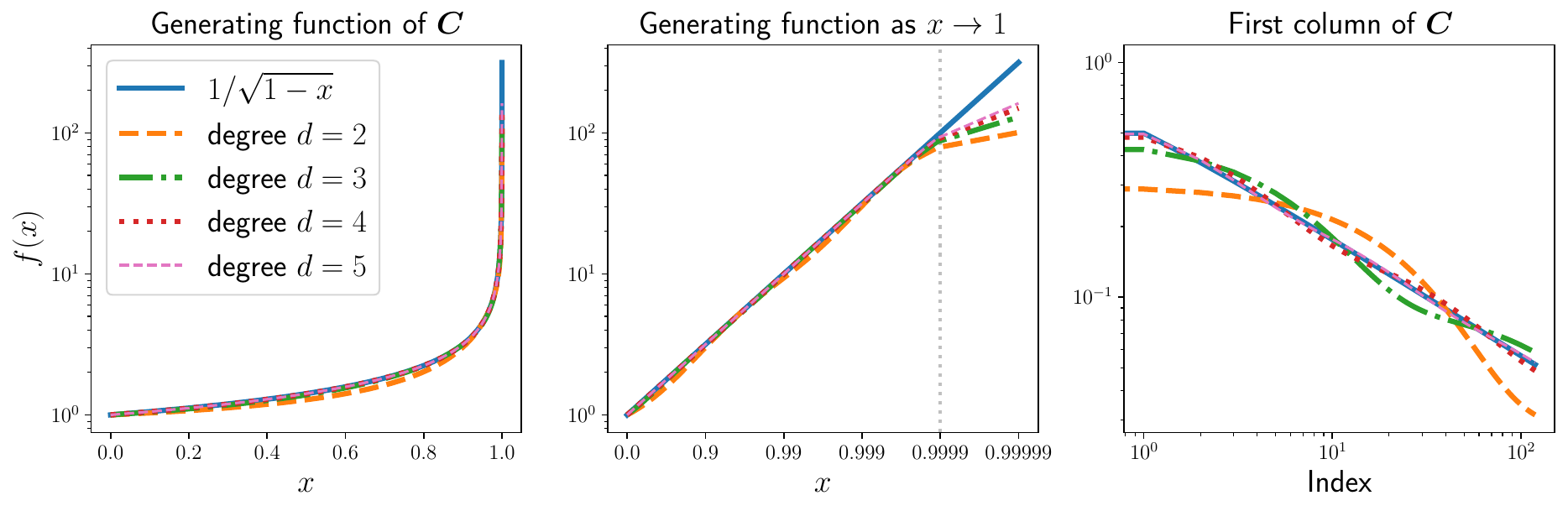}
    
    \includegraphics[width=0.95\linewidth]{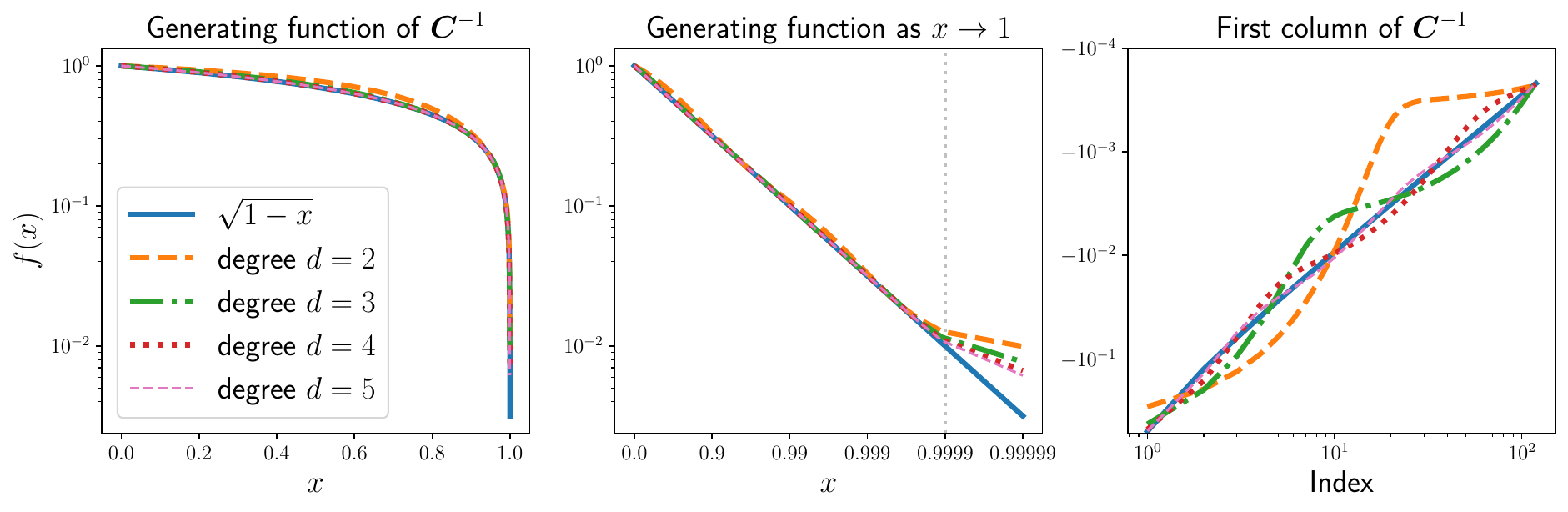}
    \caption{An illustration of the rational generating function approximation provided by the BLT mechanism compared to the optimal generating function of $1/\sqrt{1-x}$ of the max-loss-optimal Toeplitz mechanism (\textbf{top}) and the corresponding approximation its inverse provides to $\sqrt{1-x}$ (\textbf{bottom}).
    We plot the degree-$d$ BLT approximation obtained by solving Problem \ref{eq:max-err-BLT} for $n=10^4$ (using the techniques of \Cref{chap:practical}).
    We get a tight approximation for $x$ less than $\approx 1 - n^{-1}$ (shown by the dotted line in the middle plot), with tighter approximation for larger degree~$d$.
    Notice that the approximation quality of the generating function as $x \to 1$, and of the optimal Toeplitz coefficients (rightmost plots) is significantly tighter than that the banded $\strategy$ from \Cref{fig:gen_fn_bands}.
    }
    \label{fig:gen_fn_rational}
\end{figure}

We now turn to the BLT mechanism. The generating function of $\bfC = \BLT(\bfalpha, \bflambda)$ is given by
\begin{align} \label{eq:gen-fn-r}
    r(x) &= 1 + \sum_{i=1}^d \alpha_i x + \sum_{i=1}^d \alpha_i \lambda_i x^2 + \cdots
    = 1 + \sum_{i=1}^d \frac{\alpha_i x}{1 - \lambda_i x}\,.
\end{align}
This is a rational function of the form
\[
    r(x) = \frac{p(x)}{q(x)}, \quad \text{where} \quad p(x)\coloneqq\sum_{i=0}^{d} p_i x^i \quad \text{and} \quad q(x) \coloneqq{\sum_{i=0}^d q_i x^i}
\]
with $p(0) = q(0) \neq 0$.
Rational approximations generally give much tighter approximations than polynomials. For instance, rational Padé approximations are known to be much tighter than polynomial Taylor approximations. In particular, there is a rational function $r(x)$ of degree $d$ such that 
\begin{align}
\label{eq:newmanApproximation}
    \sup_{x \in [0, 1]} | r(x) - \sqrt{1-x}| \le O\big(\exp(-\sqrt{d})\big) \,.
\end{align}
The error vanishes exponentially in the degree $d$, and is significantly better than polynomial approximations. This is illustrated in \Cref{fig:gen_fn_rational}.

While this result holds only on the real line, there exists a degree-$d$ rational function $r$ over the complex plane with $\mathsf{err}(r) = O\big(\exp(-\sqrt{d})\big)$. This explains the competitive utility bound of the BLT mechanism (\Cref{thm:blt-max-err}). In particular, it suffices to take a degree of $d=O(\ln^2(n/\delta))$ so that $\mathsf{err}(r) = O(\delta / n)$, leading to an additive error of $\delta$.

\paragraph{Generating Functions and Efficient Implementation}
A broad class of Toeplitz coefficients that admit efficient noise generation algorithms correspond to the class of constant-recurrent sequences:

\begin{bdefinition} \label{def:constant-recurrent}
A sequence $(c_t)_{t=0}^\infty$ is called a constant-recurrent sequence of order $d$ if there exist numbers $q_1, \ldots, q_d \in \R$ such that
\[
    c_t + \sum_{i=1}^d q_i c_{t-i} = 0
\]
for all $t \ge d$.
\end{bdefinition}

This broad class includes many special cases such as arithmetic and geometric progressions, the Fibonacci series, and many others. More relevant to correlated noise mechanisms, banded Toeplitz coefficients (and eventually periodic sequences in general), as well as BLT coefficients are special cases. This follows from inspecting their generating function, based on the following equivalent representations:

\begin{btheorem} \label{thm:constant-recurrent}
    The following properties are equivalent:
    \begin{enumerate}[label=(\alph*)]
        \item $(c_t)_{t=0}^\infty$ is a constant-recurrent sequence of order-$d$.
        \item The generating function $f_\bfc(x) = \sum_{t=0}^\infty$ of $(c_t)_{t=0}^\infty$ is a rational function $f_c(x) = p(x) / q(x)$, where $q(0) = 1$, $\mathsf{deg}(q) \le d$ and $\mathsf{deg}(p) < d$.
        \item  \label{item:matrix-power}
        There exists a matrix $\bfLambda \in \R^{d \times d}$ and vectors $\bfu, \bfv \in \R^d$ such that $c_t = \bfu\T \bfLambda^t \bfv$ for all $t \ge 0$.
    \end{enumerate}
\end{btheorem}

For the BLT mechanism, we have that $c_0, c_1, c_2, \ldots$ forms a constant-recurrent sequence (i.e., omitting the first term $c_0=1$). As a function of the BLT parameters $\bfalpha, \bflambda$, we have $\bfLambda = \diag(\bflambda)$, $\bfv = \bfalpha$ and $\bfu=(1, \ldots, 1)$ is the vector of ones. While the sequence $c_0, c_1, \ldots$ (including the first term $c_0$) can be written as a constant-recurrent sequences of order-$(d+1)$, it is computationally advantageous to treat it separately.

A key advantage of constant-recurrent coefficients is that they are amenable to efficient recursive noise generation. Given a constant-recurrent sequence $c_t = \bfu\T \bfLambda^{t} \bfv$ (in the matrix-power representation of \Cref{thm:constant-recurrent}\ref{item:matrix-power}) that makes up the first column of the matrix $\strategy$, we can compute $\corrZnoise = \strategy^{-1} \Znoise$ with a small modification of \Cref{alg:blt-mult-by-cinv}. In particular, we replace Lines \ref{line:blt-noise-gen} and \ref{line:blt-buffer-update} with
\[
    \corrznoise_t = \znoise_t - \bfM_t \bfv\,, \quad
    \text{and} \quad
    \bfM_{t+1} = \bfM_t \bfLambda + \corrznoise_t \bfu\T \,,
\]
which requires $O(\mdim d + d^2)$ time and space. This is worse than  \Cref{alg:blt-mult-by-cinv} by an additive factor of $d^2$ on both counts.
When the matrix $\bfLambda$ is diagonalizable, this recurrence can effectively be reduced to \Cref{alg:blt-mult-by-cinv}. For example, if $\bfLambda$ has all real and unique eigenvalues, we can diagonalize it as $\bfLambda = \bfV \diag(\bflambda) \bfV^{-1}$. Then, we have 
\[
    c_t = \bfu\T \bfLambda^t \bfv = \widetilde \bfu \T \diag(\bflambda)^t \, \widetilde \bfv\,,
\]
where $\widetilde \bfu = \bfV\T \bfu$ and $\widetilde \bfv = \bfV^{-1} \bfv$.

Importantly, the optimal coefficients $(c_t^\star)_{t=0}^\infty$ of the max-loss-optimal Toeplitz mechanism defined in \Cref{thm:optimal-toeplitz} do not form a constant-recurrent sequence. Indeed, their generating function of $1/\sqrt{1-x}$ is not a rational function. Thus, they are not amenable to the efficient recursive noise generation procedure outlined above.

\section{Closed Form Expressions for the BLT Mechanism*}

Here, we give closed form expressions for the sensitivity and error for the BLT mechanism (\Cref{sec:ch2-blt}). These expressions are useful in numerically optimizing the parameters of the mechanism to minimize the \txtmaxloss; see \Cref{sec:blt-opt}.

\begin{blemma} \label{lem:blt-closed-form}
Suppose we are given parameters $\bfalpha, \tilde \bfalpha \in \R^d_+$ and $\bflambda, \tilde \bflambda \in [0, 1)^d$ such that $\bfC = \BLT(\bfalpha, \bflambda)$ and $\bfC^{-1} = \BLT(-\tilde\bfalpha, \tilde \bflambda)$ are inverses of each other, assuming they exist (see Lemma~\ref{lem:inverse-blt} for precise conditions). Further, for $\lambda \in [0, 1)$ define the $n$-term geometric sum
\[
\gamma_n(\lambda) := \sum_{t=0}^{n-1} \lambda^t = \frac{1-\lambda^n}{1-\lambda}
\,.
\]
Then, we have the following:
\begin{enumerate}[label=(\alph*)]
    \item The single epoch sensitivity is given by
    \[
        \colnorm{\strategy}^2 = 1 + \sum_{i=1}^d \sum_{j=1}^d \alpha_i \alpha_j \gamma_{n-1}(\lambda_i \lambda_j) \,.
    \]
    \item For $n \ge 2$, the maximum row norm of $\bfB
    = \workload \strategy^{-1}$ is
    \[
        \rownorm{\bfB}^2 = n + 2 \sum_{i=1}^d \tilde \alpha_i \Gamma_i^{(n)} + \sum_{i=1}^d \sum_{j=1}^d  \tilde \alpha_i \tilde \alpha_j \Gamma_{i, j}^{(n)}\,,
    \]
    where we define
    \begin{align*}
        \Gamma_i^{(n)} &= \frac{1}{1 - \tilde \lambda_i} \big(n - \gamma_{n}(\tilde \lambda_i)\big) \,, \\
        \Gamma_{i, j}^{(n)} &= \frac{1}{(1-\tilde \lambda_i) (1-\tilde \lambda_j)} \Big( n - \gamma_{n}(\tilde \lambda_i) - \gamma_{n}(\tilde \lambda_j) + \gamma_{n}(\tilde \lambda_i \tilde \lambda_j) \Big) \,.
    \end{align*}
    \item The Frobenius norm of $\bfB = \workload \strategy^{-1}$ is given by
    \[
        \frac{1}{n} \lfrob{\bfB}^2 = \frac{n+1}{2} + \frac{2}{n} \sum_{i=1}^d \tilde \alpha_i \hat \Gamma_i^{(n)} + \frac{1}{n} \sum_{i=1}^d \tilde \alpha_i \tilde \alpha_j \hat \Gamma_{i, j}^{(n)}\,,
    \]
    where we define
    \begin{align*}
        S_i^{(n)} &= 1 - \tilde \lambda_i  + \frac{\tilde \lambda_i}{1 - \tilde \lambda_i} \big(n - \gamma_{n}(\tilde \lambda_i)\big) 
        \,, \\
        S_{i, j}^{(n)} &= 1 - \tilde \lambda_i \tilde \lambda_j + \frac{\tilde \lambda_i \tilde \lambda_j}{1 - \tilde \lambda_i \tilde \lambda_j} \Big(n - \gamma_{n}(\tilde \lambda_i \tilde \lambda_j)\Big) \,, \\
        \hat \Gamma_{i}^{(n)} &= 1 + \frac{1}{1 - \tilde \lambda_i} 
        \left( 
            \frac{n(n-1)}{2} -  S_i^{(n)} 
        \right) \,, \\
        \hat \Gamma_{i, j}^{(n)} &= 
        1 + \frac{1}{(1 - \tilde \lambda_i)(1 - \tilde \lambda_j)} \left(
            \frac{n(n-1)}{2} - S_i^{(n)}  - S_j^{(n)}  
            + S_{i, j}^{(n)}   
        \right)\,.
    \end{align*}
\end{enumerate}
Further, each of these quantities can be computed in $O(d^2)$ floating point operations.
\end{blemma}

\section{Numerical Comparison in the Streaming Setting*}
\label{sec:a:detailed-numerical}

\Cref{tab:max_error,tab:rmse} give the raw numbers used to plot \Cref{fig:max-err-empirical,fig:rmse-empirical} respectively.
Each column corresponds to a different factorization, with ``Identity'' and ``Workload'' refers to the Input Perturbation and Output Perturbation mechanism introduced in \cref{sec:chap2-baselines}.  Streaming H2 and Full H2 refer to different variants of the tree aggregation approach \cref{sec:treeAgg}. BLT was presented in \cref{sec:blt-opt}.  Toeplitz and Col-Norm. Toep were presented in \cref{sec:optimial-toeplitz-mech}.  Dense was presented in \cref{sec:optimized-dense-mech}.

\begin{table}[ht]
\centering
\caption{Max Error for different matrix factorizations.}
\label{tab:max_error}
\resizebox{\textwidth}{!}{
\begin{tabular}{@{}r*{8}{c}@{}}
\toprule
 $\nd$ & \textbf{Identity} & \textbf{Workload} & \textbf{Streaming H2} & \textbf{Full H2} & \textbf{BLT} & \textbf{Toeplitz} & \textbf{Col-Norm. Toep.} & \textbf{Dense} \\
\midrule
 8 & 2.828 & 2.828 & NaN & 2.382 & 1.723 & 1.718 & 1.573 & 1.51 \\
 16 & 4.0 & 4.0 & NaN & 2.881 & 1.944 & 1.944 & 1.783 & 1.704 \\
 32 & 5.657 & 5.657 & NaN & 3.381 & 2.168 & 2.167 & 1.997 & 1.905 \\
 64 & 8.0 & 8.0 & NaN & 3.883 & 2.391 & 2.389 & 2.212 & 2.111 \\
 128 & 11.314 & 11.314 & NaN & 4.384 & 2.61 & 2.61 & 2.428 & 2.32 \\
 256 & 16.0 & 16.0 & NaN & 4.886 & 2.832 & 2.831 & 2.645 & 2.532 \\
 512 & 22.627 & 22.627 & NaN & 5.387 & 3.054 & 3.052 & 2.863 & 2.746 \\
 1024 & 32.0 & 32.0 & NaN & 5.888 & 3.273 & 3.273 & 3.081 & 2.958 \\
 2048 & 45.255 & 45.255 & NaN & 6.389 & 3.494 & 3.493 & 3.299 & 3.177 \\
 4096 & 64.0 & 64.0 & NaN & 6.89 & 3.716 & 3.714 & 3.518 & - \\
 8192 & 90.51 & 90.51 & NaN & 7.391 & 3.939 & 3.935 & 3.737 & -\\
\bottomrule
\end{tabular}}
\label{tab:maxErrorMF}
\end{table}

\begin{table}[ht]
\centering
\caption{RMSE for different matrix factorizations.}
\label{tab:rmse}
\resizebox{\textwidth}{!}{
\begin{tabular}{@{}r*{8}{c}@{}}
\toprule
 $\nd$ & \textbf{Identity} & \textbf{Workload} & \textbf{Streaming H2} & \textbf{Full H2} & \textbf{BLT} & \textbf{Toeplitz} & \textbf{Col-Norm. Toep.} & \textbf{Dense} \\
\midrule
 8 & 2.121 & 2.828 & 2.178 & 1.656 & 1.544 & 1.544 & 1.512 & 1.494 \\
 16 & 2.915 & 4.0 & 2.663 & 1.938 & 1.751 & 1.75 & 1.714 & 1.689 \\
 32 & 4.062 & 5.657 & 3.156 & 2.227 & 1.964 & 1.963 & 1.922 & 1.892 \\
 64 & 5.701 & 8.0 & 3.652 & 2.518 & 2.18 & 2.179 & 2.135 & 2.1 \\
 128 & 8.031 & 11.314 & 4.151 & 2.81 & 2.398 & 2.397 & 2.35 & 2.311 \\
 256 & 11.336 & 16.0 & 4.65 & 3.102 & 2.617 & 2.616 & 2.567 & 2.524 \\
 512 & 16.016 & 22.627 & 5.15 & 3.394 & 2.837 & 2.836 & 2.784 & 2.739 \\
 1024 & 22.638 & 32.0 & 5.65 & 3.686 & 3.057 & 3.057 & 3.003 & 2.955 \\
 2048 & 32.008 & 45.255 & 6.15 & 3.978 & 3.278 & 3.277 & 3.221 & 3.172 \\
 4096 & 45.26 & 64.0 & 6.65 & 4.269 & 3.499 & 3.498 & 3.44 & - \\
 8192 & 64.004 & 90.51 & 7.15 & 4.56 & 3.72 & 3.718 & 3.66 & - \\
\bottomrule
\end{tabular}}
\end{table}
\section{Bibliographic Notes}
\label{sec:ch2-biblio}

Bounds on the \txtmaxloss $\maxlossn(\bfB, \strategy)$ for $\prefix = \bfB \strategy$ have been studied extensively,
starting with the work of \citet{kwapien1970main}. Since then many results have improved the leading constants. \Cref{thm:accuracyMatrixMechanism} has been known in the literature from, for example, \citet*{nikolov2016geometry}, with an explicit construction appearing in \citet*{edmonds2020power}.

\paragraph{Dense Mechanism}
The bounds of \Cref{thm:dense-opt-max-err} use the property that the optimization problem \eqref{eq:max-err-optim} defines a norm $\gamma_2(\cdot)$ of a matrix:
\[
    \gamma_2(\bfM) \coloneqq 
    \min
    \left\{
        \rownorm{\bfB} \colnorm{\strategy} \, :\, 
        \bfB \strategy = \bfM
    \right\}  \,.
\]
The fact that this is a norm was first proved in an unpublished manuscript of \citet{haagerup1980decomposition}, who showed its equivalence to the Hadamard norm. (This was subsequently also shown in many works.)

The upper bound in \Cref{thm:dense-opt-max-err} 
was proved by \citet[Corollary 3.5]{mathias1993hadamard}, with the bound
\[
    \gamma_2(\prefix) \le \frac{1}{2} + \frac{1}{2n}\sum_{t=1}^n \left| \csc\left(\frac{(2t-1)\pi}{2n} \right) \right| \,,
\]
where $\prefix$ is the lower triangular matrix of all ones (cf.~\cref{eq:all-ones-workload}) and  $\csc(\cdot)$ is the trigonometric cosecant function.
This is at most\footnote{This can be shown by upper bounding the sum with an integral~\cite[cf.][Sec. 3]{mathias1993hadamard}.} $1 + \ln(n)/\pi$. The proof of \citet{mathias1993hadamard} is non-constructive and uses the triangle inequality to bound $\gamma_2(\prefix)$ with the $\gamma_2$-norms of two other matrices, which are themselves bounded by other means. Recently, \citet{henzinger2025improved} gave a constructive factorization that matches the upper bound of \citet{mathias1993hadamard} and also has the property that the columns of $\strategy$ all have the same $\ell_2$ norm. 
We note that \citet[Corollary 3.5]{mathias1993hadamard} also gives a lower bound:
\[
    \gamma_2(\prefix) \ge \frac{1}{2n} + \frac{1}{2n}\sum_{t=1}^n \left| \csc\left(\frac{(2t-1)\pi}{2n} \right) \right| \,,
\]
where the first term is $1/(2n)$ instead of $1/2$ in the upper bound. \Cref{thm:dense-opt-max-err} instead gives a tighter lower bound established by \citet*{matouvsek2020factorization} by using the fact that $|\csc(x)| \geq x^{-1}$:
\[
    \gamma_2(\prefix) \ge \frac{1}{2 n} \sum_{t=1}^n \left|{\csc\left(\frac{(2t-1)\pi}{2(2n+1)}\right)}\right|  \ge \frac{2n+1}{\pi n} \sum_{t=1}^n \frac{1}{2t-1} \ge  \frac{\ln(2n+1)}{\pi}.
\]
The dual characterization of the $\gamma_2$-norm is useful in constructing lower bounds. First shown by \citet{haagerup1980decomposition} and also appeared later in \citet*{haagerup1993bounded} and~\citet*[Theorem 9]{lee2008direct}, we have: 
\begin{align*}
    \gamma_2(\bfM)
    &= \max\left\{ \| \diag(\bfp)^{1/2} \,\, \bfM  \,\, \diag(\bfq)^{1/2} \|_{*} : 
    \begin{aligned}
    \bfp, \bfq \in \R^{n}_+ \, \text{ with }  \\
    \bfp^\top \ones = 1 = \bfq^\top \ones
    \end{aligned}
    \right\}\,,
\end{align*}
where $\norm{\bfM}_*$ denotes the nuclear norm of the matrix $\bfM$. The lower bound from \Cref{thm:dense-opt-max-err}, established by \citet*{matouvsek2020factorization}, follows from taking $\bfp = \bfq = \ones_n / n$ in the optimization problem defining the dual norm, and explicitly computing the eigenvalues of the matrix $\bfM = \prefix$.
In fact, a bound on the trace norm $\norm{\cdot}_*$ appearing in the dual norm was explicitly first computed by \citet{elliott1953characteristic} and subsequently improved by a series of works~\cite{gregory1969collection, hoffman1971linear,strang2022introduction}.

\paragraph{Tree Aggregation}
Tree aggregation, known also as the binary tree mechanism, was proposed independently by 
\citet*{hay2010boosting,dwork2010continual,chan2011private}. These were among the first correlated noise mechanisms for differential privacy.
A variant of the binary tree mechanism was also proposed by \citet*{smith2017interaction}, although that does not fit into the the class of correlated noise mechanisms we study in this monograph (\cref{eq:ch2-mechanism}). Tree aggregation is also known as the \emph{hierarchical histogram} method, especially in the context of range queries, as in \Cref{ex:range-queries}, see for example~\citet*{hay2010boosting,qardaji2013understanding,cormode2019answering,li2014data}.

Follow up work made several improvements to the tree aggregation approach of \Cref{sec:treeAgg}. For instance, \citet{honaker2015efficient} proposed to replace the noise correlation matrix $\bfB_{\mathsf{tree}}$ from \Cref{thm:treeAgg} with the matrix\footnote{
    Here, $\bfM^\dagger$ denotes the Moore-Penrose pseudoinverse of $\bfM$.
}
$\bfB_{\mathsf{tree}}^\star = \prefix \strategy_{\mathsf{tree}}^\dagger$ for its better error properties; this comes at the cost of an increased $O(mn)$ memory for noise generation.   This same construction appeared earlier in the range query literature \citet*{hay2010boosting,li2015matrix} as well, where it was shown to be the best linear unbiased estimator for the workload query answers.
As another example, \citet{andersson2023smooth} improve the error bound of \Cref{thm:treeAgg} by a factor 2 by carefully discarding rows/columns from $\bfB_{\mathsf{tree}}$/$\strategy_{\mathsf{tree}}$ with large norm. This construction preserves the $m \log_2(n)$ time and space complexity of noise generation, but the \txtmaxloss is still suboptimal by a factor of approximately $\pi / (2\ln 2) \approx 2.26$.
\citet*{dvijotham2024efficient} propose a tree construction which can get a constant factor of $1 + o(1)$, i.e., arbitrarily close to optimal. Their procedure involves building a recursive tree-like factorization over a base BLT factorization.

\paragraph{The Max-Loss-Optimal Toeplitz Mechanism}
We can trace the factorization $\prefix = \prefix^{1/2} \prefix^{1/2}$ of the prefix sum matrix (that we studied in \Cref{sec:optimial-toeplitz-mech}) at least as far back as \citet{bennett1977schur}.
\citet*{fichtenberger2023constant} were the first to utilize this factorization for differentially private computation of streaming prefix sums. 
The optimality of this mechanism for the objective \cref{eq:max-err-optim-toeplitz} among the class of all lower triangular and Toeplitz factorizations was established by \citet*{dvijotham2024efficient}. (Previously, it was only known that it is optimal in the asymptotic $n \to \infty$ regime.)

\citet*{fichtenberger2023constant} also first established the bound on the \txtmaxloss of the max-loss-optimal Toeplitz mechanism, where the optimal $1/\pi$ factor comes from \citet*{chen2005best}. Later, \citet*{henzinger2023almost} showed an almost tight bound on the \txtrmsloss. 
In \Cref{thm:toeplitz-opt-max-err}, we present the bound of \citet*[Lemma 2.1]{dvijotham2024efficient}. 
Finally, column normalization is motivated by Lemma~\ref{lem:column-norm}, which was observed by \citet*{yuan2016convex}.

\paragraph{Banded Toeplitz Mechanism}
\sloppy 
Banded strategy matrices $\strategy$ were considered by \citet*{choquette2024amplified} in the machine learning context for their compatibility with amplification by sampling, and the banded Toeplitz matrices by \citet*{kalinin2024banded,mckenna2024scaling} to improve the factorization efficiency. We discuss amplification by sampling in \Cref{chap:ml} and the factorization considerations in \Cref{chap:practical}. 
The error bound we present in \Cref{thm:max-err-banded} is due to \citet*[Theorem 6]{kalinin2024banded}.\footnote{
    While they state bounds on the $\ell_2$ error, \citet{kalinin2024banded} actually upper bound it by the \txtmaxloss and prove their bounds on the latter.
}

\paragraph{The BLT Mechanism}
The BLT mechanism was introduced by \citet*{dvijotham2024efficient}, including the error bound \Cref{thm:blt-max-err} which corresponds to their Theorem 4.6. Their Lemma 5.2 parameterized a mechanism $\strategy = \BLT(\bfalpha, \bflambda)$ and $\Cinv = \BLT(\widehat\bfalpha, \widehat\bflambda)$ via the pair $(\bflambda, \widehat\bflambda)$, allowing efficient noise generation from $\Cinv = \BLT(\widehat\bfalpha, \widehat\bflambda)$ following the approach of \cref{eq:blt-mult-by-C} (their Algorithm 1).

The BLT mechanism was extended from the streaming setting to the machine learning setting by \citet*{mcmahan2024hassle}, including the algorithm for noise generation directly from  $\strategy = \BLT(\bfalpha, \bflambda)$ (our \Cref{alg:blt-mult-by-cinv} is a restatement of their Algorithm 3).
Lemma~\ref{lem:inverse-blt} regarding the inverse BLT parameterization is due to \citet*{mcmahan2025inverse}.

For more discussion on the fact that rational Pad\'{e} approximations are known to be much tighter than polynomial Taylor approximations~\cite[e.g.,][]{baker1961pade}. 
The approximation stated in \cref{eq:newmanApproximation} was given by  \citet{newman1964rational}. 

\paragraph{The Approximation Theory Viewpoint}
The approximation theory viewpoint presented in \Cref{sec:approx-theory} was utilized by \citet*{fichtenberger2023constant} to develop the max-loss-optimal Toeplitz mechanism and by \citet*{dvijotham2024efficient} for the BLT mechanism. In particular, \Cref{thm:approx-theory-error} is a simplified (and slightly looser) version of \citet*[Proposition 4.1]{dvijotham2024efficient}.

These developments are based on classical ideas from approximation theory; we highlight a few important connections. 
The classical Prony interpolation, developed by  Gaspard Riche de Prony in 1795, is an approximate decomposition of sequence (or a function) into a sum of complex exponentials as in \cref{eq:prony}. \citet{newman1964rational} gave an approximation of $|x|$ using a rational function. This can be used to construct a rational approximation to $\sqrt{1-x}$, forming the basis for the proof of the error bound of the BLT mechanism (\Cref{thm:blt-max-err}).
Finally, \citet*{braess2005approximation} show that it is possible to approximate $1/t \approx \sum_{i=1}^d \alpha_t \lambda_i^t$ (with $\alpha_i, \lambda_i$ real) up to an error $\exp(-\Omega(\sqrt{d}))$. The BLT approximation task is closely related---we wish to approximate the function $t^{-3/2}$, as exhibited by the coefficients $(c_t')$ of $\strategy_{\mathsf{Toep}}^{-1}$, as in \Cref{thm:optimal-toeplitz}.

Finally, for more details on \Cref{thm:constant-recurrent} and the equivalent representations of constant-recurrent sequences, we refer to the excellent monograph by \citet*{corless2011concrete}.

%% file: 3_ml.tex
In this chapter, we build on the correlated noise mechanism introduced in the previous chapters to make them applicable to practical AI and machine learning settings. 
Recall from \Cref{sec:weightedPrefixSum} that we wish to find model parameters $\bftheta \in \Theta \subset \R^\mdim$ optimizing the objective
\begin{align} \label{eq:pop-risk-min}
        \min_{\bftheta \in \Theta} \,\, \mathbb{E}_{\bfx \sim \Pdata}\left[\ell\br{\bftheta, \bfx}\right],
\end{align}
where $\ell(\bftheta, \bfx)$ is the loss of making a prediction with model parameters $\bftheta$ on a datapoint $\bfx$, and $\Pdata$ is an underlying data distribution.

Throughout, we give bounds on the \emph{suboptimality} of a model $\bftheta$, also known as the \emph{excess population risk} of $\bftheta$. For a mechanism $\mathcal M$ that outputs $\theta$, 
\begin{equation} \label{eq:excess-risk}
 R(\mech) := 
 \mathbb{E}_{\bfx \sim \Pdata }\left[\ell(\model, \bfx)\right]
 -
 \min_{\bftheta^*\in \Theta} \mathbb{E}_{\bfx\sim\Pdata}\left[\ell(\bftheta^*, \bfx)\right] \, .
\end{equation}

In \Cref{chap:intro,chap:prefixsum}, we mainly focused on the streaming setting where each data point is processed only once. We relax this assumption and treat the general multiple-participation setting in this section, allowing each data point to participate multiple times in training (e.g. by making multiple passes through the dataset).

The multiple-participation setting is also highly relevant in the context of user-level DP (as opposed to our default privacy unit of example-level DP). Here, multiple participations of \emph{any} of a user's data violates the streaming assumption, even if each datapoint is processed only once (see \Cref{sec:ch1-priv-unit} for details).

The main difference between the streaming and multiple-participation settings is the sensitivity computation. As illustrated in \Cref{fig:adjacency-intro}, changing one data point (to an adjacent dataset) can change multiple gradients (even in the non-adaptive setting we use for analysis per \cref{thm:gdp-adaptivity}), increasing the sensitivity of the operation.
The key challenge in the multiple-participation setting turns out to give tight and efficient bounds on the sensitivity. We define data processing abstractions that map to practical scenarios but where the sensitivity is not too large, and give algorithms to efficiently compute this sensitivity.

\paragraph{Outline}
We start this section with the potential advantages that correlated noise mechanisms can offer in the learning setting in \Cref{sec:ml:motivation} (and hence motivating the setting of multiple participation). \Cref{sec:ml:setup} gives examples of other first order-optimization algorithms and their reduction to (weighted) prefix sum estimation as in \cref{eq:mainProblem}.
Next, \Cref{sec:ml:multi-epoch} takes a deep dive into the multiple-participation setting, including the challenges, data processing patterns for tight sensitivity calculations, and efficient computational algorithms.

Changing gears, \Cref{sec:ml:learning-guarantees} surveys learning-theoretic guarantees for correlated noise mechanisms, and how they can help over using independent noise. We end the section by discussing how the privacy guarantees of correlated noise mechanisms can be amplified by accounting for the random sampling of data points in \Cref{sec:learning:amplification}. As in \Cref{chap:prefixsum}, detailed pointers to missing proofs can be found in the bibliographic notes of \Cref{sec:ch3-biblio}.

\section{Motivation} \label{sec:ml:motivation}

Correlated noise mechanisms are generally never worse (in terms of downstream learning performance at any given privacy level) and often much better than using independent noise in learning settings. In this section, we highlight the factors behind this Pareto-dominance of correlated noise over independent noise.

\paragraph{Privacy amplification by sampling} In the centralized training setting, we can choose to form batches of data randomly. This typically has no adverse impact on the training process, and can even be helpful. However, the additional randomness from sampling can be beneficial for privacy. Intuitively, this added randomness will typically further increase the uncertainty of an adversary about whether a dataset $D$ or another adjacent dataset $D'$ was used in training. This phenomenon is also known as \emph{privacy amplification by sampling}.

\begin{figure}[t]
    \centering
    \includegraphics[width=0.7\linewidth]{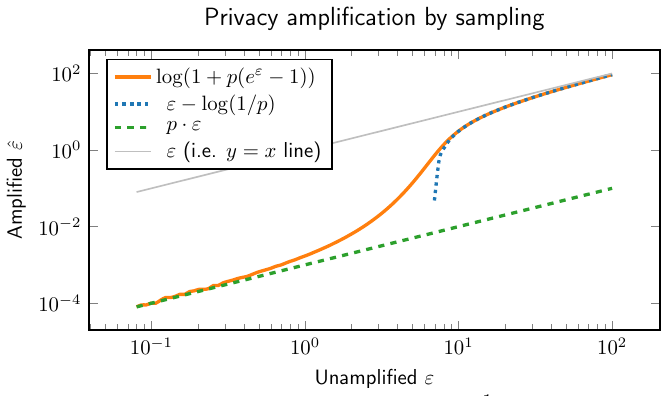}
    \caption{Plots for theoretical bounds on amplification of a generic $(\eps, \delta)$-DP with Poisson sampling (in orange), where a batch of data is formed by including each example i.i.d. with probability $p$, leading to $(\hat \eps, p \delta)$-DP with $\hat \eps = \log(1+ p(e^\eps - 1))$. This plot shows $p=10^{-3}$. Note the shape of this curve: as $\eps \to 0$, then $\hat\eps \approx p\,\eps$. On the other hand, when $\eps \to \infty$, then $\hat \eps \approx \eps - \log(1/p)$. In all cases, amplification leads to an improvement over the unamplified guarantee denoted by the solid gray line. 
    }
    \label{fig:amplification}
\end{figure}

Privacy amplification guarantees for DP-SGD (i.e., the input perturbation baseline of $\strategy = \bfI_{\nd \times \nd}$) are well understood when the data batches are chosen via \textit{Poisson sampling}: in each iteration, each example is included in the batch independently with probability $p = B / N$, chosen so that the expected batch size is $B$. While the precise details of the amplification analysis are beyond the scope of this monograph, we note that existing libraries can compute exact privacy guarantees for DP-SGD with sampling using only the number of steps $n$, the sampling probability $p$, and the noise multiplier $\nmult$. Indeed, these amplification guarantees are central to achieving practical privacy-utility trade-offs for DP-SGD, as we discuss in \Cref{sec:ml:motivation}. (We provide a deeper dive into amplification in~\Cref{sec:learning:amplification}.)

\paragraph{Drawbacks of Independent Noise Mechanisms}
Prior to the introduction of correlated noise mechanisms, the \textit{de facto} standard for differentially private deep learning was DP-SGD with independent noise (\Cref{alg:dpsgd-1}). This algorithm adds independent Gaussian noise to each stochastic gradient update, and is an instantiation of the input perturbation baseline of \cref{eq:baselineIndepZ}. 
To achieve good practical performance (in terms of downstream learning performance) at moderate privacy budgets, DP-SGD often requires privacy amplification by sampling.
Unfortunately, as we detail next, the data sampling assumptions used for theoretical analyses are often violated in practice.

Privacy amplification analyses typically assume that batches of data are constructed either by {\em Poisson sampling} or {\em fixed-size sampling with replacement}.
These sampling methods can in some regimes give (amplified) privacy guarantees competitive with correlated noise mechanisms while remaining tractable to compute. However, machine learning pipelines typically use fixed-size sampling \emph{without} replacement (usually implemented by shuffling the data).

In the academic literature, it has been a common practice to report privacy guarantees based on Poisson sampling even when the actual models are trained with fixed-size data batches sampled with replacement. This leads to a mismatch between the claimed privacy guarantee and the actual one satisfied by the trained model. Recent work has shown that there is a real gap in the level of DP offered by these two data processing patterns. That is, \textit{the gap is not due to the lack of a tight analysis for shuffling, but a real difference in the level of privacy protection}.  Therefore, this practice should not be considered acceptable, at least in real applications (see \Cref{sec:ch3-biblio} for references). 

Finally, and perhaps most importantly, privacy amplification is challenging or impossible when it is not feasible to obtain a uniformly random sample of data points (whether through Poisson sampling or otherwise) due to domain-specific constraints. We review two such examples.

\begin{bexample}[Federated Learning]
\label{ex:fl}
    Cross-device federated learning refers to a distributed learning setting where several data silos such as smartphones (referred to as ``clients'') collaborate to learn a single model while keeping their training data decentralized. The typical learning algorithm, federated averaging, samples a few \emph{available} clients and constructs a stochastic gradient estimator based on their data. In industrial federated learning systems, the availability of a client varies diurnally, and is determined by external factors such as the device being idle, connected to an unmetered Wi-Fi network, and charging. 
    
    Thus, ensuring that clients are subsampled precisely and uniformly at random from a large population is complex and hard to verify (similar concerns apply to random shuffling). This makes amplification-by-sampling arguments, and thus, amplified DP-SGD (or its federated variant), infeasible. 
    In this federated learning setting, correlated noise mechanisms can maintain provable privacy guarantees with utility comparable to or better than amplified DP-SGD, even with arbitrary changes in client availability. 
\end{bexample}

\begin{bexample}[Learning with Distribution Drift]
\label{ex:drift}
    Distribution drift in learning problems arises from non-stationary data generating distributions. This drift can be independent of the learning process, as seen in continual learning where user behavior shifts over time. For instance, discussions on social networks could change with viral trends and current affairs, while health data collected at hospitals could change with emerging health challenges and evolving medical practices.  Alternatively, drift can be driven by feedback mechanisms where agents strategically adapt their behavior to deployed decision systems for favorable outcomes. For instance, in loan application scenarios where AI models may be used to predict the likelihood of a default, applicants may manipulate model-relevant factors to improve their scores, causing data distribution drift. The model thus actively shapes the observed data rather than passively observing a static environment.
    
    In both scenarios, models should adapt to data drift. This is best achieved by treating data as a stream and continuously learning from recent data while discarding outdated information. This approach necessitates respecting the data's temporal ordering, precluding methods like random batch sampling or using randomly shuffled datasets, which disregard this crucial temporal structure.
 \end{bexample}

\begin{figure}
    \centering
    \adjincludegraphics[width=0.75\linewidth,trim={0 0 25em 0},clip]{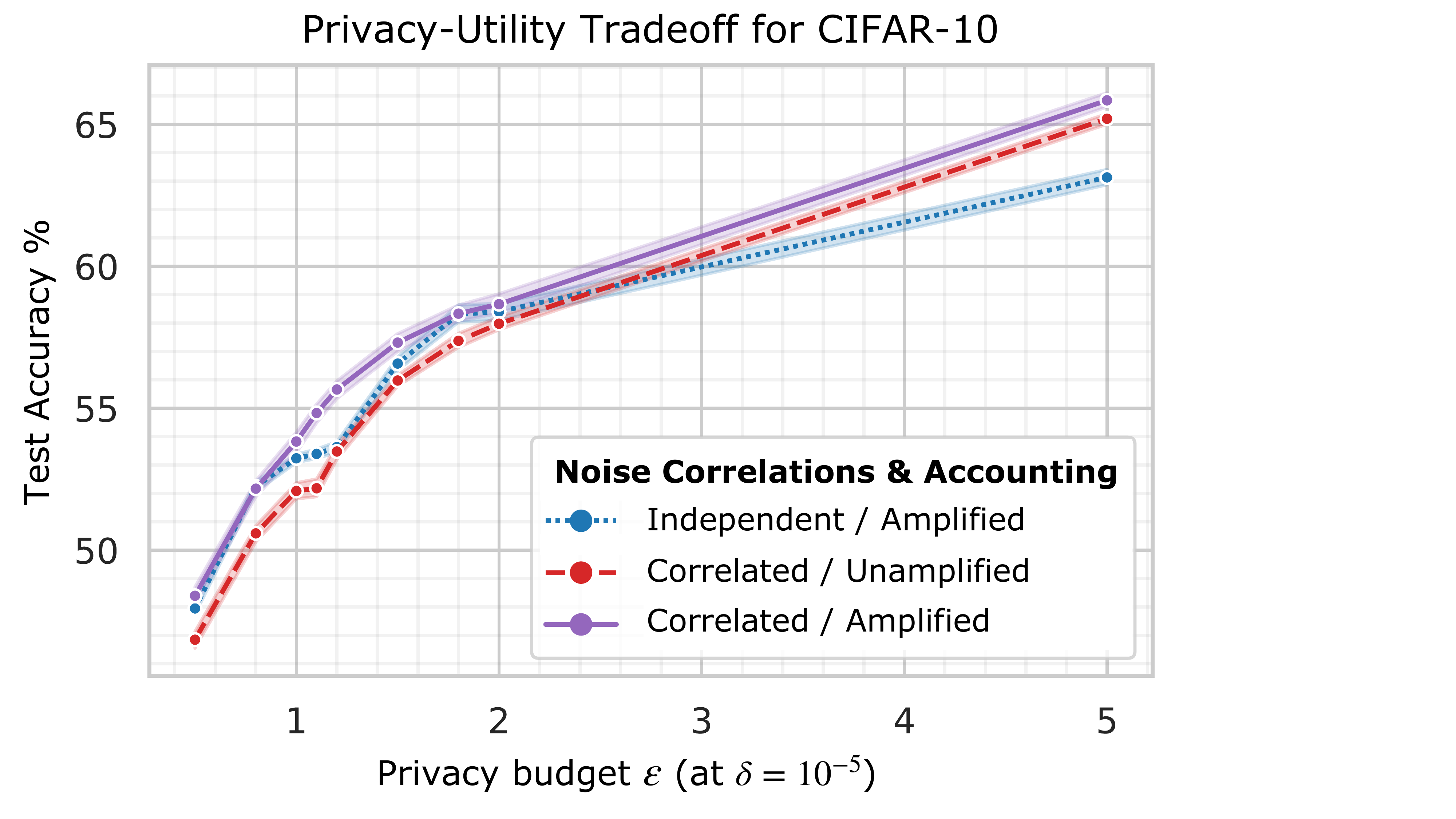}
    \caption{\textbf{Privacy-utility tradeoff of CIFAR-10}: This plot shows the classification accuracy of DP-SGD with independent noise vs. correlated noise calibrated to achieve $(\eps, 10^{-5})$-DP with or without privacy amplification by sampling at different privacy budgets $\eps$.
    We make two key observations. First, correlated noise \emph{without} amplification outperforms independent noise \emph{with} amplification at $\eps \ge 3$. Second, correlated noise \emph{with} amplification uniformly outperforms independent noise at all $\eps$.
    Figure reproduced from \citet*[Fig. 1a]{choquette2024amplified} with permission; we refer the reader there for details. \\
    \emph{Caveat}:
    While these accuracies are not state-of-the-art for the CIFAR-10 benchmark, they are representative of the general performance gain that can be expected from correlated noise. Indeed, these numbers are reported on a small convolutional neural network with 6 conv layers, one dense layer and around half a million parameters, trained with a batch size of $\bsz=500$. Significantly higher accuracies can be obtained on the same CIFAR-10 dataset under the same privacy budgets with larger batch sizes (e.g. $\bsz=65536$), larger models (e.g. a 40-layer Wide-ResNet with 9 million parameters), and data augmentation~\citep*{de2022unlocking}.
    }
    \label{fig:cifar10-pareto-opt}
\end{figure}

\paragraph{Advantages of Correlated Noise Mechanisms}
In both these examples, amplification by sampling/shuffling is not practically feasible. In the absence of amplification arguments, DP-SGD with independent noise suffers from a huge drop in downstream task utility (e.g. accuracy for classification problems) at small or moderate privacy budgets.\footnote{
    For example, privacy amplification by sampling improves the classification accuracy in the CIFAR-10 setting of \Cref{fig:cifar10-pareto-opt} by around $10$ points for $\eps\approx 7$ in \citet*[Fig. 1b]{kairouz2021practical}.
}
In such instances, DP-SGD with correlated noise (\Cref{alg:dpsgd-corr}), known also as ``DP-FTRL'' (see also \Cref{sec:dpftrl-name-background}), has been found to achieve competitive performance \emph{without} the need for privacy amplification.

In particular, correlated noise DP-SGD \emph{without} amplification outperforms independent noise DP-SGD \emph{with} amplification for moderate to large privacy budgets. We refer to \Cref{fig:cifar10-pareto-opt} for an example.
This method guarantees strict privacy regardless of sampling constraints, even in scenarios where random sampling is limited or infeasible, while delivering good downstream task performance. In cases where sampling is viable, we can provide amplified privacy guarantees for correlated noise mechanisms (\Cref{sec:learning:amplification}), offering the best of both worlds.
Indeed, correlated noise \emph{with} amplification uniformly outperforms amplified independent noise DP-SGD for all privacy budgets.

\section{Learning Problems as Weighted Prefix Sums} 
\label{sec:ml:setup}
Recall the setting of stochastic gradient descent (SGD) from \Cref{sec:weightedPrefixSum}: in step $t$ of a learning algorithm, we receive a batch 
(per-sample clipped) gradient $\gradient_t \in \R^\mdim$ calculated under the current model parameters $\bftheta_t$. The SGD updates is then  computed from the prefix sums of the gradients $\Gradients = (\gradient_0, \gradient_1, \ldots, \gradient_{\nd-1}) \in \R^{\nd \times m}$; which is obtained as linear map $\prefix \Gradients$ with the workload matrix
\begin{align} \label{eq:prefix-sum-matrix}
    \prefix = \begin{pmatrix}
    1 & 0 & \cdots &0 \\
    1 & 1 & \cdots & 0\\
    \vdots & \vdots & \ddots & \vdots  \\
    1 & 1 & \cdots & 1
    \end{pmatrix} \in \{0,1\}^{\nd \times \nd} \,.
\end{align}

More generally, SGD with momentum and weight decay can be described by a linear map as well. This update is defined by the recursion
\begin{align}
\begin{split}
    \bfv_{t+1} &= \beta \bfv_t + \eta \gradient_t \\
    \bftheta_{t+1} &= (1 - \lambda)\bftheta_t - \bfv_{t+1} \,,
\end{split}
\label{eq:momentumRecursion}
\end{align}
where $\eta > 0$ is the \textit{learning rate}, $\beta \in [0, 1)$ is the \textit{momentum parameter}, $\lambda \in [0, 1)$ is the \textit{weight decay parameter}. 

We first consider the recursion in \Cref{eq:momentumRecursion} case by case. 

\paragraph{Case $\beta=\lambda=0$} When $\beta=\lambda=0$, then \Cref{eq:momentumRecursion} reduces to the  recursion
\begin{align*}
    \bfv_{t+1} &= \eta \gradient_t \quad \text{and} \quad 
    \bftheta_{t+1} = \bftheta_t - \bfv_{t+1} \,.
\end{align*}
This is exactly the iterates of SGD, as we saw in \Cref{alg:sgd-1}.

\paragraph{Case $\beta, \lambda \neq 0$} 
Unrolling the recursion in \Cref{eq:momentumRecursion} with $\bfv_0=\zeros$ in this case, we get\footnote{
    This can be established using, e.g., an induction argument. We leave the details as an exercise to the reader.
} 
\begin{align}
\nonumber
\bftheta_t &= (1- \lambda)^t \bftheta_0 - \eta \sum_{\tau=0}^{t-1} a_{t-1-\tau} \gradient_\tau \,,
\quad\text{where} \quad \\
    a_t &= \sum_{\tau=0}^t \beta^\tau (1- \lambda)^{t-\tau} \,.
 \label{eq:sgdm:weights}
\end{align}
\noindent The weights $a_t$ are an the exponentially decaying function of $t \in \mathbb N$.
Thus, SGD with momentum and weight decay can be obtained from the weighted prefix sum $\Amom\Gradients$ with the Toeplitz workload matrix
\begin{align} \label{eq:momentum-matrix}
    \Amom = \begin{pmatrix}
    a_0 & 0 & \cdots & 0 \\
    a_1 & a_0 & \cdots & 0 \\
    \vdots & \ddots & \ddots & \vdots \\
    a_{\nd-1} & a_{\nd-2} & \cdots & a_0
    \end{pmatrix} 
\end{align}
where the $a_t$'s are defined in \Cref{eq:sgdm:weights}.

\begin{bremark}
Other flavors of momentum, such as the one arising from Nesterov's accelerated gradient method, also admit similar representations as weighted prefix sums. We leave the derivation of the corresponding workload as an exercise to the reader.
\end{bremark}

We will focus our discussion in this section mainly on the unweighted prefix sum workload matrix $\prefix$ from \Cref{eq:prefix-sum-matrix} for concreteness, although all aspects of our discussions hold for more general non-negative Toeplitz and lower triangular workload matrices $\workload$ such as the momentum matrix $\Amom$ from \Cref{eq:momentum-matrix}.

\begin{figure}
    \centering
    \includegraphics[width=\linewidth]{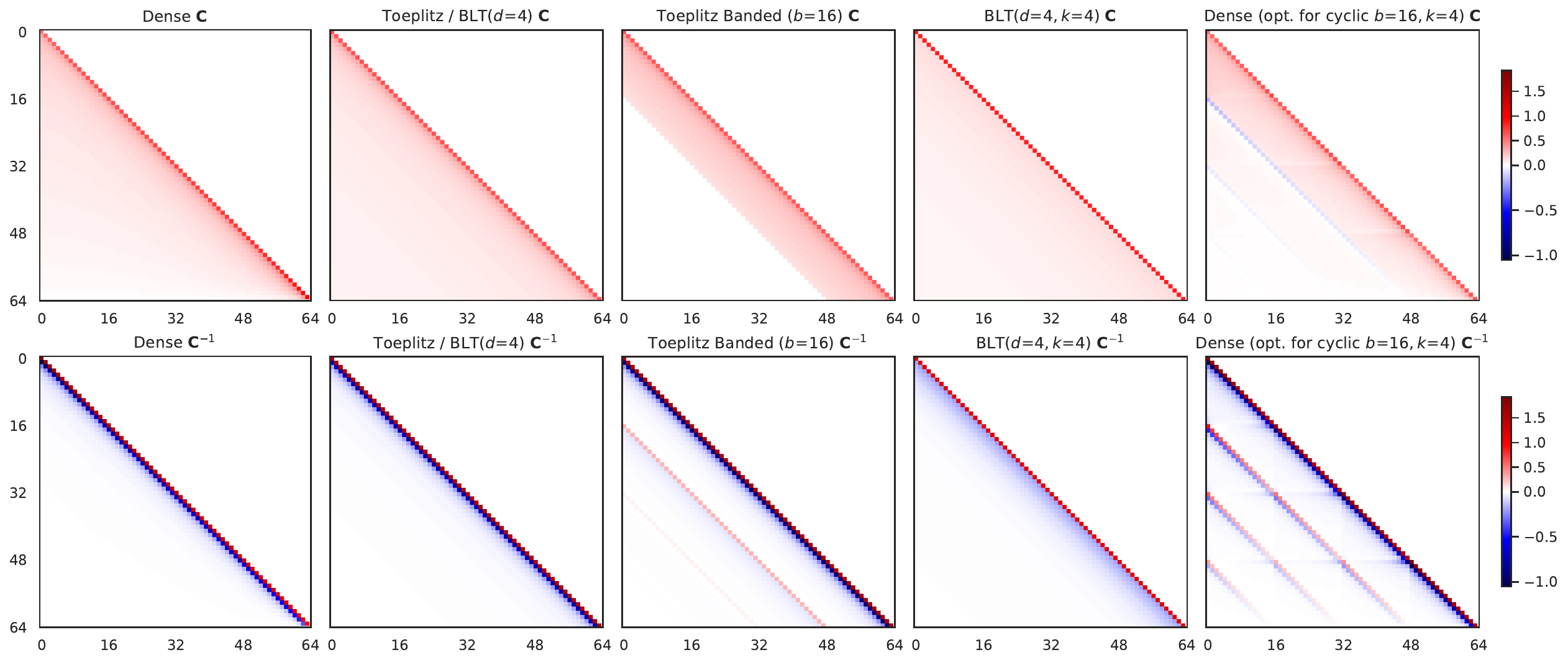}
    \caption{An expanded version of \Cref{fig:heatmaps} which includes additional mechanisms optimized for multiple participations. As before, the first three mechanisms are optimized for single-participation, though the banded mechanism is also optimal for $b \ge 16$ Min-Sep participation, as sensitivity simply scales linearly with the number of participations, \cref{eq:bandedsens}. The BLT mechanism in the fourth column is optimized to minimize \txtmaxloss under $b=16$ Min-Sep participation \cref{eq:min-sep-participation}; the final dense mechanism is optimized for $k=4$ training epochs with $b=16$, each with the same cyclic data order, \cref{eq:cyclic-schema}. Note unlike any of the other matrices, some elements of dense $\bfC$ are, in fact, negative. 
    }
    \label{fig:multi-participation-heatmaps}
\end{figure}

\begin{figure}
    \centering
    \includegraphics[width=\linewidth]{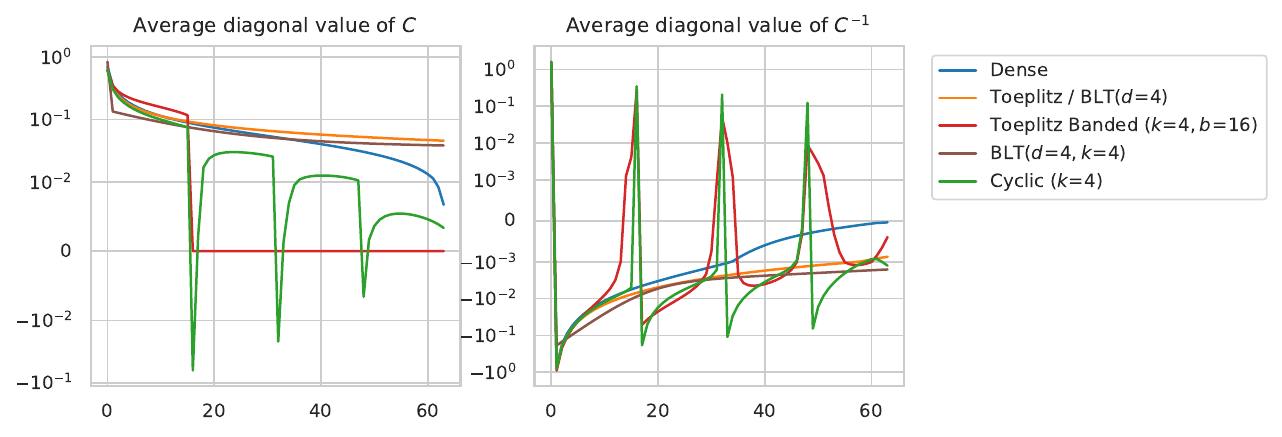}
    \caption{An alternative visualization of the matrices in \Cref{fig:multi-participation-heatmaps}; we represent each matrix by the mean value on each diagonal from the main diagonal and down. Of course, for the Toeplitz mechanisms this representation fully captures the matrix, but it is a lossy summary for the other classes. Nevertheless, these values for $\strategy^{-1}$ in particular, provide an indication of how a mechanism cancels noise; in fact, in the case of the banded and cyclic matrices, the mechanisms not only cancel previously added noise, but sometimes (on a period corresponding to $b$), re-add previously added noise as indicated by positive values below the main diagonal of $\bfC^{-1}$. }
    \label{fig:meandiags}
\end{figure}

\section{Multiple-Participation Correlated Noise Mechanisms}
\label{sec:ml:multi-epoch}

\newcommand{\batchg}{{\gradient}}
\newcommand{\dpbatchg}{\dpgradient}
\begin{figure}[t]
\begin{algorithm}[H]
\caption{Mini-batch DP-SGD with Correlated Noise}
\label{alg:dpsgd-batch}
\textbf{Inputs:} Dataset $D$, number of steps $n$, learning rate $\eta$, batch size $\bsz$, noise variance $\stddev^2$, clip norm $\clipnorm$, noise correlating matrix $\Cinv$ (for standard DP-SGD with independent noise, take $\Cinv = \bfI_{n \times n}$). 
\begin{algorithmic}[1]
\State Define $\clip_\clipnorm(\bfv) := \bfv \cdot \min\{1, \clipnorm / \ltwo{\bfv}\}$
\State Pick an initial model $\bftheta_0 \in \Theta$
\For{$t = 0, 1, \ldots, n - 1$}
\State Receive a batch $\calB_t \subseteq D$ of (expected) size $\bsz$
\State $\batchg_t \gets \frac{1}{\bsz} \sum_{\bfx \in \calB_t}  \clip_\clipnorm\big(\nabla\ell(\bftheta_{t}; \bfx)\big)$  \Comment{\Cref{eq:clippingfunction} for $\clip_\clipnorm(\cdot)$} \label{line:startnoisygrad}
\State Sample $\znoise_t \sim \normalm{0}{\stddev^2}$ \Comment{Sample i.i.d. seed noise}
\State \label{line:corrnoise-calc}
$\corrznoise_t \gets \frac{1}{\bsz} \sum_{\tau = 0}^{t-1} \Cinv\idx{t}{\tau} \znoise_{t, \tau}$ \Comment{Correlated noise $\big(\bfC^{-1} \Znoise\big)\idx{t}{:}$}
\State $\dpbatchg_t \gets \batchg_t + \corrznoise_t$ 
 \Comment{Privatized \textbf{average} gradient.}
\label{line:endnoisygrad}
\State $\bftheta_{t+1} \gets \bftheta_t - \eta \dpbatchg_t$ \Comment{Or any first-order update}
\label{line:optimizerupdate}
\EndFor
\State \textbf{Return} $\bftheta_n$
\end{algorithmic}
\end{algorithm}
\vspace{-1.2em}
\end{figure}

The results of \Cref{chap:intro,chap:prefixsum} are directly applicable for learning problems in the \emph{streaming setting} where the data is processed in a single pass.\footnote{
    A complete pass over the data is termed an \emph{epoch}.
}
Further, we also considered stochastic gradient algorithms with a batch size of one. As we discussed in \Cref{remark:large-batch}, these assumptions are typically violated in practical AI model training scenarios. Indeed, we usually process each example multiple times in the learning process; we refer to this as the \emph{multiple-participation setting}. Moreover, we process mini-batches of examples in each stochastic gradient step.

The version of DP-SGD with correlated noise we consider is given in \Cref{alg:dpsgd-batch}; it generalizes \Cref{alg:dpsgd-corr} to (expected) mini-batch sizes $\bsz$.\footnote{The batch size may only hold in expectation, as Poisson sampling or other sampling approaches \cref{sec:learning:amplification} may produce a randomly-sized batch.} 
The key difference relative to non-private SGD is that we \emph{clip the per-example gradients} before averaging them (Line~\ref{line:startnoisygrad}), and then add (correlated) noise (Line~\ref{line:endnoisygrad}). Critically, we then use the DP estimate of the \emph{average} gradient in the first-order update (Line~\ref{line:optimizerupdate}).
Following standard convention, we assume that the noise variance $\stddev^2$ is calibrated to the sum of gradients. However, since $\gradient_t$ is calculated as the average mini-batch gradient (rather than their sum), we scale down the noise $\corrznoise_t$ by a $\bsz$ factor in Line \ref{line:corrnoise-calc} to ensure correct noise calibration.

\begin{figure*}
    \centering
    \includegraphics[width=\linewidth]{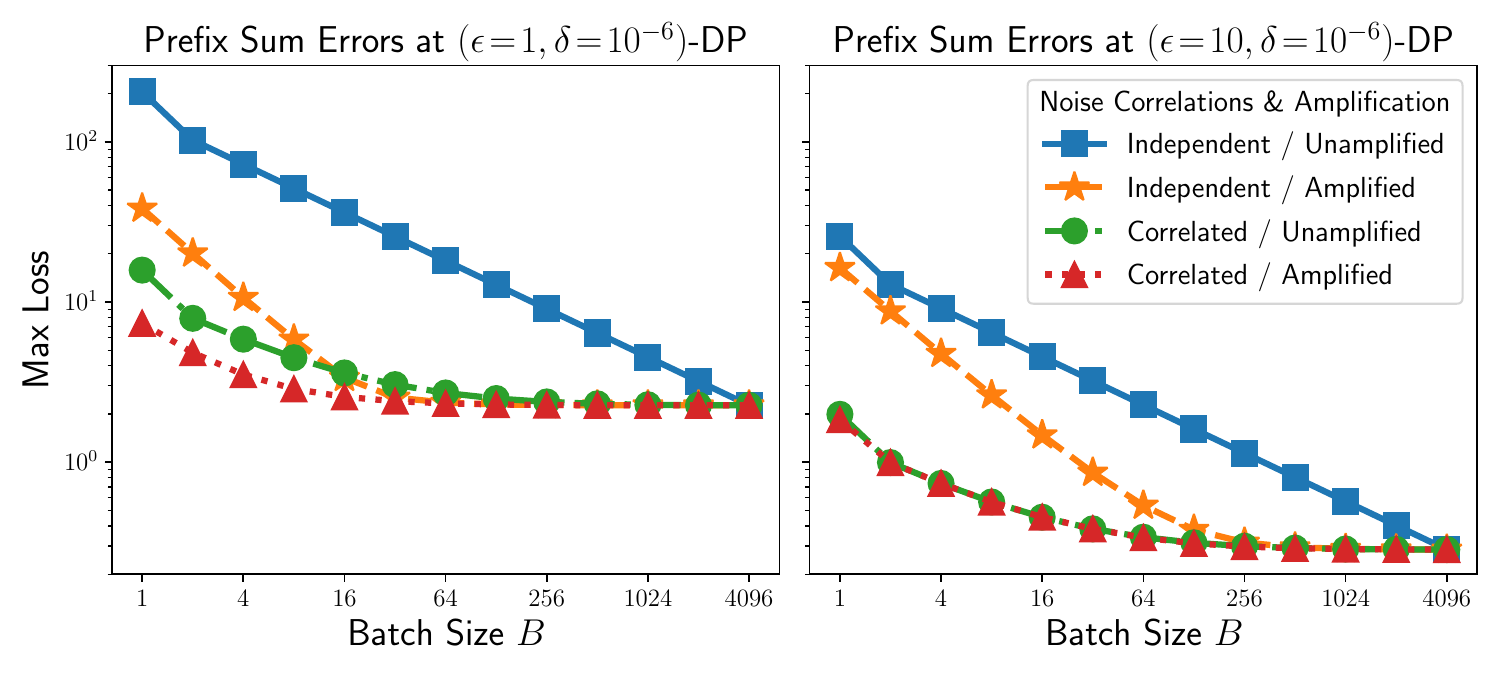} 
    \caption{ \textbf{Effect of the batch size}: Fixing the number of iterations $\nd = 2048$ and dataset size $\datasize=4069$, and increasing the batch size can substantially decrease the max loss in the average gradient of DP-SGD (\cref{alg:dpsgd-batch}, Line~\ref{line:endnoisygrad}) for both correlated noise (in this case, the banded Toeplitz mechanism of \Cref{sec:ch2-banded-Toeplitz}) and independent noise, both with privacy amplification from Poisson sampling (\cref{sec:learning:amplification}) and in unamplified scenarios. Because $\datasize/\nd = 2$, the number of epochs $\maxpart$ is equal to $\bsz/2$. Thus $\bsz=1$ and $\bsz=2$ correspond to the streaming setting, while $\bsz=4096$ corresponds to full-batch gradient descent. For the banded Toeplitz mechanism with cyclic Poisson sampling, the number of bands was empirically optimized to minimize the loss; using fewer bands increases the benefits of amplification, while more bands allows more flexibility in designing correlated noise. At $\epsilon=10$, correlated noise provides most of the benefit, but at $\epsilon=1$ amplification yields significant additional improvements for small batch sizes.
   } \label{fig:scaling_batch_size}
\end{figure*}

\begin{bremark}\label{rem:increasing-batch-size}
    Fixing the number of training iterations $\nd$, the (expected) batch size $\bsz$ is a hyperparameter that plays an essential role in privacy-utility tradeoffs as well as compute cost. The prefix-sum \txtmaxloss of the \emph{average} gradient $\dpbatchg_t$ is a reasonable proxy for learning performance. In the zero-out model, we have this \emph{unnormalized} max loss is 
    \begin{align}
      \maxloss(\dpbatchg; \bfB, \strategy) 
      &= \max_{t \in [n]} \sqrt{\mathbb{E} \norm{\sum_{\tau=0}^t \batchg_\tau - \sum_{\tau=0}^t \dpbatchg_\tau}}_2 \notag \\
      &=  \frac{\nmult \mdim}{B}\rownorm{\bfB} \sens(\strategy) \label{eq:mean-grad-loss}
    \end{align}
    following \cref{thm:accuracyMatrixMechanism} and \cref{eq:loss-error-sens}.
    Hence, if we could increase the batch size without changing anything else (e.g., if we had infinite compute and an infinite dataset), we could drive the \txtmaxloss down to be arbitrarily small. Of course, this requires $\nd \bsz$ examples, and so if our dataset is of fixed size $\datasize$, some example must participate at least $\maxpart = \ceil*{\frac{\nd \bsz}{\datasize}}$ times, and this will increase the sensitivity $\sens(\strategy)$. Formally defining and then computing $\sens(\strategy)$ in multiple-participation settings will be a principal topic in this section.
    
    Once we account for the increased sensitivity (or equivalently, the impact of the additional participations under sampling), significant benefit from increasing the batch size remains. In fact, if compute cost was not an issue, we would always simply process the entire training dataset in every batch. \emph{In the extreme of full-batch gradient descent, neither correlated noise nor privacy amplification via sampling offer any advantage.}\footnotemark  ~\Cref{fig:scaling_batch_size} is representative, showing the value of increasing batch size is pronounced for both the banded Toeplitz mechanism of \cref{sec:ch2-banded-Toeplitz} and vanilla DP-SGD with independent noise. 
    Yet, we find that both correlated noise and amplification can (independently or jointly) lead to significant improvements in compute-constrained settings, where using the full batch for each gradient update is infeasible.
    We will revisit full-batch training in \cref{remark:full-batch}.
\end{bremark}

\footnotetext{
\Cref{prop:independent-noise-optimal} shows that independent noise achieves the optimal \txtmaxloss in the full-batch setting, meaning that noise correlations do not give any additional benefits.
Similarly, there is no added uncertainty from sampling when we use the full batch, so there is no amplification by sampling in this regime.
}

In order to run this algorithm, we need to specify how to set the noise standard deviation $\stddev^2$ as a function of the problem parameters and the desired privacy level in this multiple-participation regime.
It turns out that the analysis of \Cref{chap:intro} straightforwardly generalizes to larger mini-batch sizes. These techniques are, however, insufficient to handle the multiple-participation setting.

Recall from \cref{chap:intro} (more specifically \Cref{sec:ch1-privacy-correlated-noise}) that the streaming setting combined with \cref{thm:gdp-adaptivity} made it straightforward to translate adjacency of datasets into adjacency of gradients: since each data point is processed only once,
two sequences of gradients $\Gradients = (\gradient_0, \ldots, \gradient_{\nd-1})\in \R^{\nd \times m}$ and $\Gradients' = (\gradient_0', \ldots, \gradient_{\nd-1}')\in \R^{\nd \times m}$ arising from adjacent datasets $D$ and $D'$ can differ in at most one element, i.e., $\gradient_\tau = \gradient_\tau'$ for all $\tau \in [n]$ except possibly for one index $t$ (see \cref{thm:gdp-adaptivity} for more discussion). 

However, model training typically involves making $\maxpart > 1$ passes over the data.\footnote{
    Other forms of multiple participation are possible, e.g., in federated learning where data can be processed in arbitrary order depending on device availability. We discuss this further in \Cref{sec:multi-epoch:practical-patterns}.
}
A change of one data point in an adjacent dataset leads to a change in $\maxpart$ gradients in the input $\Gradients$. The goal of this section is to extend the results of the preceding sections to the multiple-participation setting. 

\subsection{Multiple-Participation Training Setup}
\label{sec:multi-epoch:training-setup}

Suppose we have a dataset of $\datasize$ data points, which appear in $\nd$ iterations in batches of size $\bsz$. We are interested in the multiple-participation setting where $\nd\bsz/\datasize > 1$, which implies that at least one data point is used at least twice. Further, we let $\maxpart$ denote the maximum number of times a data point might participate in training. For example, $k$ can refer to the number of epochs in settings where that can be tracked.

The input to our correlated noise mechanism is a sequence of gradients $\Gradients= (\gradient_0, \ldots, \gradient_{\nd-1}) \in \R^{\nd \times \mdim}$ obtained from a dataset $D = \{\bfx_1, \ldots, \bfx_\datasize\}$ of data points. We assume that each gradient $\gradient_t$ is obtained as the total gradient over a batch $I_t \subset [\datasize]$ of $|I_t| = B$ data points (evaluated at the current model $\bftheta_t$):
\begin{align} \label{eq:minibatch-gradient}
    \gradient_t = \frac{1}{\bsz}\sum_{i \in I_t} \nabla \ell(\bftheta_t, \bfx_i) \,.
\end{align}
Throughout this section, we assume a \emph{gradient norm bound} of $1$: 
\begin{equation}\label{eq:gradnormbound}
    \norm{\nabla \ell(\bftheta, \bfx)}_2 \le 1 \quad 
    \text{for all } \bftheta \in \Theta \text{ and } \bfx \in \calX \,. 
\end{equation}  
In practice, this can be achieved by the so-called \emph{gradient clipping} operation; see \cref{alg:dpsgd-batch}, Line~\ref{line:startnoisygrad}. As mentioned in \Cref{chap:intro}, the case of clipping to a different norm can simply be handled by scaling.

\paragraph{Adjacent Datasets and Sequences} 
We define adjacent sequences of gradients based on adjacency of the underlying datasets, using the \emph{zero-out} notion of the adjacency (following much prior work in this setting; see \cref{sec:ch3-biblio}).   \Cref{sec:intro:different-adj} provides additional discussion of this choice, but for the remainder we only need the definition, which we restate for convenience:

\begin{bdefinition}[Zero-out Adjacency of Datasets]
\label{def:zeroout2}
Two datasets $D = \{\bfx_1, \ldots, \bfx_{\datasize}\}$ and $D' = \{\bfx_1', \ldots, \bfx_{\datasize}'\}$ are \textbf{zero-out adjacent} (denoted $D \simeq D'$) if $\bfx_i = \bfx_i$ for all indices $i \in [n]$ except possibly at some index $j \in [n]$. At this index $j$, we have either $\bfx_j = \perp$ or $\bfx_j' = \perp$, where ``$\perp$'' is a special \emph{null} element such that $\nabla \ell(\bftheta, \perp) = \zeros$ for all $\bftheta$.
\end{bdefinition}
Replacing an example by $\perp$ is a convenient way to capture the removal of a data point without changing the dataset size or how examples are grouped into batches. This \emph{zero-out} notion of adjacency generally yields sensitivities that are a factor of 2 smaller than \emph{replace-one adjacency} which we use in \Cref{chap:intro,chap:prefixsum}.

Recalling \cref{thm:gdp-adaptivity}, we may assume a \emph{fixed} sequence of model iterates $\thseq$ when defining adjacent sequences of gradients: 
\begin{bdefinition}[Zero-out Adjacency of Gradients]
\label{def:adjacency-of-grads}
Two sequences of gradients $\Gradients= (\gradient_0, \ldots, \gradient_{\nd-1}) \in \R^{\nd \times \mdim}$ and $\Gradients' = (\gradient_0', \ldots, \gradient_{\nd-1}') \in \R^{\nd \times \mdim}$ of gradients are \emph{adjacent} if
for each $t \in [n]$, we have that
\[
    \gradient_t = \sum_{i \in I_t} \nabla \ell(\bftheta_t, \bfx_i)
    \quad \text{and} \quad
    \gradient_t' = \sum_{i \in I_t} \nabla \ell(\bftheta_t, \bfx_i')
\]
are evaluated for the same sequence of batch indices $I_0, \ldots, I_{\nd-1} \subset [\nd]$ and the same sequence of models $\thseq$ on adjacent datasets $D = \{\bfx_i\}_{i=1}^\datasize$ and $D' = \{\bfx_i'\}_{i=1}^\datasize$. 
We say that $\Gradients, \Gradients'$ are zero-out adjacent if the underlying datasets $D \simeq D'$ are zero-out adjacent; we denote this as $\Gradients\simeq \Gradients'$.
\end{bdefinition}

\paragraph{Participation Patterns}
Let $I_t$ be the set of batch indices that is picked at time $t$. We say that a data point $\bfx_i$ \emph{participates} in step $t$ if $i \in I_t$.
Note that if a data point can participate in up to $k$ batches $I_{t_1}, \ldots, I_{t_k}$, then $\Gradients$ and $\Gradients'$ can differ in up to $k$ gradients $\gradient_{t_1}, \ldots, \gradient_{t_k}$.

\begin{figure*}
    \centering
    \includegraphics[width=0.48\linewidth]{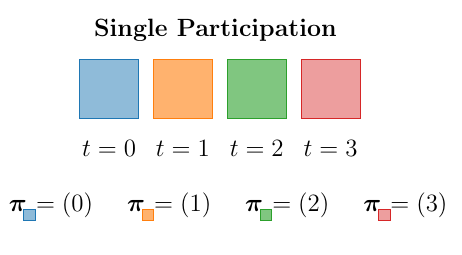} \hfill
    \includegraphics[width=0.48\linewidth]{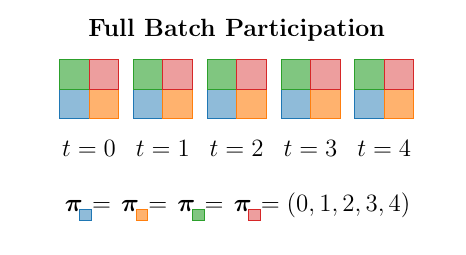}
    
    \includegraphics[width=\linewidth]{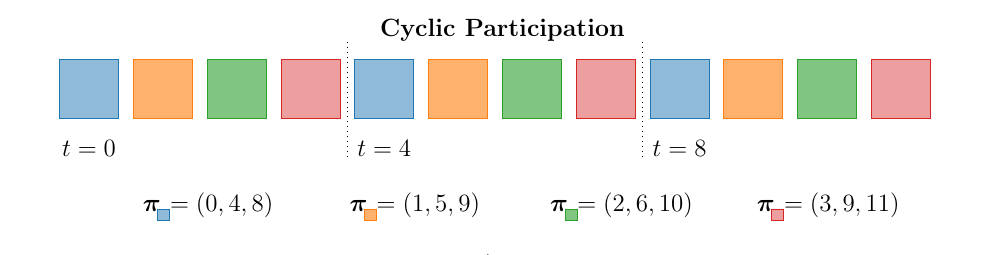}
    \caption{Illustration of participation patterns where each color represents a data point.
    \textbf{Top Left}: The streaming setting with $\datasize=4$ data points over $\nd=4$ steps with a batch size of $\bsz=1$ for one epoch ($\maxpart=1$). 
    \textbf{Top Right}: Full batch participation of $\datasize=4$ data points over $\nd=5$ steps (the batch size is $\bsz=\datasize$ and number of epochs are $\maxpart=\nd$).
    \textbf{Bottom}: Cyclic participation of $\datasize=4$ data points over $\nd=12$ steps with a batch size of $\bsz=1$, leading to $\maxpart=\nd \bsz/\datasize=3$ epochs (separated by the dotted line).
    }
    \label{fig:participation_patterns}
\end{figure*}

It is useful to define the sequence $\bfpi_i$ of all steps where $\bfx_i$ participates in training:
\begin{align} \label{eq:participation-pattern}
    \bfpi_i := \big( t \in [n] \, :\, i \in I_t\big) \,.
\end{align}

We refer to $\bfpi_i$ as the \emph{participation pattern} of data point $\bfx_i$. For instance, $\bfpi_i = [n]$ indicates that data point $i$ participates in each step, while $|\bfpi_i|=1$ corresponds to the setting where $\bfx_i$ only appears once during training.
Finally, if data point $i$ participates in the $\ell$\textsuperscript{th} step of each cyclic pass over the data, we have (assuming the batch size $\bsz$ divides the dataset size $\datasize$):
\begin{align} \label{eq:cyclical-pattern}
    \bfpi_i = \left(l, \,\, l + \frac{\datasize}{\bsz}, l + 2\frac{\datasize}{\bsz}, \, \ldots, \, l + (k-1) \frac{\datasize}{\bsz}\right) \,.
\end{align}

A natural baseline is to simply ``restart'' the mechanism at the end of each epoch (assuming the algorithm can run in epochs). Then, we can simply compose the privacy loss across epochs using Lemma~\ref{lem:gdp-composition}. We will instead derive tight sensitivity bounds for the entire multiple-participation training algorithm as one mechanism. This will recover the ``restarting mechanism'' as a special case; we return to this in \Cref{sec:multi-epoch:restart}.

As we will see in the next section (\Cref{sec:multi-epoch:sensitivity}), tight sensitivity bounds in the multiple-participation setting requires us to impose restrictions on the allowed participation patterns. We will define specific participation patterns in \Cref{sec:multi-epoch:practical-patterns} for further study.

\subsection{Multiple-Participation Sensitivity: Definition and Challenges}
\label{sec:multi-epoch:sensitivity}

Given a stream of gradients $\Gradients = (\gradient_0, \ldots, \gradient_{n-1}) \in \R^{n \times m}$ as input, the correlated noise mechanism $\mech(\Gradients) = \strategy\Gradients+ \Znoise$ is an instance of the Gaussian mechanism with i.i.d. Gaussian noise $\Znoise \sim \normalnm{0}{\stddev^2}$  (\Cref{def:gaussian-mechanism}). Thus, following \Cref{lem:gdp-of-gaussian-mechanism}, we only need to bound the sensitivity of the linear map $\Gradients\mapsto \strategy\Gradients$ to design mechanisms suited to the multiple-participation case. Recall we did this for the single-participation streaming setting in \cref{thm:gdp-of-correlated-noise-adaptive}; we now generalize this result. Similar to that case, we can simply consider a non-adaptive setting, thanks to \Cref{thm:gdp-adaptivity}. 

\begin{bdefinition}[$\ell_2$ Sensitivity Induced by the Strategy Matrix]
\label{sec:def:sens-worst-case}
The $\ell_2$-sensitivity of the matrix-valued map $\Gradients\mapsto \strategy \Gradients$ is
\begin{align} \label{eq:sens-def-no-restrictions}
    \sens(\strategy) = \sup_{\Gradients\simeq \Gradients' \in \R^{\nd \times \mdim}} \lfrob{\strategy\Gradients- \strategy\Gradients'} \,,
\end{align}
where the Frobenius norm $\lfrob{\bfM} = \sqrt{\sum_{i, j} \bfM[i, j]^2}$ generalizes the $\ell_2$-norm of \Cref{def:l2-sens} to matrix-valued maps.
\end{bdefinition}

Given a bound on this sensitivity, we can get a GDP guarantee, generalizing  \Cref{thm:gdp-of-correlated-noise} and \Cref{thm:gdp-of-correlated-noise-adaptive}:

\begin{blemma}
    \label[lemma]{lem:gdp-of-correlated-noise:multi-part:non-adaptive}
Fix a noise multiplier $\nmult > 0$.
Consider the non-adaptive multi-participation setting, i.e.
\begin{enumerate}[label=(\alph*)]
    \item \label{item:gdp-corr-ch3-a}
    we have that $\Gradients \simeq \Gradients'$ are adjacent in the zero-out sense, as defined in  \Cref{def:adjacency-of-grads}, and
    \item \label{item:gdp-corr-ch3-b}
    the rows $\gradient_t, \gradient_t'$ are clipped to norm 1 (cf. \cref{eq:ch1-grad-norm-bound} or \cref{eq:gradnormbound}).
\end{enumerate}
Then, the mechanism $\mech(\Gradients) = \bfB(\strategy \Gradients + \Znoise)$
for any matrices $\bfB, \strategy \in \R^{\nd \times\nd}$ (possibly non-lower-triangular and non-invertible) and i.i.d. Gaussian noise $\Znoise \sim \calN_{\nd \times \mdim}(0,{\stddev^2})$ satisfies $\frac{1}{\nmult}$-GDP if we choose the noise standard deviation as $\stddev = \nmult\, \sens(\strategy)$.
\end{blemma}

Recall that \Cref{thm:gdp-adaptivity}
in \Cref{chap:intro} reduces the adaptive case to the nonadaptive case. Then, similar to \Cref{thm:gdp-of-correlated-noise-adaptive}, this directly leads to a GDP bound on \Cref{alg:dpsgd-batch}:

\begin{btheorem}\label{thm:gdp-of-correlated-noise-multiple-participation}
Fix a noise multiplier $\nmult$ and consider \Cref{alg:dpsgd-batch} with an invertible lower-triangular strategy matrix $\strategy$, clip norm $\clipnorm= 1$, any batch size $\bsz > 0$, and noise standard deviation $\stddev = \nmult \, \sens(\strategy)$. Then, the privatized gradients $(\dpgradient_t)_{t\in[n]}$ and iterates $(\bftheta_t)_{t \in [n]}$ produced by \Cref{alg:dpsgd-batch} satisfy $\frac{1}{\sigma}$-GDP under the zero-out adjacency of input datasets (\Cref{def:zeroout2}), even in the multi-participation setting.
\end{btheorem}

Comparing \Cref{thm:gdp-of-correlated-noise-multiple-participation} for the multi-participation case to to \Cref{thm:gdp-of-correlated-noise-adaptive} for streaming setting, we note that the sensitivity $\sens(\strategy)$ does the heavy lifting in capturing
the multiple-participation aspect.
This sensitivity bound of \Cref{sec:def:sens-worst-case} is required to hold over all pairs of adjacent sequences of gradients $\Gradients\simeq \Gradients'$ as in \cref{def:adjacency-of-grads}. In the (non-adaptive version of the) streaming setting, $\Gradients$ and $\Gradients'$ can differ only in one row, and this difference is limited by the gradient norm bound. Thus, we recover $\sens(\strategy) = \colnorm{\strategy}$ as in \Cref{thm:gdp-of-correlated-noise-adaptive}.\footnote{
    The multiplicative factor of $2$ in \Cref{thm:gdp-of-correlated-noise-adaptive} is due to the replace-one adjacency, and vanishes if we consider the zero-out adjacency as in this section. See \Cref{sec:intro:different-adj} for a detailed discussion on different notions of adjacency. 
}
We can have $\sens(\strategy) > \colnorm{\strategy}$ in the multi-participation setting, as multiple rows of $\Gradients$ can change in an adjacent dataset (corresponding to the participation of the underlying example that changes).

If we do not impose any restrictions on the allowed participation patterns for the multiple-participation setting, a single data point can participate in every single iteration in the worst case. Thus, it is possible to have $\Gradients\simeq \Gradients'$ that differ in \emph{every row} by an $\ell_2$ distance of up to 1 (due to the gradient norm bound assumption of \cref{eq:gradnormbound}). By our definition of adjacency, it follows for the simpler case of $m=1$ dimension that two sequences $\Gradients, \Gradients' \in \R^{n \times 1}$ are \emph{adjacent} (in the absence of participation pattern restrictions) if $\norm{\Vector(\Gradients- \Gradients')}_\infty \le 1$, where $\Vector(\cdot)$ treats  its matrix input of shape $\nd \times 1$ as an $\nd$-dimensional vector.

We then have for the $\ell_2$-sensitivity
\begin{align} \label{eq:sens-worst-case}
\begin{aligned}
    \sens(\strategy) &= \max_{\norm{\Vector(\Gradients- \Gradients')}_\infty \le 1} \lfrob{\strategy (\Gradients- \Gradients')} \\
    &= \max_{\norm{\bfu}_\infty \le 1} \norm{\strategy \bfu}_2
    =: \norm{\strategy}_{\infty \to 2} \,,
\end{aligned}
\end{align}
where the Frobenius norm of \Cref{eq:sens-def-no-restrictions} reduces to the $\ell_2$ norm for $m=1$ dimension in \Cref{eq:sens-worst-case}. 
Here, the notation $\norm{\strategy}_{\infty \to 2}$ denotes the $\infty\to 2$ matrix operator norm (also known as the induced matrix norm). In general, for $p, q \in [1, \infty]$, the $p\to q$ operator norm of a matrix is defined as:
\begin{align} \label{eq:induced-norm}
    \norm{\strategy}_{p \to q} := \max_{\bfu\neq \zeros} \frac{\norm{\strategy \bfu}_q}{\norm{\bfu}_p} \,.
\end{align}

\begin{figure*}[t]
    \includegraphics[width=0.49\linewidth]{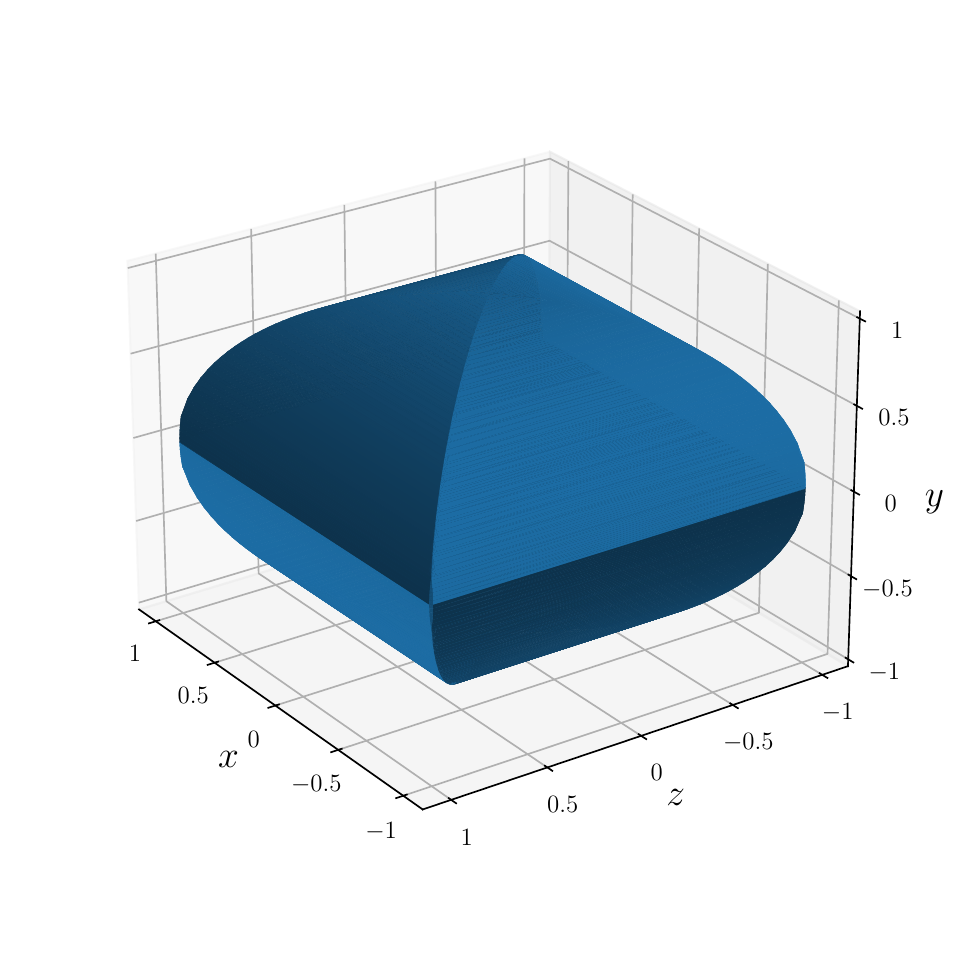} \hfill
    \includegraphics[width=0.49\linewidth]{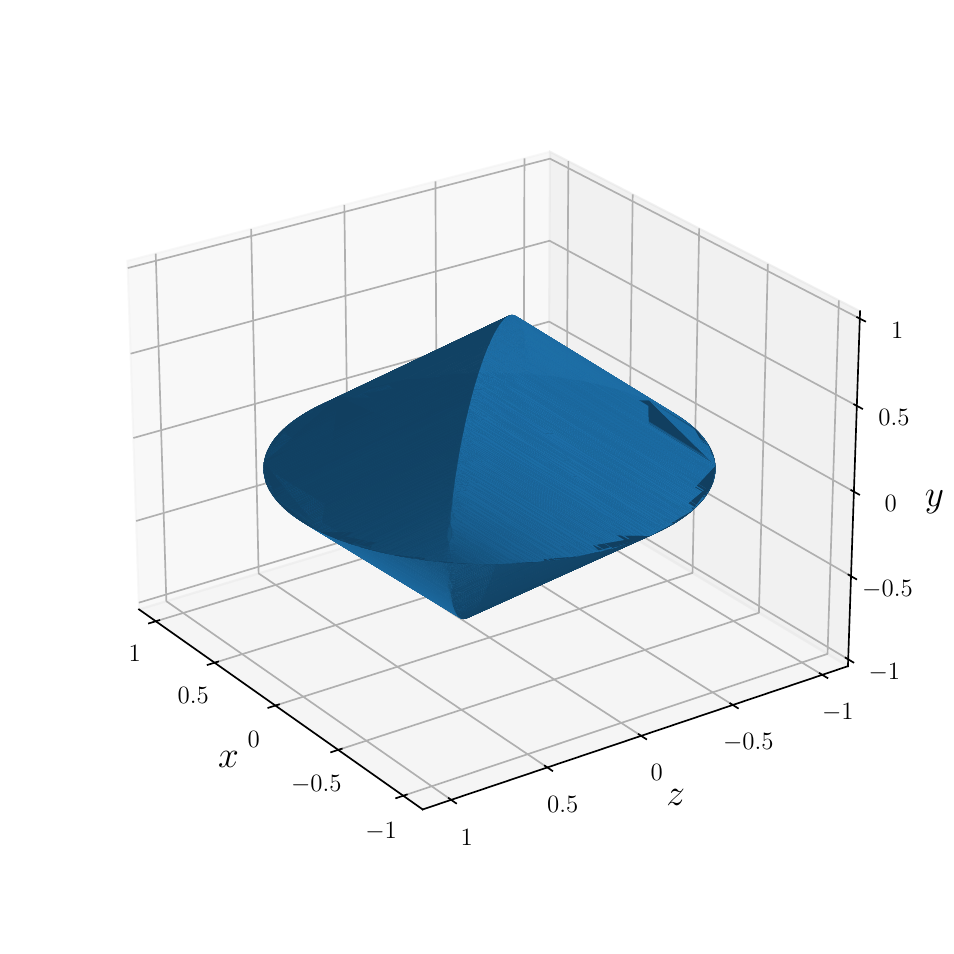}
    \caption{\textbf{Operator norm balls}: 
    The norm balls of the operator norms $\colnorm{\bfM} = \colnormop{\bfM}$ (left) and $\norm{\bfM}_{\infty \to 2}$ (right) of the symmetric matrix $\bfM = \begin{pmatrix} x & y \\ y & z \end{pmatrix}$ plotted as a function of $(x, y, z)$.
    The left norm ball $\{\bfM \, : \, \colnormop{\bfM} \le 1\}$ is much bigger than the right one $\{\bfM \, : \, \norm{\bfM}_{\infty \to 2} \le 1\}$, meaning that more matrices satisfy the constraint $\colnormop{\bfM} \le 1$ than the constraint $\norm{\bfM}_{\infty \to 2} \le 1$; see Lemma~\ref{lem:operator-norms}. Figures adapted from \citet{grimmer2022induced} with permission.
    }
    \label{fig:operator_norms}
\end{figure*}

\paragraph{Challenges}
Unfortunately, the sensitivity in \cref{eq:sens-worst-case} that we obtained in the absence of participation restrictions suffers from some major difficulties.
First, computing 
the operator norm $\norm{\strategy}_{\infty \to 2}$ for general $\strategy$ matrices is NP-hard.\footnote{
    To be precise, the corresponding decision problem is NP-hard: does there exist a vector $\bfu \in \R^n$ with $\norm{\bfu}_\infty \le 1$ such that $\norm{\strategy\bfu}_2 \le \kappa$ for some given constant $\kappa > 0$?
} 
Second, the multiple-participation sensitivity can be significantly larger than the single-participation sensitivity $\colnorm{\strategy}$. In particular, the multiple-participation sensitivity can be worse by a factor of the number $\nd$ of steps (see \Cref{sec:ch3-proofs} for a proof): 

\begin{blemma} \label[lemma]{lem:operator-norms}
    For any matrix $\strategy \in \R^{p \times \nd}$, we have,
    \[
        \colnorm{\strategy} \le \norm{\strategy}_{\infty \to 2} \le \nd \colnorm{\strategy}.
    \]
\end{blemma} 

This large spike in the sensitivity means that independent noise achieves the optimal \txtmaxloss in this setting (see \Cref{sec:ch3-proofs} for a proof):

\begin{bproposition}
\label[proposition]{prop:independent-noise-optimal}
    We have that
    \[
        \min_{\strategy \in \R^{\nd \times \nd}}
        \left\{ \rownorm{\prefix \strategy^{-1}} \, \norm{\strategy}_{\infty \to 2} \right\} = \nd\,,
    \]
    and this minimum is attained at $\strategy=\bfI_{n\times n}$.
    That is, the correlated noise mechanism $\mech(\Gradients) = \strategy \Gradients + \Znoise$ in $\mdim = 1$ dimension that achieves the optimal \txtmaxloss with the worst-case multi-participation sensitivity (\cref{eq:sens-worst-case}) is independent noise $\strategy=\bfI_{n \times n}$.
\end{bproposition}

However, in contrast to \Cref{prop:independent-noise-optimal}, empirically, we observe that carefully designed correlated noise mechanisms lead to significant practical improvements (e.g., in \Cref{fig:cifar10-pareto-opt,fig:scaling_batch_size}). A key reason for the limitations \Cref{eq:sens-worst-case} and the approach of this section is that, without participation restrictions, the matrices $\Gradients$ and $\Gradients'$ can differ arbitrarily in every row. This setup fails to reflect practical scenarios, where examples typically participate a few times during training, but not in every iteration.

In particular, as discussed in \cref{rem:increasing-batch-size}, when sufficient compute resources are available, increasing the batch size is generally beneficial, often pushing us into the multi-participation regime. So, a natural approach would be to use full-batch gradient descent  (where $\Gradients\simeq \Gradients'$ may differ in every row). However, empirical evidence suggests that the benefit of using full-batch gradient descent in terms of loss is not significant to warrant using the extra compute resource. This can be also seen in \cref{fig:scaling_batch_size}, where a batch size of $\bsz=256$ achieves nearly the same loss as using the full dataset ($\bsz=4096$) while requiring $16\times$ less compute, even without amplification.

These discussions suggests that we should seek reasonable restrictions on the allowed participation patterns (and sometimes also on the matrix $\strategy)$ to design correlated noise mechanisms for the multi-participation setting. In particular, we have the following desiderata:
\begin{enumerate}[label=(\alph*), nosep]
    \item \textbf{Implementation is practical}: The restrictions on participation patterns can be implemented efficiently in real-world ML training pipelines.
    \item \textbf{Computing $\sens(\strategy)$ is tractable}: The multiple-participation sensitivity can be computed (or tightly bounded from above) in polynomial time in the problem parameters, namely the number of training samples, $\nd$, and the model dimension, $\mdim$, for at least some classes of strategy matrices $\strategy$. Preferably, it can be computed in $O(\nd)$ time or less and is independent of the model dimension $\mdim$.
    \item \textbf{Larger batches give improvements:} Finally, we seek restrictions where the unnormalized max loss (\cref{eq:mean-grad-loss}) improves as we increase the batch size (and compute cost) --- otherwise, we would be better off just using a smaller batch size and staying in the streaming setting. (These can typically only be verified empirically, as in \cref{fig:scaling_batch_size}.)
\end{enumerate}
Next, we will define participation patterns that satisfy these criteria and derive sensitivity bounds.

\subsection{Practical Participation Patterns}
\label{sec:multi-epoch:practical-patterns}

Recall that the participation pattern $\bfpi \subset [\nd]$ of a data point is the sequence of steps in which it participates in training; see \Cref{eq:participation-pattern} for a definition. 
We define the set of all ``allowed'' participation patterns as an abstraction to impose restrictions on how we process the data:

\begin{bdefinition}[Participation Schema]
\label{def:participation-schema}
    A \textit{participation schema} $\Pi \subset 2^{[\nd]}$ is a subset of possible participation patterns, called the \textit{allowed} participation patterns. That is, 
    two sequences of gradients $\Gradients= (\gradient_0, \ldots, \gradient_{\nd-1}) \in \R^{\nd \times \mdim}$ and $\Gradients' = (\gradient_0', \ldots, \gradient_{\nd-1}') \in \R^{\nd \times \mdim}$ are adjacent under schema $\Pi$ if and only if there exists a participation pattern $\bfpi \in \Pi$ such that: 
    \begin{enumerate}[label=(\alph*)]
        \item $\gradient_t = \gradient_t'$ for all $t \notin \bfpi$;
        \item $\norm{\gradient_t - \gradient_t'}_2 \le 1$ for all $t \in \bfpi$ (following \cref{eq:gradnormbound})
    \end{enumerate}
    We denote this as $\Gradients\stackrel{\Pi}{\simeq} \Gradients'$.
\end{bdefinition}

In other words, the participation schema imposes restrictions on how many and which gradients in the stream $\gradient_0, \ldots, \gradient_{n-1}$ can be affected by changing one data point. This lets us map adjacency of datasets $D, D'$ into adjacency of the corresponding gradients $\Gradients, \Gradients' \in \R^{\nd \times \mdim}$.

\begin{figure*}
    \centering
    \adjustbox{max width = 0.7\linewidth}{
        \input{adjacency}
    }
    \caption{Two inputs $\Gradients, \Gradients' \in \R^{\nd \times \mdim}$ (with $\nd=5$) are adjacent if they can only differ in the rows contributed by one data point. In this illustration, the \textcolor{C0}{blue} data point in $\Gradients$ is replaced by the \textcolor{C2}{green} data point in $\Gradients'$.
    The participation schema $\Pi$ controls all allowed participation patterns. So, we have that $\Gradients\stackrel{\scriptscriptstyle\Pi}{\simeq} \Gradients'$ are adjacent under schema $\Pi$ only if the participation pattern $\bfpi = (1, 3)$ that encodes the rows where $\Gradients, \Gradients'$ differ is allowed as per $\Pi$ (i.e.,  only if $\bfpi = (1, 3) \in \Pi$).
    }
    \label{fig:adjacency}
\end{figure*}

We can now formally define the sensitivity of the matrix $\strategy$ (representing the linear map $\Gradients\mapsto \strategy\Gradients$) calibrated to a participation schema: 
\begin{bdefinition}[Participation-Calibrated Sensitivity]
\label{def:sens-participation}
    The $\ell_2$-sensitivity of the matrix $\strategy \in \R^{\nd \times \nd}$ restricted to participation schema $\Pi \subseteq 2^{[\nd]}$ and dimension $\mdim$ is defined as
    \begin{align}
        \sens(\strategy, \Pi, \mdim) = \max_{\Gradients, \Gradients' \in \R^{\nd \times \mdim}}\left\{
            \norm{\strategy\Gradients- \strategy\Gradients'}_{\op F} \, : \,
            \begin{gathered}
            \Gradients\stackrel{ \Pi}{\simeq} \Gradients' 
            \end{gathered}
        \right\} \,.
    \end{align}
When $\sens(\strategy, \Pi, \mdim)$ is independent of the dimension $\mdim$, we will omit $\mdim$ and simply write $\sens(\strategy, \Pi)$.
\end{bdefinition}

\paragraph{Simple Examples}
We now consider some basic participation schemas arising in the machine learning setting. 
See \Cref{fig:participation_patterns} for an illustration of the corresponding participation patterns.

\begin{enumerate}[label=(\alph*)]
\item \textbf{Streaming}: Each data point appears only once in training, so that $|\bfpi| = 1$ for each $\bfpi \in \Pi$. This conforms to the schema
\[
    \Pi^{\mathsf{single}} \coloneqq \left\{(0), (1), \ldots, (n-1) \right\} \,.
\]
Following \Cref{thm:gdp-of-correlated-noise-adaptive}, we have $\sens(\strategy, \Pi^{\mathsf{single}}) = \colnorm{\strategy}$. Crucially, this is  independent of the dimension $\mdim$. In this case, we simply use the notation $\sens(\strategy)$. 
For sufficiently large datasets and limited compute resources, this can be a very practical approach. However, when possible, we will generally achieve better privacy/utility tradeoffs by using larger batches, as we saw in \cref{rem:increasing-batch-size}. This generally requires multiple participations.

\item \textbf{Full batch}: Each data point participates in each step of training, so that $\Pi^{\mathsf{full}} = \{[n]\}$. This yields $\sens(\strategy, \Pi^{\mathsf{full}}, 1) = \norm{\strategy}_{\infty \to 2}$. As discussed in \cref{rem:increasing-batch-size}, this full-batch training approach yields optimal privacy-utility tradeoffs, but at a very large cost in compute, since the total number of gradient evaluation is $\bsz n = \bsz \datasize$). We will generally be able to do almost as well with much smaller batches with carefully chosen participation restrictions; see \cref{fig:scaling_batch_size}.

\end{enumerate}

Next, we consider two non-trivial participation schema that satisfy the desiderata of \Cref{sec:multi-epoch:sensitivity}.

\paragraph{Cyclic Participation}
The first new participation pattern we consider is tailored to centralized\footnote{
    We refer to training models in a data-center (possibly in a distributed fashion) as \emph{centralized}.  The key property of centralized training is that the full training dataset is always available, and hence can be accessed in any reasonable fashion; this is in contrast to the federated setting of \Cref{ex:fl}, where for example the data on any particular device might be unavailable for arbitrary periods of time, e.g. when the device is offline.} 
training pipelines. We loop over fixed batches of examples in some predefined fixed order.\footnote{The cyclic participation setting is practical in that it can easily be implemented in realistic centralized training pipelines. While we lose some generality as we are not allowed to shuffle the data in each epoch, we can introduce some randomness by shuffling the data \emph{once} at the start of training. It is not known if tight sensitivity bounds can be derived for the shuffle-every-epoch approach; we return to this open problem in \Cref{chap:open}.}
Therefore, each data point occurs in a fixed iteration in each epoch, so the only types of participation patterns allowed are as in \Cref{eq:cyclical-pattern}; see \Cref{fig:participation_patterns} for an illustration.
For a dataset of size $\datasize$ and batch size $\bsz$, each pass (epoch) comprises of $\minsep = \datasize / \bsz$ steps (assuming $\datasize$ divides $\bsz$), and so for $\maxpart$ epochs, we have the participation schema
\begin{align} \label{eq:cyclic-schema}
    \Pi^\cyclic_{\minsep, \maxpart}
    = \Big\{
    \big(l, l + \minsep, l+2\minsep, \ldots, l + (\maxpart-1)\minsep\big) \, :\, l \in [\minsep]
    \Big\} \,.
\end{align}

As we see in the upcoming \Cref{sec:multi-epoch:sensitivity-bounds}, it is computationally efficient to evaluate the multiple-participation sensitivity in this case under mild conditions.

\paragraph{Minimum Separation (Min-Sep) Participation}
While it is possible to make cyclic passes through the data in centralized model training, this is not possible in federated learning, where the client availability is determined by external factors (see \Cref{ex:fl}). However, it is often possible to prevent any client from participating twice in quick succession.
Each client remembers the last step $\tau$ in which it participated, and participates in training again only when the current step $t$ satisfies $t \ge \tau + \minsep$, for a minimum separation parameter $\minsep$. Importantly, no client is forced to check in with the server during a
narrow (and unknown to the device) time window for a specific step, as required by cyclic participation. This leads to a generalization of cyclic participation that inherits its favorable traits (as we will discuss further in the coming sections) but is also practical for federated learning.

\begin{figure*}
    \centering
    \includegraphics[width=\linewidth]{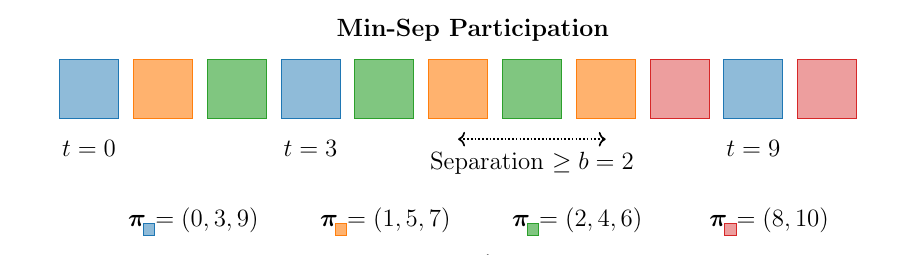}
    \caption{Illustration of the \minseppart pattern. Each color represents a data point, as in \Cref{fig:participation_patterns}. This setting corresponds to $\datasize=4$ data points participating in batches of size $\bsz=1$ over $\nd=11$. Each point participates at most $\maxpart=3$ times with a minimum separation of $\minsep=2$. Observe that any two subsequent participations of any given data point are at least $\minsep=2$ steps apart.
    }
    \label{fig:minsep_pattern}
\end{figure*}

\begin{bdefinition}[Participation with Minimum Separation]
\label{def:min-sep}
    A participation pattern $\bfpi$ satisfies \emph{$\minsep$-minimum-separation (Min-Sep)} with at most $\maxpart$ participations if
    if: (a) every pair of indices $t, \tau \in \bfpi$ satisfy $t = \tau$ or $|t-\tau| \ge \minsep$, i.e., subsequent participations of any data point are at least $\minsep$ steps apart, and (b) |$\bfpi| \le k$, i.e.,  each data point participates at most $\maxpart$ times.
    The corresponding participation schema is the collection of all participation patterns that satisfy the $\minsep$-Min-Sep condition:
    \begin{align}
    \label{eq:min-sep-participation}
        \Pi^\MinSep_{\minsep, \maxpart} = 
        \left\{
        \bfpi \subset [n] \, :\, 
        \begin{gathered}
        |\bfpi| \le k, \text{ and } \\
        \forall t, \tau \in \bfpi: \,
        t=\tau \text{ or } |t-\tau| \ge \minsep
        \end{gathered}
        \right\} \,.
    \end{align}
\end{bdefinition}
\noindent We given an illustration of the minimum separation condition in  \Cref{fig:minsep_pattern}. We usually refer to the participation pattern and schema from \Cref{def:min-sep} as ``\textit{$\minsep$-Min-Sep participation}'' with the understanding that at most $\maxpart$ participations are allowed throughout (for all participation schemas we consider).

We assume throughout that $\maxpart$ is feasible, that is, that $\maxpart$ participations are in fact possible while satisfying a minimum separation of $\minsep$. The maximum value of $\maxpart \in \N$ is then determined by the early-and-often pattern:
\begin{equation}\label{eq:piearlyoften}
\bfpi^* \coloneqq (0, \minsep, 2\minsep, \dots (\maxpart-1) b) \quad
\text{subject to $(\maxpart-1) b < \nd$}.
\end{equation}
We do not in general assume that $\minsep$ divides $\nd$, and so the last participation $(\maxpart-1)\minsep$ in $\bfpi^*$ could occur at any iteration between $t=\nd - \minsep$ and the last iteration $t=\nd - 1$, both included.

Observe that $\Pi^\cyclic_{\minsep, \maxpart} \subset \Pi^\MinSep_{\minsep, \maxpart}$. In other words, cyclic participation is a special case of \minseppart. As a consequence, if $\Gradients\simeq \Gradients$ as per the cyclic participation schema $\Pi^\cyclic_{\minsep, \maxpart}$, then $\Gradients\simeq \Gradients'$ as per the Min-Sep schema $\Pi^\MinSep_{\minsep, \maxpart}$ as well. Thus, we will focus on computing the sensitivity for the \minseppart as this directly yields bounds on the sensitivity for cyclic participation. 

We also remark that $\left| \Pi^\MinSep_{\minsep, \maxpart} \right| \gg \left|\Pi^\cyclic_{\minsep, \maxpart} \right| = \minsep$, where the number of Min-Sep participation patters can grow as $n^\maxpart$ (see \cref{eq:min-sep-bound}). This distinction will crucially matter in \Cref{sec:multi-epoch:sensitivity-algos}, when designing practical algorithms to compute the sensitivity efficiently.

\subsection{Expressions for Multiple-Participation Sensitivity}
\label{sec:multi-epoch:sensitivity-bounds}
We now give various expressions for the participation-calibrated sensitivity of \Cref{def:sens-participation}. These will directly lead to efficient algorithms to compute the sensitivity in \Cref{sec:multi-epoch:sensitivity-algos}.

\paragraph{Generic Sensitivity Bounds}
We start by explicitly relating the sensitivity to the participation schema. Recall that we use the notation $(\strategy\T\strategy)[t, \tau]$ to denote the $(t,\tau)$\textsuperscript{th} entry of the matrix $\strategy\T\strategy$.

\begin{blemma} \label[lemma]{lem:multi-epoch-sens:generic-bounds}
    For any participation schema $\Pi$ and any lower-triangular matrix $\strategy\in\R^{\nd \times \nd}$, we have:
    \begin{enumerate}[label=(\alph*)]
        \item \label{item:sens:2}
        For any dimension $m \ge 1$, we have the upper bound
        \[
            \sens\br{\strategy, \Pi, \mdim}^2
            \le \,\, \max_{\bfpi \in \Pi} \,\, \sum_{t, \tau \in \bfpi} \Big|(\strategy\T\strategy)[t, \tau] \Big| \,.
        \]
        \item \label{item:sens:3}
        Suppose that 
        \begin{equation}\label{eq:orthopi}
        \min_{t, \tau \in \bfpi}\ (\strategy\T\strategy)[t, \tau] \ge 0 \qquad \text{for all $\bfpi \in \Pi$}.
        \end{equation}
        (This is true when $\strategy\T\strategy$ is element-wise non-negative.)
        Then, we get a dimension-independent bound: 
         \begin{align} \label{eq:me-sens:non-neg}
            \sens(\strategy, \Pi)^2 = 
            \,\, \max_{\bfpi \in \Pi} \,\, \sum_{t, \tau \in \bfpi} (\strategy\T\strategy)[t, \tau] \,.
        \end{align}
        That is, Part \ref{item:sens:2} holds with equality. 
   \end{enumerate}
\end{blemma}

We discuss the interpretation of this result and its implications, before giving the proof.
Unlike the streaming setting, the sensitivity generally depends on the dimension $\mdim$. 
For general any general strategy matrix $\strategy$, we have an upper bound on the sensitivity (Part \ref{item:sens:2}). However, a fortunate exception arises when $\strategy\T\strategy$ is element-wise non-negative (Part \ref{item:sens:3}). In such cases, the sensitivity becomes independent of the dimension $\mdim$, and the upper bound holds with equality.

Fortunately, most practical scenarios involve $\strategy\T\strategy$ being non-negative. This is often because $\strategy$ itself is non-negative, a condition met by mechanisms like the max-loss-optimal Toeplitz mechanism (\Cref{sec:optimial-toeplitz-mech}), BLT (\Cref{sec:ch2-blt}), and tree aggregation (\Cref{sec:treeAgg}). Even when optimizing over $\strategy$ without enforcing non-negativity, we often find that 
$\strategy\T\strategy$ is either non-negative or has negligible negative entries. Therefore, imposing non-negativity on $\strategy\T\strategy$ results in solutions that are equally competitive (i.e., the non-negative solution attains nearly identical utility in terms of \txtmaxloss and learning performance at any given privacy level). 
We return to practical considerations around optimizing for the $\strategy$ matrix in \Cref{chap:practical}.

It is also instructional to rewrite \Cref{eq:me-sens:non-neg} as
\[
    \sens(\strategy, \Pi)^2 = 
    \max_{\bfpi \in \Pi} \sum_{t, \tau \in \bfpi}  \inp{\strategy[:, t]}{\strategy[:, \tau]} \,.
\]
This expression measures the similarity between the columns $\strategy[:, t]$ and $\strategy[:, \tau]$ corresponding to any two participations $t, \tau \in \bfpi$ of any example. For the independent noise setting of $\strategy = \bfI_{n \times n}$, this inner product is $1$ when $t=\tau$ and $0$ otherwise, leading to a (squared) sensitivity of $ \max_{\bfpi \in \Pi} |\bfpi| = k$, the maximum number of participations.

Finally, the expression of \Cref{lem:multi-epoch-sens:generic-bounds} can directly be used to compute the sensitivity with a brute-force enumeration of all possible $\bfpi \in \Pi$.
This is tractable when $|\Pi|$ is small, e.g., $O(n)$; we discuss this further in \Cref{sec:multi-epoch:sensitivity-algos}.

We now give the proof of \Cref{lem:multi-epoch-sens:generic-bounds}. 

\begin{proof}[Proof of \Cref{lem:multi-epoch-sens:generic-bounds}]

    Consider two gradient matrices $\Gradients\stackrel{\Pi}{\simeq} \Gradients' \in \R^{\nd \times \mdim}$ that are adjacent as per the schema $\Pi$ (\Cref{def:participation-schema}). By \Cref{def:sens-participation}, their difference $\bfU := \Gradients- \Gradients'$ must have non-zero rows indexed by some participation pattern $\bfpi \in \Pi$, which we denote (by slight abuse of notation) as $\bfpi(\bfU)$, so $\forall t \in [n],\quad \bfU\idx{t}{:} \neq \zeros \Rightarrow t \in \bfpi(\bfU)$. Below, we denote the rows of $\bfU$ as $\bfu_t := \bfU[t, :]$ (which satisfy $\norm{\bfu_t}_2 \le 1$ by assumption \cref{eq:gradnormbound}) and $\bfM := \bfC\T \bfC$. Then, using the fact that $\lfrob{\bfX}^2 = \tr(\bfX\T \bfX)$ for any matrix $\bfX$, we get,
    \begin{align*}
        \lfrob{\strategy \Gradients- \strategy\Gradients'}^2
        &= \lfrob{\sum_{t \in \bfpi(\bfU)} \strategy[:, t] \, \bfu_t\T}^2 \\
        &= \tr\left(\sum_{t, \tau \in \bfpi(\bfU)} \bfu_t \strategy[:,t]\T \bfC[:,\tau] \bfu_\tau\T \right)\\
        &\stackrel{(a)}{=} \sum_{t, \tau \in \bfpi(\bfU)} \bfM[t, \tau] \,\, \bfu_\tau\T \bfu_t
    \end{align*}
    {where $(a)$ used the linearity of the trace (to interchange the trace and summation) and the cyclic property (to group $\bfu_\tau$ and $\bfu_t$).
    Next, using (b) triangle inequality, (c) the Cauchy-Schwarz inequality, and (d) the norm bound \cref{eq:gradnormbound},}
    \begin{align*}
        \lfrob{\strategy \Gradients- \strategy\Gradients'}^2        &\stackrel{(b)}{\le} 
        \sum_{t, \tau \in \bfpi(\bfU)} \left| \bfM[t, \tau] \right| \cdot \left| \bfu_\tau\T \bfu_t \right| \\
        &\stackrel{(c)}{\le} 
        \sum_{t, \tau \in \bfpi(\bfU)} \left| \bfM[t, \tau] \right| \cdot \underbrace{\norm{\bfu_\tau}_2 \cdot \norm{\bfu_t}_2}_{\leq 1} \\
        &\stackrel{(d)}{\le}  \sum_{t, \tau \in \bfpi(\bfU)} \left| \bfM[t, \tau] \right|\,.
    \end{align*}
    Finally, by the definition of participation-calibrated sensitivity, we have
    \begin{align*}
        \sens(\strategy, \Pi, m)^2 &\le \max_{\Gradients\stackrel{\scriptscriptstyle \Pi}{\simeq} \Gradients'} \,\,  \sum_{t, \tau \in \bfpi(\Gradients- \Gradients')} \left| \bfM[t, \tau] \right|
        = \max_{\bfpi \in \Pi}  \sum_{t, \tau \in \bfpi} \left| \bfM[t, \tau] \right| \,,
    \end{align*}
    yielding the claimed upper bound of Part \ref{item:sens:2}.
    
    Next, we note that we can drop the absolute values in the expression of Part \ref{item:sens:2} when $\min_{t, \tau \in \bfpi} \bfM[t, \tau] \ge 0$ for each $\bfpi \in \Pi$. Thus, to show Part \ref{item:sens:3}, it suffices to exhibit a $\Gradients\simeq \Gradients'$ such that the inequalities $(b)$, $(c)$ and $(d)$ above are tight for a participation pattern $\bfpi^* \in \Pi$ that maximizes the upper bound of Part \ref{item:sens:2}. We fix $\Gradients\in \R^{n \times m}$ arbitrary and set $\Gradients' = \Gradients+ \bfU$ where the non-zero rows of $\bfU \in \R^{n \times m}$ are indexed by $\bfpi^*$ and each such row is equal a fixed (arbitrary) unit vector.
\end{proof}

\paragraph{Sensitivity for Cyclic and Min-Sep Participation}
Lemma~\ref{lem:multi-epoch-sens:generic-bounds} provides a clear connection between the sensitivity $\sens(\,\cdot\,, \Pi)$ and the participation schema $\Pi$. However, its maximization over $\Pi$ makes it less suitable for direct algorithmic implementation with $|\Pi|$ is large. 
To address this, we focus on the most practical participation schemes and mechanisms, as discussed in \Cref{chap:prefixsum} and \Cref{sec:multi-epoch:practical-patterns}. Specifically, we consider cyclic and \minseppart combined with the banded or Toeplitz/BLT mechanisms.
By simplifying the sensitivity expression for these settings, we can develop efficient algorithms to compute it.

\begin{blemma} \label[lemma]{lem:multi-epoch-sens:special-cases}
    Let $\Pi^\MinSep_{\minsep, \maxpart}$ be the participation schema as defined in \cref{eq:min-sep-participation} and $\Pi^\cyclic_{\minsep, \maxpart}$ be as defined in \cref{eq:cyclical-pattern}. For a matrix $\strategy \in \R^{\nd \times \nd}$ such that $\forall \bfpi \in \Pi,\ \min_{t, \tau \in \bfpi}\ (\strategy\T\strategy)[t, \tau] \ge 0$, as per the assumption in \cref{eq:orthopi} (for example, when $\strategy\T\strategy$ is element-wise non-negative), we have the following:
     \begin{enumerate}[label=(\alph*)]
        \item \label{part:sens:lb}
        \textbf{Lower bound}: 
        For any integers $\minsep, \maxpart > 0$, we have
        \[
            \sens\br{\strategy, \Pi^\MinSep_{\minsep, \maxpart}}^2
            \ge 
            \sens\br{\strategy, 
            \Pi^\cyclic_{\minsep, \maxpart}}^2
            \ge \sum_{i, j=0}^{\maxpart-1} (\strategy\T\strategy)[i\minsep, j\minsep] \,.
        \]
    \item \label{part:sens:banded}
    \textbf{Banded}:
    If $\strategy$ is lower-triangular and $\widetilde \minsep$-banded (\Cref{def:banded}) with $\widetilde \minsep \le \minsep$ bands (where $\minsep$ is the minimum separation parameter), we have for $\Pi \in \big\{\Pi^\cyclic_{\minsep,k}, \Pi^\MinSep_{\minsep,k} \big\}$ that
    \begin{equation} \label{eq:bandedsens}
        \sens\br{\strategy, \Pi}^2
        = \max_{\bfpi \in \Pi} \,\,\,
        \sum_{t \in \bfpi} 
        \big\|{\strategy[:, t]}\big\|_2^2
         \,\, \le \maxpart \, \colnorm{\strategy}^2
         \,.
    \end{equation}
    \item \label{part:sens:Toep}
    \textbf{Toeplitz and Monotonic}:
    If $\strategy$ is lower-triangular and Toeplitz as in \Cref{eq:Toeplitz-C} with non-negative and non-increasing entries $c_0 \ge c_1 \ge \cdots \ge c_{\nd-1} \ge 0$ in the first column, then the expression of \ref{part:sens:lb} holds with equality throughout. 
    \item \label{part:sens:bandedToep}
    \textbf{Banded and Toeplitz}: If $\strategy$ is banded (as in Part~\ref{part:sens:banded}) and Toeplitz, and $\minsep$ divides $\nd$, then the expression of \ref{part:sens:banded} also holds with equality. Regardless of whether $\minsep$ divides $\nd$, we always have for $\Pi \in \big\{\Pi^\cyclic_{\minsep,k}, \Pi^\MinSep_{\minsep,k} \big\}$ that
    \[
        (k-1) \, \colnorm{\strategy}^2 \le \sens(\strategy, \Pi)^2 \le k \, \colnorm{\strategy}^2 \,.
    \]
    \end{enumerate}
\end{blemma}

Part~\ref{part:sens:lb} gives lower bounds on the sensitivity, and these bounds hold with equality with Toeplitz and monotonic $\strategy$ matrices (Part~\ref{part:sens:Toep}). When the $\strategy$ matrix is banded, Part~\ref{part:sens:banded} gives us an expression that can be written as a linear program. As we shall see in \Cref{sec:multi-epoch:sensitivity-algos}, this expression can be evaluated efficiently with a dynamic programming algorithm.
Finally, in the special case of Part~\ref{part:sens:bandedToep}, we get a closed form expression for the sensitivity. We now give elementary proofs of these expressions.

\begin{proof}[Proof of Lemma~\ref{lem:multi-epoch-sens:special-cases}]
We prove each part in turn. 

\paragraph{Part \ref{part:sens:lb}:} First note that  $\Pi^\cyclic_{\minsep, \maxpart} \subset \Pi^\MinSep_{\minsep, \maxpart}$. This implies that the maximum over the former is a lower bound on the maximum over the latter in the expression of Lemma~\ref{lem:multi-epoch-sens:generic-bounds}\ref{item:sens:3}. That is, 
\[\sens\br{\strategy, \Pi^\MinSep_{\minsep, \maxpart}}^2 \ge  \sens\br{\strategy,   \Pi^\cyclic_{\minsep, \maxpart}}^2
\]
yielding the first inequality. 
The second inequality follows from plugging in the feasible early-and-often participation pattern $\bfpi = (0, \minsep, 2\minsep, \ldots, (\maxpart-1)\minsep)$ from \Cref{eq:piearlyoften} into the expression of Lemma~\ref{lem:multi-epoch-sens:generic-bounds} for a valid lower bound.

\paragraph{Part \ref{part:sens:banded}:}
Lemma~\ref{lem:multi-epoch-sens:generic-bounds} again gives us 
\begin{align} \label{eq:me-sens-proof-1}
    \sens(\strategy, \Pi) = \max_{\bfpi \in \Pi} \,\, \sum_{t, \tau \in \bfpi} 
    \inp*{\strategy[:, t]}{\strategy[:, \tau]} \,.
\end{align}
We make two observations:
\begin{enumerate}
    \item [(i)] for $\widetilde \minsep$-banded $\strategy$, we have $\inp{\strategy[:, t]}{\strategy[:, \tau]} = 0$ for all $| t-\tau| > \widetilde \minsep$;
    and
    \item [(ii)] for any $\minsep$-\minseppart pattern $\bfpi$, we have for every $t,\tau \in \bfpi$ that $t=\tau$ or $|t-\tau| > \minsep$.
\end{enumerate}
 
Putting these together with $\widetilde \minsep < \minsep$, we conclude that the terms corresponding to $t = \tau$ are the only possibly non-zero terms in \Cref{eq:me-sens-proof-1}. This gives the claimed equality.

We can get the claimed upper bound by using $|\bfpi| \le \maxpart$ (this applies for both $\Pi^\MinSep_{\minsep, \maxpart}$ and $\Pi^\cyclic_{\minsep, \maxpart}$):
\[
   \max_{\bfpi \in \Pi^\MinSep_{\minsep, \maxpart}} 
    \sum_{t \in \bfpi} (\strategy\T\strategy)[t, t]  \,\,
    \le \, \maxpart \, \cdot \, \max_t \, \left\{(\strategy\T\strategy)[t, t]\right\} = \maxpart \,\, \colnorm{\strategy}^2 \,
\]
by the definition of $\colnorm{\cdot}$.

\paragraph{Part \ref{part:sens:Toep}:}
The proof follows from arguing that the maximum over $\Pi^\MinSep_{\minsep, \maxpart}$ in \Cref{eq:me-sens-proof-1} is attained by the early-and-often participation pattern $\bfpi_\star = (0, \minsep, 2\minsep \ldots, (\maxpart-1)\minsep)$, i.e., the $\maxpart$ participations occur as early as possible.
Intuitively, this is because the Toeplitz coefficients $c_t$'s are monotonically non-increasing---we refer to \citet[Thm. 2]{kalinin2024banded} for a full proof.
Since $\bfpi_\star \in \Pi^\cyclic_{\minsep, \maxpart} \subset \Pi^\MinSep_{\minsep, \maxpart}$, it follows that $\bfpi_\star$ attains the maximum in \Cref{eq:me-sens-proof-1}  for $\Pi^\cyclic_{\minsep, \maxpart}$ as well.

\paragraph{Part \ref{part:sens:bandedToep}:}
Recall that $\strategy$ is a lower-triangular Toeplitz matrix. Then due to the Toeplitz structure of $\strategy$, we have that $\norm{\strategy[:, t]}_2$ (for $1 \le t < n-\widetilde b$) is equal to $\colnorm{\strategy}$ for the all but the last $\widetilde b$ columns of $\strategy$. On the other hand, the last $\widetilde \minsep$ columns of $\strategy$ can have a smaller norm.
Next, we plug this into the expression of Part \ref{part:sens:banded}.
If $\minsep$ divides $\nd$, then it is possible to have a $\bfpi$ such that it indexes only columns with norm equal to $\colnorm{\strategy}$, yielding a squared sensitivity of $k \, \colnorm{\strategy}^2$. In general, the worst-case $\bfpi$ can index at most one column whose norm is less than $\colnorm{\strategy}$, yielding the claimed lower bound.
\end{proof}

\begin{bremark}[Uniform Sensitivity Across Participation Patterns]
In the streaming setting, there exists an optimal strategy matrix $\strategy$ that is column normalized, meaning that the column norms of each column are equal (see \Cref{def:col-norm} and Lemma~\ref{lem:column-norm}). 
The natural translation to the multiple-participation setting is for the sensitivity for all $\bfpi \in \Pi$ to be the same (see the upcoming \Cref{conj:equal-participation-sensitivity}).  As in the streaming setting, Toeplitz $\strategy$ matrices in general do not satisfy this desiderata.  However, banded $\strategy$ matrices can be column normalized to satisfy this property for both cyclic and Min-Sep participation schemas. We return to this in \Cref{chap:practical}.
\end{bremark}

\subsection{Efficient Algorithms for Multiple-Participation Sensitivity}
\label{sec:multi-epoch:sensitivity-algos}

Correctly calibrating the noise magnitude $\stddev$ in the multiple-participation settings requires computing (or tightly bounding) the participation-calibrated sensitivity of \cref{def:sens-participation}. We now give various algorithms to compute this sensitivity and analyze their time complexity. We again focus on the practical settings of cyclic and \minseppart combined with the banded Toeplitz and BLT mechanisms. We also make the simplifying assumption that $\strategy\T \strategy$ is element-wise non-negative, which lets us use the dimension-independent results of Lemmata~\ref{lem:multi-epoch-sens:generic-bounds}\ref{item:sens:3} and~\ref{lem:multi-epoch-sens:special-cases}.

\begin{figure*}
    \centering
    \adjustbox{max width=0.5\linewidth}{
    \input{brute_force_sensitivity}
    }
\caption{Brute-force computation of the multi-participation sensitivity cyclic participation requires us to evaluate only $\Theta(nk)$ entries of the symmetric matrix $\bfM = \strategy\T\strategy$, highlighted here in blue. This figure illustrates the required entries for $n=9, \minsep=4, k=2$. The upper triangle does not have to be evaluated as the matrix $\bfM$ is symmetric, and has thus been omitted from this figure.}
\label{fig:sensitivity}
\end{figure*}
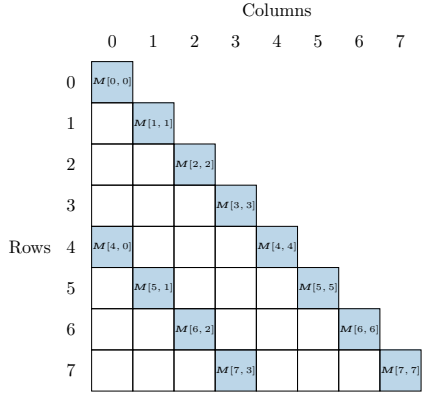

\paragraph{Brute Force Computation}
The simplest approach to compute the sensitivity is to evaluate \Cref{eq:me-sens:non-neg} of \Cref{lem:multi-epoch-sens:generic-bounds}\ref{item:sens:3} by enumerating all possible $\bfpi \in \Pi$ and returning the maximum.
Its time complexity is the sum of the time taken by each of its two steps: 
\begin{itemize}
    \item \textbf{Matrix Multiplication}:
    Computing $\strategy\T\strategy$ takes $O(\nd^3)$ time in general. This can be sped up to $O(\nd^2 \ln(\nd))$ for a Toeplitz matrix $\strategy$ using Fast Fourier Transform (FFT). Both of these approaches are only feasible for a small number of steps $\nd$. If $\strategy$ is $\widetilde\minsep$-banded (and possibly non-Toeplitz), then $\strategy\T\strategy$ can be computed in $O(\nd \widetilde \minsep^2)$ making it practical for moderate $\nd$ (as long as $\widetilde \minsep$ is small).
    
    Further improvements are possible for cyclic participation: we only need to the compute the entries $(\strategy\T\strategy)[t, \tau]$ such that $t, \tau \in \bfpi$ for some pattern $\bfpi \in \Pi$. Indeed, these are the only entries that appear in \Cref{lem:multi-epoch-sens:generic-bounds}\ref{item:sens:3}. For cyclic participation, these $\Theta(nk)$ entries can be computed $O(n^2 k)$ time (see \Cref{fig:sensitivity}) in general making it practical for moderate $n$ (as long as the number of epochs $k$ is small).
    \item \textbf{Exhaustive search}: 
    Searching for the best $\bfpi \in \Pi$ takes $|\Pi|$ time. This step is feasible for cyclic participation on small to moderate sized datasets, since $|\Pi^\cyclic_{\minsep, \maxpart}| = \minsep = \datasize / \bsz$, which is the number of steps in each epoch. (Recall that $\datasize$ is the dataset size and $\bsz$ is the batch size.) 
    Unfortunately, for \minseppart, we have
    \begin{align}
    \label{eq:min-sep-bound}
      |\Pi^\MinSep_{\minsep,\maxpart}| \le
        n (n-\minsep) (n-2\minsep) \cdots (n-(\maxpart-1)\minsep) \le O(n^\maxpart) \,.
    \end{align}
    \noindent This is not tractable for even moderate values of $n$ and $k$.
\end{itemize}
Overall, the brute force approach can be tractable only for moderately sized problems with cyclic participation.

\begin{figure*}
    \centering
    \includegraphics[width=\linewidth]{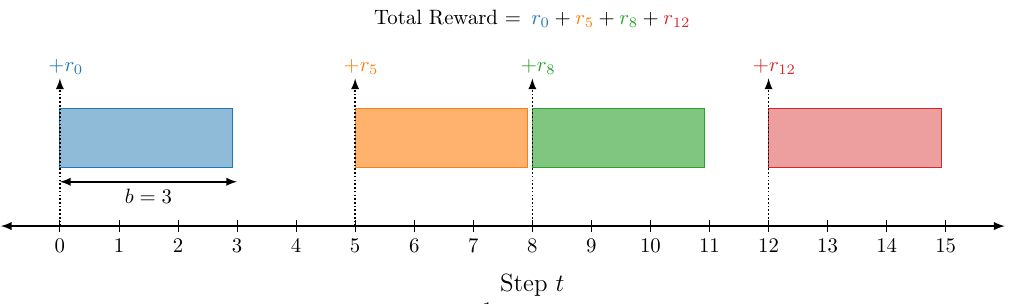}
    \caption{
    \textbf{Multiple-participation sensitivity of banded matrices via dynamic programs}:
    Given here is an illustration of Problem~\eqref{eq:dynamic-program-sens}. The goal here is to place $\maxpart=4$ non-overlapping blocks of width $\minsep=3$ over $\nd=16$ steps such that we maximize the total reward, where we receive reward $r_t$ from placing a block starting at step $t$.
    The (squared) multiple-participation sensitivity of a banded matrix $\strategy$ under \minseppart can be solved by such a dynamic program with $r_t = \norm{\strategy[:, t]}_2^2$, as per Lemma~\ref{lem:multi-epoch-sens:special-cases}\ref{part:sens:banded}.
    This problem can be solved in $O(\maxpart\nd)$ time and space with the dynamic program of \Cref{alg:banded-sens-dp}.
    }
    \label{fig:sens_dynamic_program}
\end{figure*}

\paragraph{Dynamic programming for banded matrices}
In the case of multiple-participation, we can compute the sensitivity exactly for a $\widetilde \minsep$-banded $\strategy$ matrix under $\minsep$-\minseppart $\Pi^\MinSep_{\minsep, \maxpart}$ (where $\widetilde \minsep \le \minsep$) defined by \Cref{eq:min-sep-participation} using a dynamic program. Consider the linear maximization problem
\begin{align} \label{eq:dynamic-program-sens}
    h(\bfr) := \max_{\bfpi \in \Pi^\MinSep_{\minsep,\maxpart}} \sum_{t \in \bfpi} r_t \,, 
\end{align}
for some non-negative  $\bfr = (r_0, r_1, \ldots, r_{\nd-1})$. In particular, the expression of Lemma~\ref{lem:multi-epoch-sens:special-cases}\ref{part:sens:banded} is a special case of this problem with $r_t = \norm{\strategy[:, t]}_2^2$.

The term $\sum_{t\in\bfpi} r_t$ is known as the {\em reward}. Maximizing the objective of \cref{eq:dynamic-program-sens} requires us to ``select'' $\maxpart$ indices $\bfpi \subset [n]$ to maximize the total reward $\sum_{t\in\bfpi} r_t$ such that any two distinct indices $t, \tau \in \bfpi$ are at least $\minsep$ apart (as required by the minimum separation constraint). This is a classical scheduling problem (see \Cref{fig:sens_dynamic_program}), which can be efficiently solved by the textbook dynamic program of \Cref{alg:banded-sens-dp} in $O(\nd \maxpart)$ time and space. %

\begin{algorithm}[t]
\caption{Dynamic program for maximizing \cref{eq:dynamic-program-sens}}
\label{alg:banded-sens-dp}
\begin{algorithmic}[1]
\Require{Non-negative numbers $r_0, \ldots, r_{\nd-1}$, separation $\minsep$, number of indices $\maxpart$}
\State Initialize buffer $\bfM \in \R^{\nd \times (\maxpart+1)}$ and set its first column $\bfM[:, 0] = \zeros$
\Statex \Comment{$\bfM[t, \ell]$ will store the maximum reward from selecting $l$ indices over steps $t, \ldots, \nd-1$. We treat $\bfM[t, \ell] = 0$ for $t, \ell$ out of bounds.}

\For{$l = 1, \ldots, \maxpart$}
\For{$t= n-1, \ldots, 0$}
\State $R_1 = r_t + \bfM[t + b, \ell-1]$ \Comment{Select $t$ as the $l$\textsuperscript{th} index}
\State $R_2 = \bfM[t+1, \ell]$  \Comment{Ignore $t$ for the $l$\textsuperscript{th} index}
\State $\bfM[t, \ell] = \max\big\{R_1, R_2\}$ 
\EndFor
\EndFor
\Ensure{$\bfM[0, \maxpart]$}
\end{algorithmic}
\end{algorithm}

\paragraph{Closed-form expression for Toeplitz \& monotonic matrices}
Consider a Toeplitz matrix $\strategy$ whose first column coefficients are monotonically non-increasing and non-negative as in the max-loss-optimal Toeplitz mechanism (\Cref{sec:optimial-toeplitz-mech}) or the BLT mechanism (\Cref{sec:ch2-blt}). Lemma~\ref{lem:multi-epoch-sens:special-cases}\ref{part:sens:Toep} lets us compute the sensitivity in these cases for both cyclic and \minseppart. First note that every entry of $\strategy^\top \strategy$ is non-negative. Therefore, $\sens$ is independent of $m$ and we have 
\begin{align} \label{eq:sens-blt-fn}
    \sens\br{\strategy, \Pi^\MinSep_{\minsep,\maxpart}}^2 = \sum_{i, j=0}^{\maxpart-1} (\strategy\T \strategy)[i\minsep, j\minsep] = \norm{\textstyle \sum_{j=0}^{\maxpart-1}\strategy[:, j\minsep]}_2^2 \,.
\end{align}
This can be evaluated in $O( \nd \maxpart)$ time, making it highly practical.

\subsection{Restarted Mechanisms*}
\label{sec:multi-epoch:restart}

We can extend a streaming correlated noise mechanism to multiple epochs by simply restarting the mechanism at the end of each epoch. In practice, this is inferior to directly designing a multi-epoch correlated noise mechanism using the machinery in this section (see \Cref{sec:ch3-biblio} for more discussion); however, restarted mechanism provides a simple way to prove bounds for multiple participation. 

In particular, in the case of restarted mechanism, 
the privacy loss then simply composes (sequentially) over the $k$ epochs (as per the upcoming \Cref{lem:gdp-composition}). This requires a noise multiplier (and hence, the \txtmaxloss) that is $\sqrt{k}$ times larger, when compared to the single-epoch setting. 
Our goal in this section is to demonstrate that the multiple-participation sensitivity can, by interpreting a restarted mechanism as a special case of cyclic participation, obtain the same bounds. 

We note that general multi-participation correlated noise algorithms, such as the banded Toeplitz mechanism, are not restarted mechanisms (whose strategy matrix has a sawtooth shape as in \Cref{fig:restarted_mechanism}).
Thus, using the multi-participation sensitivity directly lets us design substantially better mechanisms.\footnote{
    Restarted mechanisms can apply in some settings not covered by existing multi-participation theory, such as shuffled passes, i.e., where the dataset is shuffled at the start of the epoch. It is not known if multiple-participation sensitivity can be extended to this setting, as we discuss in \Cref{chap:open}.
}

Mathematically, we can describe the restarted mechanism by a \emph{tensorization} operation, as illustrated in \Cref{fig:restarted_mechanism}:

\begin{figure}
    \centering
    \includegraphics[width=\linewidth]{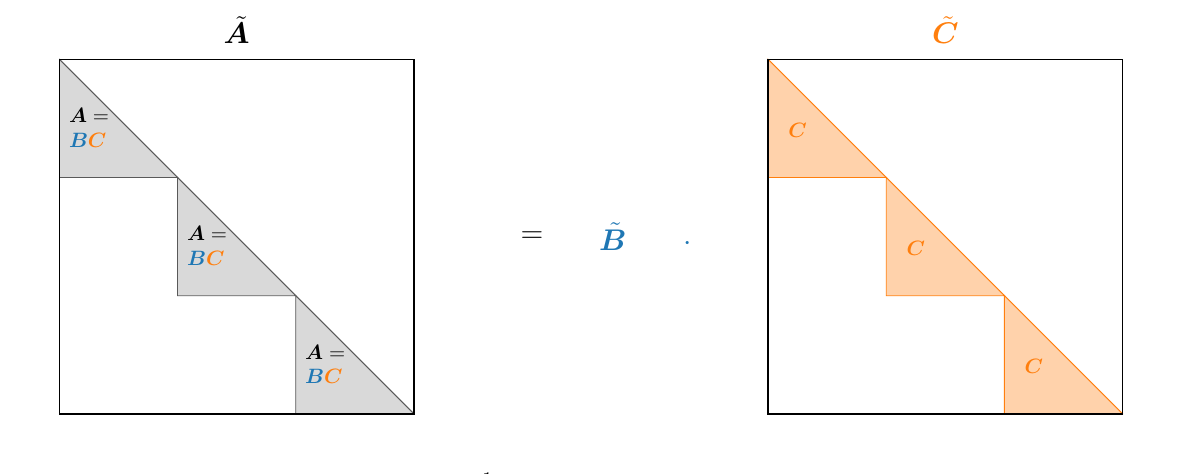}
    \caption{A restarted mechanism with $\maxpart$ restarts can, as per \Cref{def:restarted_mechanism}, be constructed from factorizations of the block diagonals components.}
    \label{fig:restarted_mechanism}
\end{figure}

\begin{bdefinition}[Restarted Mechanism]
\label{def:restarted_mechanism}
    Let $\prefix$ denote a $\maxpart n \times \maxpart n$ workload matrix.
    The $\maxpart$-restarted version of a correlated noise mechanism over $n$ steps based on the factorization $\prefix[:\nd, :\nd] = \bfB \bfC$ of the first $n$ rows and $n$ columns of $\prefix$ is given by $\restart_{\maxpart\nd \times \maxpart\nd} = \widetilde \bfB \widetilde \bfC$ with
    \[
        \widetilde \bfC = \bfC \otimes \bfI_{\maxpart \times \maxpart},
        \quad\text{and}\quad
        \widetilde \bfB = \prefix (\bfC^{-1} \otimes \bfI_{\maxpart \times \maxpart}) \,,
    \]
    where $\bfI_{\maxpart \times \maxpart}$ is the $\maxpart \times \maxpart$ identity matrix, and ``$\otimes$'' denotes the Kronecker product.
    If $\bfC$ is not invertible, $\widetilde \bfB$ can be defined using a suitable pseudoinverse $\bfC^\dagger$ instead.
\end{bdefinition}

We can expand out the definition to see why it corresponds operationally to restarting the mechanism: the noise $\widetilde \bfC^{-1} \Znoise$ injected by the restarted mechanism is correlated by the matrix
\[
    \widetilde \bfC^{-1} =  \bfC^{-1} \otimes \bfI_{k \times k} = \begin{pmatrix}
    \bfC^{-1} \\
    & \ddots \\
    & & \bfC^{-1}
    \end{pmatrix}\,,
\]
which is simply a block-diagonal matrix. Thus, the noise injected in any given epoch is independent of the noise injected in any other epoch---this is equivalent to freezing the results of the previous epoch and restarting the mechanism.

\paragraph{Restarts vs. Multiple-participation Sensitivity}
We compare the restarted mechanisms with directly computing the multiple-participation sensitivity under cyclic participation (the case of Min-Sep participation is similar). We show through a simple calculation that the multiple-participation framework we have developed so far can recover the special case of restarted mechanisms.

The key tool to give a differential privacy guarantee for a $k$-restarted mechanism is composition. We recall the standard composition property of GDP:

\begin{blemma}[Adaptive Composition of GDP]
\label[lemma]{lem:gdp-composition}
Let $\mech_1 : \calX^* \to \calY_1$ be $\mu_1$-GDP, and $\mech_2: \calX^* \times \calY_1 \to \calY_2$ be $\mu_2$-GDP mechanism with respect to its first argument for any fixed second argument. Then, the mechanism $\mech(D) = \big(Y_1, \mech_2(D, Y_1)\big)$ for $Y_1 = \mech_1(D)$ is $\sqrt{\mu_1^2 + \mu_2^2}$-GDP. 
\end{blemma}
While we stated Lemma~\ref{lem:gdp-composition} for two mechanisms, by induction it is easy to extend to $n$ adaptive mechanisms which have GDP parameters $\mu_0, \ldots, \mu_{n-1}$, in which case releasing all $n$ outputs satisfies $\mu$-GDP with $\mu = \sqrt{\sum_{t=0}^{n-1} \mu_t^2}$.

Now, returning to the restarted correlated noise mechanism, suppose we wish to obtain a $\mu$-GDP guarantee over $k$ epochs, where each epoch corresponds to $b = N/B$ steps with a batch size of $B$ over $N$ examples (see the setting in \Cref{sec:multi-epoch:training-setup} for notation).
Then, by the GDP composition property of Lemma~\ref{lem:gdp-composition}, each epoch must satisfy $\hat \mu$-GDP with $\hat \mu = \mu/\sqrt{k}$. 
Thus, the seed noise $\Znoise$ per-epoch mechanism $\mech(\Gradients) = \workload(\Gradients + \strategy^{-1} \Znoise)$ has to be distributed as $\Znoise \sim \calN_{b \times \mdim}(0, \colnorm{\strategy}^2 / \hat \mu^2)$, where $\mdim$ is the dimension of the model space. That is, its component-wise variance is $\sigma_{\text{restart}}^2 := k \colnorm{\strategy}^2 / \mu^2$.

On the other hand, we can view all $kb$ steps as one run of a multi-epoch mechanism with strategy matrix $\widetilde \strategy$. Indeed, this multiple-participation correlated noise mechanism corresponds to cyclic participation schema $\Pi^\cyclic_{b, k}$ of $n$ steps per epoch for $k$ epochs. 
Thus, we can also derive privacy guarantees using the multiple-participation sensitivity of all $kn$ steps. In this case, the component-wise variance of the seed noise $\Znoise$ for the correlated noise mechanism to satisfy $\mu$-GDP is $\sigma_{\text{ME}}^2 := \sens(\widetilde\strategy, \Pi^\cyclic_{b, k})^2 / \mu^2$. 

Now, it follows from the definition that both approaches turn out to be equivalent. Indeed, assuming that $\strategy\T\strategy$ is element-wise non-negative, we have
by Lemma~\ref{lem:multi-epoch-sens:special-cases}\ref{part:sens:banded} that
\begin{align*}
    \sens(\widetilde\strategy, \Pi^\cyclic_{b, k})^2 &= \max_{\bfpi \in \Pi^\cyclic_{b, k}} \,\,\,
        \sum_{t \in \bfpi} 
        \big\|{\widetilde\strategy[:, t]}\big\|_2^2 \\
    &\stackrel{(a)}{=} \max_{l \in [b]} \sum_{j=0}^{k-1} \big\|\widetilde\strategy[:, l + jb]\big\|^2 \\
    &\stackrel{(b)}{=} \max_{l \in [b]} k \big\| \strategy[:, l]\big\|_2^2 = k \cdot \colnorm{\strategy}^2 \,,
\end{align*}
where $(a)$ follows from the definition of the schema $\Pi^\cyclic_{n, k}$ (\Cref{eq:cyclic-schema}), while $(b)$ follows from $\widetilde \strategy = \strategy \otimes  \bfI_{k \times k}$ (see also \Cref{fig:restarted_mechanism}).
Thus, we have that $\sigma^2_{\text{ME}}= \sigma^2_\text{restart}$, meaning that both approaches are equivalent.

\input{3a_amplification}

%% file: adjacency.tex
\begin{tikzpicture}

\draw[white] (-0.2, 2.75) rectangle (10, 5.8) ; 

\tikzstyle{boxA}=[draw=black!65,fill=C0!50, pattern=north west lines, pattern color=C0!5, preaction={fill=C0!50}]
\tikzstyle{boxB}=[draw=black!65,fill=C1!40]
\tikzstyle{boxC}=[draw=black!65,fill=C2!60, pattern=dots, pattern color=white, preaction={fill=C2!60}]
\tikzstyle{boxD}=[draw=black!65,fill=C3!30]

\node at (1, 6) {\large $\Gradients \in \mathbb{R}^{5 \times \mdim}$};

\node at (5, 6) {\large $\Gradients' \in \mathbb{R}^{5 \times \mdim}$};

\filldraw[boxB] (0, 5) rectangle ++(2, 0.5) node[pos=0.5] {\large $\gradient_0$} ;
\filldraw[boxA] (0, 4.5) rectangle ++(2, 0.5) node[pos=0.5] {\large $\gradient_1$} ;
\filldraw[boxD] (0, 4) rectangle ++(2, 0.5) node[pos=0.5] {\large $\gradient_2$} ;
\filldraw[boxA] (0, 3.5) rectangle ++(2, 0.5) node[pos=0.5] {\large $\gradient_3$} ;
\filldraw[boxB] (0, 3) rectangle ++(2, 0.5) node[pos=0.5] {\large $\gradient_4$} ;

\node at (3, 4.4) {\Large $\stackrel{\Pi}{\simeq}$} ;

\filldraw[boxB] (4, 5) rectangle ++(2, 0.5) node[pos=0.5] {\large $\gradient_0' = \gradient_0$} ;
\filldraw[boxC] (4, 4.5) rectangle ++(2, 0.5) node[pos=0.5] {\large $\gradient_1'$} ;
\filldraw[boxD] (4, 4) rectangle ++(2, 0.5) node[pos=0.5] {\large $\gradient_2' = \gradient_2$} ;
\filldraw[boxC] (4, 3.5) rectangle ++(2, 0.5) node[pos=0.5] {\large $\gradient_3'$} ;
\filldraw[boxB] (4, 3) rectangle ++(2, 0.5) node[pos=0.5] {\large $\gradient_4' = \gradient_4$} ;

\node at (8.5, 4.25) {\Large if \,\, $\bm{\pi} = (1, 3) \in \Pi$};

\end{tikzpicture}

%% file: brute_force_sensitivity.tex
\begin{tikzpicture}[scale=0.8]

    \def\n{8}  
    \def\b{4}  
    \def\k{2}  

    \foreach \d in {0,...,1} { 
        \foreach \i in {0,...,7} {
            \pgfmathtruncatemacro{\row}{\i + \d * \b}
            \ifnum\row>7
            \else
                \fill[C0!30] (\i+1,-\row-1) rectangle ++(1,-1);
                \node at (\i+1.5,-\row-1.5) {\tiny $\bm{M}\idx{\row}{\i}$};
            \fi
        }
    }
    
    \foreach \i in {0,...,7} { 
        \foreach \j in {0,...,7} {
            \ifnum\i<\j
            \else
                \draw (\j+1,-\i-1) rectangle ++(1,-1);
            \fi
        }
    }

    \foreach \i in {0,...,7} {
        \node at (0.5,-\i-1.5) {\small $\i$};
        \node at (\i+1.5,-0.5) {\small $\i$};
    }

    \node at (-0.5, -5.5) {\small Rows};
    \node at (5.5, 0.25) {\small Columns};

\end{tikzpicture}

%% file: 3a_amplification.tex
\section{Privacy Amplification by Sampling: A Deeper Dive}
\label{sec:learning:amplification}

Recall also from \Cref{sec:ml:motivation} that DP-SGD with correlated noise 
can give much better utility than DP-SGD with independent noise in scenarios such as federated learning where amplification by sampling is not applicable. In fact, correlated noise can also be competitive with \emph{independent noise with amplification} in some regimes as we elaborate on below.

The benefit from privacy amplification gets better as the noise multiplier $\nmult$ gets larger, or equivalently, when the privacy budget  $\eps$ is small\footnote{
    We consider amplification in the framework of $(\eps, \delta)$-DP, as it admits a tighter description of the amplified privacy guarantee than Gaussian DP.
}; see \Cref{fig:amplification}.
In particular, if the noise multiplier $\nmult$ is sufficiently large, the benefits of privacy amplification for DP-SGD are larger than the benefits of using correlated noise (as opposed to independent noise). That is, amplified independent noise mechanisms can outperform un-amplified correlated noise mechanisms in the high-privacy regime. 

However, the data processing and noise addition are independent components of the algorithm.  This raises a key question: 

\begin{quote}
    Can correlated noise mechanisms benefit from privacy amplification by sampling (when it is feasible)?
\end{quote}

This turns out to be technically challenging. The main reason is that the analysis of privacy amplification often relies heavily on independence of the randomness in each iteration of DP-SGD, both for the noise generation and in the sampling process. This independence lets us tightly combine the per-round privacy guarantees of DP-SGD (each step is just a Gaussian mechanism) using standard composition results such as  Lemma~\ref{lem:gdp-composition} to get a privacy guarantee for the entire training process. Without independence, the guarantees given by composition are not valid, which prevents a straightforward amplification analysis of correlated noise under sampling. It remains an open question to give efficiently computable and near-tight privacy guarantees for general correlated noise mechanisms under amplification by sampling.\footnote{
    Here, we only tackle the problem of giving an amplified privacy guarantee for a given strategy matrix $\strategy$. Ultimately, we wish to co-design the strategy matrix and the sampling scheme to attain the best utility at any privacy level; we get back to this in \Cref{chap:open}.
}

We overview a simple construction that allows amplification of privacy guarantees with $b$-banded $\strategy$ matrices, i.e., $\strategy[t, \tau] = 0$ for all $t \geq \tau + b$ and $t < \tau$ (see \Cref{def:banded} of \Cref{sec:ch2-banded-Toeplitz}).
This construction uses a modification of Poisson sampling that creates a Poisson subsample from a different block (where we cycle through the blocks in fixed order): 

\begin{bdefinition}[Block-Cyclic Poisson Sampling] \label{def:cyclic_poisson}
    Given a number $b$ of blocks and a expected batch size $B$, we first partition the dataset $D$ of size $|D|=N$ into $b$ arbitrary subsets $D_0, D_1, D_2, \ldots D_{b-1}$ of equal size $N/b$. In iteration $t$, we (Poisson) sample from the dataset $D_{t \pmod b}$, with sampling probability $p' = Bb/N$, giving a batch of expected size $B$.
\end{bdefinition}

We can now give an amplified privacy guarantee using this sampling process for banded $\strategy$ matrices:

\begin{btheorem}[Informal]
\label{thm:banded-amplif}
Consider the block-cyclic Poisson sampling strategy with $b$ blocks and an expected batch size $B$ and suppose the gradients satisfy the norm bound of \Cref{eq:gradnormbound}.
In this setting, the correlated noise mechanism $\mech(\Gradients) = \strategy \Gradients + \Znoise$ with $\Znoise \sim \normalnm{0}{\stddev^2}$ for some variance $\stddev^2 > 0$ satisfies any privacy guarantee satisfied by independent noise DP-SGD (i.e. \Cref{alg:dpsgd-batch} with $\strategy$ as the identity matrix) using $n' = n/b$ steps, sampling probability $p' = Bb/N$ (i.e., so that batches are formed with an expected size of $B$ from a dataset of size $N/b$), and noise standard deviation $\stddev' = \stddev / \colnorm{\strategy}$. 
\end{btheorem}
\begin{proof}[Proof Sketch]
Without loss of generality we can assume that adjacent datasets $D$ and $D'$ differ in the example $x_0$ which is in $D_0$. Since the gradient $\bfg_0$ is drawn from dataset $D_0$, it could differ between $D$ and $D'$. However, 
since the matrix $\strategy$ is $b$-banded, changing $\bfg_0$ only affects rows $0, 1, \ldots, b-1$ of the output $\strategy \Gradients + \Znoise$, and $\bfg_{b}, \bfg_{2b}, \ldots$ do not affect these rows. Similarly, the gradient $\bfg_{b}$ is also computed on a subsample from $D_0$ and could differ when we move to an adjacent dataset. This only affects rows $b, b+1, \ldots, 2b-1$ of $\strategy \Gradients$, and $\bfg_{0}, \bfg_{2b}, \bfg_{3b}, \ldots$ do not affect these rows. 

In other words, each batch drawn from $D_0$ affects a disjoint part of the output: the $j$\textsuperscript{th} batch can only affect rows $R_j := \{jb, \ldots, (j+1)b - 1\}$ for a total of $n' = n/b$ such groups.
Then the mechanism $\mech_j(\Gradients) := (\strategy \Gradients + \Znoise)[R_j]$ generating corresponding rows of the $\mech(\Gradients)$ can be viewed as one run of the Gaussian mechanism with subsampling. In particular, the $\ell_2$-sensitivity (assuming $\Gradients'$ is generated as above) is
\[
    \lfrob{\big(\strategy(\Gradients - \Gradients')\big)[R_j, :]} = \lfrob{\strategy[R_j, 0] (\bfg_0 - \bfg_0')\T} \le \colnorm{\strategy} \,.
\]
This would satisfy the same privacy guarantee as a mechanism with unit $\ell_2$-sensitivity, but with noise multiplier $\nmult = \stddev / \colnorm{\strategy}$. 
Moreover, $\bfg_0$ is generated by sampling an expected $\bsz$ elements from $|D_0| = N/\minsep$ choices; so the sampling probability is $p' = B / (N/b) = Bb/N$.
Thus, $\mech_j$ satisfies the same privacy guarantee as DP-SGD (whose $\ell_2$-sensitivity is 1) with a noise multiplier $\nmult$ and sampling probability $p'$. Finally, the complete mechanism $\mech$ is an adaptive composition of $\mech_0, \ldots, \mech_{n'-1}$, meaning that the original correlated noise mechanism $\mech$ satisfies the same privacy guarantee as $n'$ steps of DP-SGD.
\end{proof}

\section{Learning Guarantees for Correlated Noise Mechanisms*}
\label{sec:ml:learning-guarantees}

In this section, we will focus on the learning theoretic guarantees enjoyed by correlated noise mechanisms, and how they compare to independent noise mechanisms (i.e., DP-SGD or differentially private gradient descent). Note that the full batch noisy gradient descent (DP-GD) achieves optimal accuracy-privacy trade-off for empirical risk minimization, something that can also be achieved using a proper instantiation of DP-SGD (for example, with appropriate choice of minibatching and subsampling) for both stochastic convex optimization and empirical risk minimization. The benefit of using DP-SGD over DP-GD is that the former has significantly less overhead in terms of computational time. 

The main message in this section is that, while we can design optimal algorithms for  stochastic convex optimization (SCO) using multiple pass over the data,  it is not known that it is possible in the single-epoch setting, i.e., with a single pass over the data. On the other hand,  correlated noise mechanisms are known to be the best in terms of the  trade-offs between privacy, utility, and computation time in the single-pass setting. 
In the following, we make this claim more precise, surveying generic bounds for stochastic convex problems, and more fine-grained bounds for specific 
problems like linear regression.

\subsection{Generic Learning Bounds}
\label{sec:generic-learning-bounds}

Stochastic convex optimization with differential privacy has been studied in a variety of settings, we focus on the case of Lipschitz functions and convexity to illustrate the effect of introducing the noise correlations. Specifically, we assume $\Theta\subseteq\mathbb{R}^\mdim$ is a convex set (where $\mdim$ is the dimensionality of the model space) and:
\begin{enumerate}[label=(\alph*)]
    \item \textbf{Convex}: The loss function $\ell(\cdot\,, \bfx)$ is convex in its first parameter for all $\bfx$.
    
    \item \textbf{Lipschitz}: The loss function $\ell(\cdot\,, \bfx)$ is $1$-Lipschitz (in the Euclidean norm). That is, for all pairs $\bftheta_1, \bftheta_2 \in \Theta$ and all $\bfx \in \calX$, we have 
    \[
    | \ell(\bftheta_1,\bfx) - \ell(\bftheta_2,\bfx) | \leq \, \norm{\bftheta_1 - \bftheta_2}_2.
    \]
\end{enumerate}

The first mechanism to exploit the correlated noise mechanism was the \textit{DP-follow-the-regularized-leader} (DP-FTRL), which we collectively refer as correlated noise DP-SGD. We refer the readers to \Cref{sec:dpftrl-name-background} for details. The analysis of correlated noise DP-SGD goes via the standard regret analysis of online learning, while accounting for the additional noise added due to privacy. One can show the following results for the algorithm $\calA_{\texttt{cor-noise}}$ defined in~\Cref{sec:dpftrl-name-background}:

\begin{btheorem}
\label{thm:ch3-dp-sco}
Given $\nd$ data samples, $\calA_{\texttt{DP-FTRL}}$ is $(\eps,\delta)$-DP and outputs a $\theta \in \real^\mdim$ such that  
\begin{itemize}
    \item In the general setting, we have   $$R(\calA_{\texttt{cor-noise}})=\widetilde O_\delta\left({\frac {\mdim^{1/4}}{\sqrt{\epsilon n}}}\right),$$ 
    where $\widetilde O_\delta(\cdot)$ hides polylog factors in $n$ and $\delta$.
    \item In the \emph{realizable} setting, i.e., when $\min\limits_{\theta\in\Theta} \mathbb{E}_{\bfx\sim\distribution}\left[\ell(\model;\bfx)\right]=0$, the SCO guarantee can be improved to $$R(\calA_{\texttt{cor-noise}})= \widetilde \Theta_\delta\left(\frac{1}{\sqrt n}+\frac{\sqrt \mdim }{\epsilon n}\right).$$
\end{itemize}
\end{btheorem}

\subsection{Detailed Learning Bounds on Specific Problems} 
\label{sec:detailed-learning-bounds}

We now review some bounds demonstrating that correlated noise can help \cref{alg:dpsgd-corr} attain a better objective value than independent noise in the streaming setting with an infinite clipping norm of $\zeta = \infty$.
That is, the correlated noise mechanism receives as input the unclipped gradient $\gradient_t = \nabla \ell\big(\bftheta_t, \bfx_t \big)$. Hence the norm, $\norm{\gradient_t}_2$, and, thus the sensitivity, can potentially be unbounded, and so this algorithm does not satisfy differential privacy.\footnote{
    Clipping can also impact the optimization dynamics of the learning algorithm; the bibliographic notes of \Cref{sec:ch3-biblio} provides some pointers.
}    
Still, studying the suboptimality bounds on the resulting algorithms with noise calibrated to a desired privacy level under the \emph{false} assumption that $\norm{\gradient_2}_2 \le 1$ sheds light on the precise effect of \emph{introducing noise correlations} into the learning process (while avoiding technicalities due to clipping), and how it subtly differs from the prefix sum estimation.

\subsubsection{Linear Regression}
 \label{sec:ch3-linear-regression}
One of the simplest learning problems is linear regression. We aim to predict a target $y \in \R$ from input $\bfx \in \R^\mdim$ by solving
\begin{align} \label{eq:least-squares-obj}
    \min_{\bftheta \in \R^\mdim} \,\, \tfrac{1}{2} \, \mathbb{E}_{(\bfx, y) \sim \Pdata} (\bfx\T\bftheta - y)^2 \,.
\end{align}
This is an instantiation of the learning problem of \Cref{sec:weightedPrefixSum} where $\ell\big(\bftheta, (\bfx, y)\big) = (1/2) (\bfx\T \bftheta - y)^2$ is the mean squared loss. This multiplicative constant $1/2$ in the squared loss ensures that the input covariance matrix $\bfH = \mathbb{E}[\bfx\bfx\T] \in \real^{\mdim \times \mdim}$ is also the Hessian matrix of the objective of \cref{eq:least-squares-obj}.  

We make several simplifications to illustrate the effect of the noise correlations. First, we assume that input is Gaussian
vector with full-rank covariance matrix $\bfH$, i.e., $\bfx \sim \calN(\zeros, \bfH)$;\footnote{The notation $\calN(\zeros,\bfH)$ denote the multivariate Gaussian with $\zeros$-mean and covariance $\bfH \in \R^{m \times m}$. Its probability density function is $(2\pi)^{-\mdim/2}\det (\bfH)^{-1/2} \, \exp \left( -\frac{1}{2} \bfx^\top  \bfH^{-1}\bfx \right)$} second, we assume a realizable model, i.e., the average loss at the global minimizer $\bftheta^\star$ is zero. This can only happen if $\ell\big(\bftheta^\star, (\bfx, y)\big) = (1/2)(\bfx\T \bftheta^\star - y)^2 = 0$ almost surely for $(\bfx, y) \sim \Pdata$. In other words, with probability one, we have that $(\bfx, y) \sim \Pdata$ satisfies $y = \bfx\T \bftheta^\star$.\footnote{
    The realizability assumption is easily lifted by assuming instead that $y = \bfx\T\bftheta^\star + \xi$ for i.i.d. Gaussian noise $\xi \sim \normal(0, s^2)$. This yields an extra additive term that is independent of the DP noise in each of the bounds we give below.
} 
Finally, as we discussed above, we omit gradient clipping, so the correlated noise mechanisms receives as input the unclipped gradient $\gradient_t = \nabla \ell\big(\bftheta_t, (\bfx_t, y_t)\big)$. We calibrate the noise to a desired privacy level under the \emph{false} assumption that $\norm{\gradient_2}_2 \le 1$.

\begin{figure*}
    \centering
    \begin{subfigure}[b]{0.5\textwidth}
        \centering
        \includegraphics[width=\linewidth]{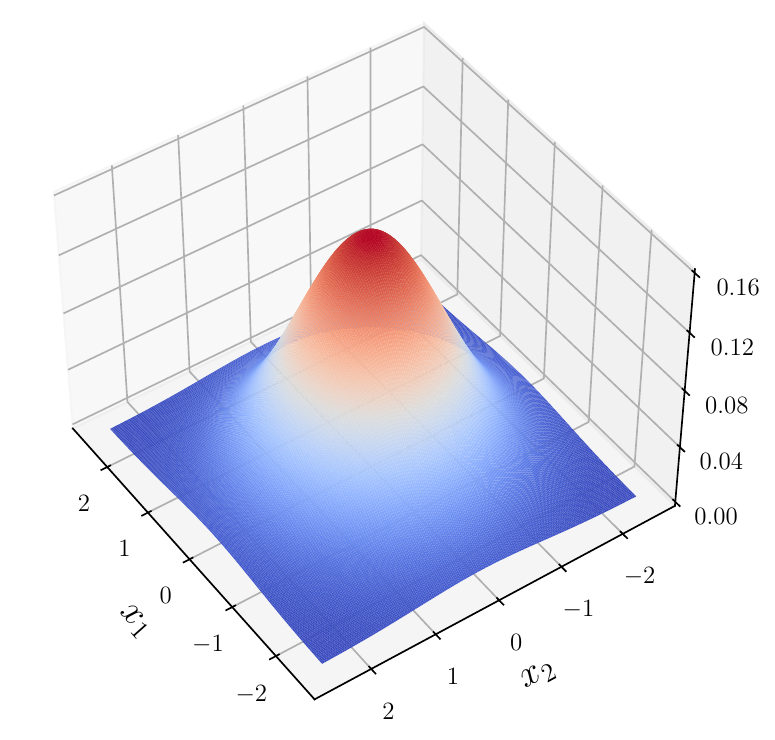}
        
        \includegraphics[width=0.75\linewidth]{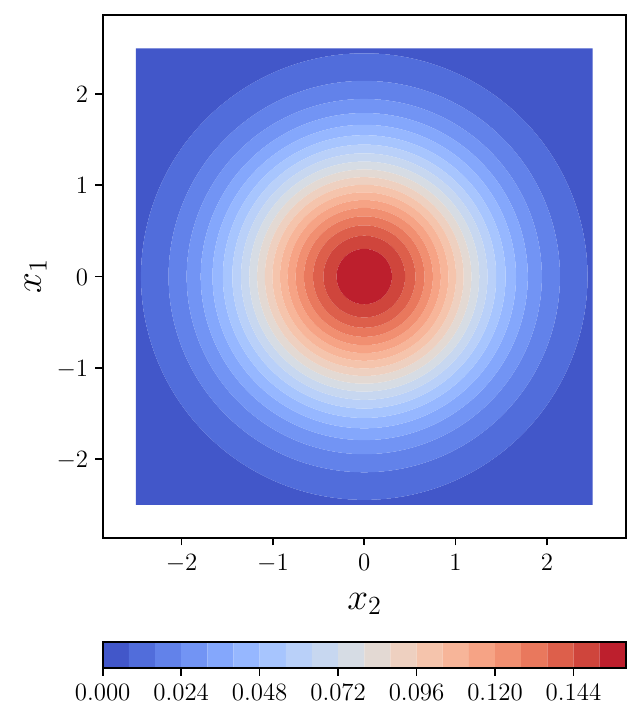}
        \caption{Gaussian with high effective dimension.}
    \end{subfigure}%
    ~ 
    \begin{subfigure}[b]{0.5\textwidth}
        \centering
        \includegraphics[width=\linewidth]{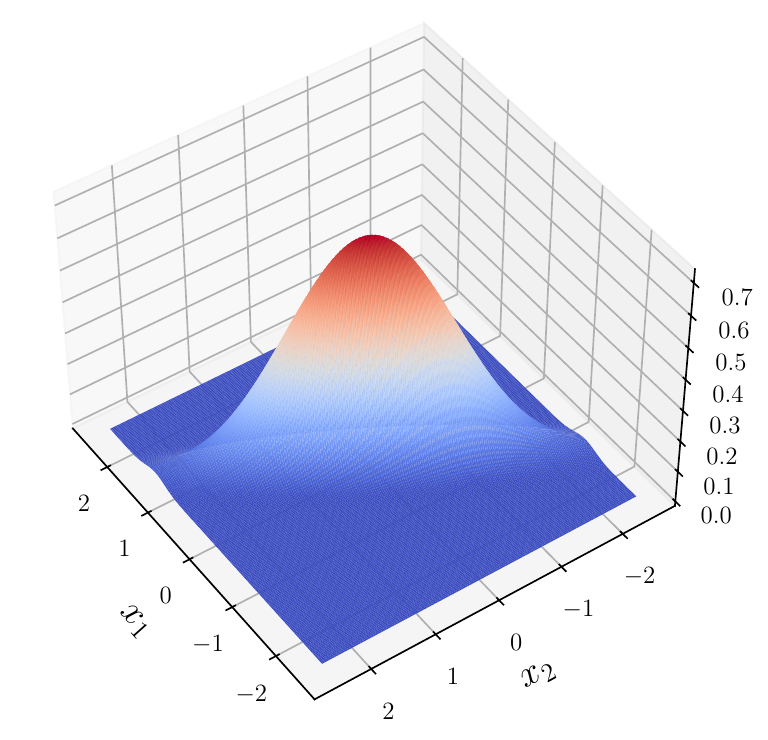}
        
        \includegraphics[width=0.75\linewidth]{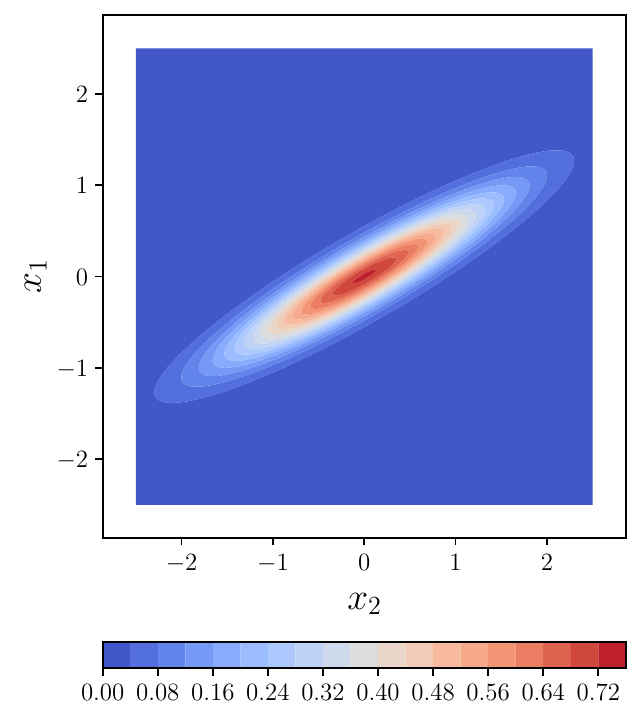}
        \caption{Gaussian with low effective dimension.}
    \end{subfigure}
    \caption{The densities of two Gaussian distributions in $\R^2$ and their effective dimensions $\mdim_{\mathsf{eff}}$. 
    The left plot depicts an isotropic Gaussian, meaning that its covariance matrix has equal eigenvalues. Its effective dimension is then $\mdim_{\mathsf{eff}} = 2 = \mdim$. The right plot shows a nearly low rank Gaussian, whose covariance matrix has eigenvalues $(1.2, 0.04)$. Its effective dimension $\mdim_{\mathsf{eff}} \approx 1.03$ is strictly smaller than the ambient dimension $\mdim=2$.
    }
    \label{fig:eff_dim}
\end{figure*}

The bounds depend on the ratio
$\mdim_\mathsf{eff} = {\tr(\bfH)} / {\norm{\bfH}_{2\to 2}}$ of the trace of the $\mdim \times \mdim$ input covariance matrix $\bfH$ to its spectral norm; 
this is also known as its \emph{effective dimension}. We have the alternative expression in terms of the eigenvalues $\lambda_1, \ldots, \lambda_m > 0$ of $\bfH$:
\[
    \mdim_\mathsf{eff} = \frac{\sum_{i=1}^m \lambda_i}{\max_{i=1, \ldots, m} \lambda_i} \,.
\]
As illustrated in \Cref{fig:eff_dim}, we have $1 \le \mdim_\mathsf{eff} \le \mdim$, with a smaller $\mdim_\mathsf{eff}$ for approximately low rank problems, and equal to the ambient dimension $\mdim$ when all the eigenvalues of $\bfH$ are equal.
It is desirable for the error of numerical algorithms to scale with effective dimension $\mdim_\mathsf{eff}$ of the problem, rather than the ambient dimension $\mdim$, which can be significantly larger. In particular, it is quite common for over-parameterized problems on real-world data to exhibit an effective dimension that is much smaller than the ambient dimension.

The advantage of correlated noise can be seen as follows. Suppose, without loss of generality that $\norm{\bfH}_{2 \to 2} = 1$. We have the following bounds in the asymptotic $n \to \infty$ regime in terms of the learning rate $0 < \eta < 1$ (assumed small enough),
the GDP parameter $\mu$ (so the component-wise variance of the injected noise scales as $1/\mu^2$), and the effective dimension $\mdim_\mathsf{eff}$:
\begin{itemize}
    \item Using independent noise, we have the matching upper and lower bounds on the excess population risk \eqref{eq:excess-risk} (up to absolute constants):
    \[
        \lim_{n \to \infty} R(\mech_{\mathsf{indep}, n})
        = \Theta\left( \frac{\eta\mdim}{\mu^2} \right) \,,
    \]
    where $\mech_{\mathsf{indep}, n}$ is the output of $n$ steps of DP-SGD where the gradients are perturbed with independent noise.
    \item Given a parameter $\nu > 0$, consider a variant of the max-loss-optimal Toeplitz mechanism (\Cref{sec:optimial-toeplitz-mech}) where the first column $c_0, c_1, \ldots$ of the strategy matrix $\strategy$ is given by
    \begin{align}
    \label{eq:opt-toeplitz-c-regr}
    c_t
    = (-1)^t \binom{-1/2}{t}  (1-\nu)^t \,.
    \end{align}
    This coincides with the Toeplitz coefficients in \cref{eq:opt-toeplitz-c} up to a factor of $(1-\nu)^t$. (For intuition on why we have this extra factor, see the paragraph on ``Interpretation'' below.)  
    The mechanism $\mech_{\mathsf{Toep}, n}$ that uses the correlated noise mechanism over $n$ steps with this strategy matrix $\strategy\in\R^{n \times n}$ satisfies the excess risk bound
    \[
        \lim_{n \to \infty} R(\mech_{\mathsf{Toep}, n}) 
        \le 
        O\left( \frac{\eta^2 \mdim_{\mathsf{eff}}}{\mu^2} \, \ln^2\frac{1}{\nu} \right) 
    \]
    for all $0 < \nu \le \eta \lambda_{\min}(\bfH)$, where $\lambda_{\min}(\bfH)$ denotes the smallest eigenvalue of the covariance $\bfH$.
    \item For any correlated noise mechanism $\mech_n$ over $n$ steps with a Toeplitz strategy matrix $\strategy \in \R^{n \times n}$, we have the lower bound on the excess risk:
    \[
        \inf_\mech \,\, \left\{ \lim_{n \to \infty} R(\mech_n) \right\}
        \ge \Omega\left( \frac{\eta^2\mdim_\mathsf{eff}}{\mu^2} \right) \,.
    \]
\end{itemize}

Interestingly, using independent noise in this case is strictly sub-optimal, while the use of  correlated noise is (almost) optimal: it matches the lower bound up to log factors. The bound for correlated noise is better than that of using independent noise by a problem-dependent factor that can be as large as $\mdim / \ln \mdim$.  
The gap is significant when the covariance $\bfH$ is approximately low rank. This is observed empirically in, e.g., overparameterized models where the features are highly correlated.

\paragraph{Interpretation}
The improvement from the ambient dimension $\mdim$ to the effective dimension $\mdim_{\mathsf{eff}}$ sheds some light on the role played by the correlated noise in learning problems. The gradient $\gradient_t$ aims to counteract the effect of the noise added in previous iterations and move the trajectory back towards the minimizer $\bftheta^\star$.

Unfortunately, the gradients cannot effectively cancel the noise along eigenvectors of the covariance $\bfH$ with small eigenvalues (i.e., low signal directions). This leads to accumulation of (independent) noise across iterations. However, this can be remedied by partial cancellation of this noise due to \emph{anti-correlations}, as illustrated in \Cref{fig:noise_cancellation} of \Cref{chap:intro}.

In particular, the suboptimality $R(\mech)$ decomposes along the directions of the eigenvectors of the input covariance $\bfH$ as  follows. 
Let $\lambda_j$ and $\bfv_j$ for $j=1, \ldots, m$ denote the eigenvalues and eigenvectors of $\bfH$. Then, we can decompose the suboptimality as
\[
R(\mech) = \sum_{j=1}^m R_j(\mech)\,,
\]
where $R_j$ is the suboptimality incurred in the direction given by $\bfv_j$, i.e., this depends on $\lambda_j$ alone and not on $\{\lambda_i\}_{i \neq j}$.
It can be shown that the contribution of each direction is the same if using independent noise:
\[
    \lim_{n \to \infty} R_j(\mech_{\mathsf{indep}, n}) = 
        \Theta(1)\,.
\]
This also holds for low signal directions with $\lambda_j$ small.
On the other hand, with correlated noise as in \Cref{eq:opt-toeplitz-c-regr}, we get that the contribution of a direction $j$ scales with the eigenvalue $\lambda_j$: 
\[
    \lim_{n \to \infty} R_j(\mech_{\mathsf{Toep}, n}) \le \widetilde O(\lambda_j) \,,
\]
where $\widetilde O(\cdot)$ hides constants and log factors of problem-dependent constants.
That is, the contribution of lower signal directions $j$ with small eigenvalue $\lambda_j$ reduces proportionally due to the noise cancellation effect.

Finally, since the gradients of a strongly convex objective function can already provide some noise cancellation effect, the max-loss-optimal Toeplitz mechanism of \Cref{eq:opt-toeplitz-c-regr} requires less aggressive anti-correlations than in the prefix sum estimation of \cref{eq:opt-toeplitz-c}. This explains the a damping factor of $(1-\nu)^t$ in the former.

In summary, anti-correlated DP noise can prevent noise accumulation by repeatedly cancelling out a part of the noise in low signal directions.

\section{Proofs of Technical Results*}
\label{sec:ch3-proofs}

We give here proofs of technical results of \Cref{lem:operator-norms} and \Cref{prop:independent-noise-optimal}.

\begin{proof}[Proof of \Cref{lem:operator-norms}]
    The claim follows from plugging in the inequality
    \[
        \norm{\bfu}_\infty \le \norm{\bfu}_1 \le \nd \, \norm{\bfu}_\infty \quad \text{for all } \bfu \in \R^\nd \,,
    \]
    into the definition of the operator norm (\cref{eq:induced-norm}) and noting that $\colnorm{\cdot} = \colnormop{\cdot}$. In particular, let $\bfu_1$ be such that 
    \begin{align}
        \forall \bfu \in \R^{\nd}, \quad \frac{\norm{\strategy \bfu}_2}{\norm{\bfu}_1} \leq \frac{\norm{\strategy \bfu_1}_2}{\norm{\bfu_1}_1}
    \label{eq:1to2norm}        
    \end{align}
    and $\bfu_\infty$ be such that 
    \begin{align}
        \forall \bfu \in \R^{\nd}, \quad \frac{\norm{\strategy \bfu}_2}{\norm{\bfu}_\infty} \leq \frac{\norm{\strategy \bfu_\infty}_2}{\norm{\bfu_\infty}_\infty}
    \label{eq:inftyto2norm}        
    \end{align}
    Setting $\bfu:=\bfu_\infty $  in \cref{eq:1to2norm}, we get 
    \[
        \frac{1}{\nd} \norm{\strategy}_{\infty \to 2} = \frac{\norm{\strategy \bfu_\infty}_2}{n\norm{\bfu_\infty}_\infty} \leq \frac{\norm{\strategy \bfu_\infty}_2}{\norm{\bfu_\infty}_1} \leq \frac{\norm{\strategy \bfu_1}_2}{\norm{\bfu_1}_1} = \colnormop{\strategy} = \colnorm{\strategy}.
    \]
    The other inequality follows similarly.
\end{proof}

\begin{proof}[Proof of \Cref{prop:independent-noise-optimal}]
    The crux of the proof is to establish
\begin{align} \label{eq:proof-multipart-worst}
    f(\strategy) := \rownorm{\prefix \strategy^{-1}} \, \norm{\strategy}_{\infty \to 2} \ge \norm{\prefix}_{\infty \to \infty} \,.
\end{align}
We will prove this later. Assuming that it holds for now, the next step is to note that $\norm{\bfM}_{\infty \to \infty} = \max_{t \in [n]} \norm{\bfM[t, :]}_1$ is the maximum absolute row sum of the matrix $\bfM$.
For the matrix $\bfM = \prefix$, this is attained by the last row and we have $\norm{\prefix}_{\infty \to \infty} = \nd$. Thus, we have, $f(\strategy) \ge \nd$.

The next step of our proof is to establish that the identity matrix $\strategy = \bfI_{n \times n}$ achieves the lower bound with equality. First, $\rownorm{\prefix} = \sqrt{\nd}$ is the largest row norm of $\prefix$, which is attained by the last row of $\prefix$. Second, we have that $\norm{\bfI}_{\infty \to 2} = \sqrt{n}$, because of the matching upper and lower bounds
\begin{align*}
    \norm{\bfI}_{\infty \to 2} &= \max_{\bfu \ne \zeros} \norm{\bfu}_2 / \norm{\bfu}_\infty\le \sqrt{n} \, \quad \text{and} \\
    \norm{\bfI}_{\infty \to 2} &\ge \norm{\bm{1}_n}_2 / \norm{\bm{1}_n}_\infty = \sqrt{n} \,.
\end{align*}
Here, the upper bound is based on
    the vector norm inequality $\norm{\bfu}_2 \le \sqrt{n} \norm{\bfu}_\infty$ and we plugged in $\bm{1}_n := (1, \ldots, 1)\in\R^n$ for the lower bound.
Thus, we have that 
\[
    f(\bfI) = \rownorm{\prefix} \, \norm{\bfI}_{\infty \to 2} = n \,,
\]
establishing the required claim $\min_\strategy f(\strategy) \ge n = f(\bfI)$.

\paragraph{Proving \cref{eq:proof-multipart-worst}}
To complete the proof, we have to show \cref{eq:proof-multipart-worst}.
Fix a vector $\bfu$ such that $\strategy\bfu \ne \zeros$. Noting that the max row norm can be written as an induced matrix norm, we have
\begin{align*}
    \rownorm{\prefix \strategy^{-1}}
    &= \rownormop{\prefix \strategy^{-1}}
    \ge \frac{\norm{\prefix \strategy^{-1} (\strategy\bfu)}_\infty}{\norm{\strategy\bfu}_2} = \frac{\norm{\prefix \bfu}_\infty}{\norm{\strategy\bfu}_2} \,.
\end{align*}
Rearranging (and noting that it holds trivially for $\bfu$ such that $\strategy\bfu = \zeros$), we get the following inequality that holds for all $\bfu \in \R^n$:
\begin{align}   \label{eq:proof-multipart-worst-2}
   \rownorm{\prefix \strategy^{-1}} \, \norm{\strategy\bfu}_2 \ge \norm{\prefix \bfu}_\infty \,.
\end{align}
Fix a vector $\bfu$ with $\norm{\bfu}_\infty \le 1$. Continuing on with the definition of $\norm{\cdot}_{\infty \to 2}$, we get
\begin{align*}
    f(\strategy) &= \rownorm{\prefix \strategy^{-1}} \, \left( \max_{\norm{\bfv}_\infty \le 1} \norm{\strategy \bfv}_2 \right)  \\
    &\ge \rownorm{\prefix \strategy^{-1}} \,\norm{\strategy\bfu}_2  \\
    &\ge \norm{\prefix \bfu}_\infty \,,
\end{align*}
where the last inequality followed from \cref{eq:proof-multipart-worst-2}. Since this inequality holds for all $\bfu$ such that $\norm{\bfu}_\infty \le 1$, we can take the largest lower bound to get
\[
 f(\strategy) \ge \max_{\norm{\bfu}_\infty \le 1} \norm{\prefix \bfu}_\infty =: \norm{\prefix}_{\infty \to \infty} \,.
\]
This establishes \cref{eq:proof-multipart-worst}.
\end{proof}

\section{Bibliographic Notes}
\label{sec:ch3-biblio}

\paragraph{Motivation for Correlated Noise in AI}
The challenges in proper subsampling and its impact on privacy accounting is covered recently by \citet*{annamalai2024shuffle, chua2024private, chua2024scalable}. 
For various aspects of federated learning and privacy, refer to the monograph by \citet{kairouz2019flsurvey}.
For a survey on continual learning problems where user behavior can shift over time, see \citet*{wang2024comprehensive}. The observation of distribution drift due to feedback mechanisms where users strategically adapt their behavior in response to a deployed system has been studied by~\citet*{perdomo2020performative}.

\paragraph{Multi-Epoch Sensitivity}
Much of the material on multi-epoch correlated noise mechanisms in \Cref{sec:ml:multi-epoch} is due to \citet*{choquette2022multi,choquette2024amplified}.
We give precise attributions for each part of the preceding section below.

\paragraph{Challenges with Multi-Epoch Sensitivity}
Recall from \Cref{sec:multi-epoch:sensitivity} that the multi-epoch sensitivity is equal to $\norm{\strategy}_{\infty \to 2}$ in the absence of meaningful restrictions on the participation patterns.
\citet[Sec. 4.3]{tropp2003topics} showed that computing the $\norm{\cdot}_{\infty \to 2}$ induced matrix norm is NP-hard, and \citet{steinberg2005computation} showed that any general $\norm{\cdot}_{p \to q}$ norm for $p > q$ is NP-hard to compute.

\paragraph{Practical Participation Patterns}
The cyclical participation schema was introduced in \citet*{choquette2022multi} under the name ``$(k,b)$-participation'' while the minimum separation participation schema was introduced in \citet*{choquette2024amplified} as its relaxation suitable to federated learning.

\paragraph{Multi-Epoch Sensitivity: Bounds}
The sensitivity bounds of \Cref{sec:multi-epoch:sensitivity-bounds} can be attributed as follows:
\begin{itemize}
    \item Lemma~\ref{lem:multi-epoch-sens:generic-bounds}: Part \ref{item:sens:2} is due to \citet*[Prop. E.1]{choquette2024amplified}, while \ref{item:sens:3} is taken from \citet*[Eq. 3, Cor. 2.1]{choquette2022multi}.
    \item Lemma~\ref{lem:multi-epoch-sens:special-cases}: Part \ref{part:sens:lb} is elementary; its second inequality has been noted, for instance in \citet*[Remark 3.2]{mcmahan2024hassle}. Part \ref{part:sens:banded} previously appeared in \citet*[Thm. 2]{choquette2024amplified}. Finally, Part \ref{part:sens:Toep} was established in \citet*[Thm. 2]{kalinin2024banded}, while Part \ref{part:sens:bandedToep} is a direct corollary of \ref{part:sens:banded} and \ref{part:sens:bandedToep}.
\end{itemize}

\paragraph{Multi-Epoch Sensitivity: Algorithms}
Next, we move on to the algorithms presented in \Cref{sec:multi-epoch:sensitivity-algos}.
The brute force computation of sensitivity for cyclic participation was proposed in \citet*{choquette2022multi}, while the dynamic program for banded matrices was used by \citet*{choquette2024amplified}. The closed-form expression was, as we previously noted, established in \citet*
{kalinin2024banded} and was used later, for instance, in \citet*{mcmahan2024hassle}. Recently, \citet*{kalinin2025back} introduced a new explicit factorization method, {\em Banded Inverse Square Root}, which imposes a banded structure on the inverse correlation matrix. 

Finally, the tensorization process of restarted mechanisms in \Cref{sec:multi-epoch:restart} was called as ``Stamping'' in \citep[App. D.4]{choquette2022multi}; restarting mechanisms is a commonly used trick, particularly for baselines.

\paragraph{Amplification by Sampling for Correlated Noise Mechanisms}
The amplification result of \Cref{thm:banded-amplif} is due to \citet*{choquette2024amplified}. Unfortunately, this resulting has two drawbacks: it is applicable only to banded matrices, and the result guarantees can be worse than the unamplified ones at large $\eps$. The follow-up work of \citet*{choquette2024privacyamplif} lifted the first drawback (but in the setting of Poisson sampling), giving a procedure that gives amplified guarantees for non-banded correlated noise mechanisms. \citet*{choquette2024near} consider a different sampling scheme which allows tight guarantees through Monte Carlo sampling.

\paragraph{Privacy Preserving Learning} 
Differentially private empirical risk minimization (DP-ERM) and DP stochastic convex optimization (DP-SCO)~\citep*{BST14,bassily2019private,feldman2019private,bassily2020stability,asi2021private,bassily2021non,zhang2022differentially,asi2021adapting,kulkarni2021private,gopi2022private,chaudhuri2011differentially,kifer2012private} are probaly the most studied problems in the theoretical DP literature.  This body of work captures the optimal privacy/utility trade-offs for a large class of convex optimzation problems. 
A comprehensive survey article for various methods used in practice to instantiate DP-SGD (and detailed best practices) can be found in \citet*{ponomareva2023dp}, including how one instantiate subsampling to achieve better accuracy guarantee.

\paragraph{Theoretical Analysis of Learning with Correlated Noise Mechanisms}
The generic bounds of \Cref{sec:generic-learning-bounds} were shown by \citet*{kairouz2021practical} for the non-realizable (agnostic) setting, while the one in the realizable setting was shown by \citet*{asi2023near}.
The detailed bounds of \Cref{sec:detailed-learning-bounds} were shown by \citet*{choquette2023correlated}. Finally, the currently best known rate for single-pass algorithms for stochastic convex optimization is in~\citet*{choquette2024optimal}.  We refer to these original works for further details and proofs.

The analysis of \Cref{sec:detailed-learning-bounds} is based on \cref{alg:dpsgd-corr} without clipping. However, this is only an approximation as clipping does impact the optimization dynamics, see \citet*{zhang2020gradient,chen2020understanding,zhang2022understanding,xiao2023theory,koloskova2023revisiting,schaipp2024sgd,marshall2024clip} and the references therein.
Using anti-correlated DP noise to  prevent noise accumulation by repeatedly cancelling out a part of the noise in low signal directions is discussed in more details in  \citet*[Remark C.16]{choquette2023correlated}.

%% file: 4_practical.tex
We now turn to practical considerations in implementing correlated noise mechanisms. 
In particular, we discuss how to solve the optimization problems involved in constructing correlated noise mechanisms for the streaming setting (\Cref{chap:prefixsum}) and the multiple-participation setting (\Cref{chap:ml}). We also discuss the nuances involved in efficient noise generation, including in distributed environments.

While our discussions apply more broadly to correlated noise mechanisms based on any lower triangular and Toeplitz workload matrix $\workload \in \R^{n \times n}$, we use the prefix sum workload matrix, encountered in stochastic gradient descent (SGD; see \Cref{sec:weightedPrefixSum}), as a concrete example:
\begin{align} 
    \prefix = \begin{pmatrix}
    1 & 0 & \cdots &0 \\
    1 & 1 & \cdots & 0\\
    \vdots & \vdots & \ddots & \vdots  \\
    1 & 1 & \cdots & 1
    \end{pmatrix} \in \{0,1\}^{\nd \times \nd} \,.
\end{align}

The main focus of this section is two fold. First, we discuss numerical optimization techniques to find a factorization $\prefix = \bfB\strategy$ with lower-triangular factors $\bfB,\strategy$. In particular, we setup the problem in \Cref{sec:strategy-opt}, focusing on specific methods for the optimizing the dense mechanism in \Cref{sec:ch4-dense} and parameterized mechanism (e.g. BLT) in \Cref{sec:parameterized-mech}. In both cases, we handle the streaming and multi-participation settings.
The second objective of this section is to provide practical recommendations on the design choices in correlated noise mechanisms (\Cref{sec:recommendations}).

\section{Mechanism Optimization Using Numerical Methods}
\label{sec:strategy-opt}

Recall that we considered two types of mechanisms in \Cref{chap:prefixsum}: (a) those that involved numerically optimizing the strategy matrix $\strategy$, such as the dense mechanism of \Cref{sec:optimized-dense-mech} or the Buffered Linear Toeplitz (BLT) mechanism of \Cref{sec:ch2-blt}, and (b) mechanisms with handcrafted matrices, such as the max-loss-optimal Toeplitz mechanism of \cref{thm:optimal-toeplitz} ( \Cref{sec:optimial-toeplitz-mech}).

While numerical optimization of the mechanisms requires additional up-front computation cost (compared to the handcrafted mechanisms), they have a key practical advantage: they can be configured for a wider variety of settings, particularly varied participation patterns in the multiple-participation setting. Such settings are particularly relevant in the learning setting, as we discussed in \Cref{chap:ml}.

\paragraph{Setting}
Our goal is find a lower triangular strategy matrix $\strategy \in \R^{n\times n}$, such that some loss induced by the factors $\bfB = \prefix \strategy^{-1}$ and $\strategy$ is minimized. While we focused primarily on the \txtmaxloss (\Cref{def:max-loss}) in \Cref{chap:intro,chap:prefixsum}, it is customary to use the root-mean square loss (\txtrmsloss) as an objective for numerical optimization (\Cref{def:rms-loss}). For the streaming setting, we showed in \Cref{thm:rmse:closedform} that:
\begin{align} \label{eq:ch4-rmse}
    \rmslossn(\strategy) :=
    \rmslossn(\prefix\strategy^{-1}, \strategy) = \frac{1}{\sqrt{\nd}} \lfrob{\prefix\strategy^{-1}} \colnorm{\strategy} \,,
\end{align}
where we take $\bfB = \prefix \strategy^{-1}$ if not specified otherwise.
In the multiple-participation setting, the term $\colnorm{\strategy}$ is replaced with the participation calibrated sensitivity as discussed in \Cref{sec:multi-epoch:practical-patterns}.

\begin{figure*}
    \centering
    \begin{subfigure}[b]{0.5\textwidth}
        \centering
        \includegraphics[width=\linewidth]{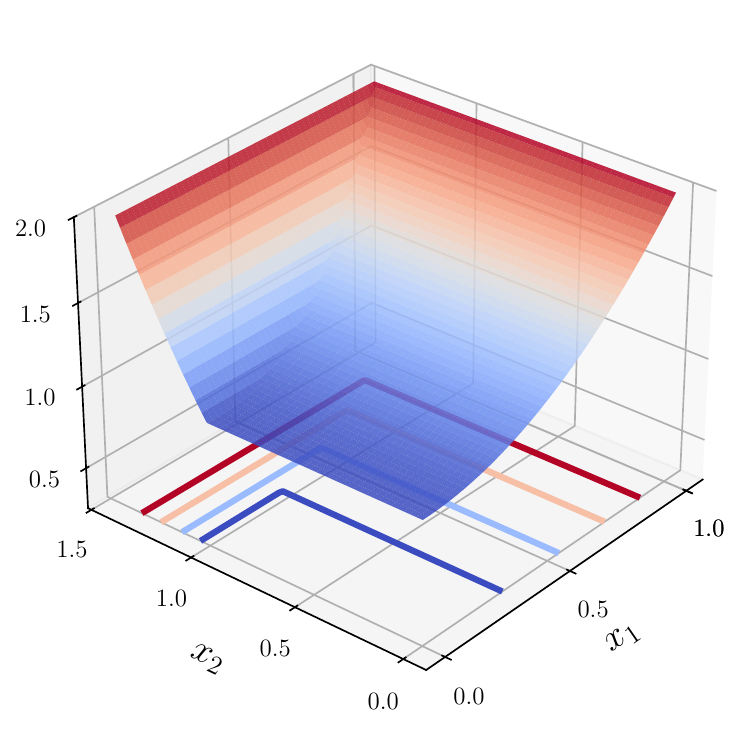} 
        \caption{$\rownorm{\bfB}^2$, encountered in the \txtmaxloss.}
    \end{subfigure}%
    ~ 
    \begin{subfigure}[b]{0.5\textwidth}
        \centering
        \includegraphics[width=\linewidth]{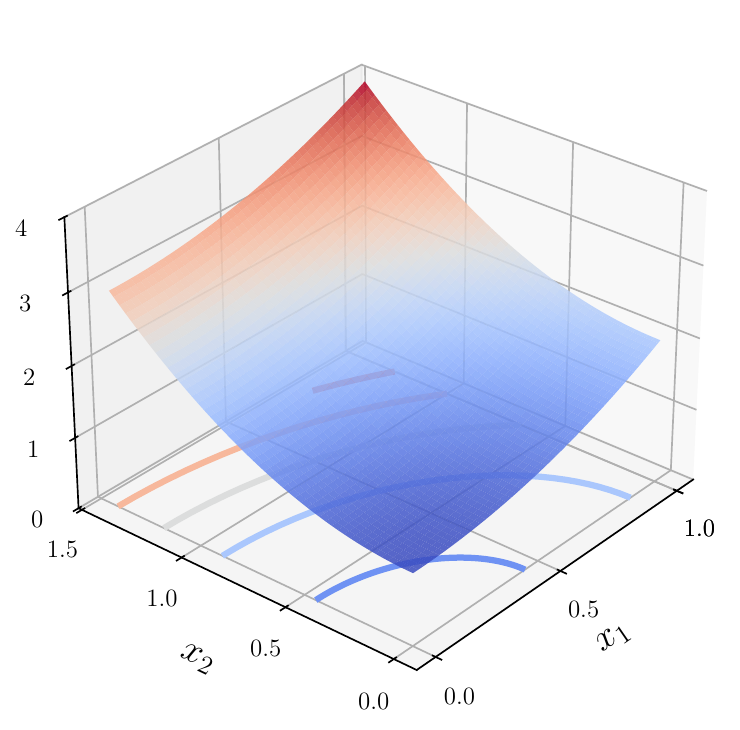} 
        \caption{$\lfrob{\bfB}^2$, encountered in the \txtrmsloss.}
    \end{subfigure}
    \caption{We plot $\rownorm{\bfB}^2$ and $\lfrob{\bfB}^2$ for the matrix
    $\bfB = \begin{pmatrix} x_1 & 0 \\ 1 & x_2 \end{pmatrix}$ as a function of $x_1, x_2$. Notice the non-smoothness in the left plot. 
       }
    \label{fig:errors}
\end{figure*}

The \txtrmsloss has traditionally been the objective of choice to optimize correlated noise mechanisms, both for streaming prefix sums and for machine learning, as it better captures the ``overall'' \txtloss across all prefix sums than the \txtmaxloss. Moreover, the squared norm $\lfrob{\cdot}^2$ encountered when optimizing the (square of the) \txtrmsloss is a smooth function,\footnote{
    We say a function $f$ is smooth if it is continuously differentiable and its gradient $\nabla f$ is Lipschitz.
}
while $\rownorm{\cdot}^2$ encountered in the (square of the) \txtmaxloss of \cref{eq:norm-max-err} is not; see \Cref{fig:errors} for an illustration. Practically, this makes the optimization of \txtrmsloss more stable.

In the upcoming sections, we focus on optimizing mechanisms whose strategy matrix $\strategy$ can be of two types: 
\begin{enumerate}
    \item \textbf{Dense strategies}, which are represented explicitly as a matrix $\strategy$. Dense strategies provide full generality and coverage over the space of strategies.  
    \item \textbf{Parameterized strategies}, which represents the strategy implicitly in terms of a parameter vector $\bfphi \in \R^p$ via the parameterization $\strategy(\bfphi)$. Examples include the banded Toeplitz strategies (\Cref{sec:ch2-banded-Toeplitz}) and the BLT strategies (\Cref{sec:ch2-blt}).
\end{enumerate}

\begin{bremark}[Scalability]
\label{remark:Scalability}
While explicit dense strategy matrices $\strategy$ are more general, their $O(n^2)$ space and $O(n^3)$ optimization time complexity can be prohibitive for large number of steps $n$.
Indeed, prior work has scaled up the dense mechanism only to $n \approx 10^4$ steps.
For example, the numerical mechanism optimization in the open-source Jax Privacy library for dense mechanisms takes less than $10$ seconds (on a GPU) for $n=1024$ steps, but the running time increases $100\times$ to around $15$ minutes for $n=8192$.
When appropriately designed, implicitly represented strategies tend to have compact tractable representations and can be optimized efficiently for much larger values of $n$. 
\end{bremark}

\section{Optimizing the Dense Mechanism}
\label{sec:ch4-dense}

The dense mechanism attempts to directly optimize the strategy matrix $\strategy$ to minimize the \txtrmsloss objective. We treat the streaming and multiple-participation settings separately.

\subsection{Optimization in the Streaming Setting}
\label{sec:practical-opt-streaming}

The \txtrmsloss objective takes the simple form of \Cref{eq:ch4-rmse} in the streaming setting.
The key challenge arises from this objective being a non-convex function of $\strategy$. Fortunately, it can be rewritten as a convex optimization problem as a function of the Gram matrix $\bfM = \strategy\T\strategy$. The key property is the following:

\begin{blemma}\label[lemma]{lem:rmse-gram-matrix}
For any two matrices $\bfA, \bfC \in \mathbb{R}^{n \times n}$ with $\bfC$ invertible, and $\bfM = \bfC\T \bfC$, we have 
\[
\colnorm{\bfC}^2 = \norm{\diag{(\bfM)}}_\infty,
\quad\text{and}\quad
\lfrob{\bfA \bfC^{-1}}^2 = \trace{\bfA \bfM^{-1} \bfA\T} \,.
\]
\end{blemma}
\begin{proof}
The proof follows from direct calculation using the following two facts: (1) $\bfM[t, t] = \norm{\bfC[:, t]}_2^2$, and (2) $\lfrob{\bfQ}^2 = \trace{\bfQ\T \bfQ}$ for any real matrix $\bfQ$.
\end{proof}

The next ingredient to a convex reformulation is that the objective $\rmslossn(\strategy)$ is scale-invariant: $\rmslossn(\strategy) = \rmslossn(\alpha \strategy)$ for any constant $\alpha \neq 0$. Thus, we can fix a scale by imposing the constraint that $\colnorm{\strategy} = 1$, or equivalently, that $\diag(\bfM) \le \ones$ element-wise. Together, we end up with the following convex optimization problem:

\begin{bproblem}\label{prob:singleepoch}
Find the matrix $\bfM_\star$ that solves the optimization problem
\begin{equation} \label{eq:singleepoch}
\begin{aligned}
& {\mathrm{minimize}}
& & \trace{\bfA \bfM^{-1} \bfA\T} \\
& \text{subject to}
& & \diag(\bfM) = \ones \quad \text{and} \quad \bfM \succ 0
\end{aligned}
\end{equation}
and then find $\bfC_\star$ so that $\bfC_\star\T \bfC_\star = \bfM_\star$ via e.g., Cholesky decomposition.
\end{bproblem}

Here, the positive-definiteness constraint $\bfM \succ \zeros$ ensures that there exists a matrix $\strategy_\star$ such that $\strategy_\star\T \strategy_\star = \bfM_\star$ and the objective $\trace{\workload\bfM^{-1} \workload}$ is finite and convex. 
 Finally, note that we replace the \emph{inequality} constraint $\diag(\bfM) \le \ones$ with the \emph{equality} constraint $\diag(\bfM) = \ones$, as this does not change the solution: As we showed in \cref{lem:column-norm}  the set of optimal strategies always includes a column-normalized one. This has the added benefit of yielding an unconstrained optimization problem, as we will soon discuss.
Together, we get the equivalence:

\begin{btheorem}
    The solution $\strategy_\star$ obtained from \Cref{prob:singleepoch} minimizes the \txtrmsloss, i.e.,
    $$
    \rmslossn(\strategy_\star) = \min_{\strategy\in \R^{n \times n}} \rmslossn(\strategy).
    $$
\end{btheorem}

See \cref{fig:multi-participation-heatmaps} (left) for an example of a mechanism optimized using this approach.

\begin{figure*}
    \centering
    \includegraphics[width=0.7\linewidth]{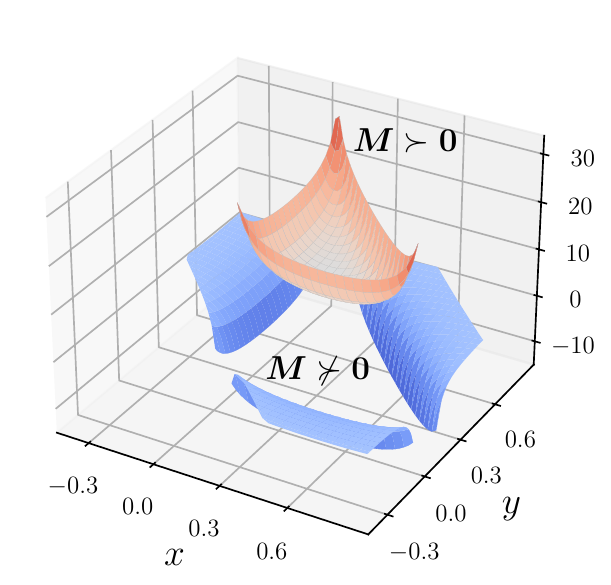}
    \caption{Dropping the positive-definiteness constraint of \Cref{prob:singleepoch}: we plot $f(\bfM) = \trace{\bfM^{-1}}$ with $\bfM = \diag\big( x, y, 1-x-y\big)$, plotted as a function of $x, y$. (We have $\trace{\bfM} = 1$ by construction.)
    Notice that the red region with $\bfM \succ \zeros$ (where the objective is convex) is disconnected from the blue regions where $\bfM$ fails to be positive definite. Indeed, $f(\bfM)$ is discontinuous whenever one of the eigenvalues of $\bfM$ is zero.
    A gradient-based optimizer initialized in the red region with $\bfM \succ \zeros$  with appropriate safeguards (such as line search or a small enough learning rate) will generally not leave that region. Thus, this positive-definiteness constraint can be heuristically dropped in practical implementations.
    }
    \label{fig:optim_problem}
\end{figure*}

\paragraph{Practical Algorithms}
While \Cref{prob:singleepoch} can be written as a semi-definite program, it is possible to develop more efficient solutions using unconstrained quasi-Newton algorithms.
The equality constraint $\diag(\bfM) = \ones$ amounts to fixing the diagonal elements of $\bfM$ to one and not optimizing over them. (In contrast, the inequality constraint $\diag(\bfM) \le \ones$ is more challenging to handle, and requires constrained optimizers.)

Next, prior work has generally found it to be safe to ignore the positive-definiteness constraint $\bfM \succ \zeros$ as well (once $\diag(\bfM) = \ones$ is imposed), as long as the optimization of $\bfM$ is initialized at a feasible point. The intuition behind this is illustrated in \Cref{fig:optim_problem}.  Ignoring this constraint as a heuristic, \Cref{prob:singleepoch} is then a \emph{smooth, unconstrained, and convex} optimization problem.  These three properties are crucial as they allow us to leverage rapidly convergent off-the-shelf optimization algorithms to solve \Cref{prob:singleepoch}. We recommend L-BFGS, a limited memory quasi-Newton algorithm, for its rapid empirical convergence and highly optimized numerical implementations.

\paragraph{Time and Space Complexity}
The $n \times n$ matrix $\bfM$ requires $O(n^2)$ memory. Thus, the total memory requirement of L-BFGS is $O(n^2)$ (assuming a small constant number of memory buffers for L-BFGS). Since the gradient $\nabla_\bfM \trace{\bfA \bfM^{-1} \bfA\T} = - \bfM^{-1} \bfA\T\bfA \bfM^{-1}$ requires computing $\bfM^{-1}$, the per-step time complexity is $O(n^3)$.  In practice, when utilizing GPUs, it is feasible to solve this problem up to $n \approx 10^4$.

\subsection{Optimization in the Multiple-Participation Setting}
\label{sec:optimization_multi_participation}
The main difference between the streaming and the multiple-participation setting lies in how sensitivity is calculated. As discussed in \Cref{sec:multi-epoch:practical-patterns,sec:multi-epoch:sensitivity-bounds}, we need specialized algorithms tailored to particular participation schema to tightly compute the sensitivity. Indeed, in this case, we have that the \txtrmsloss of a strategy $\strategy$ under a participation schema $\Pi$ and model dimension $\mdim$ is
\begin{align} \label{eq:rmse-multi-epoch}
    \rmslossn(\strategy \, |\, \Pi, m) = \sens(\strategy, \Pi, \mdim)  \,\, \lfrob{\prefix\strategy^{-1}} \,,
\end{align}
where $\sens(\cdot, \cdot, \cdot)$ is the participation-calibrated sensitivity (also see \Cref{def:sens-participation}) and the other factor $\bfB$ is again understood to be  $\bfB=\prefix\strategy^{-1}$, and is omitted for brevity.

From an optimization perspective, it is convenient to use the following corollary of \cref{lem:multi-epoch-sens:generic-bounds}:

\begin{bcorollary}
Let $\bfC \in \mathbb{R}^{n \times n}$ denote a lower-triangular matrix and let $\bfM = \bfC^\top \bfC$ denote its Gram matrix.
\begin{enumerate}[label=(\alph*)]
    \item For any participation schema $\Pi$, we have we  
     \begin{align}
        \sens\br{\strategy, \Pi, \mdim}^2
        \le \,\, \max_{\bfpi \in \Pi} \,\, \sum_{t, \tau \in \bfpi} \Big|\bfM[t, \tau] \Big| \,.
    \end{align}
    \item This upper bound is tight (and the sensitivity is independent of the dimensions $\mdim$) if $\bfM{[t,\tau]} = 0$ for all $t \neq \tau \in \bfpi \in \Pi$.
\end{enumerate}
\end{bcorollary}

\cref{lem:multi-epoch-sens:generic-bounds}\ref{item:sens:3} gives more general non-negativity conditions for the upper bound to be tight. However, as discussed in the streaming setting, it is numerically easier to handle equality constraints rather than inequality constraints; we just fix $\bfM[t, \tau] = 0$ and do not optimize that variable. Thus, in the interest of scalability of numerical optimization, it is common to opt for this more restrictive formulation with equality constraints.

The constraint that $\bfM{[t,\tau]} = 0$ means that the noise added in two iterations $t, \tau$ where the same datapoint (or user) $i$ can participate is \emph{uncorrelated}. While this is a modest constraint for the cyclic participation, it imposes a $b$-banded structure (\Cref{def:banded}) for $b$-MinSep participation (\Cref{def:min-sep}).
We design optimization algorithms under this assumption. In particular, since the sensitivity is then dimension-independent, we denote it by $\sens(\strategy, \Pi)$, as in \Cref{def:sens-participation}.

Similar to the streaming setting, this objective is scale-invariant, so we can impose the bound $\sens(\strategy, \Pi) \le 1$. This leads us to the following optimization problem: 

\begin{bproblem}\label{prob:multiepochopt}
Given a participation schema $\Pi$, (i) find the matrix $\bfM_\star$ that solves the optimization problem
\begin{equation} \label{eq:multiepochopt}
\begin{aligned}
& \underset{\bfM \succ 0}{\mathrm{minimize}}
& & \trace{\bfA \bfM^{-1} \bfA\T}  \\
& \text{subject to}
& & \sum_{t \in \bfpi} \bfM{[t,t]} = 1~ \forall \bfpi \in \Pi \\
& & & \bfM{[t, \tau]} = 0 \:\:\forall \:\: t, \tau \in \bfpi, \, \bfpi \in \Pi, t \neq \tau \\
\end{aligned}
\end{equation}
and then (ii) find $\bfC_\star$ so that $\bfC_\star\tp \bfC_\star = \bfM_\star$ via e.g., Cholesky decomposition.
\end{bproblem}

Note that $\sens(\bfC, \Pi) \le 1$ is equivalent to the inequality constraint $\sum_{t \in \bfpi} \bfM[t, t] \le 1$ for each $\bfpi \in \Pi$. Like in the streaming setting, we have replaced the inequality constraint with an \emph{equality} in \cref{eq:multiepochopt}. Unlike the streaming setting, however, we do not have a proof of optimality of this step. This substitution is based on the following conjecture, which is strongly supported by empirical evidence:

\begin{bconjecture} \label{conj:equal-participation-sensitivity}
For every participation schema $\Pi$ and workload $\workload$, the unique solution $\bfM_\star$ of \Cref{prob:multiepochopt} remains optimal when the first constraint of \Cref{eq:multiepochopt} is replaced with the inequality constraint $\sum_{t \in \bfpi} \bfM{[t,t]} \le 1~ \forall \bfpi \in \Pi$.
\end{bconjecture}

Practically, \Cref{prob:multiepochopt}, with its equality constraints, has several advantages over the inequality constraints.  First, it simplifies the problem significantly, making it closer to \Cref{prob:singleepoch} from the streaming setting, which is well-understood.  Second, imposing these constraints improves the convergence of the optimization algorithm, allowing it to find better solutions in less time.  Third, it reveals interesting and interpretable structures about the behavior of optimal strategies with different participation schemas (e.g., \Cref{prop:dpgd-mf}).

\paragraph{Practical Algorithms}
\Cref{prob:multiepochopt} is only slightly harder {to solve} than its streaming version in \Cref{prob:singleepoch}.  The individual entries of $\bfM$ where $\bfM{[t,\tau]} = 0$ can be removed as variables and constraints from the optimization problem,\footnote{
    One way to achieve this is to initialize $\bfM[t,\tau] = 0$ and never update it during the course of optimization.
} 
leaving the linear equality constraint $\sum_{t \in \bfpi} \bfM{[t, t]} = 1$ as the main technical challenge to overcome. Ideally, we solve this problem in a similar fashion as in the streaming case, i.e. using an off-the-shelf rapidly-convergent optimization algorithm like L-BFGS.  
Since L-BFGS cannot natively handle equality constraints, some modifications are necessary, as discussed next for the case of cyclic participation.

\paragraph{Algorithms for Cyclic Participation}
Constraints can be incorporated into gradient descent by \emph{projecting} the iterates onto the constraint set.  Because the constraints are linear, this can be equivalently achieved by projecting the gradients (instead of the iterates) onto the \emph{level sets} of the constraints. Heuristically, we recommend using the same orthogonal projection strategy for L-BFGS.\footnote{
    L-BFGS can natively handle box constraints. However, more complex constraints require projections in the Mahalanobis norm defined by L-BFGS's Hessian approximation. Instead, our heuristic uses a Euclidean projection, making it directly compatible with existing highly-optimized L-BFGS implementations (see \Cref{sec:ch4-biblio} for references).
}  That is, to use an out-of-the-box implementation of L-BFGS, but pass in the projected gradients (onto the level sets of the constraints) in place of the true gradients.  This ensures that the iterates of the optimization respect the desired constraints, as long as the initial value does.  

We emphasize that this is a \emph{heuristic}, and we are not aware of convergence guarantees for this approach.  Specifically, we use the projected gradient $\mathsf{proj}(\bfG^+)$ in place of the actual gradient $\bfG^+$ in the L-BFGS update step, as follows:
\begin{align} \label{eq:proj-lbfgs}
    \mathsf{proj}(\bfG^+) = \argmin_\bfG\left\{ 
    \lfrob{\bfG - \bfG^+}^2 \, :\, 
    \sum_{t \in \bfpi} \bfG{[t,t]} = 0~ \forall \bfpi \in \Pi
    \right\} \,.
\end{align}

This projection is easy in the case of cyclic participation $\Pi^\cyclic_{\minsep, \maxpart}$, where the equality constraints are non-overlapping, i.e., each index $t$ only belongs to one $\bfpi \in \Pi^\cyclic_{\minsep, \maxpart}$.
In this case,  $\mathsf{proj}(\bfG^+)$ simply modifies the diagonal entries of $\bfG^+$ for $|\Pi^\cyclic_{\minsep, \maxpart}| = k$ non-overlapping subsets of entries.  We refer the readers to \cref{fig:multi-participation-heatmaps} (far right) for an example of a mechanism optimized using this approach. In this example, some entries in the corresponding $\strategy$ are in fact negative, unlike with most of the other mechanisms we consider.

\paragraph{Algorithms for Min-Sep Participation}
The projection of \Cref{eq:proj-lbfgs} cannot be efficiently implemented with Min-Sep participation.
In particular, it has overlapping groups (i.e. $\bfpi \cap \bfpi' \neq \emptyset$), so an exact projection may not be possible. 
Approximation projections (e.g., based on Dykstra's alternating projection algorithm) would also be infeasible, as their cost grows with $|\Pi|$ and $|\Pi|$ is exponentially large for the Min-Sep schema.  

An alternate heuristic employed  is to impose the constraint that $\diag(\bfM) = \ones$:

\begin{bproblem}\label{prob:multiepoch-heuristic}
Given a participation schema $\Pi$, (i) find the matrix $\bfM_\star$ that solves the optimization problem
\begin{equation} \label{eq:multiepoch-heuristic}
\begin{aligned}
& \underset{\bfM \succ 0}{\mathrm{minimize}}
& & \trace{\bfA \bfM^{-1} \bfA\T}  \\
& \text{subject to}
& & \diag(\bfM) = \ones \\
& & & \bfM{[t, \tau]} = 0 \:\:\forall \:\: t, \tau \in \bfpi, \, \bfpi \in \Pi, t \neq \tau \\
\end{aligned}
\end{equation}
and then (ii) find $\bfC_\star$ so that $\bfC_\star\tp \bfC_\star = \bfM_\star$.
\end{bproblem}

In the setting where every example participates an equal number of times, i.e. $|\bfpi| = |\bfpi'|$ for all $\bfpi, \bfpi' \in \Pi$, then $\diag(\bfM) = c \, \ones$ implies the constraint of \Cref{eq:multiepochopt} for an appropriate constant $c$.\footnote{We can take $c=1$ due to scale invariance of the objective.} 
This is true for cyclic participation but not for Min-Sep participation.

Despite this difference,
\Cref{prob:multiepoch-heuristic} naturally interpolates between the streaming setting and the full batch setting.
Indeed, it is nearly identical to the \Cref{prob:singleepoch} from the streaming setting, with the only difference being a smaller set of free variables.  As a result, the same optimization routines used in the streaming setting are applicable here with minor modifications.  
On the other hand, analogous to \Cref{prop:independent-noise-optimal}, \Cref{prob:multiepoch-heuristic} recovers independent noise under full-batch participation:\footnote{
    While computing the sensitivity in the full-batch case is NP-hard (\Cref{sec:multi-epoch:sensitivity}), we can nonetheless solve Problems \ref{prob:multiepochopt} and \ref{prob:multiepoch-heuristic} efficiently. This is because we additionally impose $\bfM[t, \tau] = 0$ for all $t \neq \tau$.
}

\begin{bproposition}
\label{prop:dpgd-mf}
    Under the full-batch setting (i.e., under  participation schema $\Pi^{\mathsf{full}} = \{[n]\}$), we have:
    \begin{enumerate}[label=(\alph*)]
        \item The minimizers $\bfM_\star$, $\bfC_\star$ of \Cref{prob:multiepochopt} are diagonal matrices.
        \item \Cref{prob:multiepoch-heuristic} is solved by $\bfM = \bfI_{n\times n} = \bfC$.
    \end{enumerate}
\end{bproposition}
\begin{proof}
    The proof follows from noting that only diagonal matrices are allowed under the constraints of \Cref{prob:multiepochopt} for $\Pi = \{[n]\}$, and that we require these diagonal elements to be 1 in \Cref{prob:multiepoch-heuristic}.
\end{proof}

\begin{bremark}[Full-batch correlated noise mechanisms]
\label{remark:full-batch}
In the full batch setting, where every example participates in every round,
the diagonal entries of the optimal strategy $\strategy_\star$ solving \Cref{prob:multiepochopt} (as described in \Cref{prop:dpgd-mf}) are generally decreasing, which means \emph{more noise is added in the later iterations}.\footnotemark
\, From the machine learning perspective, this is counter-intuitive and possibly even undesirable.  On the other hand, from the perspective of minimizing \txtrmsloss on the prefix queries, this is natural, as the earlier elements of the stream are required by more of the prefix queries, and hence need less noise. 

Beyond this abnormality, \Cref{prop:dpgd-mf} reiterates the observation of \Cref{prop:independent-noise-optimal}
that correlated noise offers no further improvements over independent noise in the full-batch setting.  Indeed, the most notable improvements of correlated noise mechanisms over independent noise mechanisms occur in the streaming (i.e., the  single-epoch) setting, and significant gains can still be obtained in large-scale compute-limited settings. 
\end{bremark}
\footnotetext{
    The decreasing ordering of diagonal elements is a property of \txtrmsloss. Recall for \txtmaxloss that the diagonal elements are equal, as we established in \Cref{prop:independent-noise-optimal}.
}

\paragraph{Time and Space Complexity}
The time and space complexity of solving \Cref{prob:multiepochopt} with projected L-BFGS for both cyclic and $\minsep$-minimum separated participation remains unchanged from the streaming setting. That is, we require $O(n^3)$ time per L-BFGS step (with a small constant number of total steps), and $O(n^2)$ space when representing the strategies as dense matrices.

For $\minsep$-minimum separated participation,  \Cref{prob:multiepoch-heuristic} imposes a bandedness constraint that $\bfM[t, \tau]$ = 0 for all $|t-\tau| \ge \minsep$. By using parameterized strategies instead, we can reduce the time and space complexity to $O(n \minsep^2)$ and $O(n\minsep)$ respectively, which we discuss further in \Cref{sec:parameterized-mech}.

\begin{bremark}[Primal vs. Dual Optimization]
Problems \ref{prob:singleepoch}-\ref{prob:multiepoch-heuristic} can also be effectively solved via their dual problems.\footnotemark 
We recommend using the primal approach described earlier with L-BFGS for multiple reasons. First, it generally converges very rapidly (possibly due to the use of limited second order information about the curvature of the objective function). Second, the primal problem formulation can more naturally handle additional constraints on $\bfM$. Finally, parameterized mechanisms can only be tackled by primal-based approaches (as we see in the next section), allowing for a unified approach.
\end{bremark}
\footnotetext{It turns out that these problems satisfy strong duality, so the primal and dual optimal values coincide.}

\section{Optimizing Parameterized Mechanisms}
\label{sec:parameterized-mech}

While dense strategies provide the best utility, they can be expensive or infeasible to optimize and deploy in practice for $n \gtrsim 10^4$ steps (see \Cref{remark:Scalability}). This precludes their use in some regimes of practical interest. We now discuss approaches to optimizing over parameterized strategy classes such as Banded Toeplitz and Buffered Linear Toeplitz.  These strategy classes have sufficient expressive capacity to represent near-optimal strategies (obtaining expected errors close to the optimal dense strategies), while their structure allows for efficient optimization and noise generation even with a large number of steps $\nd$.

Let $\bfphi \in \Phi \subset \R^p$ represent arbitrary parameters and let $\bfC(\bfphi) \in \R^{n \times n}$ denote a lower-triangular strategy matrix parameterized by $\bfphi$. We focus in particular on two parameterized mechanisms introduced in \Cref{chap:prefixsum}:
\begin{enumerate}[label=(\alph*)]
    \item \textbf{The Banded Toeplitz mechanism} (\Cref{sec:ch2-banded-Toeplitz}): Given parameters $\bfphi=(c_0, \ldots, c_{\minsep-1})$, we parameterize the first column of $\strategy(\bfphi)$ as
    \[ 
        \big(\strategy(\bfphi)\big)[:, 0] = (c_0, \ldots, c_{\minsep-1}, 0, \ldots, 0)\,.
    \]
    Further, we take $\strategy(\bfphi)$ to be Toeplitz, so all other columns can be determined from the first one, e.g. for $\minsep=3$ and $\nd=4$,
    \[
    \strategy = \begin{bmatrix}
    c_0 &   0 &   0 &   0 &   0 \\
    c_1 & c_0 &   0 &   0 &   0 \\
    c_2 & c_1 & c_0 &   0 &   0 \\
    0   & c_2 & c_1 & c_0 &   0 \\
    0   &   0 & c_2 & c_1 & c_0 \\
    \end{bmatrix} \,.
    \]
    \item \textbf{The Buffered Linear Toeplitz (BLT) mechanism} (\Cref{sec:ch2-blt}): Given parameters $\bfphi = (\bfalpha, \bflambda)$, where  $\bfalpha \in \R^d_+$ is a scale parameter and $\bflambda \in [0, 1)^d$ is a decay parameter, we parameterize $\strategy(\bfphi) = \BLT(\bfalpha, \bflambda)$ as in \Cref{eq:blt-param}.
\end{enumerate}

The goal here is then to find parameters $\bfphi$ so as to minimize the \txtrmsloss or the \txtmaxloss.   We consider the \txtmaxloss in addition to the \txtrmsloss because for Toeplitz strategies (both banded and BLT), the \txtmaxloss is equal to the last-iterate \txtloss, and is hence a smooth function that is amenable to numerical methods.\footnote{
    In all the cases we consider, we have that $\sens(\strategy(\bfphi), \Pi)$ is independent of the model dimension $\mdim$, as it satisfies the conditions of \cref{lem:multi-epoch-sens:generic-bounds}\ref{item:sens:3}. 
}

\begin{bproblem} \label{prob:param-opt}
Given a participation schema $\Pi$, find the parameters $\bfphi_\star \in \Phi$ that solve the optimization problem
\begin{align} \label{eq:param-multipeoch-opt}
\underset{\bfphi \in \Phi}{\mathrm{minimize}} \quad
\sens\big(\strategy(\bfphi), \Pi \big)^2 \,\, \norm{\bfA \bfC(\bfphi)^{-1} }_{\calE}^2  \,,
\end{align}
where $\norm{\cdot}_{\calE} = \lfrob{\cdot}/\sqrt{n}$ for the \txtrmsloss and $\norm{\cdot}_{\calE} = \rownorm{\cdot}$ for the \txtmaxloss. 
For the streaming setting where $\Pi = \Pi^{\mathsf{single}} = \left\{(0), (1), \ldots, (n-1) \right\}$, the problem simplifies to 
\begin{align} \label{eq:param-singleepoch-opt}
\underset{\bfphi \in \Phi}{\mathrm{minimize}} \quad
\colnorm{\strategy(\bfphi)}^2 \,\, \norm{\bfA \bfC(\bfphi)^{-1} }_{\calE}^2  \,.
\end{align}
\end{bproblem}

\paragraph{Practical Algorithms}
We tackle \Cref{prob:param-opt} with automatic differentiation coupled with off-the-shelf gradient-based optimization. Specifically, it suffices to have a function that can efficiently evaluate the objective function with differentiable operations. In order to scale to a large number of steps $\nd$, it is crucial to be able to evaluate the objective function \emph{without ever materializing the matrices $\bfC(\bfphi)$ or $\bfA$ explicitly} by exploiting their special structure. Details for how this can be achieved with specific strategy classes are given in the subsections below.

In general, \Cref{prob:param-opt} is a non-convex optimization problem with respect to either the strategy matrix $\bfC$ or its parameters $\bfphi$ for the banded Toeplitz and BLT mechanisms.  Thus, gradient-based optimization is not guaranteed to converge to a global optimal solution.  
However, in practice, we find that appropriate initialization and parameter tuning lead to high-quality solutions.\footnote{
    For example, we find in the streaming setting that the gradient-based solutions obtained are competitive with the Toeplitz mechanism (\Cref{sec:optimial-toeplitz-mech}), which is an upper bound on how well both the banded Toeplitz and BLT mechanisms can perform in the streaming setting.  See \Cref{sec:ch2-empirical} for empirical comparisons. 
}

Next, we describe how to evaluate the objective of \Cref{prob:param-opt} for both the banded Toeplitz and BLT mechanisms.

\subsection{Optimizing the Banded Toeplitz Mechanism}\label{sec:optimizeblts}

When the number of bands $b'$ of the banded Toeplitz mechanism is chosen to be no greater than the minimum separation $b$ under MinSep participation $\Pi^\MinSep_{\minsep, \maxpart}$, we have by \cref{lem:multi-epoch-sens:special-cases}\ref{part:sens:bandedToep} that the multiple-participation sensitivity $\sens(\strategy, \Pi^\MinSep_{\minsep, \maxpart})$ is $\sqrt{\maxpart}$ times the streaming sensitivity $\colnorm{\strategy}$, where $\maxpart$ is the maximum number of partitions.
The same relation also holds for the cyclic participation setting, where $b=N/B$ is then the number of steps in an epoch and $k$ is the number of epochs.
In both cases, $k$ is a constant and can be ignored for optimization.

It remains to efficiently compute the error $\norm{\bfA \bfC(\bfphi)^{-1} }_{\calE}^2$.
Because the matrix $\bfB = \bfA \strategy(\bfphi)^{-1}$ is Toeplitz (as long as $\bfA$ is Toeplitz) and has coefficients $\bfb = \strategy(\bfphi)^{-1} \bf{1} \in \R^\nd$, the expected error can be computed efficiently as
\begin{align} \label{eq:err:computation-graph}
    \rownorm{\prefix \strategy(\bfphi)^{-1}}^2 &= \norm{\bfb}_2^2 \,, \quad 
    \lfrob{\prefix \strategy(\bfphi)^{-1}}^2 = \sum_{t=0}^{n-1} (n-t) b_t^2 \,.
\end{align}
These expressions can be computed efficiently.
First, computing the vector $\bfb \in \R^n$ requires $O(n b)$ time and $O(n)$ space by leveraging the banded structure of $\strategy(\bfphi)$ (see \Cref{alg:banded-mult-by-cinv} in \Cref{chap:prefixsum} for details).
Second, given the vector $\bfb$, we can evaluate the expressions of \Cref{eq:err:computation-graph} in $O(n)$ time.

The final step to optimize for \Cref{prob:param-opt} using first-order optimization is to calculate the gradients of the objective w.r.t. the parameters.
This can be done with automatic differentiation. 
The sensitivity $\colnorm{\bfC(\bfphi)}^2 = \sum_{t=0}^{b-1} c_t^2$ is clearly a differentiable function of the parameters $\bfphi=(c_0, \ldots, c_{b-1})$.
The error $\norm{\bfA \bfC(\bfphi)^{-1} }_{\calE}^2$ is a function of $\bfb$, 
which is in turn a function of the $\bfphi$. 

\paragraph{Practical Considerations}
We find that this non-convex optimization problem is somewhat sensitive to its initialization. In practice, initializing the optimizer with $\bfphi_0 = (c_0^\star, \ldots, c_{\minsep-1}^\star)$, which are the optimal Toeplitz coefficients from \Cref{thm:optimal-toeplitz}, is highly effective across a wide range of settings.
Then, an off-the-shelf implementation of L-BFGS with default parameters returns a high quality solution.  

\begin{figure*}
    \centering
    \includegraphics[width=0.85\linewidth]{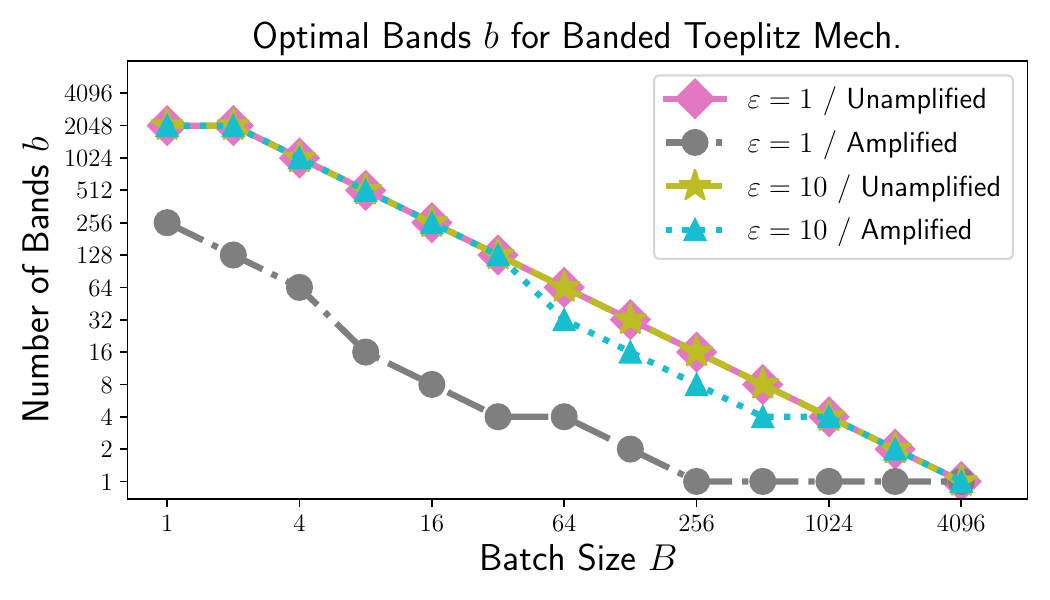} 
    \caption{
    \textbf{Tuning the Number of Bands}:
    We plot the optimal number of bands $b$ of the banded Toeplitz mechanism as a function of the batch size $\bsz$. The number of bands is chosen to empirically minimize the \txtmaxloss with and without amplification at various values of the privacy budget $\eps$ for $(\eps, \delta)$-DP.
    Throughout, we fix the number of iterations $\nd = 2048$ and dataset size $\datasize=4069$, as in \Cref{fig:scaling_batch_size}.
    The value of $\eps$ does not matter in the unamplified scenarios, as the unnormalized max loss is simply a scaled version of $\sens(\strategy) \, \rownorm{\prefix\strategy^{-1}}$ (see \Cref{thm:accuracyMatrixMechanism,thm:gdp-of-correlated-noise-multiple-participation}).
    On the other hand, when accounting for amplification by sampling, the optimal number of bands varies with $\eps$: this is because the effect of amplification is determined by the noise multiplier, as we discussed in \Cref{fig:amplification}.
    Recall from \Cref{fig:scaling_batch_size} that significant amplification is obtained at $\eps=1$: this figure shows that this requires a smaller number of bands. On the other hand, 
    there is almost no amplification at $\eps=10$, so the optimal number of bands is almost identical to the unamplified one.
   } 
   \label{fig:scaling_batch_size_bands}
\end{figure*}

{Furthermore}, we find that it is important to tune the number of bands. One can do so without paying a privacy cost of working with real data by using the \txtmaxloss or \txtrmsloss as a proxy for learning performance. See \Cref{fig:scaling_batch_size_bands} which shows how the optimal number of bands varies with the batch size.

\subsection{Optimizing the BLT Mechanism}
\label{sec:blt-opt}

Our high-level approach is to express the \txtrmsloss/\txtmaxloss objective as a differentiable function of the parameters $\bfphi$ that we wish to optimize over. This enables us to leverage automatic differentiation to optimize the objective with gradient-based optimization.

We first map the parameters $\bfphi$ to the BLT/inverse-BLT parameters $\bfalpha, \widehat \bfalpha \in \R^n$ and $\bflambda, \widehat \bflambda \in [0, 1)^d$ using a differentiable function such that the inverse of $\strategy = \BLT(\bfalpha, \bflambda)$ is given by $\strategy^{-1} = \BLT(\widehat\bfalpha, \widehat\bflambda)$.\footnote{
    This is possible under the conditions imposed by \cref{lem:inverse-blt}.
}
The different ways of achieving this are summarized in \Cref{subfig:comp:1}.
Next, we map $(\bfalpha, \bflambda, \widehat \bfalpha, \widehat \bflambda)$ to the sensitivity and the \txtrmsloss\ or \txtmaxloss via a differentiable function, as summarized in \Cref{subfig:comp:2}. By the chain rule, the composition of both these steps gives the loss as a differentiable function of the parameters $\bfphi$.

We consider the streaming setting, cyclic participation, and $\minsep$-minimum separated participation with at most $\maxpart$ participations. The conditions of \cref{lem:multi-epoch-sens:special-cases}\ref{part:sens:Toep} will hold so that the sensitivity for the latter two multiple-participation settings will coincide, as in the banded case.

\paragraph{Parameterization Choices: From $\bfphi$ to BLT}
Two possible BLT parameterizations $\bfphi$ have been proposed such that the mapping $\bfphi \mapsto (\bfalpha, \bflambda, \widehat \bfalpha, \widehat \bflambda)$ is a differentiable mapping.

The first approach is to take $\bfphi = (\bfalpha, \bflambda)$ as the parameters of $\strategy = \BLT(\bfalpha, \bflambda)$ and reconstruct the inverse BLT parameters $\widehat\bfalpha, \widehat\bflambda$ from it as per \cref{lem:inverse-blt}; see \Cref{subfig:comp:1} (left) for an illustration.
This requires finding the roots of a degree-$d$ polynomial, a step that can be performed via an eigen-decomposition of a non-symmetric matrix (known as the \emph{companion matrix}) in $O(d^3)$ time. This subroutine is available as a differentiable function in automatic differentiation frameworks like JAX and PyTorch.

An alternate approach is to take $\bfphi=(\bflambda, \widehat\bflambda)$, and reconstruct $\bfalpha, \widehat \bfalpha \in \R^d$, see \Cref{subfig:comp:1} (right) for an illustration.\footnote{We allow the scale parameters $\bfalpha, \widehat \bfalpha$ to take both positive or negative values here. We circumvent this issue in practice with appropriate barrier functions.} This can be achieved in $O(d^2)$ time and memory as per the following lemma:

\begin{figure*}
    \centering
        \begin{subfigure}[b]{\textwidth}
        \hspace{2em}
       \adjustbox{max width=0.35\linewidth}{\input{graph_1a}}
       \hfill
       \adjustbox{max width=0.35\linewidth}{\input{graph_1b}}
       \hspace{2em}
      \caption{
      Computation graph of the differentiable mapping $\bfphi \mapsto(\bfalpha, \bflambda, \widehat\bfalpha, \widehat \bflambda)$ for different choices of the parameters $\bfphi$, denoted by the double bordered nodes. \textbf{Left}: 
      $\bfphi=(\bflambda, \bfalpha)$. \textbf{Right}: $\bfphi=(\bflambda, \widehat\bflambda)$.
      }
      \label{subfig:comp:1}
    \end{subfigure} \vspace*{0.5em}
    
    \begin{subfigure}[b]{\textwidth}
       \adjustbox{max width=\linewidth}{\input{graph_2}}
      \caption{
      Computation graph of the differentiable mapping from the BLT/inverse-BLT parameters $(\bfalpha, \bflambda, \widehat\bfalpha, \widehat \bflambda)$ to the \txtrmsloss/\txtmaxloss objectives.
    }
      \label{subfig:comp:2}
    \end{subfigure}
   
    \caption{
    \textbf{BLT Computation Graph}:
    We show the computation graph to compute the objective of \Cref{prob:param-opt} starting with different choices of the parameters $\bfphi$ of the BLT.
    The blue nodes denote variables (with the double bordered nodes denoting  $\bfphi$) while the orange nodes denote operations.
    }
    \label{fig:blt-computation-graph}
\end{figure*}
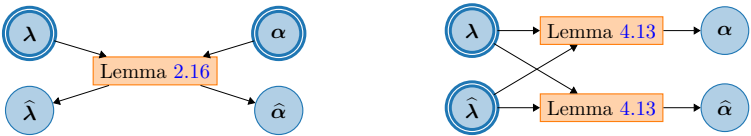
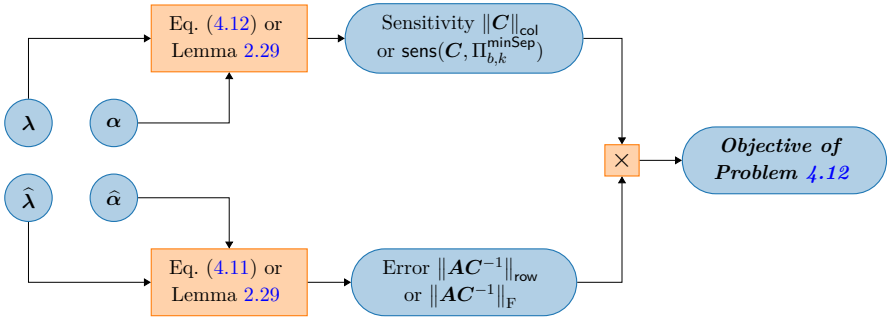

\begin{blemma} \label[lemma]{lem:calc-output-scale}
    Consider non-zero decay parameters $\bflambda, \widehat \bflambda \in \R^d$ that are pairwise distinct.\footnotemark 
    \, Then, the unique parameters $\bfalpha, \widehat\bfalpha \in \R^d$ that achieve $\BLT(\bfalpha, \bflambda) = \BLT(\widehat\bfalpha, \widehat \bflambda)^{-1}$ are given by:
    \begin{align*}
        \alpha_i &=  \frac{\prod_{j=1}^d \lambda_i - \widehat \lambda_j}{\prod_{j \neq i} \lambda_i - \lambda_j}\,,
        \quad\text{and} \quad
        \widehat\alpha_i = \frac{\prod_{j=1}^d \widehat\lambda_i - \lambda_j}{\prod_{j \neq i} \widehat\lambda_i - \widehat\lambda_j} \,.
    \end{align*}
\end{blemma}

\footnotetext{Specifically, $\lambda_i \ne \lambda_j$ and $\widehat \lambda_i \ne \widehat \lambda_j$ for all $i \ne j$, and $\lambda_i \ne \widehat \lambda_j$ for all $i, j \in [d]$.}

\noindent\cref{subfig:comp:1} compares these different parameterizations. 
From a theoretical perspective, both these parameterizations are equivalent, in that they represent the same class of BLT/inverse-BLT systems:

\begin{blemma} \label[lemma]{lem:blt-param-equivalent}
    Consider the following two BLT parameterizations:
    \begin{enumerate}[label=(\alph*), nosep]
        \item Let $\Phi_1$ denote the set of $(\bfalpha, \bflambda) \in \R^d_{++} \times (0, 1)^d$ that satisfy $\sum_{i=1}^d\alpha_i / \lambda_i < 1$, in addition to the conditions of \Cref{lem:inverse-blt};
        \item Let $\Phi_2$ denote the set of $(\bflambda, \widehat \bflambda) \in (0, 1)^d \times (0, 1)^d$ that satisfy the strict interlacing condition
        \[
        \bflambda_1 > \widehat \bflambda_1 > \bflambda_2 > \widehat \bflambda_2 > \cdots > \widehat \bflambda_{d-1} > \bflambda_d > \widehat \bflambda_d\,.
        \]
    \end{enumerate}
    Then, the set of BLT/inverse-BLT systems represented $\Phi_1$ and $\Phi_2$ are identical. That is, for every $\bfphi_1 \in \Phi_1$, there exists $\bfphi_2 \in \Phi_2$ such that $\BLT(\bfphi_1) = \BLT(\bfphi_2)$ and vice versa.
\end{blemma}

\paragraph{BLT to Loss}
Having obtained a differentiable mapping $\bfphi \mapsto (\bfalpha, \bflambda, \widehat \bfalpha, \widehat \bflambda)$, we now turn to expressing the loss as a function of these scale and decay parameters.
We consider two approaches to achieve this:

\begin{itemize}
\item \textbf{Toeplitz Coefficient Materialization}:
For a given time horizon $\nd$, we can materialize the first column $c_0 = 1$ and $c_t = \sum_{i=1}^d \alpha_i \lambda_i^{t-1}$ for $t\ge 1$ of the strategy matrix $\strategy=\BLT(\bfalpha, \bflambda)$. Then, denoting $\bfc = (c_0, \ldots, c_{n-1}) \in \R^\nd$, the sensitivity can simply be calculated as (see \Cref{eq:sens-blt-fn}):
\begin{align} \label{eq:sens:computation-graph}
    \colnorm{\strategy}^2 = \norm{\bfc}_2^2\,, \quad\text{and} \quad
    \sens(\strategy, \Pi^\MinSep_{b, k}) &=  \norm{\sum_{j=0}^{k-1} \calB^{jb}(\bfc)  }_2^2
    \,,
\end{align}
where $\calB^l(\bfc) = \strategy[:, l]$ is given by the \emph{backshift} operator:
\[
    \calB^l(\bfc) := \left(\underbrace{0, \ldots, 0}_{l \text{ zeros}}, c_0, \ldots, c_{n-l-1} \right) \in \R^n\quad\text{for } l \ge 0\,.
\]
These can be programmed as differentiable functions in automatic differentiation frameworks such as JAX or PyTorch that execute in $O(\nd d \maxpart)$ time. Reverse-mode automatic differentiation creates a computation graph that effective stores $\bfc$ in memory, leading to a $O(\nd)$ space complexity; see \cref{fig:blt-computation-graph} (a).

We can then materialize the first column of $\bfb = (\prefix \strategy^{-1})[:, 0]$ since we have that $\bfb$ is the vector of prefix sums of $\bfc' = (\bfC^{-1})[:, 0]$. Finally, we use \Cref{eq:err:computation-graph} to compute the error $\norm{\bfA \bfC(\bfphi)^{-1} }_{\calE}^2$.
These quantities can be computed in $O(\nd d)$ time and $O(\nd)$ memory, as can their derivatives with respect to the parameters (using automatic differentiation).
This can get prohibitively large, especially when the number $\nd$ of steps is in the order of billions.
 
\item \textbf{Closed-Form Expression}:
In the streaming (single-participation) setting, we can do better with a little more effort---all the sensitivity and error terms in \Cref{prob:param-opt} can be computed in closed form in $O(d^2)$ time and memory. This improvement is significant because, in practice,  $d<10$ typically provides sufficient accuracy for the BLT mechanism. The exact expressions are given in the appendix in \Cref{lem:blt-closed-form}.

\end{itemize}

\paragraph{Practical Considerations}
We recommend optimizing over the parameterization $\bfphi=(\bfalpha, \bflambda)$, as in \Cref{subfig:comp:1}. While the $O(d^3)$ eigen-decomposition is marginally more expensive, we find that it is numerically more stable.
To calculate the loss, we recommend using the closed-form expressions whenever available, i.e., for the streaming setting. This is especially advantageous with a large number $\nd$ of steps. For the multi-participations setting where closed-form expressions are not available, we recommend materializing the Toeplitz coefficients. 

Regardless of the parameterization (from \Cref{subfig:comp:1,subfig:comp:1}), it is crucial to impose a log-barrier function so that $\bflambda \in (0, 1)^d$, and $\bfalpha$ is coordinate-wise positive:
\[
    h(\bfalpha, \bflambda) = -\sum_{i=1}^d \left(\log(\lambda_i) + \log(1-\lambda_i) + \log(\alpha_i) \right) \,.
\]

Multiple sets of BLT parameters $\bfalpha$, $\bflambda$ can produce the numerically similar strategy matrix $\strategy$, with different (random) initializations resulting in different parameters. Interestingly, we observe that a smaller number of buffers $d$ leads to greater numerical stability. While theory suggests that a larger $d$ should always result in lower error, our empirical findings show that there is no benefit (in terms of \txtrmsloss and \txtmaxloss) in taking $d$ larger than $5$ for this reason. In fact, $d > 10$ can actually hurt empirical performance. This also suggests that reducing $d$ may help mitigate over-parameterization issues in this highly non-convex optimization landscape.

Using double-precision floating point arithmetic is crucial.\footnote{This requires an explicit adjustment in JAX and PyTorch, which default to single-precision.}
This is necessary because decay parameters $\lambda_i$ that are very close to, but strictly less than, $1$ are frequently encountered in practice. Finally, careful tuning of L-BFGS parameters helps improve robustness.

\begin{table}[t]
    \centering
\adjustbox{width=\linewidth}{
    \begin{tabular}{lccc}
       \toprule
       & \multicolumn{3}{c}{\textbf{Strategy Optimization Complexity}} \\
       \cmidrule(lr){2-4}
       \textbf{Mechanism} & \textbf{Runtime} & \textbf{Memory} & \textbf{Practical} \\
       & \textbf{(Per Step)} & \textbf{} & \textbf{Limitations} \\\midrule
       Dense & $\nd^3$ & $\nd^2$ & $\nd \lesssim 10^4$ \\
       $\bands$-Banded & $\bands \nd^2$ & $\bands \nd$ & $\nd \lesssim 10^5$ \\
       $\bands$-Banded Toeplitz & $\bands \nd$ & $ \nd$ & $\nd \lesssim 10^7$\\
       $d$-Buffered Linear Toeplitz, $k > 1$ & $d \nd$ & $\nd$ & $ \nd \lesssim  10^{8}$ \\
       $d$-Buffered Linear Toeplitz, $k=1$ & $d$ & $d$ & $n \lesssim  10^{10}$ \\
       \bottomrule
    \end{tabular}}
    \caption{Time and space complexity of optimizing over different strategy classes in terms of the number of steps $\nd$, the maximum number of participations $\maxpart$, the number of bands $\bands$ for banded strategy matrices (\Cref{sec:ch2-banded-Toeplitz}), and the order $d$ for the BLT mechanism (\Cref{sec:ch2-blt}). The optimal value of $\bands$ for Banded and Banded Toeplitz mechanisms depends on the problem parameters, see  \cref{fig:scaling_batch_size} for an example. The $\maxpart=1$ optimization of BLTs uses the closed-forms of the loss from \cref{sec:blt-opt}, while for $\maxpart > 1$ we use the Toeplitz coefficient materialization approach (necessitating time and space $O(\nd)$, but scaling very well in practice).
    }
    \label{tab:noise_complexity}
\end{table}

\section{Choosing a Correlated Noise Mechanism: Recommendations} 
\label{sec:recommendations}

We now provide some concrete rules of thumb to choose various design aspects of correlated noise mechanisms in the context of AI model training problem based on various practical considerations we have discussed so far. These recommendations are summarized in \Cref{fig:recommendations}.

\begin{figure}[tbhp]
    \centering
    \adjustbox{max width=0.99\linewidth}{\input{recommendations}}
    \caption{A summary of the practical recommendations for each of the design considerations highlighted in \Cref{fig:outline} in the context of AI model training.}
    \label{fig:recommendations}
\end{figure}

\subsection{Choosing a Workload Matrix}

Recall that the workload matrix does not explicitly appear in an implementation of DP-SGD with correlated noise (\cref{alg:dpsgd-corr} or the batch version \cref{alg:dpsgd-batch}). Indeed, the mechanism is fully specified by the noise-correlating matrix $\Cinv$ (whereas the $\strategy$ matrix is only needed to calibrate the noise for a particular $\mu$-GDP guarantee). At the end of the day, our goal is good learning performance for the final trained model, for example the empirical risk $\expectation_{\bfx \sim D}[\ell(\bfx, \bftheta_{n})]$ (cf. \cref{eq:empiricalrisk}). Unfortunately, we lack a rigorous theory to estimate this performance for a given noise-correlating matrix $\Cinv$ used in \cref{alg:dpsgd-batch}.\footnote{
    The learning guarantees in \Cref{sec:ml:learning-guarantees} are too loose for general problems, or only apply to specific problems such as linear regression.
}

In \cref{chap:intro}, we discussed a heuristic of selecting the workload matrix $\workload$ implied by the choice of first-order optimizer, e.g. $\prefix$ from \cref{eq:all-ones-workload} for vanilla SGD and $\Amom$ from \cref{eq:momentum-matrix} for SGD with momentum. Unfortunately, this heuristic does not apply to adaptive optimizers such as Adam or AdaGrad.

We recommend an alternative heuristic: choose $\workload=\prefix$ regardless of the base (non-private) optimizer.
That is, we optimize the mechanism to achieve low error on \emph{unweighted} prefix sum estimates. 
We find that this heuristic yields a $\Cinv$ that works very well in \cref{alg:dpsgd-batch} even if the actual first-order update rule (Line~\ref{line:optimizerupdate} of \cref{alg:dpsgd-batch}) corresponds to a different workload such as SGD with momentum, or even an adaptive gradient algorithm such as Adam or AdaGrad. Conversely, there is no guarantee that a mechanism selected for good performance on a particular momentum matrix $\Amom$ will outperform one optimized for prefix sums $\prefix$, even if the actual update rule corresponds exactly to the momentum workload matrix $\Amom$.

\begin{recommendation}[Workload Matrix]
    Choose $\workload=\prefix$ (irrespective of the base non-private optimizer).
\end{recommendation}

There has been some effort in getting theoretical guarantee for workload matrices arising from momentum methods, but they do not lead to space and time efficient mechanisms. Moreover, while we recommend the above, an important research question is whether there are better choices when we consider adaptive learning algorithms, like Adam, RMSProp, etc. We return to open questions in this space in \Cref{sec:ch5-adaptiveOptimizer}.
\subsection{Choosing a Surrogate Loss}

We primarily focused on the \txtmaxloss objective in this monograph, although the \txtrmsloss is another common choice in the literature.  In general, any differentiable function of the per query squared errors can be used as the objective with minor modifications to the optimization techniques discussed in this section. However, these more flexible alternatives have not been carefully studied previously. Prior work empirically comparing \txtmaxloss with mean squared error is also slim, but in the experiments that have been done the training-time performance characteristics have been similar.

\begin{recommendation}[Surrogate Loss for Mechanism Optimization]
    Choose the max loss (\Cref{def:max-loss}) or the RMS loss (\Cref{def:rms-loss}).
\end{recommendation}
As in the case of workload matrices, the choice of a better surrogate loss still remains an interesting research question. We discuss it in more detail in Section~\ref{sec:ch5-surrogateLoss}.

\subsection{Choosing a Participation Schema}
To a large extent, the possible participation schemas will be determined by the training environment and infrastructure.  
Generally, we recommend choosing the first schema in the following list that is fully supported by the infrastructure:

\begin{itemize}
    \item \textbf{Block Cyclic Poisson Sampling}, defined in \Cref{def:cyclic_poisson} should be preferred when possible, as it endows the mechanism with both the benefits of privacy amplification by sampling and noise correlation. Recall that block-cyclic sampling with $\bands=1$ blocks recovers the usual data processing pattern for DP-SGD with Poisson sampling. This schema is generally conceptually feasible in centralized training scenarios. Of course, if this sampling pattern is assumed for mechanism design and privacy accounting, it is essential that the training infrastructure correctly follows this sampling scheme; unfortunately, currently no major (DP) ML training platforms directly support such data processing.

    \item \textbf{Cyclic Participation}, described in \Cref{fig:participation_patterns}, should be used in centralized training scenarios where strictly enforcing Poisson subsampling is infeasible or too inconvenient. In the low privacy regime, cyclic participation will only be slightly worse than block cyclic Poisson sampling; see also \Cref{fig:amplification}.
    
    \item \textbf{Min-Sep Participation}, which was also described in \Cref{def:min-sep} and \Cref{fig:participation_patterns}, should be used in federated training scenarios where it is difficult to control the precise order of examples or users.  It may also be applicable in centralized training scenarios under the multi-attribution model of user-level DP. 
    
    \item \textbf{Single-participation} If none of the above are possible, it is generally relatively straightforward to ensure that each example contributes at most once to training (for example, training for a single epoch in an arbitrary order). For user-level privacy, we additionally need that each user contributes at most one example to training. 
\end{itemize}

\begin{recommendation}[Participation Schema]
    Depending on the setting, choose the following:
    \begin{itemize}
        \item Centralized (non-federated) learning: Choose block-cyclic Poisson sampling as a default. 
        \item Federated learning: Choose the Min-Sep participation schema.
    \end{itemize}
     For the centralized (i.e. non-federated) setting choose the max loss (\Cref{def:max-loss}) or the RMS loss (\Cref{def:rms-loss}).
\end{recommendation}

\subsection{Choosing a Class of Strategy Matrices}

Our default recommendation is to use banded Toeplitz strategies with column normalization in most settings.  These strategies offer the best expected errors in both federated settings (under a $b$-min-sep participation schema) and centralized settings (under a cyclic participation schema with Poisson sub-sampling).  Moreover, with a careful implementation, their memory overhead is generally small compared to the overall cost of the automatic differentiation to compute per-example gradients and gradient clipping + accumulation. 

On the other hand, when the number of participations $\maxpart$ is small, and a larger $\epsilon$ provides sufficient privacy protection, then amplification offers relatively {little} benefit, and so the mechanisms for the centralized un-amplified setting below may be sufficient (and allow e.g. much lower runtime costs via BLTs).

Below we provide some additional suggestions for scenarios where the default recommendation may not apply or may not be necessary. 
    \paragraph{Centralized training with many epochs and high privacy} In settings where we are training a model for many epochs over the dataset (typically using a batch size that consists of a significant fraction of the total training data set size), particularly when a high privacy bar is required (e.g., $1$-GDP or stronger),  correlated noise offers little to no improvement over a careful implementation of DP-SGD when accounting for amplification by sampling. Full-batch gradient descent (when the full dataset is used to compute each gradient) is an extreme example of this setting.
    However, since DP-SGD is just a special case of matrix factorization with banded strategies, there is no need to deviate from the default recommendation here in principle.
    
    \paragraph{Centralized training without amplification-via-sampling} 
    In the single-participation setting, column-normalized BLTs offer near-optimal loss, with the best-known noise-generation efficiency. Under \minseppart, banded matrices with $\bands=\frac{\nd}{\maxpart}$ bands can offer small performance improvements over BLTs, but the noise generation cost for large $\bands$ can be prohibitive.
    
    In centralized training scenarios without amplification, where cyclic participation can be enforced, using dense strategies can offer a slight improvement in the expected error compared to BLTs or banded matrices, but mechanism optimization is only feasible for $n \leq 4000$ or so.  Further, this approach has even more expensive noise generation for large $n$. In this case,  the cost of noise generation, which we discuss in greater detail in \Cref{sec:noise_generation_impls}, may start to outweigh the other training costs. 
        
    \paragraph{More than $\nd = 10^7$ training iterations} Optimizing banded Toeplitz strategies using the techniques we have described only scales up to $n=10^7$. Beyond this point, we recommend using BLTs instead, although the compute budget may be better spent on larger batch sizes or model sizes instead.

\begin{recommendation}[Strategy Matrix Class]
Depending on the setting, choose:
\begin{itemize}
    \item Centralized learning: choose the banded Toeplitz mechanism.
    \item Federated learning: choose the BLT mechanism.
\end{itemize}
\end{recommendation}

\subsection{Implementation Considerations for Noise Generation} \label{sec:noise_generation_impls}

There are several ways one can generate the correlated noise at training time, and these implementations have varying performance characteristics which we discuss in this section.  \Cref{alg:banded-mult-by-cinv} and \Cref{alg:blt-mult-by-cinv} from \Cref{chap:prefixsum} give two algorithms, specialized for banded Toeplitz and BLT strategies.  Here, we provide a more comprehensive overview of the implementation options and their trade-offs.

\paragraph{Approach 1: Naive Noise Generation}  
The simplest approach is to sample the full matrix of independent noise $\Znoise \sim \normalnm{0}{\stddev^2}$ in memory, and then compute $\bfC^{-1} \Znoise$ explicitly.  The rows of this matrix are the correlated noise vectors and can be iteratively indexed during model training.  This implementation of noise generation has an $n \times m$ time and memory overhead, and for large $n$ and $m$ this overhead can be prohibitive. This approach may be feasible in very small scale problems but should otherwise be avoided.

\paragraph{Approach 2: Memoryless Noise Regeneration} A better approach avoids the $n \times m$ memory overhead by regenerating the rows of $\Znoise$ in each iteration.  That is, on iteration $t$ we need to compute $\sum_{\tau=0}^{t-1} \bfC^{-1}{[t, \tau]} \Znoise{[\tau,:]}$.  By keeping track of the random seeds used to sample each row of $\Znoise$ we can avoid keeping the entire matrix in memory at once, in favor of generating the rows on demand exactly when 
needed.  This approach crucially relies on the fact that a pseudo-random number generator is used to sample the noise.  If the random keys are discarded after use or otherwise unavailable (e.g. for security purposes), this approach would not be viable.  The running time complexity of Approach 2 is the same as Approach 1, but it only has a $O(m)$ memory overhead.  Prior work have used this approach and demonstrated scalability up to about $n \approx 2000$ for models with a few million parameters. 

\paragraph{Approach 3: Low-Memory Streaming Linear Operators} A third approach applies for strategy classes with specific structure that enable efficient multiplication-by-inverse algorithms, like banded Toeplitz and BLT strategies.  We considered these approaches earlier in \Cref{alg:banded-mult-by-cinv} and \Cref{alg:blt-mult-by-cinv} of \Cref{chap:prefixsum}.  In both cases, the noise generator receives i.i.d. noise vectors as a stream of inputs, and returns appropriately correlated noise vectors as outputs in a streaming fashion.  To accomplish this, the streaming algorithm stores a ``buffer state'' in memory. For the banded Toeplitz strategy, the buffer consists of the correlated noise vectors for the previous $\bands - 1$ iterations, incuring a $O(b \mdim)$ time and memory overhead. For the BLT mechanism, we require $d$ buffers for a BLT of order $d$, incuring a $O(d \mdim)$ time and memory overhead.

\paragraph{{Distributed Noise Generation}}

As mentioned in \Cref{sec:recommendations}, correlated noise mechanisms are highly advantageous in large-scale training scenarios which typically occur in distributed environments with many accelerators working in tandem.  By carefully splitting up the work, we can greatly reduce the time and (per-machine) memory required to generate the correlated noise. There are several ``sharding'' strategies one could consider, which describe how the work should be partitioned across machines.
\newcommand{\grp}{\calG}
\begin{enumerate}

    \item \textbf{Sharding noise like the model} is one natural option.  There, each row $\bfz \in \R^\mdim$ of the noise matrix $\Znoise \in \R^{\nd \times \mdim}$, which has the same size and structure as the model, is distributed across machines in the same way as the model.  This approach is natural and can leverage  existing model sharding rules that are known to be compatible with the model shape.  Moreover, once the noise is generated, it can be added to the model gradient without any communication.  However, for models trained with only data parallelism (where a full copy of the model is replicated on every machine), this approach does not leverage the additional machines effectively, as it duplicates the work of generating noise on each machine.  
    
    \item \textbf{Sharding noise across all machines} is a better option that ensures no duplicate work is done across machines.  There, each row $\bfz \in \R^\mdim$ of the noise matrix $\Znoise \in \R^{\nd \times \mdim}$ is evenly distributed across all machines in the environment, and all computations on these vectors respect this sharding. That is, if we have $M$ machines, we (as evenly as possible) partition the indices $[\mdim]$ into $M$ groups $\{\grp_0, \dots, \grp_{M-1}\}$, with each machine $i$ on iteration $t$ being responsible for computing the entries 
    \begin{equation}\label{eq:allshard}
            \big(\corrZnoise\idx{t}{j}\ |\ j \in \grp_i \big) 
            \qquad \text{for} \qquad
            \corrZnoise = \Cinv \Znoise \,.   
    \end{equation}    
    For all three approaches above, 
    this sharding strategy requires no communication between machines to compute the values of \cref{eq:allshard}. 
    Hence, noise computation is an entirely embarrassingly parallel computation and its running time is therefore inversely proportional to the number of machines in the training environment.  
    Finally, the per-iteration noise vector $\corrznoise_t$ must be assembled from \cref{eq:allshard} so that it can be added to the (clipped and aggregated) gradient which in general uses different sharding.  
    
    \item \textbf{Generating noise on CPU} is a third option worth considering, where noise is generated on the CPUs while the forward/backward pass is done on an accelerator. At the surface, this may seem appealing because it allows the noise generation to happen asyncronously rather than sequentially. However, communicating large matrices from CPU to accelerators is typically slow on current hardware, and simple preliminary tests rule this out as a viable approach.
\end{enumerate}

\begin{bremark}[Handling Complex Model Structures]
In the discussion above, we assumed that the model and noise are generated as flat vectors in $\R^\mdim$, and that $\mdim$ is divisible by the number of machines.  In practice, the model parameters are often represented as a collection of smaller arrays of various shapes, whose sizes may not all be divisible by the number of machines.  This can be handled by appropriate flattening and padding. (The open-source software of \Cref{sec:oss} automatically implements such sub-routines.)
\end{bremark}

\begin{bremark}[Relative Cost of Noise Generation]
\label{rem:relativecomplexity}
Until this section, we have discussed the costs of different correlated noise generation strategies relative to each other (e.g. in \Cref{tab:mechanisms}).
While the Dense and Toeplitz mechanisms have a $n$ times higher time complexity than Input and Output perturbation, they are not actually $n$   times slower in practice, since correlated noise generation is only one part of the cost associated with training a model with differential privacy.  The other primary cost is computing (and clipping) per-example gradients. 
Perhaps somewhat unintuitively, it turns out that the overhead of correlated noise generation is small relative to the other costs of DP training, \emph{even when the number of buffers is large}. 
This can be attributed to careful implementation (e.g. described above) coupled with the need to use very large batch sizes to get the best utility with DP.

While the relative costs of each step are dependent on the model and compute environment, let us next consider a concrete example. For large transformer-based language models, the cost of gradient computation and clipping can be approximated as $\approx 6 \cdot m \cdot B \cdot S$ where $B$ is the batch size and $S$ is the sequence length. For common language models, $S$ is typically taken to be in the thousands and $B$ is at least as large as $S$ for DP training. Hence this cost can often outweigh the $b \cdot m$ time complexity of noise generation required for e.g., a banded Toeplitz strategy for reasonable values of $b$, $B$, and $S$.  As a concrete example, if $b=100, B = S = 1000$, then noise generation requires $\leq 0.01\%$ of the total step time.  On the other hand, if $S=1$, which essentially reduces to a fully connected neural network, then noise generation requires closer to $\sim 10\%$ of the total step time.
\end{bremark}

\section{Open-Source Software}
\label{sec:oss}

As discussed in this section, correctly and efficiently optimizing for correlated noise mechanisms requires some care to correctly handle subtleties. Jax Privacy provides implementations of the techniques described in this section. At the time of writing, Jax Privacy provides well-tuned strategy optimization routines for dense, banded, banded toeplitz, and buffered linear toeplitz strategies. It also provides implementation of the various noise addition strategies discussed in \cref{sec:noise_generation_impls} that can run efficiently in large distributed environments.

\section{Bibliographic Notes}
\label{sec:ch4-biblio}

\paragraph{Dense Strategies}

\Cref{lem:rmse-gram-matrix}, that underlies all numerical optimizations for the dense mechanism was established by \citet*{li2015matrix}.  Many approaches have been proposed to optimize the dense mechanism in the streaming setting, including primal approaches \cite*{yuan2016convex,mckenna2021hdmm,choquette2024amplified} and dual approaches \cite*{denisov2022improved,choquette2022multi}.  \citet*{mckenna2021hdmm} demonstrated the effectiveness of solving  \Cref{prob:singleepoch} with L-BFGS, including the heuristic to ignore $\bfM \succ \zeros$.
Using memoryless noise regeneration has used to demonstrate scalability up to about $n \approx 2000$ for models with a few million parameters by several works like \citet*{denisov2022improved,choquette2022multi,choquette2024amplified}.

\paragraph{A note on L-BFGS}
L-BFGS can natively handle box constraints~\citet*{zhu1997algorithm}. However, more complex constraints require projections in the Mahalanobis norm defined by L-BFGS's Hessian approximation~\cite{schmidt2011projected}. Instead, the heuristic used in this monograph is a Euclidean projection, making it directly compatible with existing highly-optimized L-BFGS implementations.

\paragraph{Parameterized Strategies}

In the counting query literature, there have been many prior works that optimize carefully parameterized strategies to overcome scalability limitations of the dense representation.  \citet*{qardaji2013understanding} optimize the branching factor over generalized hierarchical strategies, while \citet*{li2014data} optimize the weighting coefficients for each level of the hierarchy.  \citet*{mckenna2021hdmm} propose optimizing so-called $p$-Identity strategies specifically for the pure-DP version of the problem, and strategies built from smaller building blocks via the Kronecker product.  The approaches to optimize the BLT parameters with a $(\bflambda, \widehat \bflambda)$ parameterization are due to \citet*{dvijotham2024efficient,mcmahan2024hassle}. In particular, \cref{lem:calc-output-scale} results from a simplification of the result of \citet*[Lem. 5.2]{dvijotham2024efficient}; the exact statement we give can be found in \citet*{mcmahan2025inverse}. On the other hand, the $(\widehat \bfalpha, \widehat \bflambda)$ parameterization approach, and the equivalence between the two parameterizations (i.e., \Cref{lem:blt-param-equivalent}) are from \citet*{mcmahan2025inverse}.

\paragraph{Matrix mechanism in other settings}

\citet*{xiao2023optimal,ding2011differentially,mckenna2021hdmm} propose matrix mechanisms specifically tailored for marginal query workloads.  \citet*{mckenna2021hdmm} further provided optimized strategies for the broad class of conjunctive query workloads, that are applicable for domains as large as $10^9$. \citet*{edmonds2020power} showed that the matrix mechanism can be applied under local DP with favorable theoretical guarantees, and \citet*{mckenna13workload} proposed a more practical variant that generalizes randomized response rather than the Gaussian mechanism.

%% file: graph_1a.tex
\begin{tikzpicture}[
node distance = 5mm and 7mm,
   arr/.style = {-Triangle},
   box/.style = {rounded rectangle, draw, semithick,
                 minimum size=9mm},
    boxA/.style = {rounded rectangle, draw, semithick, minimum size=9mm, color=C0, fill=C0!35},
    boxAA/.style = {rounded rectangle, draw, double, ultra thick, minimum size=9mm, color=C0, fill=C0!35},
    boxB/.style = {rounded rectangle, draw, semithick, minimum size=9mm, color=C0, fill=C0!35},
    op/.style = {rectangle, draw, semithick,
                 color=C1, fill=C1!35},
                        ]
\node (n1) [boxAA] {\textcolor{black}{${\bflambda}$}};
\node (op1) [op, right=of n1, yshift=-0.7cm] {{\textcolor{black}{Lemma~\ref{lem:inverse-blt}}}};
\node (n3) [boxAA, right=of n1, xshift=3cm]
{\textcolor{black}{${\bfalpha}$}}; 

\node (n2) [boxB, below=of n1, yshift=0cm] {\textcolor{black}{$\widehat{{\bflambda}}$}};
\node (n4) [boxB, below=of n3, yshift=0cm] {{\textcolor{black}{$\widehat{{\bfalpha}}$}}};



\draw[arr]   (n1) -- (op1);
\draw[arr]   (n3) -- (op1);
\draw[arr]  (op1) -- (n2);
\draw[arr]  (op1) -- (n4);






\end{tikzpicture}

%% file: graph_1b.tex
\begin{tikzpicture}[
node distance = 5mm and 7mm,
   arr/.style = {-Triangle},
   box/.style = {rounded rectangle, draw, semithick,
                 minimum size=9mm},
    boxA/.style = {rounded rectangle, draw, semithick, minimum size=9mm, color=C0, fill=C0!35},
    boxAA/.style = {rounded rectangle, draw, double, ultra thick, minimum size=9mm, color=C0, fill=C0!35},
    boxB/.style = {rounded rectangle, draw, semithick, minimum size=9mm, color=C0, fill=C0!35},
    op/.style = {rectangle, draw, semithick,
                 color=C1, fill=C1!35},
                        ]
\node (n1) [boxAA] {\textcolor{black}{${\bflambda}$}};
\node (n2) [boxAA, below=of n1] {\textcolor{black}{$\widehat{{\bflambda}}$}};
\node (op1) [op, right=of n1, xshift=0.1cm] {{\textcolor{black}{Lemma~\ref{lem:calc-output-scale}}}};
\node (op2) [op, right=of n2, xshift=0.1cm] {{\textcolor{black}{Lemma~\ref{lem:calc-output-scale}}}};
\node (n3) [boxA, right=of op1] {{\textcolor{black}{${\bfalpha}$}}};
\node (n4) [boxB, right=of op2] {{\textcolor{black}{$\widehat{{\bfalpha}}$}}};



\draw[arr]   (n1) -- (op1);
\draw[arr]   (n2) -- (op1);
\draw[arr]   (n1) -- (op2);
\draw[arr]   (n2) -- (op2);
\draw[arr]  (op1) -- (n3);
\draw[arr]  (op2) -- (n4);






\end{tikzpicture}

%% file: graph_2.tex
\begin{tikzpicture}[
node distance = 5mm and 7mm,
   arr/.style = {-Triangle},
   box/.style = {rounded rectangle, draw, semithick,
                 minimum size=9mm},
    boxA/.style = {rounded rectangle, draw, semithick, minimum size=9mm, color=C0, fill=C0!35},
    boxAA/.style = {rounded rectangle, draw, double, ultra thick, minimum size=9mm, color=C0, fill=C0!35},
    boxB/.style = {rounded rectangle, draw, semithick, minimum size=9mm, color=C0, fill=C0!35},
    op/.style = {rectangle, draw, semithick,
                 color=C1, fill=C1!35},
                        ]
\node (n1) [boxA] {\textcolor{black}{${\bflambda}$}};
\node (n2) [boxA, below=of n1] {\textcolor{black}{$\widehat{{\bflambda}}$}};
\node (n3) [boxA, right=of n1] {{\textcolor{black}{${\bfalpha}$}}};
\node (n4) [boxB, right=of n2] {{\textcolor{black}{$\widehat{{\bfalpha}}$}}};

\node (op3) [op, above right=of n3] {\textcolor{black}{
\begin{tabular}{c}
Eq.~\eqref{eq:sens:computation-graph}  or \\
Lemma~\ref{lem:blt-closed-form}
\end{tabular}
} };
\node (op4) [op, below right=of n4] {\textcolor{black}{
\begin{tabular}{c}
Eq.~\eqref{eq:err:computation-graph}  or \\
Lemma~\ref{lem:blt-closed-form}
\end{tabular}
} };

\node (n5) [boxA, right=of op3] {\textcolor{black}{
\begin{tabular}{c} 
Sensitivity $\colnorm{\strategy}$ \\ or $\sens(\strategy, \Pi^\MinSep_{\minsep, \maxpart})$
\end{tabular}
}};
\node (n6) [boxB] at (n5|- op4) {\textcolor{black}{
\begin{tabular}{c}
Error $\rownorm{\workload\strategy^{-1}}$ \\ 
or $\lfrob{\workload\strategy^{-1}}$
\end{tabular}
} };


\draw[arr] (n1.north) |- (op3.west) ;
\draw[arr] (n3.east) -| (op3.south) ;
\draw[arr] (n2.south) |- (op4.west) ;
\draw[arr] (n4.east) -| (op4.north) ;

\draw[arr] (op3) -- (n5);
\draw[arr] (op4) -- (n6);

\node (op5) [op, xshift=3cm] at ($(n5)!0.5!(n6)$) {\Large \textcolor{black}{$\times$} };

\draw[arr] (n5) -| (op5);
\draw[arr] (n6) -| (op5);

\node (n7) [boxA, right=of op5, xshift=0.1cm] {\textcolor{black}{
\begin{tabular}{c}
\textit{\textbf{Objective of}} \\
\textit{\textbf{\Cref{prob:param-opt}}}
\end{tabular}
} };

\draw[arr] (op5) -- (n7) ;
\end{tikzpicture}

%% file: recommendations.tex
\begin{tikzpicture}[
node distance = 5mm and 7mm,
box1/.style={rectangle, semithick, minimum size=1.25cm},
box2/.style={rectangle, semithick, minimum height=4cm},
boxWL/.style = {box1, fill=gray!25},
boxErr/.style = {box1, fill=C0!25},
boxSens/.style = {box1, fill=C3!25},
boxMech/.style = {box1, fill=C2!25},
boxOpt/.style = {box1, fill=C1!25},
boxWL2/.style = {box2, fill=gray!25},
boxErr2/.style = {box2, fill=C0!25},
boxSens2/.style = {box2, fill=C3!25},
boxMech2/.style = {box2, fill=C2!25},
boxOpt2/.style = {box2, fill=C1!25},
]

\node at (29, 2) {\Large\underline{\textbf{Finding a Correlated Noise Mechanism}}:};

\node[boxOpt] (min) at (25, 0) {\Large $\underset{\bm{B}, \bm{C}} {\mathrm{minimize}}$};
\node[boxErr, right=of min, xshift=3mm] (err) {\Large \,$\mathsf{Error}(\bm{B})$};
\node[right=of err, xshift=-7mm] (times) {\Large $\times$};
\node[boxSens,right= of times, xshift=-7mm] (sens) {\Large $\mathsf{Sensitivity}(\bm{C})$};
\node[below=of min] (st) {\Large $\mathrm{subject\,\, to}$};
\node[boxWL] (wl) at (err|- st) {\Large $\bm{B} \bm{C} = \bm{A}$};
\node[right=of wl, xshift=-5mm] (Andbox) {\Large and};
\node[right=of sens, xshift=2cm] {\Large $(*)$};
\node[boxMech, below=of wl, xshift=19.5mm] (constr) {\Large $\bm{C}$ satisfies some constraints};

\node[below=of constr, xshift=-1.5cm,yshift=-0.5cm] {\Large\underline{\textbf{Practical Recommendations}}:};

\node[boxWL2, below left=of constr, yshift=-1.6cm, xshift=-1cm] (wlbox)
{
\renewcommand{\arraystretch}{1.4}
\begin{tabular}{c}
\textbf{Workload} (Ch. \ref{chap:intro}, \ref{chap:ml}) \\
\midrule 
Use $\workload=\prefix$ \\
(irrespective of base optimizer) \\
\end{tabular}
};

\node[boxErr2, right=of wlbox] (errbox)
{
\renewcommand{\arraystretch}{1.4}
\begin{tabular}{c}
\textbf{Error/Utility} (Ch. \ref{chap:prefixsum}) \\
\midrule 
Use max loss \\
or RMS loss \\
(not a critical choice)
\end{tabular}
};

\node[boxSens2, right=of errbox] (sensbox)
{
\renewcommand{\arraystretch}{1.4}
\begin{tabular}{c}
\textbf{Participation Pattern} (Ch. \ref{chap:ml}) \\
\midrule 
Block-Cyclic Poisson Sampling (centralized, high privacy) \\
Cyclic Order (centralized, low privacy) \\
Min-Sep (federated learning)
\end{tabular}
};

\node[boxMech2, yshift=-5cm, xshift=-0.5cm] (mechbox) at ($(wlbox)!0.5!(errbox)$)
{
\renewcommand{\arraystretch}{1.4}
\begin{tabular}{c}
 
\textbf{Mechanism Constraints} (Ch. \ref{chap:prefixsum}) \\
\midrule 
Use \\
Banded Toeplitz (centralized) or \\
BLT (federated)
\end{tabular}
};

\node[boxOpt2, yshift=-5cm, xshift=1.5cm] (optbox) at ($(errbox)!0.5!(sensbox)$)
{
\renewcommand{\arraystretch}{1.4}
\begin{tabular}{c}

\textbf{Mechanism Optimization} (Ch. \ref{chap:practical}) \\
\midrule 
Non-convex optimization with \\
Gradient descent \\
or Quasi-Newton (L-BFGS)
\end{tabular}
};

\end{tikzpicture}

%% file: 5_open_new.tex
This monologue has explored the landscape of correlated noise mechanisms for privacy preserving machine learning, providing a structured overview of their theoretical foundations, practical applications, and privacy implications. Looking ahead, correlated noise mechanisms offer a promising frontier in privacy-preserving data analysis. As data privacy continues to grow in importance, correlated noise mechanisms will likely play a critical role in balancing the need for data utility with the imperative for individual privacy.

In the rest of this section, we explore some of the important open problems. 

\section{Client Participation in Weighted Prefix Sums}\label{sec:codesign_mech_and_sampling}
In practice and as shown in \Cref{chap:ml} and \Cref{chap:practical}, private learning has two primary sources of stochasticity: (1) \emph{client participation}, which arises through random sampling, system-driven availability, or shuffling; and (2) \emph{noise added to ensure privacy}, such as correlated noise, i.i.d. noise, etc. 

Most existing approaches in designing correlated noise mechanisms treat these two components somewhat independently. In particular, \textit{participation} is typically handled through heuristics such as defining participation schema or uniform random sampling, without regard to the implications for noise design. Conversely, correlated noise mechanisms are often built assuming idealized participation patterns, like $b$-MinSep, etc. This separation leads to inefficiencies: either the variance-reduction benefits of correlation are lost, or the system must be artificially constrained to preserve fragile noise structures. While the latter can be enforced, it makes the architecture a little more complicated than desired. 

This raises a natural question: 
\begin{bquestion}
{Can we jointly design client participation strategies and correlated noise mechanisms to exploit their interaction for improved privacy-utility tradeoffs in the computation of weighted prefix sums?}
\end{bquestion}

\section{Defining Factorization Losses that Reflect Learning Performance}\label{sec:betterProxyLoss}
The use of linearly-correlated noise mechanisms in (weighted) prefix sum estimation is well-understood, especially for \txtmaxloss (\Cref{def:max-loss}) or \txtrmsloss (\Cref{def:rms-loss}). In these two cases, the mechanism design reduces to a tractable optimization problem (more precisely, a semi-definite program). However, extending these methods to learning problems presents few challenges. 

Currently, in all the works that uses correlated noise mechanisms for learning (see \Cref{sec:ch3-biblio} for references), proxy losses are introduced that are typically derived from {a related} linear estimation problem with the \textit{hope} that minimizing them leads to better learning performance. However, how close these proxy loss are in theoretical and empirical terms to the true learning dynamics remains an unresolved open question. In particular, while these proxies often serve as upper bounds on the learning error, whether they faithfully capture the behavior of real-world models, particularly across diverse function classes and worst-case dynamics, is still unclear.

 Recent work 
 has begun to investigate this gap between proxy losses and actual learning performance. For example, \citet*{koloskova2023correlated} showed that there was not a monotonic relationship between losses imposed on typical factorization problems and learning performance, even for simple models. They proposed alternative analyses and adjusted loss functions that better track true performance. As discussed in \Cref{sec:detailed-learning-bounds}, \citet*{choquette2023correlated}  have provided an asymptotic characterization of performance under quadratic losses, relating them directly to properties of the noise design. These insights suggest that more faithful modeling of noise-influenced learning dynamics, especially using the linearly correlated noise mechanism, could guide improved mechanism design. By developing loss measures that more accurately reflect real learning behavior, one can hope to further improve privacy-preserving training algorithms.

An intriguing direction is to view noise mechanism design through the lens of learned optimizers. Much like learning an optimizer that performs well across tasks, we can interpret the design of noise mechanisms (e.g., DP-SGD with correlated noise) as an optimization problem over a space of algorithmic parameters (such as noise covariance structures). This analogy points to a promising opportunity: designing losses and optimization targets that better capture learning performance across tasks, without inheriting the challenges of overfitting faced by learned optimizers. By exploring richer loss parameterizations (e.g., quadratic or convex losses), one can aim to build a more robust and expressive framework for noise mechanism design. 

\begin{bquestion}
    How can we design proxy loss functions for linearly-correlated noise mechanisms that more faithfully capture and improve actual learning performance across a broad class of machine learning problems?
\end{bquestion}

Solving this research question could ultimately lead to more effective and generalizable private learning algorithms, grounded in a better theoretical understanding of the connection between proxy objectives and learning performance.

\section{Adaptive Private Optimization with Correlated Noise Mechanisms}
\label{sec:ch5-adaptiveOptimizer}
The idea of casting learning algorithm design as an optimization problem over algorithmic spaces naturally leads to adaptive optimization methods such as AdaGrad, Adam, and RMSProp. These optimizers were originally motivated by meta-optimization or regret minimization frameworks, and remain central to training large-scale models like Transformer-based architectures, which benefit from per-parameter adaptive learning rates. Despite their widespread use and perceived necessity in modern machine learning workflows, integrating differentially private mechanism design, especially correlated noise mechanisms, into adaptive optimization remains an open challenge.

At the heart of this challenge lies the fundamental non-linearity introduced by adaptive optimizers: they typically perform pointwise division of gradients by functions of past gradients, breaking the linear stream-to-stream relationship between gradients and model updates that the design of correlated noise mechanisms relies on. This linearity is critical, as it ensures that loss functions (such as those quantifying the distance between noised and unnoised training trajectories) are independent of the data itself. Once this assumption fails, the proxy loss used in the optimization problem (e.g., \cref{eq:mainProblem}) becomes data-dependent, complicating both theoretical analysis and practical implementation. 

\begin{bquestion}
    How can linearly-correlated noise mechanisms be effectively extended to support adaptive optimization methods, either through novel regret-based analytical formulations or learned-optimizer-style numerical solutions, while preserving privacy guarantees and achieving competitive learning performance?
\end{bquestion}

\section{Stochastic Convex Optimization and Correlated Noise Mechanisms}
A significant open question in differentially private optimization is whether the \textit{stochastic convex optimization} (SCO) guarantees of correlated noise DP-SGD 
in the general setting can match the minimax optimal bounds known for the realizable regime. As discussed in \Cref{sec:detailed-learning-bounds}, \citet*{choquette2024optimal} and \citet*{zhang2022differentially} have shown promising advances in this direction by modifying the correlated noise DP-SGD to operate over sequences of \emph{gradient differences} instead of raw gradients. This reformulation enables the algorithm to achieve the optimal error rates for DP-SCO previously believed to require multi-pass training or more complex mechanisms.

In particular, these works demonstrate that not only is it possible to attain the optimal DP-SCO guarantee using a \emph{single-epoch} algorithm, but that this can be done with batch sizes as large as $\Omega(\sqrt{n})$. This is a notable improvement over standard instantiations of correlated noise DP-SGD, such as in \Cref{sec:dpftrl-name-background}, where each update is based on a batch of size one. The ability to scale batch sizes in this way is critical for practical applications and also reduces the number of noise-injected updates required during training.

A crucial component enabling this improvement is the use of correlated noise mechanisms applied to the sequence of gradient differences. Bounding the noise injected in this transformed domain becomes more tractable and can be done in a way that aligns with optimal privacy-utility tradeoffs. This innovation suggests that revisiting classic DP optimization pipelines with alternative sequence parameterizations may unlock further improvements.  

Difference estimator has been used a lot in statistics and streaming algorithms. For example, using difference estimator has been used in the streaming algorithm literature to design more efficient robust streaming algorithms~\citep{woodruff2022tight}. In the context of differential privacy, by observing the difference sequence has bounded sensitivity, \citet{fichtenberger2021differentially} and \citet{song2018differentially} gave a differentially private algorithm for various graph statistics in the continual release model. However, application of these techniques are limited. This raises the following question:

\begin{bquestion}
    Can the transformation to gradient-difference sequences be generalized beyond DP-SGD with correlated noise mechanism to  differentially private learning algorithms for other statistics or settings (like low-space privacy preserving algorithms or continual release algorithms for other statistics and settings), and under what conditions does this transformation preserve or improve privacy-utility tradeoffs? 
   \end{bquestion}

\section{Better Instantiation of DP-SGD with Correlated Noise}
\label{sec:ch5-surrogateLoss}
Extending the question of \cref{sec:codesign_mech_and_sampling}, and bringing together \cref{sec:betterProxyLoss,sec:ch5-adaptiveOptimizer}, the bigger goal is to holistically design private training methods.

DP-SGD and its variant algorithms, such as the one outlined in \cref{alg:dpsgd-corr}, present a flexible template for private model training. However, to turn this into a fully specified and high-performing learning algorithm, several critical design choices must be addressed as discussed in \Cref{chap:ml} and \Cref{chap:practical}:
\begin{enumerate} 
	\item the data processing pattern---particularly the participation schema and its impact on privacy amplification, as seen in techniques like block Poisson sampling; 
	\item the structure of the strategy matrix $\strategy$ (or equivalently the noise-correlating matrix $\Cinv$), which governs how noise is injected and correlated across iterations; and 
	\item the choice of the first-order optimizer, which translates noisy gradient estimates into model updates. 
	
\end{enumerate} 

As we saw in \Cref{chap:ml} and \Cref{chap:practical}, each of these components interacts in subtle and sometimes nontrivial ways. This suggests that a piecemeal design approach may leave substantial performance gains that are not fully realized, and, hence, a lot of effort has been dedicated to understand these subtle interactions. For example, a recent thrust in the literature of privacy preserving learning has been motivated by the thesis that a \emph{co-design} approach is necessary, one that jointly optimizes the data processing scheme, noise correlation strategy, and learning algorithm. 
For instance, the structured, banded design of $\strategy$ is essential for theoretical guarantees under block-cyclic Poisson sampling, while the effectiveness of a proxy loss function (e.g., \txtrmsloss\ with respect to a prefix-sum workload matrix) likely depends on both the strategy matrix and the optimizer used. 

Such interdependence suggests that better proxy losses, e.g. losses that are tailored to specific optimizer dynamics and workload structures, could drive more effective noise mechanism design. Furthermore, simple instantiations of DP-SGD may not suffice when moving toward adaptive optimization methods, which introduce additional complexity and opportunities for tuning noise injection and update behavior.

\begin{bquestion}
    How can we develop a unified framework for the co-design of data sampling schemes, noise correlation strategies, and optimizers in DP-SGD, to achieve improved learning performance under differential privacy constraints---especially in the presence of adaptive optimization methods? 
\end{bquestion}

\section{Resolving Conjecture~\ref{conj:equal-participation-sensitivity}}
In designing optimal correlated noise mechanisms under a given participation schema $\Pi$, a central step involves solving an optimization problem over positive definite matrices $\bfM$, subject to sensitivity constraints. In analogy to the streaming setting, we consider a scale-invariant objective and normalize sensitivity via the constraint $\sum_{t \in \bfpi} \bfM[t, t] \le 1$ for each $\bfpi \in \Pi$. In practice, however, this inequality is commonly replaced with an equality: $\sum_{t \in \bfpi} \bfM[t, t] = 1$, which simplifies analysis and implementation.

While empirical evidence strongly supports that this substitution preserves optimality, no theoretical guarantee currently exists for the general case. Notably, in the streaming setting, this replacement is provably without loss of generality. Whether the same holds in the multi-epoch setting remains an open question. If the equality-constrained problem indeed yields an optimal solution for the original inequality-constrained formulation, it would justify current practices and simplify both the analysis and deployment of such mechanisms.

\begin{bquestion}
    {Does the equality-constrained problem in \Cref{prob:multiepochopt} yield a solution that is also optimal for the corresponding inequality-constrained version? In particular, under what conditions on the participation schema $\Pi$ and workload matrix $\workload$ does the solution $\bfM_\star$ to the equality-constrained formulation remain valid for the relaxed inequality-constrained problem?}
\end{bquestion}

As discussed in  \Cref{chap:practical}, resolving the above question would not only simplify the problem of multi-participation, but also improve the convergence rate of the optimization algorithms. 

\section{Correlated Noise Mechanisms for Streaming Prefix Sums}\label{sec:open_theory}
In this monograph, we primarily focused on one central application of differentially private prefix sum estimation: the training of machine learning models, particularly via private variants of stochastic gradient descent. However, the prefix sum primitive has far-reaching applications beyond private learning. It serves as a foundational tool in numerous other domains, including but not limited to histogram estimation, range query answering, shortest-path estimation on structured graphs, non-interactive local learning, graph spectral analysis, and matrix computation. In these settings, accuracy is typically assessed through metrics such as the root mean square loss (\txtrmsloss) or the maximum error (\txtmaxloss) over the query outputs.

For the canonical prefix sum workload matrix $\workload = \prefix$, strong theoretical foundations exist. In particular, tight upper and lower bounds on the \txtrmsloss are known not only in asymptotic terms but also up to constants. More generally, \citet*{liu2024optimality} showed that, for any {positive} constant $p$, we have asymptotically tight characterizations of the $\ell_p$ error defined as
\[
\max_{\bfx \in \mathbb{R}^n} \left( \mathbb{E} \left[ \|\mech(\bfx) - \workload \bfx \|_p^p \right] \right)^{1/p},
\]
where $\mech$ is a differentially private mechanism. These bounds reflect a mature understanding of the trade-offs between privacy and accuracy under a range of loss functions.

In contrast, for the \txtmaxloss\ metric (or for $p=\omega(1))$, the picture is more nuanced. Using standard packing arguments, \citet*{dwork2010continual} showed that any $(1, o(1/\sqrt{n}))$-differentially private mechanism for the prefix sum problem must incur a \txtmaxloss\  error of at least $\Omega(\log(n)/\varepsilon)$ in the worst case. Recently, \citet*{cohen2024lower} showed that, when the input vector is $s$-sparse, this lower bound becomes $\Omega(\min\{\log(s), \log(n)\})$. On the other hand, the best known upper bound on \txtmaxloss\ for $(\varepsilon,\delta)$-differential private mechanism is $O\left( \log^{3/2}(n) \cdot \sqrt{\log(1/\delta)} / \varepsilon \right)$ for input streams bounded in $[-1, 1]$.

\begin{bquestion}
The gap between best known upper and lower bound on \txtmaxloss\ for differentially private prefix sum for $n$ updates is $\log^{1/2}(n)$ dependence. Can we close this gap? Further, what is the optimal accuracy guarantee if the stream is $s$-sparse with respect to the sparsity parameter.
\end{bquestion}

\section{Amplification of Correlated Noise Mechanism Under Sliding Window}
\label{sec:ch5-amplificationslidingwindow}
This monograph has explored the theory and practice of correlated noise mechanisms for differentially private learning, starting with their application to prefix sum estimation and culminating in their use for training large-scale models under privacy constraints. 

One compelling direction for future work, which naturally extends the models discussed in this monograph, is the \textbf{sliding window model}. In many modern applications of machine learning—especially in online learning, federated learning, or real-time analytics—\emph{more recent data are more relevant} to current predictions and model updates. In these settings, it may be desirable to estimate statistics (e.g., means, prefix sums, or gradients) over only the most recent $w$ observations, rather than over the full history. This calls for differentially private mechanisms tailored to \emph{sliding window estimation}, where the goal is to release private estimates of functions computed over a moving window of size $w$, rather than over an accumulating prefix.

From a privacy mechanism design standpoint, this introduces new challenges and opportunities. Mechanisms must now adapt not only to the prefix structure but also to \emph{temporal locality}—ensuring that noise is injected in a way that respects the decay in importance of older data. Correlated noise mechanisms, already well-suited to structured time dependencies, provide a promising starting point. For instance, designing strategy matrices $\strategy$ or workload matrices $\workload$ with banded or decaying influence patterns could yield effective mechanisms for sliding window analytics. Moreover, privacy accounting in the sliding window model may require novel analyses, as traditional notions of sensitivity and amplification must be redefined for moving-window adjacency relations.

\begin{bquestion}
    {How can we extend correlated noise mechanisms to the sliding window model in a way that reflects the heightened importance of recent data while maintaining rigorous privacy guarantees and high utility in private estimation and training?}
\end{bquestion}

This question opens a rich design space that intersects with time-series privacy, streaming algorithms, and adaptive learning. Addressing it could lead to more temporally aware privacy mechanisms—particularly relevant for real-time deployment scenarios where recent user interactions carry more signal than stale historical data.

\section{A Functional Analysis Perspective on Sensitivity}

The computation of sensitivity (e.g. defined in terms of participation schema by~\cref{def:sens-participation}) can be viewed as an operator norm with interesting structure.

\sloppy 
Most of the sensitivity calculations in~\Cref{chap:ml} factor through~\cref{lem:multi-epoch-sens:generic-bounds}, which can be significantly suboptimal in the presence of cancellation (e.g., via entries of mixed sign) in the matrix $\strategy^\top\strategy$. In fact, precise control of this operator-norm cancellation was a problem posed by Grothendieck in the 1950's, who showed a uniform bound in all dimensions:

\begin{btheorem}\label{thm:grothendieck}
Consider a matrix $\bfW \in \R^{n \times n}$ with entries $w_{ij} := \bfW[i, j]$ and suppose that it has unit $\ell_\infty$ to $\ell_1$ operator norm (defined in \Cref{eq:induced-norm}):
\[
\norm{\bfW}_{\infty \to 1} \le 1
\quad \iff \quad 
\left| \sum_{i, j=0}^{n-1} w_{ij} t_i s_j \right| \leq 1 \,\, \forall \,\, |t_i| \le 1, |s_j| \le 1 \,.
\]
Then, for every set of unit vectors $\{\bfx_i\}_{i=0}^{n-1}$, $\{\bfy_j\}_{j=0}^{n-1}$ in a Hilbert space $\mathbb{H}$ (endowed with an inner product $\langle \cdot, \cdot \rangle_{\mathbb{H}}$), we have the uniform bound
$$
\left| \sum_{i, j=0}^{n-1} w_{ij} \langle \bfx_i, \bfy_j\rangle_{\mathbb{H}} \right| \leq K,
$$
where $K$ is an absolute constant known as Grothendieck's constant.
\end{btheorem}

Grothendieck's constant $K$ is known to be between $1.57$ and $1.7822$ in general, though in the application of Grothendieck's theorem to calculation of sensitivity one may leverage symmetry to achieve the value $\pi / 2$. As has been noted previously, Grothendieck's theorem provides an alternate means of estimating the $\ell_2$ sensitivity of a strategy matrix $\strategy$, as one may rewrite the trace which appears in the proof of~\cref{lem:multi-epoch-sens:generic-bounds} into a form to which Grothendieck's theorem applies. 

If we restrict the matrices $\bfW$ in~\cref{thm:grothendieck}, $K$ may be significantly improved. For example, the arguments of~\cref{lem:multi-epoch-sens:generic-bounds} show that for elementwise nonnegative $\bfW$, one may take $K = 1$ (though this represents a tiny fraction of invertible matrices). In some sense, one may view the restrictions to nonnegative $\strategy^\top\strategy$ while optimizing for nontrivial $\Pi$ as primarily playing the role of \emph{avoiding} the need to pay the $\pi/2$ cost that would otherwise be implied by~\cref{thm:grothendieck}.

Whether optimal or near-optimal $\strategy$ satisfy $\strategy^\top\strategy \geq 0$ is in many cases not known, and is in principle be dependent on the workload matrix $\workload$. The presence of the Grothendieck constant in vector-to-scalar reductions is itself $\Pi$-dependent; this concern disappears in the streaming (single participation) setting. Thus the current state of theory relating $\workload$, $\Pi$, and structure in $\strategy$ is somewhat unsatisfying: we know that in many cases of interest, we may suppress the factor of $\pi/2$ by imposing appropriate structure on $\strategy$, but we do not in principle know what we are losing in the optimization problem by enforcing this restriction.

\begin{bquestion}
    {Are there settings of $\workload, \Pi$ pairs in which loosening structure on $\strategy$ permits significantly improved solutions? In the case of nontrivial $\Pi$, are there large classes of matrices for which $K$ can be lowered and yield useful extensions of the space of permissible $\strategy$ matrices? If not, why not?}
\end{bquestion}

Given the practical success of banded, Toeplitz, and BLT mechanisms for which \cref{lem:multi-epoch-sens:generic-bounds} applies, this question is perhaps slightly academic, but represents a tempting gap in our theoretical understanding with connections to a deep result in functional analysis.

\section{Characterizing the error of optimal BLTs}
\citet*{dvijotham2024efficient} showed that BLTs constructed via rational function approximation can achieve (up to a small additive factor) the optimal \txtmaxloss of $\log(\nd)/\pi$ using $\nbuf = \Theta(\log^2 \nd)$ buffers, using the same BLT (for a given $\nbuf$) for all possible $\nd$. However, direct numerical optimization of the BLT parameters for a specific $\nd$ (that is, the approach of \cref{sec:optimizeblts}) yields substantially better mechanisms. This naturally yields the following questions:
\begin{bquestion}
Can the BLT parameters that minimize \txtmaxloss for a specific $\nd$ and specific number of buffers $\nbuf$ be characterized in closed form (without numerical optimization)? Can a tight bound on \txtmaxloss be given for these parameters? And most importantly, how many buffers $\nbuf$ are necessary to achieve the optimal rate of $\log(\nd)/\pi$? Empirically evidence leads us to conjecture $\nbuf = \Theta(\log \nd)$ is sufficient. 
\end{bquestion}

%% file: main.bbl
\begin{thebibliography}{113}
\providecommand{\natexlab}[1]{#1}
\providecommand{\url}[1]{\texttt{#1}}
\expandafter\ifx\csname urlstyle\endcsname\relax
  \providecommand{\doi}[1]{doi: #1}\else
  \providecommand{\doi}{doi: \begingroup \urlstyle{rm}\Url}\fi

\bibitem[Abadi et~al.(2016)Abadi, Chu, Goodfellow, McMahan, Mironov, Talwar,
  and Zhang]{abadi2016deep}
Martin Abadi, Andy Chu, Ian Goodfellow, H~Brendan McMahan, Ilya Mironov, Kunal
  Talwar, and Li~Zhang.
\newblock {Deep Learning with Differential Privacy}.
\newblock In \emph{Proceedings of the 2016 ACM SIGSAC conference on computer
  and communications security}, pages 308--318, 2016.

\bibitem[Andersson and Pagh(2023)]{andersson2023smooth}
Joel~Daniel Andersson and Rasmus Pagh.
\newblock {A Smooth Binary Mechanism for Efficient Private Continual
  Observation}.
\newblock \emph{Advances in Neural Information Processing Systems}, 36, 2023.

\bibitem[Annamalai et~al.(2024)Annamalai, Balle, De~Cristofaro, and
  Hayes]{annamalai2024shuffle}
Meenatchi Sundaram Muthu~Selva Annamalai, Borja Balle, Emiliano De~Cristofaro,
  and Jamie Hayes.
\newblock To shuffle or not to shuffle: Auditing dp-sgd with shuffling.
\newblock \emph{arXiv preprint arXiv:2411.10614}, 2024.

\bibitem[Apple(2017)]{appledp}
{Differential Privacy Team} Apple.
\newblock {Learning with Privacy at Scale}.
\newblock 2017.

\bibitem[Asi et~al.(2021{\natexlab{a}})Asi, Feldman, Koren, and
  Talwar]{asi2021private}
Hilal Asi, Vitaly Feldman, Tomer Koren, and Kunal Talwar.
\newblock Private stochastic convex optimization: Optimal rates in l1 geometry.
\newblock In \emph{International Conference on Machine Learning}, pages
  393--403. PMLR, 2021{\natexlab{a}}.

\bibitem[Asi et~al.(2021{\natexlab{b}})Asi, Levy, and Duchi]{asi2021adapting}
Hilal Asi, Daniel Asher~Nathan Levy, and John Duchi.
\newblock Adapting to function difficulty and growth conditions in private
  optimization.
\newblock In \emph{Advances in Neural Information Processing Systems},
  2021{\natexlab{b}}.

\bibitem[Asi et~al.(2023)Asi, Feldman, Koren, and Talwar]{asi2023near}
Hilal Asi, Vitaly Feldman, Tomer Koren, and Kunal Talwar.
\newblock Near-optimal algorithms for private online optimization in the
  realizable regime.
\newblock In \emph{International Conference on Machine Learning}, pages
  1107--1120. PMLR, 2023.

\bibitem[Baker~Jr and Gammel(1961)]{baker1961pade}
George~A Baker~Jr and John~L Gammel.
\newblock The {P}ad{\'e} {A}pproximant.
\newblock \emph{Journal of Mathematical Analysis and Applications}, 2\penalty0
  (1):\penalty0 21--30, 1961.

\bibitem[Bassily et~al.(2014{\natexlab{a}})Bassily, Smith, and Thakurta]{BST14}
Raef Bassily, Adam Smith, and Abhradeep Thakurta.
\newblock Private empirical risk minimization: Efficient algorithms and tight
  error bounds.
\newblock In \emph{Proc. of the 2014 IEEE 55th Annual Symp. on Foundations of
  Computer Science (FOCS)}, pages 464--473, 2014{\natexlab{a}}.

\bibitem[Bassily et~al.(2014{\natexlab{b}})Bassily, Smith, and
  Thakurta]{bassily2014private}
Raef Bassily, Adam Smith, and Abhradeep Thakurta.
\newblock Private empirical risk minimization: Efficient algorithms and tight
  error bounds.
\newblock In \emph{2014 IEEE 55th annual symposium on foundations of computer
  science}, pages 464--473. IEEE, 2014{\natexlab{b}}.

\bibitem[Bassily et~al.(2019)Bassily, Feldman, Talwar, and
  Thakurta]{bassily2019private}
Raef Bassily, Vitaly Feldman, Kunal Talwar, and Abhradeep Thakurta.
\newblock Private stochastic convex optimization with optimal rates.
\newblock In \emph{Advances in Neural Information Processing Systems}, pages
  11279--11288, 2019.

\bibitem[Bassily et~al.(2020)Bassily, Feldman, Guzm{\'a}n, and
  Talwar]{bassily2020stability}
Raef Bassily, Vitaly Feldman, Crist{\'o}bal Guzm{\'a}n, and Kunal Talwar.
\newblock Stability of stochastic gradient descent on nonsmooth convex losses.
\newblock \emph{arXiv preprint arXiv:2006.06914}, 2020.

\bibitem[Bassily et~al.(2021)Bassily, Guzm{\'a}n, and Nandi]{bassily2021non}
Raef Bassily, Crist{\'o}bal Guzm{\'a}n, and Anupama Nandi.
\newblock Non-euclidean differentially private stochastic convex optimization.
\newblock In \emph{Conference on Learning Theory}, pages 474--499. PMLR, 2021.

\bibitem[Bennett(1977)]{bennett1977schur}
Grahame Bennett.
\newblock Schur multipliers.
\newblock 1977.

\bibitem[Bolot et~al.(2013)Bolot, Fawaz, Muthukrishnan, Nikolov, and
  Taft]{bolot2013private}
Jean Bolot, Nadia Fawaz, Shanmugavelayutham Muthukrishnan, Aleksandar Nikolov,
  and Nina Taft.
\newblock Private decayed predicate sums on streams.
\newblock In \emph{Proceedings of the 16th International Conference on Database
  Theory}, pages 284--295, 2013.

\bibitem[Braess and Hackbusch(2005)]{braess2005approximation}
Dietrich Braess and Wolfgang Hackbusch.
\newblock Approximation of $1/x$ by exponential sums in $[1,\infty)$.
\newblock \emph{IMA journal of numerical analysis}, 25\penalty0 (4):\penalty0
  685--697, 2005.

\bibitem[Bun and Steinke(2016)]{bun16concentrated}
Mark Bun and Thomas Steinke.
\newblock Concentrated differential privacy: Simplifications, extensions, and
  lower bounds.
\newblock In Martin Hirt and Adam Smith, editors, \emph{Theory of
  Cryptography}, pages 635--658, Berlin, Heidelberg, 2016. Springer Berlin
  Heidelberg.
\newblock ISBN 978-3-662-53641-4.

\bibitem[Chan et~al.(2011)Chan, Shi, and Song]{chan2011private}
T.{-}H.~Hubert Chan, Elaine Shi, and Dawn Song.
\newblock {Private and Continual Release of Statistics}.
\newblock \emph{{ACM} Trans. Inf. Syst. Secur.}, 14\penalty0 (3):\penalty0
  26:1--26:24, 2011.

\bibitem[Charles et~al.(2025)Charles, Ganesh, McKenna, McMahan, Mitchell,
  Pillutla, and Rush]{charles2024fine}
Zachary Charles, Arun Ganesh, Ryan McKenna, H~Brendan McMahan, Nicole Mitchell,
  Krishna Pillutla, and Keith Rush.
\newblock {Fine-Tuning Large Language Models with User-Level Differential
  Privacy}.
\newblock In \emph{SaTML}, 2025.

\bibitem[Chaudhuri et~al.(2011)Chaudhuri, Monteleoni, and
  Sarwate]{chaudhuri2011differentially}
Kamalika Chaudhuri, Claire Monteleoni, and Anand~D Sarwate.
\newblock Differentially private empirical risk minimization.
\newblock \emph{Journal of Machine Learning Research}, 12\penalty0
  (Mar):\penalty0 1069--1109, 2011.

\bibitem[Chen and Qi(2005)]{chen2005best}
Chao-Ping Chen and Feng Qi.
\newblock The best bounds in wallis' inequality.
\newblock \emph{Proceedings of the American Mathematical Society}, 133\penalty0
  (2):\penalty0 397--401, 2005.

\bibitem[Chen et~al.(2020)Chen, Wu, and Hong]{chen2020understanding}
Xiangyi Chen, Zhiwei~Steven Wu, and Mingyi Hong.
\newblock {Understanding Gradient Clipping in Private SGD: A Geometric
  Perspective}.
\newblock In \emph{NeurIPS}, 2020.

\bibitem[Choquette-Choo et~al.(2023{\natexlab{a}})Choquette-Choo, Ganesh,
  McKenna, McMahan, Rush, Guha~Thakurta, and Xu]{choquette2024amplified}
Christopher~A Choquette-Choo, Arun Ganesh, Ryan McKenna, H~Brendan McMahan,
  John Rush, Abhradeep Guha~Thakurta, and Zheng Xu.
\newblock {(Amplified) Banded Matrix Factorization: A unified approach to
  private training}.
\newblock \emph{Advances in Neural Information Processing Systems}, 36,
  2023{\natexlab{a}}.

\bibitem[Choquette-Choo et~al.(2023{\natexlab{b}})Choquette-Choo, McMahan,
  Rush, and Thakurta]{choquette2022multi}
Christopher~A Choquette-Choo, H~Brendan McMahan, Keith Rush, and Abhradeep
  Thakurta.
\newblock {Multi-Epoch Matrix Factorization Mechanisms for Private Machine
  Learning}.
\newblock In \emph{International Conference on Machine Learning}, volume 202,
  pages 5924--5963. {PMLR}, 2023{\natexlab{b}}.

\bibitem[Choquette-Choo et~al.(2024{\natexlab{a}})Choquette-Choo, Dvijotham,
  Pillutla, Ganesh, Steinke, and Thakurta]{choquette2023correlated}
Christopher~A Choquette-Choo, Krishnamurthy Dvijotham, Krishna Pillutla, Arun
  Ganesh, Thomas Steinke, and Abhradeep Thakurta.
\newblock {Correlated Noise Provably Beats Independent Noise for Differentially
  Private Learning}.
\newblock In \emph{ICLR}, 2024{\natexlab{a}}.

\bibitem[Choquette-Choo et~al.(2024{\natexlab{b}})Choquette-Choo, Ganesh,
  Haque, Steinke, and Thakurta]{choquette2024near}
Christopher~A Choquette-Choo, Arun Ganesh, Saminul Haque, Thomas Steinke, and
  Abhradeep Thakurta.
\newblock {Near Exact Privacy Amplification for Matrix Mechanisms}.
\newblock \emph{arXiv Preprint}, 2024{\natexlab{b}}.

\bibitem[Choquette{-}Choo et~al.(2024)Choquette{-}Choo, Ganesh, Steinke, and
  Thakurta]{choquette2024privacyamplif}
Christopher~A. Choquette{-}Choo, Arun Ganesh, Thomas Steinke, and
  Abhradeep~Guha Thakurta.
\newblock {Privacy Amplification for Matrix Mechanisms}.
\newblock In \emph{International Conference on Learning Representations}, 2024.

\bibitem[Choquette-Choo et~al.(2024)Choquette-Choo, Ganesh, and
  Thakurta]{choquette2024optimal}
Christopher~A Choquette-Choo, Arun Ganesh, and Abhradeep Thakurta.
\newblock Optimal rates for dp-sco with a single epoch and large batches.
\newblock \emph{arXiv preprint arXiv:2406.02716}, 2024.

\bibitem[Chua et~al.(2024{\natexlab{a}})Chua, Ghazi, Kamath, Kumar, Manurangsi,
  Sinha, and Zhang]{chua2024private}
Lynn Chua, Badih Ghazi, Pritish Kamath, Ravi Kumar, Pasin Manurangsi, Amer
  Sinha, and Chiyuan Zhang.
\newblock How private are dp-sgd implementations?
\newblock \emph{arXiv preprint arXiv:2403.17673}, 2024{\natexlab{a}}.

\bibitem[Chua et~al.(2024{\natexlab{b}})Chua, Ghazi, Kamath, Kumar, Manurangsi,
  Sinha, and Zhang]{chua2024scalable}
Lynn Chua, Badih Ghazi, Pritish Kamath, Ravi Kumar, Pasin Manurangsi, Amer
  Sinha, and Chiyuan Zhang.
\newblock Scalable dp-sgd: Shuffling vs. poisson subsampling.
\newblock \emph{Advances in Neural Information Processing Systems},
  37:\penalty0 70026--70047, 2024{\natexlab{b}}.

\bibitem[Cohen et~al.(2024)Cohen, Lyu, Nelson, Sarl{\'o}s, and
  Stemmer]{cohen2024lower}
Edith Cohen, Xin Lyu, Jelani Nelson, Tam{\'a}s Sarl{\'o}s, and Uri Stemmer.
\newblock Lower bounds for differential privacy under continual observation and
  online threshold queries.
\newblock In \emph{The Thirty Seventh Annual Conference on Learning Theory},
  pages 1200--1222. PMLR, 2024.

\bibitem[Corless et~al.(2011)Corless, Ida, and Hong]{corless2011concrete}
Robert Corless, Tetsuo Ida, and Hoon Hong.
\newblock \emph{{The Concrete Tetrahedron: Symbolic Sums, Recurrence Equations,
  Generating Functions, Asymptotic Estimates}}.
\newblock Springer, 2011.

\bibitem[Cormode et~al.(2019)Cormode, Kulkarni, and
  Srivastava]{cormode2019answering}
Graham Cormode, Tejas Kulkarni, and Divesh Srivastava.
\newblock {Answering Range Queries Under Local Differential Privacy}.
\newblock \emph{VLDB}, 12\penalty0 (10):\penalty0 1126--1138, 2019.

\bibitem[De et~al.(2022)De, Berrada, Hayes, Smith, and Balle]{de2022unlocking}
Soham De, Leonard Berrada, Jamie Hayes, Samuel~L. Smith, and Borja Balle.
\newblock {Unlocking High-Accuracy Differentially Private Image Classification
  through Scale}.
\newblock \emph{arXiv Preprint}, 2022.

\bibitem[Denisov et~al.(2022)Denisov, McMahan, Rush, Smith, and
  Guha~Thakurta]{denisov2022improved}
Sergey Denisov, H~Brendan McMahan, John Rush, Adam Smith, and Abhradeep
  Guha~Thakurta.
\newblock {Improved Differential Privacy for SGD via Optimal Private Linear
  Operators on Adaptive Streams}.
\newblock In \emph{Advances in Neural Information Processing Systems},
  volume~35, pages 5910--5924, 2022.

\bibitem[Ding et~al.(2011)Ding, Winslett, Han, and Li]{ding2011differentially}
Bolin Ding, Marianne Winslett, Jiawei Han, and Zhenhui Li.
\newblock Differentially private data cubes: optimizing noise sources and
  consistency.
\newblock In \emph{Proceedings of the 2011 ACM SIGMOD International Conference
  on Management of data}, pages 217--228, 2011.

\bibitem[Dong et~al.(2022)Dong, Roth, and Su]{dong22gaussian}
Jinshuo Dong, Aaron Roth, and Weijie~J. Su.
\newblock {Gaussian Differential Privacy}.
\newblock \emph{Journal of the Royal Statistical Society Series B: Statistical
  Methodology}, 84\penalty0 (1):\penalty0 3--37, 02 2022.
\newblock ISSN 1369-7412.
\newblock \doi{10.1111/rssb.12454}.
\newblock URL \url{https://doi.org/10.1111/rssb.12454}.

\bibitem[Duchi et~al.(2011)Duchi, Agarwal, and Wainwright]{duchi2011dual}
John~C Duchi, Alekh Agarwal, and Martin~J Wainwright.
\newblock Dual averaging for distributed optimization: Convergence analysis and
  network scaling.
\newblock \emph{IEEE Transactions on Automatic control}, 57\penalty0
  (3):\penalty0 592--606, 2011.

\bibitem[Dupr{\'e}~la Tour et~al.(2024)Dupr{\'e}~la Tour, Henzinger, and
  Saulpic]{Dupre2024}
Max Dupr{\'e}~la Tour, Monika Henzinger, and David Saulpic.
\newblock Making old things new: a unified algorithm for differentially private
  clustering.
\newblock In \emph{Proc. 41th {ICML}}, 2024.

\bibitem[Dvijotham et~al.(2024)Dvijotham, McMahan, Pillutla, Steinke, and
  Thakurta]{dvijotham2024efficient}
Krishnamurthy Dvijotham, H.~Brendan McMahan, Krishna Pillutla, Thomas Steinke,
  and Abhradeep Thakurta.
\newblock {Efficient and Near-Optimal Noise Generation for Streaming
  Differential Privacy}.
\newblock In \emph{IEEE Symposium on Foundations of Computer Science (FOCS)},
  2024.

\bibitem[Dwork and Rothblum(2016)]{dwork2016concentrated}
Cynthia Dwork and Guy~N Rothblum.
\newblock Concentrated differential privacy.
\newblock \emph{arXiv preprint arXiv:1603.01887}, 2016.

\bibitem[Dwork et~al.(2006)Dwork, Kenthapadi, McSherry, Mironov, and
  Naor]{dwork06our}
Cynthia Dwork, Krishnaram Kenthapadi, Frank McSherry, Ilya Mironov, and Moni
  Naor.
\newblock Our data, ourselves: Privacy via distributed noise generation.
\newblock In Serge Vaudenay, editor, \emph{Advances in Cryptology - EUROCRYPT
  2006}, pages 486--503, Berlin, Heidelberg, 2006. Springer Berlin Heidelberg.

\bibitem[Dwork et~al.(2010)Dwork, Naor, Pitassi, and
  Rothblum]{dwork2010continual}
Cynthia Dwork, Moni Naor, Toniann Pitassi, and Guy~N. Rothblum.
\newblock Differential privacy under continual observation.
\newblock In \emph{STOC}, pages 715--724, 2010.

\bibitem[Dwork et~al.(2016)Dwork, McSherry, Nissim, and
  Smith]{dwork2016calibrating}
Cynthia Dwork, Frank McSherry, Kobbi Nissim, and Adam Smith.
\newblock Calibrating noise to sensitivity in private data analysis.
\newblock \emph{Journal of Privacy and Confidentiality}, 7\penalty0
  (3):\penalty0 17--51, 2016.

\bibitem[Edmonds et~al.(2020)Edmonds, Nikolov, and Ullman]{edmonds2020power}
Alexander Edmonds, Aleksandar Nikolov, and Jonathan Ullman.
\newblock The power of factorization mechanisms in local and central
  differential privacy.
\newblock In \emph{Proceedings of the 52nd Annual ACM SIGACT Symposium on
  Theory of Computing}, pages 425--438, 2020.

\bibitem[Elliott(1953)]{elliott1953characteristic}
Joseph~Frederick Elliott.
\newblock The characteristic roots of certain real symmetric matrices.
\newblock 1953.

\bibitem[Erlingsson et~al.(2014)Erlingsson, Pihur, and
  Korolova]{erlingsson2014rappor}
{\'U}lfar Erlingsson, Vasyl Pihur, and Aleksandra Korolova.
\newblock {RAPPOR: Randomized Aggregatable Privacy-Preserving Ordinal
  Response}.
\newblock In \emph{ACM SIGSAC}, pages 1054--1067, 2014.

\bibitem[Feldman et~al.(2020)Feldman, Koren, and Talwar]{feldman2019private}
Vitaly Feldman, Tomer Koren, and Kunal Talwar.
\newblock Private stochastic convex optimization: Optimal rates in linear time.
\newblock In \emph{Proc. of the Fifty-Second {ACM} Symp. on Theory of Computing
  ({STOC}'20)}, 2020.

\bibitem[Fichtenberger et~al.(2021)Fichtenberger, Henzinger, and
  Ost]{fichtenberger2021differentially}
Hendrik Fichtenberger, Monika Henzinger, and Lara Ost.
\newblock Differentially private algorithms for graphs under continual
  observation.
\newblock \emph{arXiv preprint arXiv:2106.14756}, 2021.

\bibitem[Fichtenberger et~al.(2023)Fichtenberger, Henzinger, and
  Upadhyay]{fichtenberger2023constant}
Hendrik Fichtenberger, Monika Henzinger, and Jalaj Upadhyay.
\newblock {Constant Matters: Fine-grained Error Bound on Differentially Private
  Continual Observation}.
\newblock In \emph{International Conference on Machine Learning}, 2023.

\bibitem[Gopi et~al.(2022)Gopi, Lee, and Liu]{gopi2022private}
Sivakanth Gopi, Yin~Tat Lee, and Daogao Liu.
\newblock Private convex optimization via exponential mechanism.
\newblock \emph{arXiv preprint arXiv:2203.00263}, 2022.

\bibitem[Gregory and Karney(1969)]{gregory1969collection}
Robert~Todd Gregory and David~L Karney.
\newblock A collection of matrices for testing computational algorithms.
\newblock \emph{(No Title)}, 1969.

\bibitem[Grimmer(2022)]{grimmer2022induced}
Benjamin Grimmer.
\newblock {A Collection of Induced Norm Balls}.
\newblock 2022.
\newblock URL \url{https://www.ams.jhu.edu/~grimmer/Induced.pdf}.

\bibitem[Haagerup(1980)]{haagerup1980decomposition}
Uffe Haagerup.
\newblock Decomposition of completely bounded maps on operator algebras, 1980.

\bibitem[Haagerup and Pisier(1993)]{haagerup1993bounded}
Uffe Haagerup and Gilles Pisier.
\newblock Bounded linear operators between $c^*$-algebras.
\newblock \emph{Duke Mathematical Journal}, 71\penalty0 (3):\penalty0 889--925,
  1993.

\bibitem[Hay et~al.(2010)Hay, Rastogi, Miklau, and Suciu]{hay2010boosting}
Michael Hay, Vibhor Rastogi, Gerome Miklau, and Dan Suciu.
\newblock {Boosting the Accuracy of Differentially Private Histograms Through
  Consistency}.
\newblock \emph{VLDB}, 3\penalty0 (1):\penalty0 1021--1032, 2010.

\bibitem[Henzinger and Upadhyay(2025)]{henzinger2025improved}
Monika Henzinger and Jalaj Upadhyay.
\newblock {Improved Differentially Private Continual Observation Using Group
  Algebra}.
\newblock In \emph{Proceedings of the 2025 Annual ACM-SIAM Symposium on
  Discrete Algorithms (SODA)}. SIAM, 2025.

\bibitem[Henzinger et~al.(2023)Henzinger, Upadhyay, and
  Upadhyay]{henzinger2023almost}
Monika Henzinger, Jalaj Upadhyay, and Sarvagya Upadhyay.
\newblock {Almost Tight Error Bounds on Differentially Private Continual
  Counting}.
\newblock In \emph{Proceedings of the 2023 Annual ACM-SIAM Symposium on
  Discrete Algorithms (SODA)}, pages 5003--5039. SIAM, 2023.

\bibitem[Hoffman(1971)]{hoffman1971linear}
Kenneth Hoffman.
\newblock \emph{Linear algebra}.
\newblock 1971.

\bibitem[Honaker(2015)]{honaker2015efficient}
James Honaker.
\newblock Efficient use of differentially private binary trees.
\newblock \emph{Theory and Practice of Differential Privacy (TPDP 2015),
  London, UK}, 2:\penalty0 26--27, 2015.

\bibitem[Jain et~al.(2012)Jain, Kothari, and Thakurta]{jain2012differentially}
Prateek Jain, Pravesh Kothari, and Abhradeep Thakurta.
\newblock {Differentially Private Online Learning}.
\newblock In \emph{Conference on Learning Theory}, pages 24--1, 2012.

\bibitem[Kairouz et~al.(2021{\natexlab{a}})Kairouz, McMahan, Song, Thakkar,
  Thakurta, and Xu]{kairouz2021practical}
Peter Kairouz, Brendan McMahan, Shuang Song, Om~Thakkar, Abhradeep Thakurta,
  and Zheng Xu.
\newblock {Practical and Private (Deep) Learning without Sampling or
  Shuffling}.
\newblock In \emph{International Conference on Machine Learning}, pages
  5213--5225. PMLR, 2021{\natexlab{a}}.

\bibitem[Kairouz et~al.(2021{\natexlab{b}})Kairouz, McMahan, Avent, Bellet,
  Bennis, Bhagoji, Bonawitz, Charles, Cormode, Cummings, D'Oliveira, Eichner,
  Rouayheb, Evans, Gardner, Garrett, Gasc{\'{o}}n, Ghazi, Gibbons, Gruteser,
  Harchaoui, He, He, Huo, Hutchinson, Hsu, Jaggi, Javidi, Joshi, Khodak,
  Kone{\v{c}}n{\'y}, Korolova, Koushanfar, Koyejo, Lepoint, Liu, Mittal, Mohri,
  Nock, {\"{O}}zg{\"{u}}r, Pagh, Qi, Ramage, Raskar, Raykova, Song, Song,
  Stich, Sun, Suresh, Tram{\`{e}}r, Vepakomma, Wang, Xiong, Xu, Yang, Yu, Yu,
  and Zhao]{kairouz2019flsurvey}
Peter Kairouz, H.~Brendan McMahan, Brendan Avent, Aur{\'{e}}lien Bellet, Mehdi
  Bennis, Arjun~Nitin Bhagoji, Kallista~A. Bonawitz, Zachary Charles, Graham
  Cormode, Rachel Cummings, Rafael G.~L. D'Oliveira, Hubert Eichner, Salim~El
  Rouayheb, David Evans, Josh Gardner, Zachary Garrett, Adri{\`{a}}
  Gasc{\'{o}}n, Badih Ghazi, Phillip~B. Gibbons, Marco Gruteser, Za{\"{\i}}d
  Harchaoui, Chaoyang He, Lie He, Zhouyuan Huo, Ben Hutchinson, Justin Hsu,
  Martin Jaggi, Tara Javidi, Gauri Joshi, Mikhail Khodak, Jakub
  Kone{\v{c}}n{\'y}, Aleksandra Korolova, Farinaz Koushanfar, Sanmi Koyejo,
  Tancr{\`{e}}de Lepoint, Yang Liu, Prateek Mittal, Mehryar Mohri, Richard
  Nock, Ayfer {\"{O}}zg{\"{u}}r, Rasmus Pagh, Hang Qi, Daniel Ramage, Ramesh
  Raskar, Mariana Raykova, Dawn Song, Weikang Song, Sebastian~U. Stich, Ziteng
  Sun, Ananda~Theertha Suresh, Florian Tram{\`{e}}r, Praneeth Vepakomma, Jianyu
  Wang, Li~Xiong, Zheng Xu, Qiang Yang, Felix~X. Yu, Han Yu, and Sen Zhao.
\newblock {Advances and Open Problems in Federated Learning}.
\newblock \emph{Found. Trends Mach. Learn.}, 14\penalty0 (1-2):\penalty0
  1--210, 2021{\natexlab{b}}.

\bibitem[Kalinin and Lampert(2024)]{kalinin2024banded}
Nikita Kalinin and Christoph Lampert.
\newblock {Banded Square Root Matrix Factorization for Differentially Private
  Model Training}.
\newblock \emph{arXiv Preprint}, 2024.

\bibitem[Kalinin et~al.(2025)Kalinin, McKenna, Upadhyay, and
  Lampert]{kalinin2025back}
Nikita~P Kalinin, Ryan McKenna, Jalaj Upadhyay, and Christoph~H Lampert.
\newblock Back to square roots: An optimal bound on the matrix factorization
  error for multi-epoch differentially private sgd.
\newblock \emph{arXiv preprint arXiv:2505.12128}, 2025.

\bibitem[Kifer et~al.(2012)Kifer, Smith, and Thakurta]{kifer2012private}
Daniel Kifer, Adam Smith, and Abhradeep Thakurta.
\newblock Private convex empirical risk minimization and high-dimensional
  regression.
\newblock In \emph{Conference on Learning Theory}, pages 25--1, 2012.

\bibitem[Koloskova et~al.(2023{\natexlab{a}})Koloskova, Hendrikx, and
  Stich]{koloskova2023revisiting}
Anastasia Koloskova, Hadrien Hendrikx, and Sebastian~U. Stich.
\newblock {Revisiting Gradient Clipping: Stochastic bias and tight convergence
  guarantees}.
\newblock In \emph{ICML}, volume 202, pages 17343--17363, 2023{\natexlab{a}}.

\bibitem[Koloskova et~al.(2023{\natexlab{b}})Koloskova, McKenna, Charles, Rush,
  and McMahan]{koloskova2023correlated}
Anastasiia Koloskova, Ryan McKenna, Zachary Charles, John Rush, and H.~Brendan
  McMahan.
\newblock Gradient descent with linearly correlated noise: Theory and
  applications to differential privacy.
\newblock In A.~Oh, T.~Naumann, A.~Globerson, K.~Saenko, M.~Hardt, and
  S.~Levine, editors, \emph{Advances in Neural Information Processing Systems},
  volume~36, pages 35761--35773. Curran Associates, Inc., 2023{\natexlab{b}}.
\newblock URL
  \url{https://proceedings.neurips.cc/paper_files/paper/2023/file/70255afc962aca0930327c090eb7d8c5-Paper-Conference.pdf}.

\bibitem[Kulkarni et~al.(2021)Kulkarni, Lee, and Liu]{kulkarni2021private}
Janardhan Kulkarni, Yin~Tat Lee, and Daogao Liu.
\newblock Private non-smooth erm and sco in subquadratic steps.
\newblock \emph{Advances in Neural Information Processing Systems}, 34, 2021.

\bibitem[Kwapie{\'n} and Pe{\l}czy{\'n}ski(1970)]{kwapien1970main}
Stanis{\l}aw Kwapie{\'n} and Aleksander Pe{\l}czy{\'n}ski.
\newblock The main triangle projection in matrix spaces and its applications.
\newblock \emph{Studia Mathematica}, 34\penalty0 (1):\penalty0 43--67, 1970.

\bibitem[Lee et~al.(2008)Lee, Shraibman, and {\v{S}}palek]{lee2008direct}
Troy Lee, Adi Shraibman, and Robert {\v{S}}palek.
\newblock {A Direct Product Theorem for Discrepancy}.
\newblock In \emph{2008 23rd Annual IEEE Conference on Computational
  Complexity}, pages 71--80. IEEE, 2008.
\newblock URL
  \url{https://web.archive.org/web/20230109145533/https://www2.mta.ac.il/~adish/Pubs/Papers/DPTDiscrepancy.pdf}.

\bibitem[Li et~al.(2014)Li, Hay, Miklau, and Wang]{li2014data}
Chao Li, Michael Hay, Gerome Miklau, and Yue Wang.
\newblock A data-and workload-aware algorithm for range queries under
  differential privacy.
\newblock \emph{arXiv preprint arXiv:1410.0265}, 2014.

\bibitem[Li et~al.(2015)Li, Miklau, Hay, McGregor, and Rastogi]{li2015matrix}
Chao Li, Gerome Miklau, Michael Hay, Andrew McGregor, and Vibhor Rastogi.
\newblock The matrix mechanism: optimizing linear counting queries under
  differential privacy.
\newblock \emph{The VLDB journal}, 24:\penalty0 757--781, 2015.

\bibitem[Liu et~al.(2024)Liu, Upadhyay, and Zou]{liu2024optimality}
Jingcheng Liu, Jalaj Upadhyay, and Zongrui Zou.
\newblock Optimality of matrix mechanism on $\ell_p^{p}$-metric.
\newblock \emph{arXiv preprint arXiv:2406.02140}, 2024.

\bibitem[Marshall et~al.(2024)Marshall, Xiao, Agarwala, and
  Paquette]{marshall2024clip}
Noah Marshall, Ke~Liang Xiao, Atish Agarwala, and Elliot Paquette.
\newblock {To Clip or not to Clip: the Dynamics of SGD with Gradient Clipping
  in High-Dimensions}.
\newblock \emph{arXiv Preprint}, 2024.

\bibitem[Mathias(1993)]{mathias1993hadamard}
Roy Mathias.
\newblock The hadamard operator norm of a circulant and applications.
\newblock \emph{SIAM journal on matrix analysis and applications}, 14\penalty0
  (4):\penalty0 1152--1167, 1993.

\bibitem[Matou{\v{s}}ek et~al.(2020)Matou{\v{s}}ek, Nikolov, and
  Talwar]{matouvsek2020factorization}
Ji{\v{r}}{\'\i} Matou{\v{s}}ek, Aleksandar Nikolov, and Kunal Talwar.
\newblock {Factorization Norms and Hereditary Discrepancy}.
\newblock \emph{International Mathematics Research Notices}, 2020\penalty0
  (3):\penalty0 751--780, 2020.
\newblock URL \url{https://arxiv.org/abs/1408.1376}.

\bibitem[McKenna(2024)]{mckenna2024scaling}
Ryan McKenna.
\newblock {Scaling up the Banded Matrix Factorization Mechanism for
  Differentially Private ML}.
\newblock \emph{arXiv preprint arXiv:2405.15913}, 2024.

\bibitem[McKenna et~al.(2020)McKenna, Maity, Mazumdar, and
  Miklau]{mckenna13workload}
Ryan McKenna, Raj~Kumar Maity, Arya Mazumdar, and Gerome Miklau.
\newblock A workload-adaptive mechanism for linear queries under local
  differential privacy.
\newblock \emph{Proceedings of the VLDB Endowment}, 13\penalty0 (11), 2020.

\bibitem[McKenna et~al.(2021)McKenna, Miklau, Hay, and
  Machanavajjhala]{mckenna2021hdmm}
Ryan McKenna, Gerome Miklau, Michael Hay, and Ashwin Machanavajjhala.
\newblock Hdmm: Optimizing error of high-dimensional statistical queries under
  differential privacy.
\newblock \emph{arXiv preprint arXiv:2106.12118}, 2021.

\bibitem[McMahan(2011)]{mcmahan2011follow}
Brendan McMahan.
\newblock Follow-the-regularized-leader and mirror descent: Equivalence
  theorems and l1 regularization.
\newblock In \emph{Proceedings of the Fourteenth International Conference on
  Artificial Intelligence and Statistics}, pages 525--533. JMLR Workshop and
  Conference Proceedings, 2011.

\bibitem[McMahan and Pillutla(2025)]{mcmahan2025inverse}
Brendan McMahan and Krishna Pillutla.
\newblock {An Inversion Theorem for Buffered Linear Toeplitz (BLT) Matrices and
  Applications to Streaming Differential Privacy}.
\newblock \emph{Preprint}, 2025.

\bibitem[McMahan et~al.(2018)McMahan, Ramage, Talwar, and
  Zhang]{mcmahan2017learning}
H.~Brendan McMahan, Daniel Ramage, Kunal Talwar, and Li~Zhang.
\newblock {Learning Differentially Private Recurrent Language Models}.
\newblock In \emph{ICLR}, 2018.

\bibitem[McMahan et~al.(2024)McMahan, Xu, and Zhang]{mcmahan2024hassle}
H~Brendan McMahan, Zheng Xu, and Yanxiang Zhang.
\newblock {A Hassle-free Algorithm for Private Learning in Practice: Don't Use
  Tree Aggregation, Use BLTs}.
\newblock In \emph{EMNLP}, 2024.

\bibitem[Nesterov(2009)]{nesterov2009primal}
Yurii Nesterov.
\newblock Primal-dual subgradient methods for convex problems.
\newblock \emph{Mathematical Programming}, 120\penalty0 (1):\penalty0 221--259,
  2009.

\bibitem[Newman(1964)]{newman1964rational}
D.~J. Newman.
\newblock {Rational approximation to $| x| $.}
\newblock \emph{Michigan Mathematical Journal}, 11\penalty0 (1):\penalty0 11 --
  14, 1964.
\newblock \doi{10.1307/mmj/1028999029}.
\newblock URL \url{https://doi.org/10.1307/mmj/1028999029}.

\bibitem[Nikolov et~al.(2016)Nikolov, Talwar, and Zhang]{nikolov2016geometry}
Aleksandar Nikolov, Kunal Talwar, and Li~Zhang.
\newblock {The Geometry of Differential Privacy: the Sparse and Approximate
  Cases}.
\newblock \emph{SIAM Journal on Computing}, 45\penalty0 (2):\penalty0 575--616,
  2016.

\bibitem[Perdomo et~al.(2020)Perdomo, Zrnic, Mendler-D{\"u}nner, and
  Hardt]{perdomo2020performative}
Juan Perdomo, Tijana Zrnic, Celestine Mendler-D{\"u}nner, and Moritz Hardt.
\newblock {Performative Prediction}.
\newblock In \emph{International Conference on Machine Learning}, pages
  7599--7609. PMLR, 2020.

\bibitem[Ponomareva et~al.(2023)Ponomareva, Hazimeh, Kurakin, Xu, Denison,
  McMahan, Vassilvitskii, Chien, and Thakurta]{ponomareva2023dp}
Natalia Ponomareva, Hussein Hazimeh, Alex Kurakin, Zheng Xu, Carson Denison,
  H~Brendan McMahan, Sergei Vassilvitskii, Steve Chien, and Abhradeep~Guha
  Thakurta.
\newblock {How to DP-fy ML: A Practical Guide to Machine Learning with
  Differential Privacy}.
\newblock \emph{Journal of Artificial Intelligence Research}, 77:\penalty0
  1113--1201, 2023.

\bibitem[Qardaji et~al.(2013)Qardaji, Yang, and Li]{qardaji2013understanding}
Wahbeh Qardaji, Weining Yang, and Ninghui Li.
\newblock Understanding hierarchical methods for differentially private
  histograms.
\newblock \emph{Proceedings of the VLDB Endowment}, 6\penalty0 (14):\penalty0
  1954--1965, 2013.

\bibitem[Schaipp et~al.(2024)Schaipp, Garrigos, Simsekli, and
  Gower]{schaipp2024sgd}
Fabian Schaipp, Guillaume Garrigos, Umut Simsekli, and Robert Gower.
\newblock {SGD with Clipping is Secretly Estimating the Median Gradient}.
\newblock \emph{arXiv Preprint}, 2024.

\bibitem[Schmidt et~al.(2011)Schmidt, Kim, and Sra]{schmidt2011projected}
Mark Schmidt, Dongmin Kim, and Suvrit Sra.
\newblock {Projected Newton-type Methods in Machine Learning}.
\newblock 2011.

\bibitem[Smith et~al.(2017)Smith, Thakurta, and Upadhyay]{smith2017interaction}
Adam~D. Smith, Abhradeep Thakurta, and Jalaj Upadhyay.
\newblock {Is Interaction Necessary for Distributed Private Learning?}
\newblock In \emph{IEEE Symposium on Security and Privacy}, pages 58--77, 2017.

\bibitem[Song et~al.(2013)Song, Chaudhuri, and Sarwate]{song2013stochastic}
Shuang Song, Kamalika Chaudhuri, and Anand~D Sarwate.
\newblock Stochastic gradient descent with differentially private updates.
\newblock In \emph{2013 IEEE global conference on signal and information
  processing}, pages 245--248. IEEE, 2013.

\bibitem[Song et~al.(2018)Song, Little, Mehta, Vinterbo, and
  Chaudhuri]{song2018differentially}
Shuang Song, Susan Little, Sanjay Mehta, Staal Vinterbo, and Kamalika
  Chaudhuri.
\newblock Differentially private continual release of graph statistics.
\newblock \emph{arXiv preprint arXiv:1809.02575}, 2018.

\bibitem[Steinberg(2005)]{steinberg2005computation}
Daureen Steinberg.
\newblock Computation of matrix norms with applications to robust optimization.
\newblock \emph{Research thesis, Technion-Israel University of Technology}, 2,
  2005.

\bibitem[Steinke(2022)]{steinke2022compositiondifferentialprivacy}
Thomas Steinke.
\newblock {Composition of Differential Privacy \& Privacy Amplification by
  Subsampling}, 2022.
\newblock URL \url{https://arxiv.org/abs/2210.00597}.

\bibitem[Strang(2022)]{strang2022introduction}
Gilbert Strang.
\newblock \emph{Introduction to linear algebra}.
\newblock SIAM, 2022.

\bibitem[Thakurta and Smith(2013)]{guha2013nearly}
Abhradeep~Guha Thakurta and Adam Smith.
\newblock {(Nearly) Optimal Algorithms for Private Online Learning in
  Full-information and Bandit Settings}.
\newblock \emph{Advances in Neural Information Processing Systems}, 26, 2013.

\bibitem[Tropp(2003)]{tropp2003topics}
Joel~Aaron Tropp.
\newblock {Topics in Sparse Approximation}.
\newblock 2003.

\bibitem[{US Census Bureau}(2021)]{uscensus}
{US Census Bureau}.
\newblock {Census Bureau Sets Key Parameters to Protect Privacy in 2020 Census
  Results}.
\newblock
  \url{https://www.census.gov/newsroom/press-releases/2021/2020-census-key-parameters.html},
  2021.
\newblock Accessed: 2025-03-21.

\bibitem[Vadhan(2017)]{vadhan2017complexity}
Salil~P. Vadhan.
\newblock {The Complexity of Differential Privacy}.
\newblock In \emph{Tutorials on the Foundations of Cryptography}, pages
  347--450. Springer International Publishing, 2017.

\bibitem[Wang et~al.(2024)Wang, Zhang, Su, and Zhu]{wang2024comprehensive}
Liyuan Wang, Xingxing Zhang, Hang Su, and Jun Zhu.
\newblock {A Comprehensive Survey of Continual Learning: Theory, Method and
  Application}.
\newblock \emph{IEEE Transactions on Pattern Analysis and Machine
  Intelligence}, 2024.

\bibitem[Woodruff and Zhou(2022)]{woodruff2022tight}
David~P Woodruff and Samson Zhou.
\newblock Tight bounds for adversarially robust streams and sliding windows via
  difference estimators.
\newblock In \emph{2021 IEEE 62nd Annual Symposium on Foundations of Computer
  Science (FOCS)}, pages 1183--1196. IEEE, 2022.

\bibitem[Xiao et~al.(2023{\natexlab{a}})Xiao, Xiang, Wang, and
  Devadas]{xiao2023theory}
Hanshen Xiao, Zihang Xiang, Di~Wang, and Srinivas Devadas.
\newblock {A Theory to Instruct Differentially-Private Learning via Clipping
  Bias Reduction}.
\newblock In \emph{{IEEE} Symposium on Security and Privacy (SP)}, pages
  2170--2189. {IEEE}, 2023{\natexlab{a}}.

\bibitem[Xiao(2009)]{xiao2009dual}
Lin Xiao.
\newblock Dual averaging method for regularized stochastic learning and online
  optimization.
\newblock \emph{Advances in Neural Information Processing Systems}, 22, 2009.

\bibitem[Xiao et~al.(2023{\natexlab{b}})Xiao, He, Zhang, and
  Kifer]{xiao2023optimal}
Yingtai Xiao, Guanlin He, Danfeng Zhang, and Daniel Kifer.
\newblock An optimal and scalable matrix mechanism for noisy marginals under
  convex loss functions.
\newblock \emph{Advances in Neural Information Processing Systems},
  36:\penalty0 20495--20539, 2023{\natexlab{b}}.

\bibitem[Xu et~al.(2023)Xu, Zhang, Andrew, Choquette{-}Choo, Kairouz, McMahan,
  Rosenstock, and Zhang]{xu2023federated}
Zheng Xu, Yanxiang Zhang, Galen Andrew, Christopher~A. Choquette{-}Choo, Peter
  Kairouz, H.~Brendan McMahan, Jesse Rosenstock, and Yuanbo Zhang.
\newblock {Federated Learning of Gboard Language Models with Differential
  Privacy}.
\newblock In \emph{ACL (Industry Track)}, pages 629--639, 2023.

\bibitem[Yuan et~al.(2016)Yuan, Yang, Zhang, and Hao]{yuan2016convex}
Ganzhao Yuan, Yin Yang, Zhenjie Zhang, and Zhifeng Hao.
\newblock Convex optimization for linear query processing under approximate
  differential privacy.
\newblock In \emph{Proceedings of the 22nd ACM SIGKDD International Conference
  on Knowledge Discovery and Data Mining}, pages 2005--2014, 2016.

\bibitem[Zhang et~al.(2020)Zhang, He, Sra, and Jadbabaie]{zhang2020gradient}
Jingzhao Zhang, Tianxing He, Suvrit Sra, and Ali Jadbabaie.
\newblock {Why Gradient Clipping Accelerates Training: A Theoretical
  Justification for Adaptivity}.
\newblock In \emph{ICLR}, 2020.

\bibitem[Zhang et~al.(2022{\natexlab{a}})Zhang, Tran, and
  Cutkosky]{zhang2022differentially}
Qinzi Zhang, Hoang Tran, and Ashok Cutkosky.
\newblock Differentially private online-to-batch for smooth losses.
\newblock In \emph{NeurIPS}, 2022{\natexlab{a}}.

\bibitem[Zhang et~al.(2022{\natexlab{b}})Zhang, Chen, Hong, Wu, and
  Yi]{zhang2022understanding}
Xinwei Zhang, Xiangyi Chen, Mingyi Hong, Zhiwei~Steven Wu, and Jinfeng Yi.
\newblock {Understanding Clipping for Federated Learning: Convergence and
  Client-Level Differential Privacy}.
\newblock In \emph{ICML}, 2022{\natexlab{b}}.

\bibitem[Zhu et~al.(1997)Zhu, Byrd, Lu, and Nocedal]{zhu1997algorithm}
Ciyou Zhu, Richard~H Byrd, Peihuang Lu, and Jorge Nocedal.
\newblock {Algorithm 778: L-BFGS-B: Fortran subroutines for large-scale
  bound-constrained optimization}.
\newblock \emph{ACM Transactions on mathematical software (TOMS)}, 23\penalty0
  (4):\penalty0 550--560, 1997.

\end{thebibliography}
